\documentclass{article} %
\usepackage{iclr2025_conference,times}

\iclrfinalcopy

\usepackage{amsmath,amsfonts,bm}

\def\eqref#1{equation~\ref{#1}}

\def\1{\bm{1}}

\DeclareMathAlphabet{\mathsfit}{\encodingdefault}{\sfdefault}{m}{sl}
\SetMathAlphabet{\mathsfit}{bold}{\encodingdefault}{\sfdefault}{bx}{n}

\usepackage{xspace}
\usepackage{hyperref}
\usepackage{url}
\usepackage{adjustbox}
\usepackage{colortbl}
\usepackage{makecell}
\usepackage{float}
\usepackage{url}
\usepackage{balance}
\usepackage{bbding}
\usepackage{hyperref}
\usepackage{graphicx}
\usepackage{sidecap}
\usepackage{microtype}
\usepackage{graphicx}
\usepackage{subfigure}
\usepackage{booktabs} 
\usepackage{multirow}
\usepackage{array}
\usepackage{subcaption}
\usepackage{graphicx}
\usepackage{makecell}
\usepackage{amsmath}
\usepackage{amssymb}
\usepackage{booktabs}
\usepackage{float}
\usepackage{multirow}
\usepackage{adjustbox} 
\usepackage{amsmath}
\usepackage{amssymb}
\usepackage{widetext}
\usepackage{mathtools}
\usepackage{amsthm}
\usepackage{caption}
\usepackage{tikz}
\usepackage{dashrule}
\usepackage{wrapfig}
\usepackage[capitalize,noabbrev]{cleveref}

\usepackage{ifthen}
\newboolean{showchanges}
\setboolean{showchanges}{false}  %

\newcommand{\add}[1]{%
    \ifthenelse{\boolean{showchanges}}%
        {\textcolor{blue}{#1}}
        {#1\relax}
}

\title{Overcoming False Illusions in Real-World Face Restoration with Multi-Modal Guided Diffusion Model}

\author{
    Keda Tao\textsuperscript{1},\quad Jinjin Gu\textsuperscript{2}\thanks{Corresponding authors},\quad Yulun Zhang\textsuperscript{3},\quad Xiucheng Wang\textsuperscript{1},\quad Nan Cheng\textsuperscript{1, 4}\footnotemark[1] \\
    \textsuperscript{1}Xidian University, \textsuperscript{2}The University of Sydney, \textsuperscript{3}Shanghai Jiao Tong University\\ \textsuperscript{4}State Key Laboratory of Integrated Services Networks \\
    \texttt{KD.TAO@outlook.com, jinjin.gu@sydney.edu.au, yulzhang@sjtu.edu.cn} \\
    \texttt{xcwang\_1@stu.xidian.edu.cn, dr.nan.cheng@ieee.org}
}

\begin{document}

\maketitle

\begin{widetext}
\vspace{-8mm}
\begin{figure}
  \captionsetup{font={small}, skip=14pt}
  \includegraphics[width=1\textwidth]{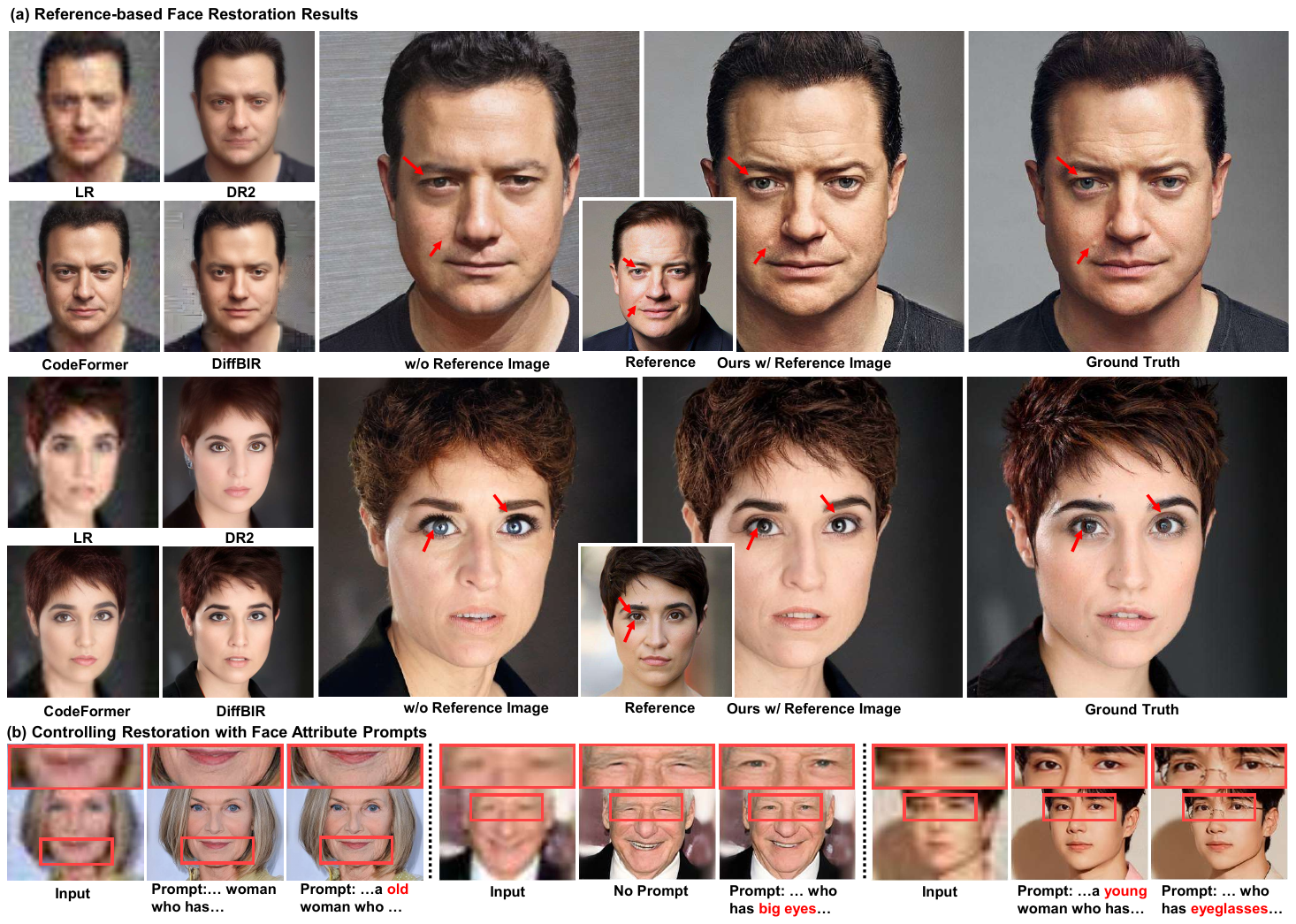}
  \vspace{-8mm}
  \caption{The proposed MGFR model demonstrates an exceptional ability in restoring low-quality face images, yielding more outstanding visual effects with the addition of reference images, particularly in situations of extreme degradation, shown in \textbf{(a)}. Furthermore, the model is capable of target-specific restoration in \textbf{(b)}, directed by facial attribute prompts. This encompasses defining facial age characteristics (Case 1), adjusting the restoration process based on attribute prompts (Case 2), and executing precise modifications to facial elements (Case 3). \emph{w/o Reference Image} means the results of our model without introducing reference image.}
  \label{figt}
\end{figure}
\vspace{0mm}
\end{widetext}

\begin{abstract}
We introduce a novel Multi-modal Guided Real-World Face Restoration (MGFR) technique designed to improve the quality of facial image restoration from low-quality inputs.
Leveraging a blend of attribute text prompts, high-quality reference images, and identity information, MGFR can mitigate the generation of false facial attributes and identities often associated with generative face restoration methods. By incorporating a dual-control adapter and a two-stage training strategy, our method effectively utilizes multi-modal prior information for targeted restoration tasks. We also present the Reface-HQ dataset, comprising over 21,000 high-resolution facial images across 4800 identities, to address the need for reference face training images. Our approach achieves superior visual quality in restoring facial details under severe degradation and allows for controlled restoration processes, enhancing the accuracy of identity preservation and attribute correction. Including negative quality samples and attribute prompts in the training further refines the model's ability to generate detailed and perceptually accurate images. 
\end{abstract}

\section{Introduction}
\label{Intro}
\begin{figure}[t]
\captionsetup{font={small}, skip=14pt}
\scriptsize
\centering
\begin{tabular}{ccc}
\hspace{-0.5cm}
\begin{adjustbox}{valign=t}
\begin{tabular}{c}
\end{tabular}
\end{adjustbox}
\begin{adjustbox}{valign=t}
\begin{tabular}{ccccccccc}
\includegraphics[width=0.12\linewidth]{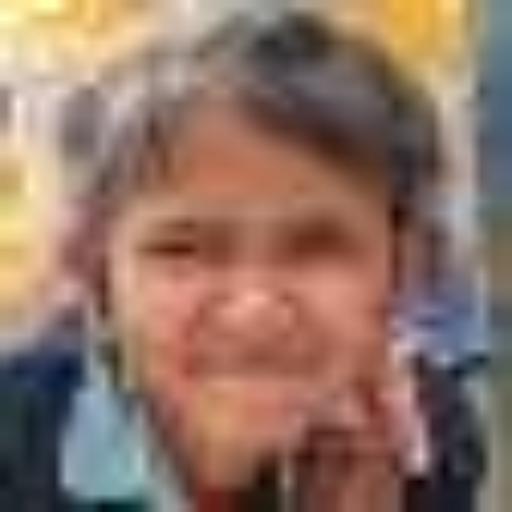} \hspace{-4mm} &
\includegraphics[width=0.12\linewidth]{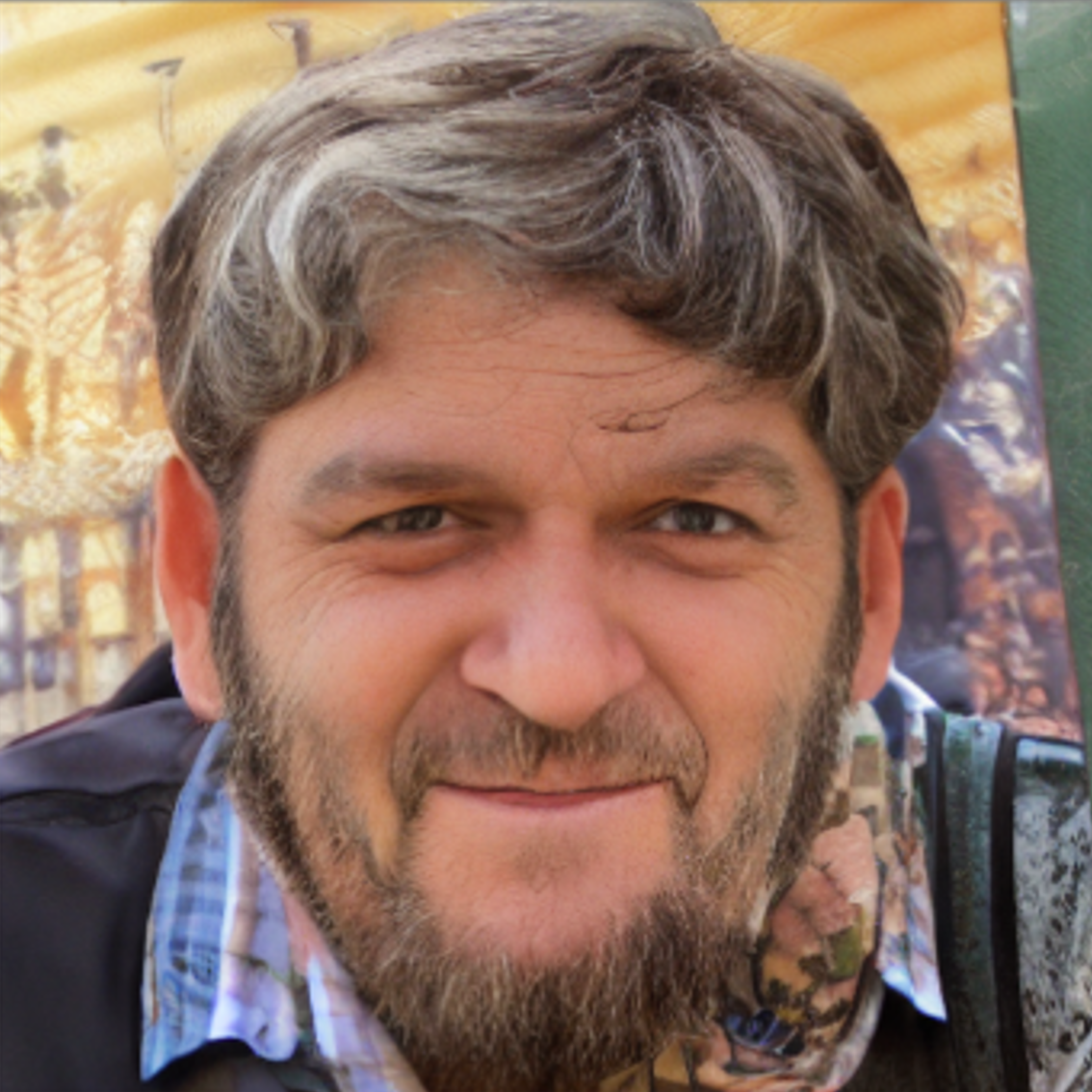} \hspace{-4mm} &
\includegraphics[width=0.12\linewidth]{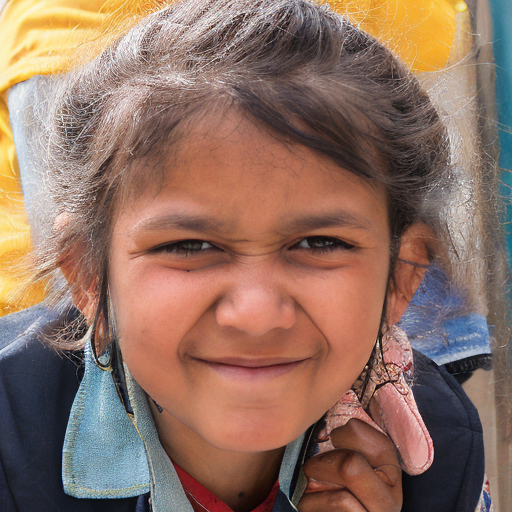} \hspace{-4mm} &
\includegraphics[width=0.12\linewidth]{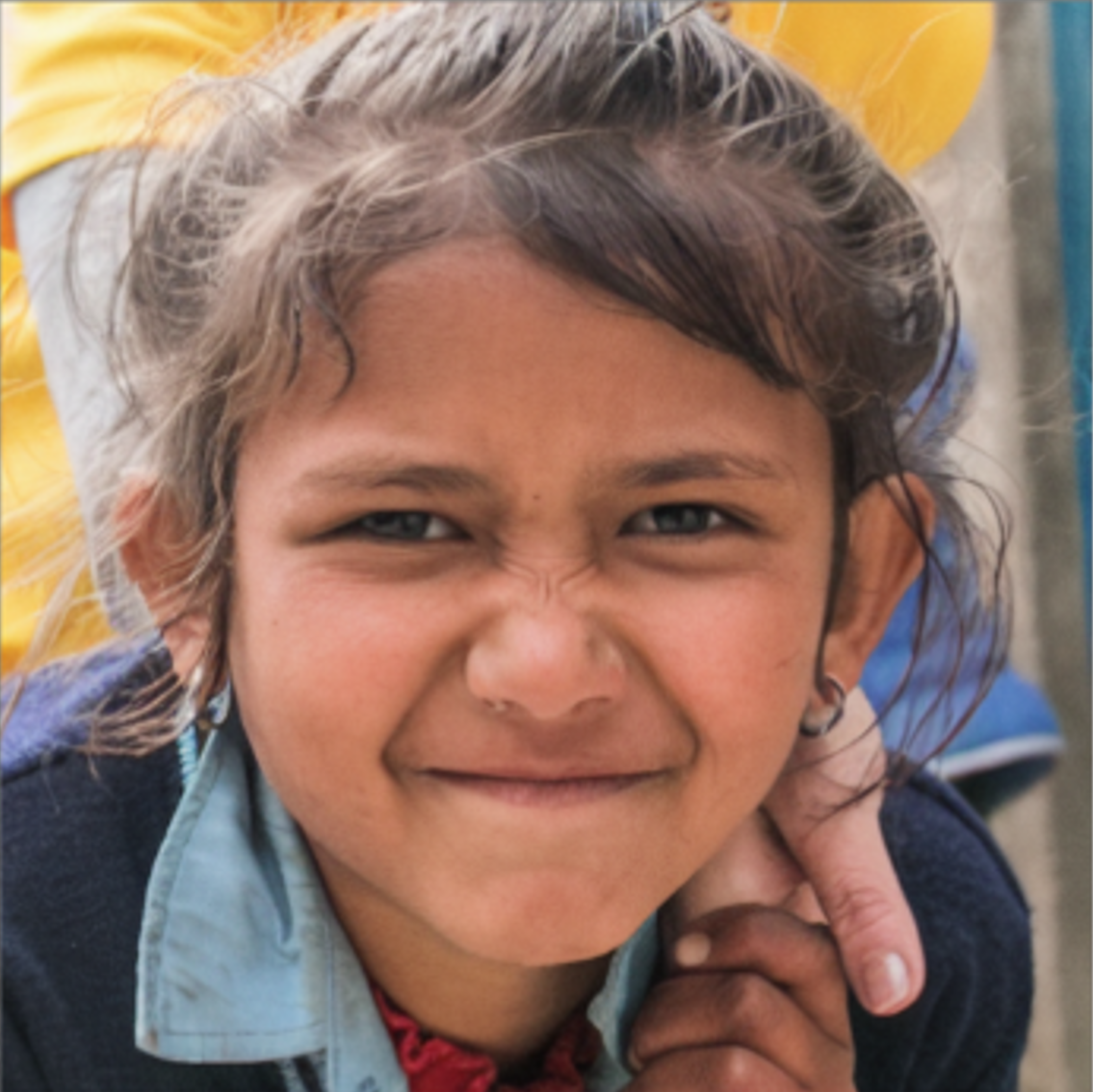} \hspace{-4mm} &
\tikz[baseline,overlay] \draw[dashed, thick] (0,0.12\linewidth) -- (0,-\tabcolsep); 
\includegraphics[width=0.12\linewidth]{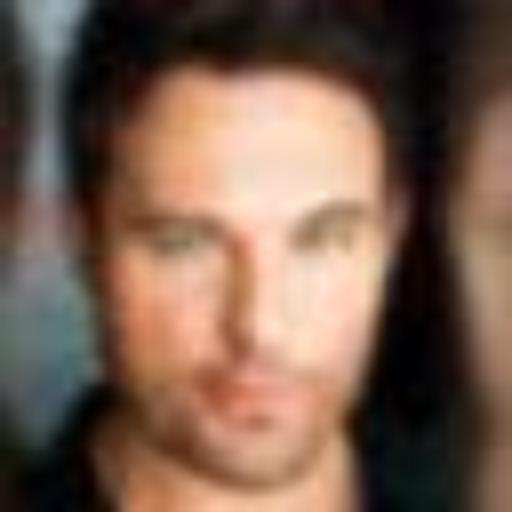} \hspace{-4mm} &
\includegraphics[width=0.12\linewidth]{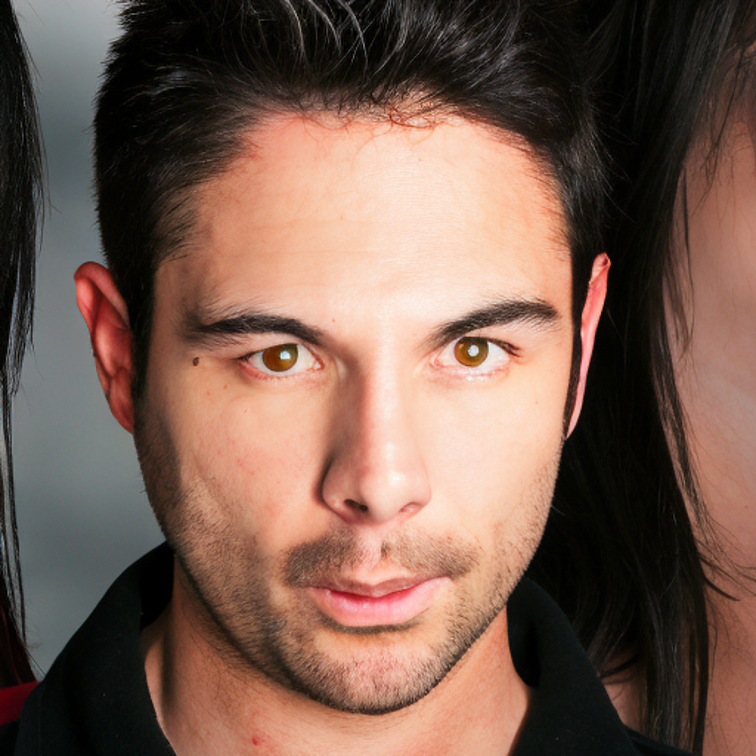} \hspace{-4mm} &
\includegraphics[width=0.12\linewidth]{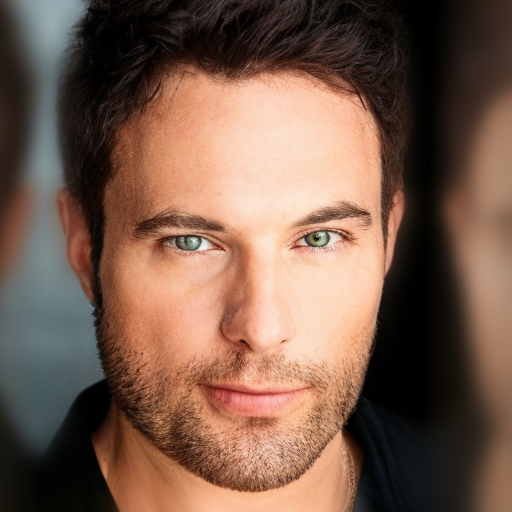} 
\hspace{-4mm} &
\includegraphics[width=0.12\linewidth]{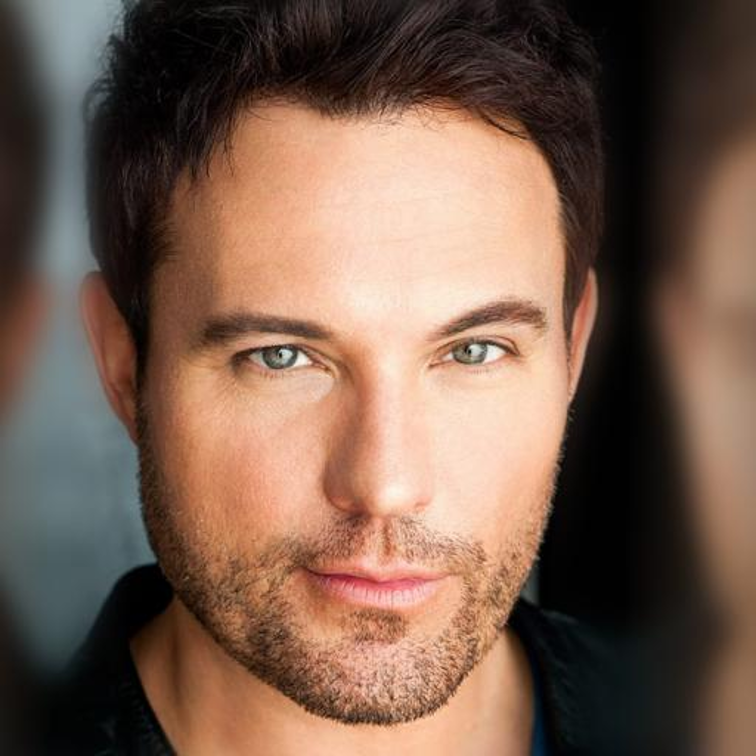} \hspace{-4mm} &
\\
\textbf{(a)} \hspace{3mm} LQ \hspace{0mm} &
No Prompt \hspace{-4mm} &
\parbox{0.12\linewidth}{\emph{`woman, young, no beard...'}} \hspace{-4mm} &
GT \hspace{-4mm} &

\textbf{(b)} \hspace{3mm} LQ \hspace{0mm} &
w/o Reference \hspace{-4mm} &
w/ Reference \hspace{-4mm} &
GT \hspace{-4mm} &
\vspace{1mm}
\end{tabular}
\end{adjustbox}
\hspace{-0.55cm}
\end{tabular}
\vspace{-6mm}
\caption{\textbf{Motivation.} In conditions of severe degradation, the loss of facial identity information becomes profoundly pronounced without a reference image. During the face restoration process, distortions of facial attributes, including gender and age, are commonly encountered. Appropriate attribute prompts can offer additional reference points and exert control in the recovery process.}
\label{fig2}
\vspace{-4mm}
\end{figure}

Real-World Face Restoration (FR) aims to reconstruct high-resolution, high-quality (HQ) facial images from their degraded, low-resolution observations.
Recent works, leveraging powerful generative priors and diffusion models, have achieved significant progress \citep{menon2020pulse, yang2021gan, wang2021towards, diffbir, dr2}, particularly in addressing severely degraded facial images.
However, the information contained in the low-quality (LQ) inputs is limited.
FR inevitably introduces the illusion of generation, producing results with different facial attributes or even different identities from the target image. 
For example, in \cref{figt} (a) and \cref{fig2}, we cannot effectively predict the eye colour and skin characteristics of the person in the LQ input, resulting in the output results -- even the quality can be improved -- having an apparent perceptual distance from the target image.
Many applications find this unacceptable, as humans can readily identify these flaws. Achieving optimal facial image recovery requires effectively tackling false hallucinations.

Practically, we find that for the restoration of specific face images, we can obtain a lot of prior information.
For example, we may know this person's various attributes and identity, and there may even be other clear images of this person in the photo album.
Suppose we can use this information as additional guidance to guide the restoration. In that case, we can alleviate the impact of false illusions on key issues, thus helping to generate facial details that better suit our needs.
For example, in \cref{fig2} (a), when we provide an additional key description of gender and age, we can correct the illusion. In \cref{figt} (a) and \cref{fig2} (b), additional high-quality images are used as reference, and the details of the eyes and skin texture can be accurately generated.
What is even more gratifying is that this kind of prior information can be widely obtained, making this problem of application significant.

This work proposes a method called Multi-modal Guided Real-World Face Restoration (MGFR).
We aim to use multiple control methods to consider diverse multi-modal prior information in FR to restore face images in a targeted manner.
Specifically, MGFR uses attribute text prompts, HQ reference images, and identity information as priors for collaborative guidance during restoration.
We designed a dual-control adapter with a two-stage training strategy to balance the complex multi-modal and multi-source prior information.
This dual controller is compatible with pre-trained generative diffusion models \citep{rombach2022high} and prioritizes restoration tasks while incorporating additional multi-modal guidance.
In addition, we collect the \textbf{Reface-HQ} dataset to address the scarcity of reference image samples containing over 4800 identities and 21000 high-resolution facial images.
Based on the FFHQ \citep{ffhq} and the proposed Reface-HQ datasets, we develop a high-quality synthetic dataset for model training enriched with attribute text prompts.
Furthermore, we adopt a counterintuitive strategy to integrate negative-quality samples with negative-quality prompts and negative-attribute prompts into training to enhance perceptual quality and detail generation.

The proposed MGFR model shows exemplary performance in the FR task, achieving superior visual quality in facial details, especially under severe degradation conditions.
MGFR can take a high-resolution reference image as prior information and restore important details based on the reference image that cannot be displayed in the LQ input.
The identity information provided by the reference image will also be considered in FR to ensure that the restoration does not change the identity characteristics.
In addition, MGFR can also provide a certain degree of control over the restoration process through attribute text prompts, significantly enhancing the feasibility of FR.
As shown in \cref{figt} (b), textual prompts fulfil a dual function: they significantly reduce facial attribute illusions, such as ``big eyes'' or ``old'', and also guide the restoration of specific facial features, such as ``wearing glasses'' and ``young''.

\section{Related Works}

\paragraph{Real-World Face Image Restoration}
\vspace{-2mm}
Real-world face restoration (FR) concentrates on the challenging task of reconstructing HQ face images from LQ inputs.
These LQ inputs are often blemished by various forms of quality degradation, such as low-resolution \citep{chen2018fsrnet, dong2014learning, lim2017enhanced}, blur \citep{kupyn2018deblurgan, shen2018deep}, noise \citep{zhang2017beyond}, and JPEG compression artifacts \citep{dong2015compression}, \emph{etc}.
FR heavily relies on facial priors, such as facial landmarks \citep{chen2018fsrnet}, parsing maps \citep{chen2018fsrnet, chen2021progressive}, and facial component heatmaps \citep{yu2018face}.
Generative priors \citep{karras2020analyzing,rombach2022high, gu2020image, shen2020interpreting} have also emerged as fundamental elements in providing vibrant textures and details in FR \citep{menon2020pulse, hu2023dear, zhu2022blind}.
Advanced techniques like GPEN \citep{gpen}, GFP-GAN \citep{gfp}, and GLEAN \citep{chan2021glean} are recognized for more effectively incorporating these priors within encoder-decoder structures. 
There are also works that considerably reduce the uncertainty commonly associated with generative priors \citep{gu2022vqfr,coderformer,wang2022restoreformer}, which are trained on discrete feature codebooks for high-quality facial images.
Recently, diffusion models like DiffBIR \citep{diffbir} have revitalized interest in this area, leveraging the generative power of pre-trained LDM as a prior.
DR2 \citep{dr2} also contributes by transforming input images into noisy states and then denoising them to capture the essential semantic information.

\paragraph{Reference-Based Face Image Restoration}
\vspace{-2mm}
Reference-based face restoration utilizes HQ images of the same identity as references. This concept was first introduced in \citep{Li_2018_ECCV}. To address discrepancies in poses and expressions, GWAInet \citep{Dogan_2019_CVPR_Workshops} and the later work of Li et al. \citep{Li_2020_CVPR, li2018learning} focused on more effectively directing deformations or choosing the optimal reference image for reconstruction. MyStyle \citep{nitzan2022mystyle} adopts a unique approach by refining StyleGAN \citep{ffhq} with numerous reference images based on personal appearance. DMDNet \citep{9921338} employs a dictionary constructed from diverse, high-quality facial images to rehabilitate degraded images using its high-quality components. In the MGFR framework, incorporating a single reference image is vital for tailoring the restoration to individual faces. Unlike conventional methods, MGFR does not require strict alignment constraints on expressions or postures.

\paragraph{Multi-modal Guided Generation}
\vspace{-2mm}
Diffusion models have shown significant effectiveness in a broad range of image processing tasks.
Current methods \citep{chen2023hierarchical,zhang2023adding,yu2024scaling,chen2023image} employ pre-trained text-to-image diffusion models \citep{rombach2022high} for image processing, demonstrating the potential of language as a comprehensive input for image reconstruction tasks.
Concurrently, approaches like ControlNet \citep{zhang2023adding}, T2I-adapter \citep{mou2023t2i}, and ControlNet-XS \citep{zavadski2023controlnet} have further developed the integration of more intricate condition controls within the text-to-image framework, facilitating more precise and tailored image generation.
Nevertheless, the field of FR, particularly in the utilization of natural language prompts, continues to be an area of untapped potential.

\section{Methodology}
\vspace{-2mm}
The proposed MGFR method is able to take face attribute text prompts, reference images, and identity information as input to alleviate illusions and improve visual effects.
MGFR involves controlling information from multiple modalities.
However, we found that if the model is directly trained to process control information from multiple sources and of different importance, it is not easy to utilize all the information effectively. The model may ignore the more complex information to utilize.
This causes some of the controls to fail and reduces image quality.
In our method, text prompts are the most complex control information because they involve understanding text and the correspondence between text and face attributes.
Therefore, we divide the training into two stages. In the first stage, the training focuses on the basic text-guided restoration model (\cref{sec:Text-guided}).
This allows the model to restore high-quality images and understand facial attributes.
Then, we introduce other control information on this basis.
The second stage introduces the HQ reference image and face identity information as the control means (\cref{sec:Multimodal}).
To improve the image effect further, \cref{sec:Negative} describes negative examples and the adopted prompting strategy.

\begin{figure*}
  \centering
  \captionsetup{font={small}, skip=12pt}
  \includegraphics[width=1\textwidth]{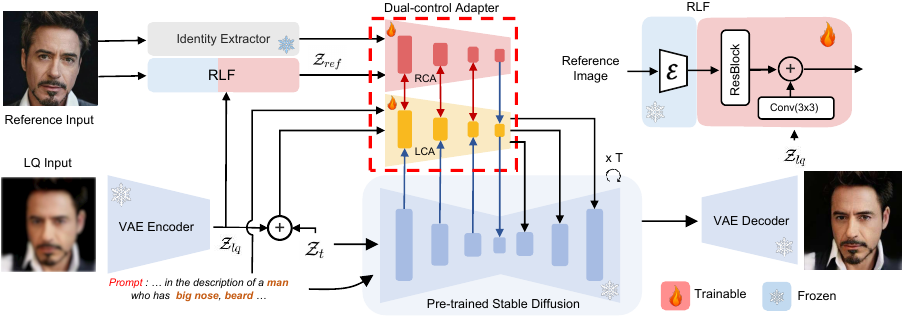}
  \vspace{-5mm}
  \caption{\textbf{Framework Overview.} This figure illustrates the overall workflow of the proposed MGFR model.}
  \label{fig4}
  \vspace{-4mm}
\end{figure*}

\subsection{Stage One: Text-guided Face Restoration}
\label{sec:Text-guided}
\vspace{-2mm}
This stage trains the face image restoration model that accepts text prompt as input, as shown in \cref{fig3}.
We use the pre-trained Stable Diffusion (SD) \citep{rombach2022high} model as our generative prior and train an additional adapter to extend it to the face restoration applications.
The pre-trained SD generative prior has the ability to understand face image attributes and text and generate high-quality face images.
In this stage, our model restores the image $x$ according to the condition $\{y, c_a\}$, where $y$ represents the degraded LQ image, and $c_a$ constitutes the facial attribute prompts describing the face attributes.
We first use the CLIP text encoder \citep{radford2021learning} to calculate the text embedding $e_r=\operatorname{CLIP}(c_a)$.
The LQ input $y$ is also mapped to a latent representation $z_{lq}$ using the VAE encoder in SD \citep{rombach2022high}.
We then perform diffusion generation on this latent representation.
In the framework of SD, the model uses UNet \citep{ronneberger2015u} denoising model ${\mathcal{E}}_{\theta}\left({z}_{t},t,e_r\right) $ to perform the diffusion generation process, where $t$ is the time stamp in diffusion model and $z_t$ is the intermediate results at time $t$.
Based on the ControlNet \citep{zhang2023adding} framework, we use an external adapter that takes the LQ input $y$ and text prompts embedding $e_r$ as input to provide guidance for the fixed UNet $\mathcal{E}_\theta$.
We call this adapter the LQ Control Adapter (LCA).
Specifically, the UNet model contains the encoder, intermediate blocks, and the decoder.
The decoder receives features from the encoder and fuses them at each corresponding scale.
The LCA contains the same encoder and intermediate blocks as the UNet model.
The feature output of each scale in LCA is integrated with the corresponding scale of the UNet decoder to achieve the effect of output control.
However, we found that simply using the above ControlNet framework has a key limitation -- the lack of information exchange from the UNet encoder to the LCA.
This gap means that the LCA is unaware of the processes that are performed in the UNet encoder, thereby limiting its ability to generate effective control features.
In order to solve this problem, we add the feature output of each scale in the UNet encoder to the corresponding scale in the LCA.
The LQ controller part of \cref{fig5} illustrates this operation.
In this way, the capability of the LCA is greatly enhanced, so better visual effects and control results can be achieved.

\subsection{Stage Two: Multi-modal-guided Face Restoration}
\label{sec:Multimodal}
\begin{figure}[t]
\vspace{-6mm}
\begin{minipage}[t]{0.485\textwidth}
\captionsetup{font={small}, skip=8pt}
\centering
    \includegraphics[width=\linewidth]{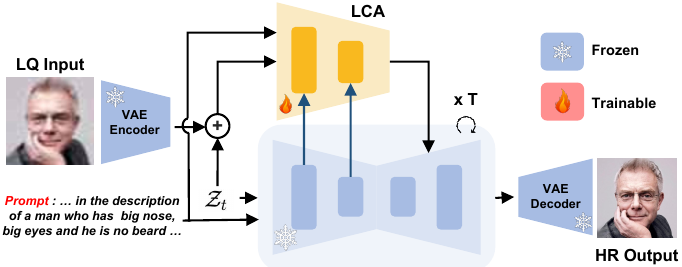}
    \vspace{-5mm}
    \caption{The model architecture employed during the initial training stage is discussed. In the article, 'Ours w/o Reference Image' refers to the outcome of the model trained following this stage.}
    \label{fig3}
\end{minipage}
\hfill
\begin{minipage}[t]{0.485\textwidth}
\captionsetup{font={small}, skip=8pt}
    \centering
    \includegraphics[width=\linewidth]{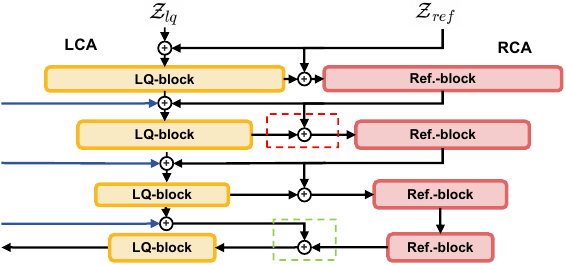}
    \vspace{-5mm}
    \caption{\textbf{Dual-control Adapter.} LQ-blocks are from the LQ control adapter (LCA), and Ref.-blocks are from the reference control adapter (RCA). $\oplus$ represents the element-wise add operation.}
    \label{fig5}
\end{minipage}
\vspace{-6mm}
\end{figure}

\vspace{-2mm}
After the first stage of training, the model can already reconstruct high-quality images from the LQ inputs guided by text prompts.
Next, we further enrich the guidance and introduce high-quality reference images and face identity information as additional control means based on the first-stage model.
We design a new Dual-Control Adapter (DCA), as shown in \cref{fig4}.
In DCA, we introduce a Reference Control Adapter (RCA) specifically for reference image processing.
RCA has the same architecture as LCA, and its role is to extract related and useful information from reference images and identity information and provide additional details to LCA.
The input of RCA includes an HQ reference face image $r$ containing the same identity as the LQ input and its identity information embedding $e_{f}$.
For the reference image $r$, we first use the VAE encoder consistent with the SD model for feature extraction to obtain $z_{ref}$.
Next, we fuse the LQ latent representation $z_{lq}$ with $z_{ref}$ using a reference and LQ feature fusion module (RLF).
This module allows RCA to identify the high-frequency details missing in the LQ input and perform targeted information extraction for restoration guidance.
For identity information embedding, we calculate $e_{f}=\operatorname{Proj}(\operatorname{Arcface}(r))$, where $\operatorname{Arcface}(r)$ is the face recognition model Arcface \citep{deng2019arcface} to extract the identity feature from the reference image $r$.
We align it to the space that RCA can handle through a trainable linear projection layer $\operatorname{Proj}(\cdot)$.
Due to the function of the RCA extracting information from the reference image according to the LQ input, the RCA requires the information of the LCA branch as input.
At the same time, RCA needs to provide the extracted information back to LCA in reverse.
Therefore, we designed a dual-way interaction mechanism for RCA and LCA, as shown in \cref{fig5}.
In this design, RCA runs in parallel with LCA.
At each scale, the LQ block in LCA first processes the fused information of both two branches and then hands the intermediate features to RCA.
RCA performs feature extraction and processing based on these intermediate results and reference conditions and finally uses the same operation to apply the processing results to the next layer of LCA processing.
Finally, the output of each scale of LCA is applied to the corresponding position of the UNet decoder.
RCA directly affects LCA and, therefore, also affects the calculation of UNet.
Since then, we have had a dual-control adapter that can accept multiple control inputs.

\subsection{Negative Samples and Prompt}
\label{sec:Negative}
\vspace{-2mm}
Classifier-Free Guidance (CFG) \citep{ho2022classifier} introduces a novel control mechanism utilizing negative prompts to delineate unwanted content for the model.
This feature can be leveraged to inhibit the generation of low-quality images by the model and to enhance the precision of facial detail reconstruction.
Throughout the inference phase, at each step of diffusion, three distinct predictions are generated: one employing the positive prompt $pos$, another using the negative quality prompt $nq$, and a third via the negative attribute prompt $na$ (the negation sentence described by $pos$).
We combine the results generated from these different prompts to form the final output:
\begin{equation}
\tilde{z}_{t-1} =z_{t-1}^{pos}+\lambda_{nq}\times(z_{t-1}^{pos}- z_{t-1}^{nq}) +\lambda_{na}\times(z_{t-1}^{pos}-z_{t-1}^{na}), 
\end{equation}
where $\lambda_{na}$ and $\lambda_{nd}$ is the hyperparameters, $z_{t-1}^{pos}=\mathcal{E}_{\theta}(z_{t},z_{lq},z_{ref},t,pos)$, $z_{t-1}^{nq}=\mathcal{E}_{\theta}(z_{t},z_{lq},z_{ref},t,nq)$, $z_{t-1}^{na}=\mathcal{E}_{\theta}(z_{t},z_{lq},z_{ref},t,na)$.
$pos$ represents a standard description of a facial attribute. $nq$ is the negative words of quality, e.g., ``\textit{oil painting, cartoon, blur, dirty, messy, low quality, deformation, low resolution, over-smooth}''.
$na$ is used for a negative description of a facial attribute, implying complete negation. 
Accurate prediction in both positive and negative directions is essential for the CFG technique.
The lack of negative-quality samples and prompts in our training might cause the model to misinterpret negative prompts, leading to artifacts.
To resolve this, we generated 16K images with negative-quality prompts using the original SD generative model and included these low-quality images in our training to enable the model to learn the concept of negative quality.
\cref{fig:data} (a) shows an example of the negative quality sample and prompt.

\begin{figure*}[t]
  \captionsetup{font={small}, skip=14pt}
  \includegraphics[width=1\textwidth]{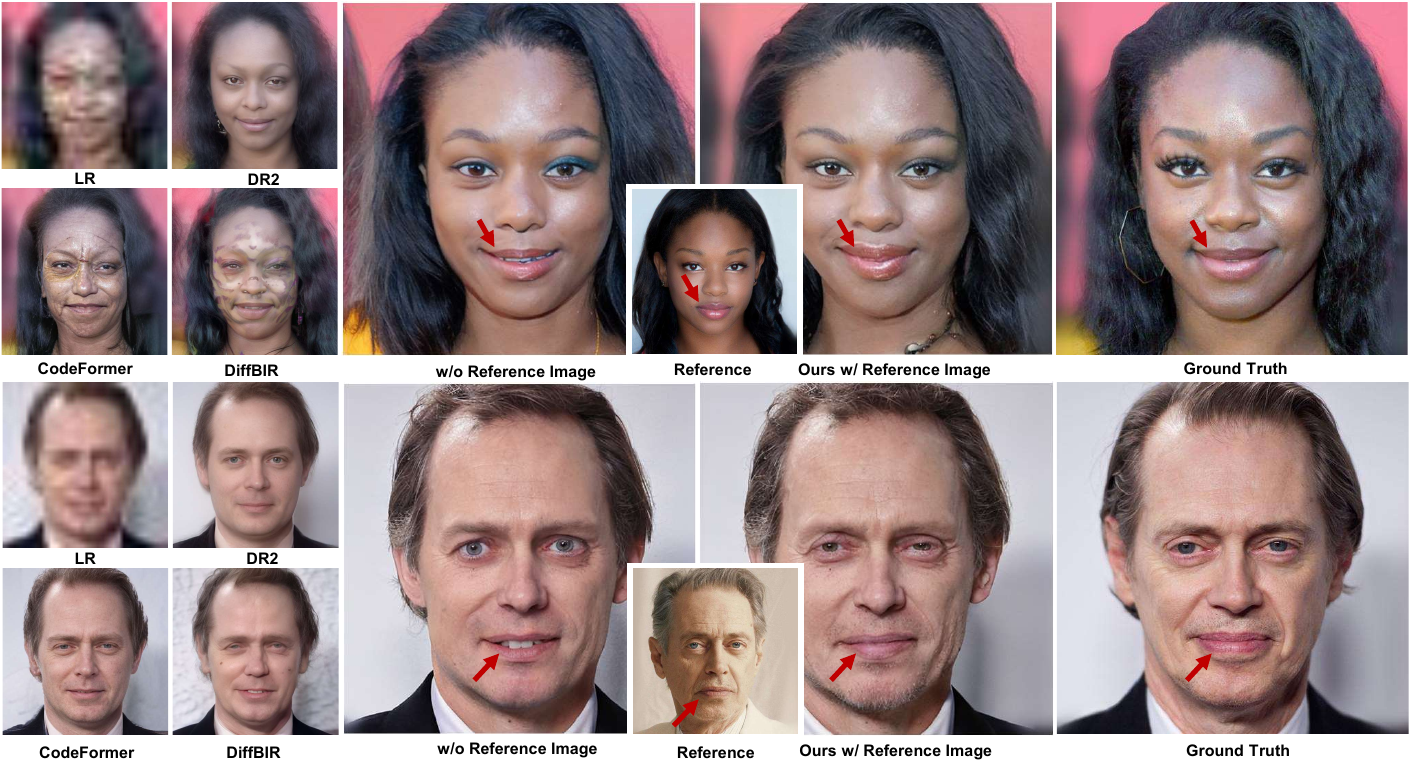}
  \vspace{-7mm}
  \caption{The MGFR model demonstrates a remarkable capacity for restoring LQ images. Upon integrating the reference image, particularly in instances of severe degradation, the model significantly enhances the restoration of facial details and overall image quality.}
  \label{fig:result}
  \vspace{-2mm}
\end{figure*}

\begin{figure*}

\begin{minipage}{0.60\textwidth}
\centering
\scriptsize
\begin{tabular}{ccc}
\hspace{-15.5mm} 
\begin{adjustbox}{valign=t}
\begin{tabular}{c}
\end{tabular}
\end{adjustbox}
\begin{adjustbox}{valign=t}
\begin{tabular}{cccccc}
\includegraphics[width=0.2\linewidth]{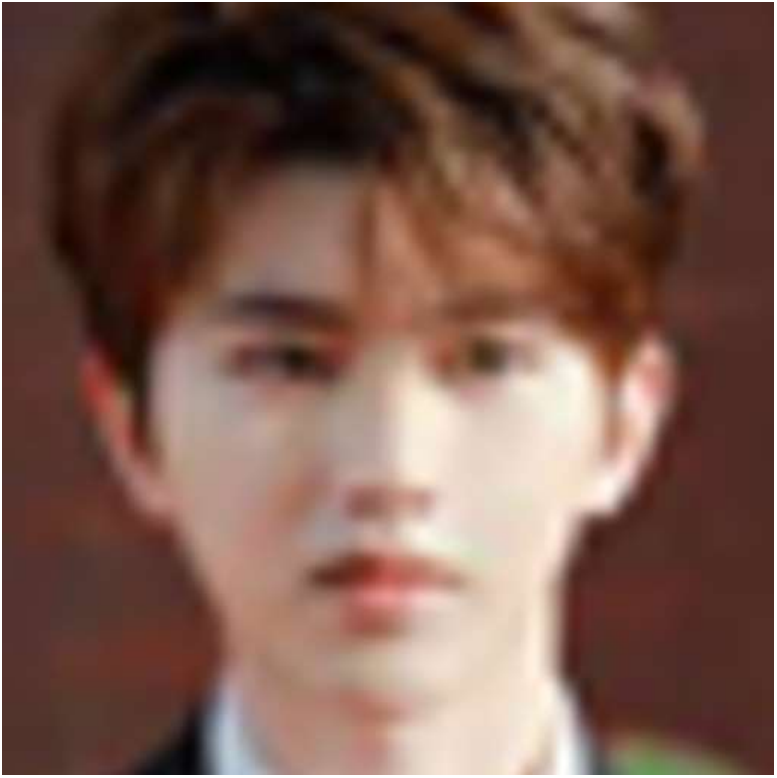} \hspace{-4mm} &
\includegraphics[width=0.2\linewidth]{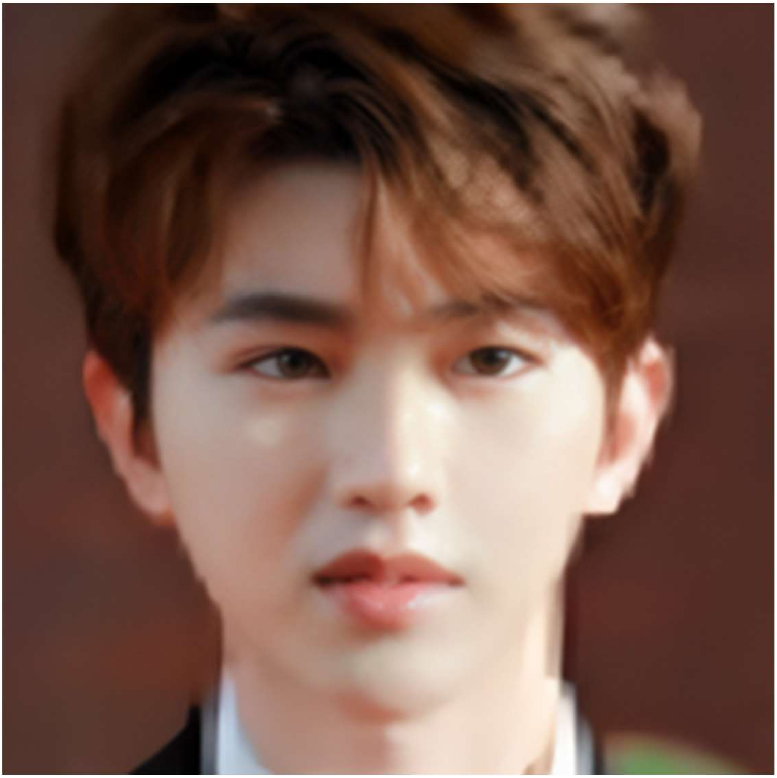} \hspace{-4mm} &
\includegraphics[width=0.2\linewidth]{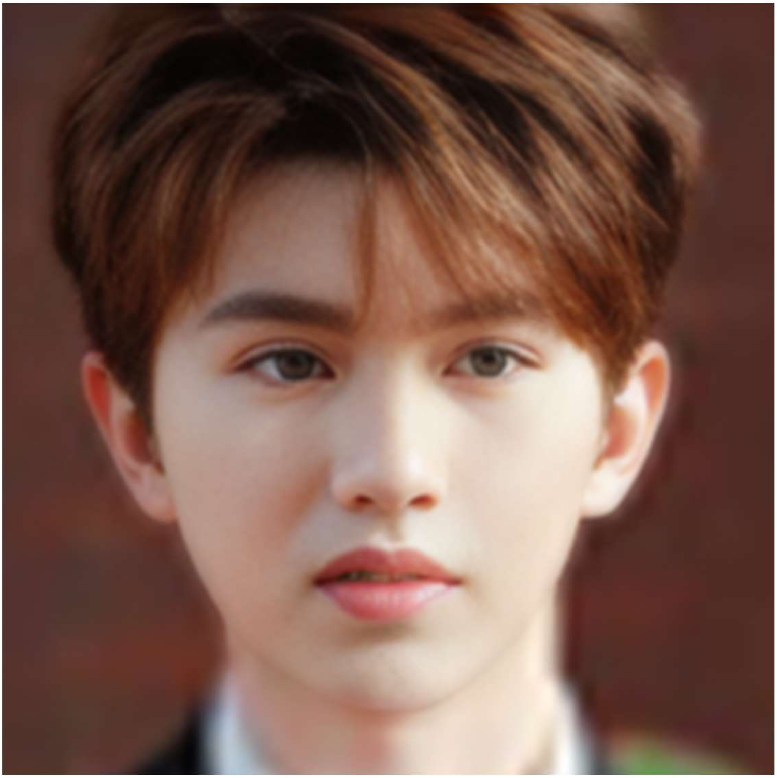} \hspace{-4mm} &
\includegraphics[width=0.2\linewidth]{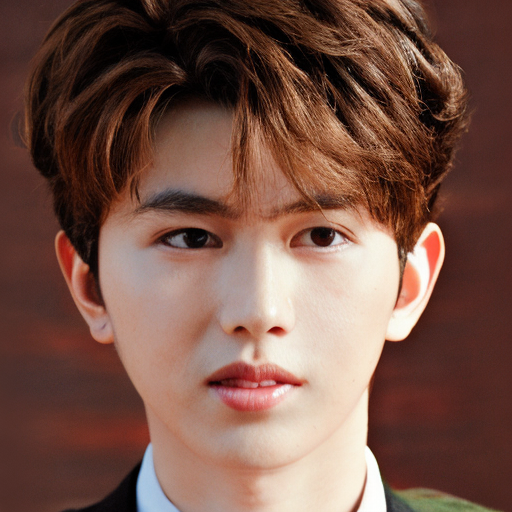} \hspace{-4mm} &
\includegraphics[width=0.2\linewidth]{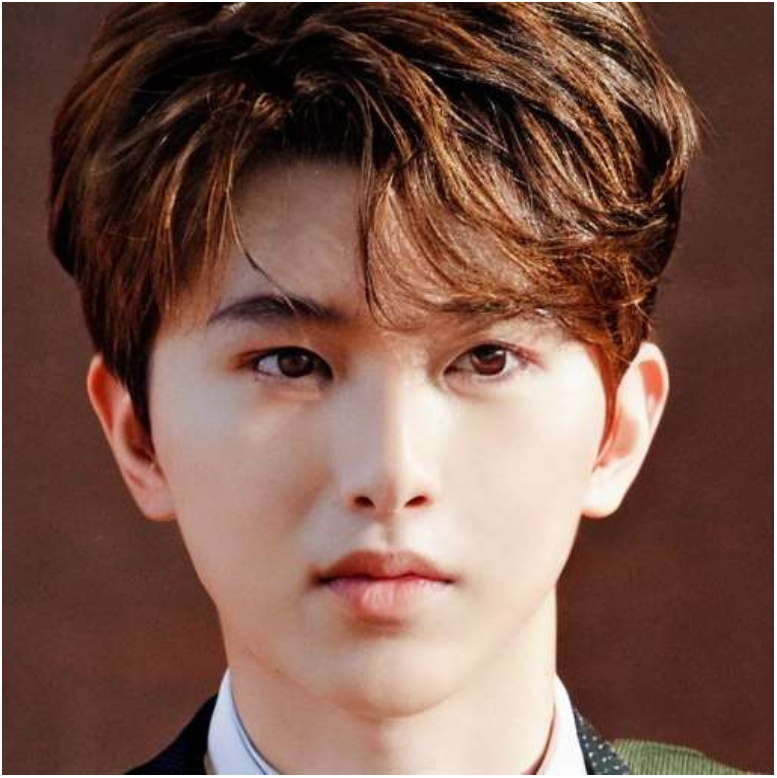} \hspace{-4mm} &
\includegraphics[width=0.2\linewidth]{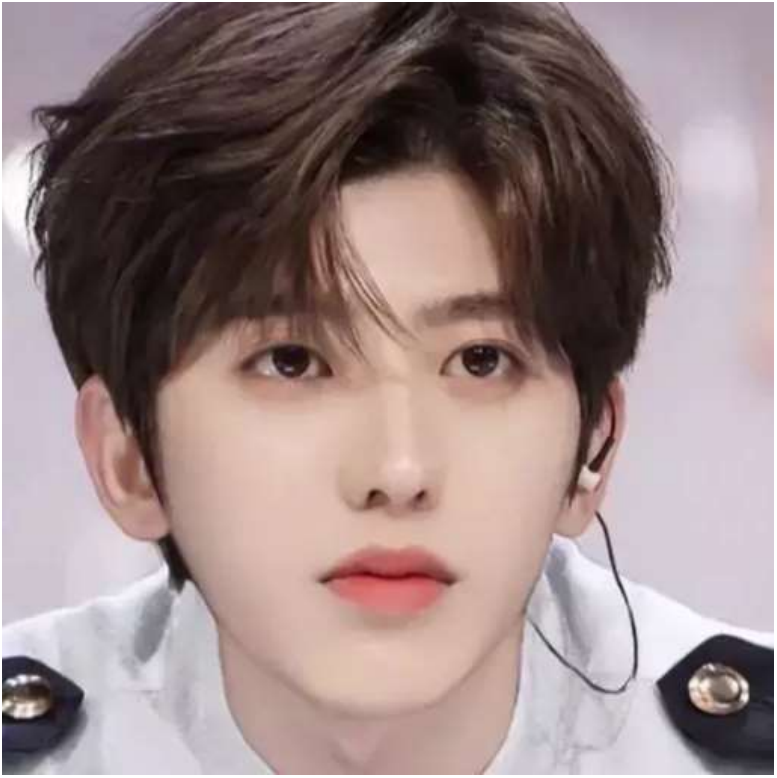} 

\end{tabular}
\end{adjustbox}
\\
\begin{adjustbox}{valign=t}
\begin{tabular}{c}
\end{tabular}
\end{adjustbox}

\hspace{-5mm} 
\begin{adjustbox}{valign=t}
\begin{tabular}{cccccc}
\includegraphics[width=0.2\linewidth]{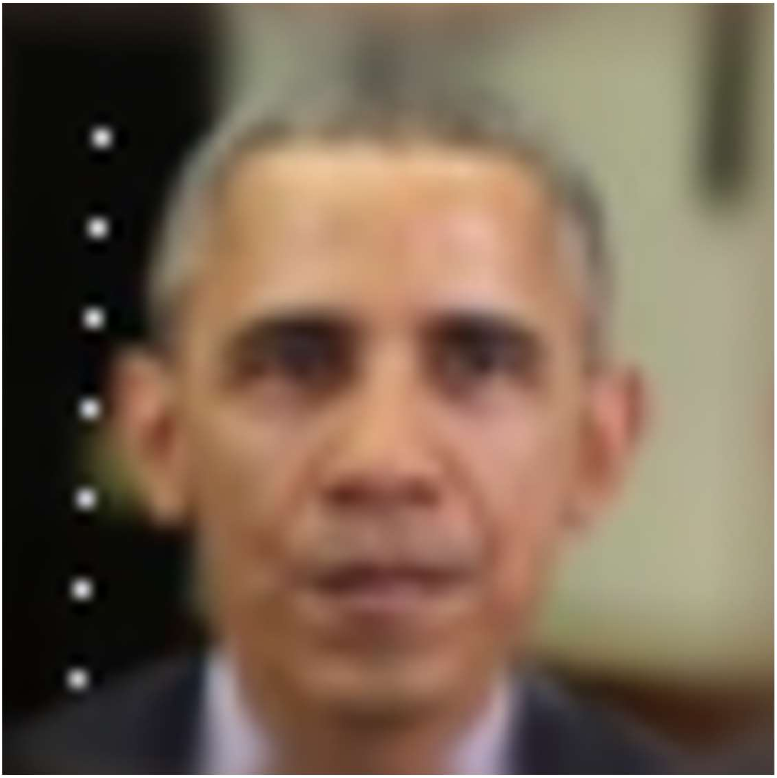} \hspace{-4mm} &
\includegraphics[width=0.2\linewidth]{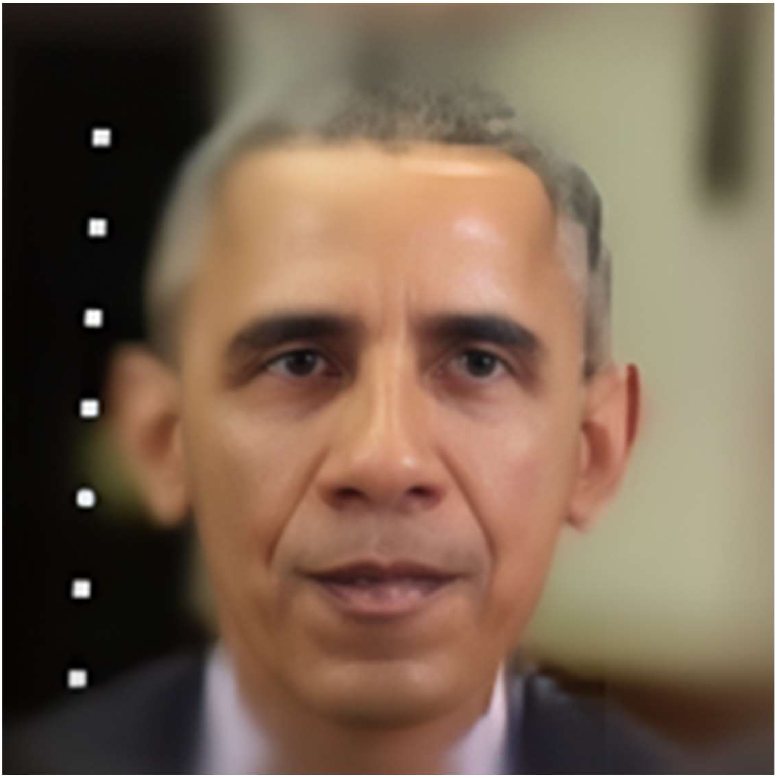} \hspace{-4mm} &
\includegraphics[width=0.2\linewidth]{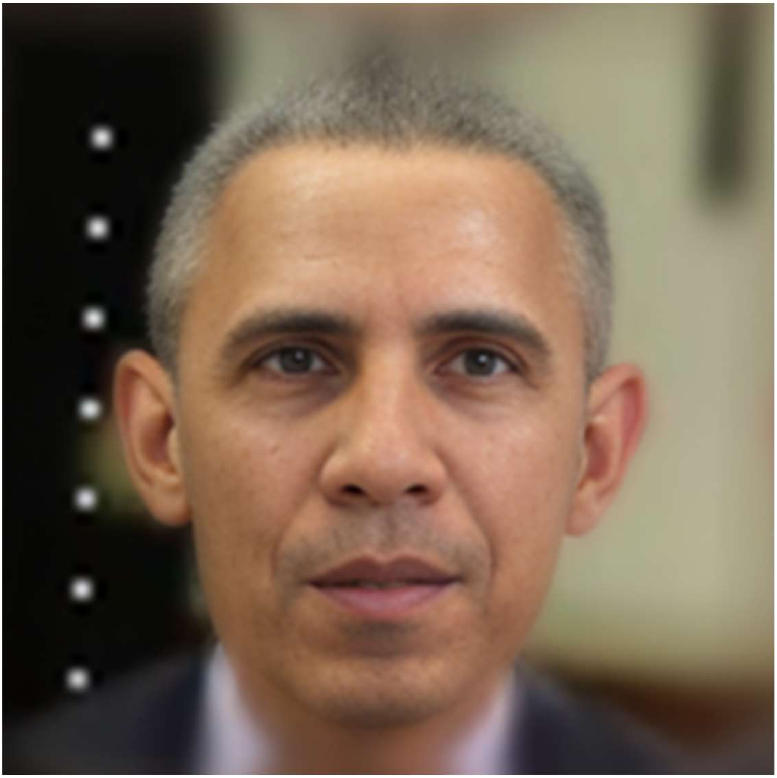} \hspace{-4mm} &
\includegraphics[width=0.2\linewidth]{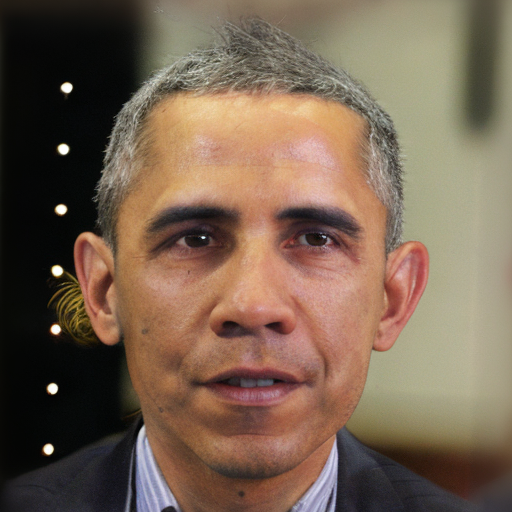}    \hspace{-4mm} &
\includegraphics[width=0.2\linewidth]
{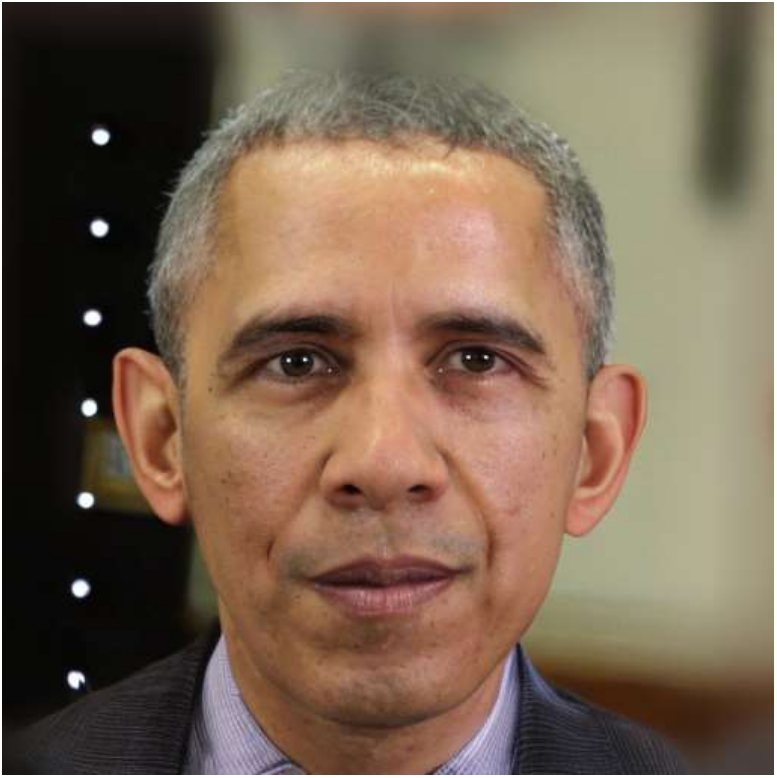} \hspace{-4mm} &
\includegraphics[width=0.2\linewidth]{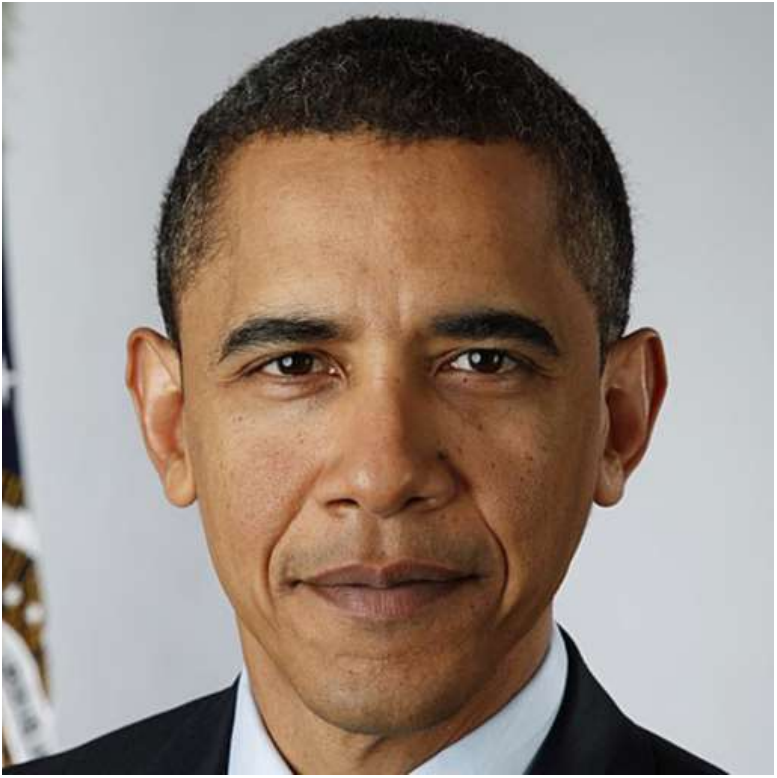} 

\\
Input \hspace{-4mm} &
DiffBIR\hspace{-4mm} &
CodeFormer \hspace{-4mm} &
w/o Reference \hspace{-4mm} &
Ours \hspace{-4mm} &
Reference 
\\
\end{tabular}
\end{adjustbox}
\hspace{10mm} 
\end{tabular}
\hspace{10mm} 
\end{minipage}%
\hspace{22mm}
\begin{minipage}{0.23\textwidth}
\vspace{-1mm}
\captionsetup{font={small}}
\caption{In the qualitative comparison of real-world low-quality (LQ) images, MGFR demonstrates success in recovering facial details without false illusion and preserving identity from. Please zoom in for a better view.}
\label{fig:real-world}
\end{minipage}
\vspace{-8mm}

\end{figure*}

\begin{figure*}[t]
\captionsetup{font={small}, skip=14pt}
\scriptsize
\centering
\begin{tabular}{ccc}
\hspace{-0.5cm}
\\
\hspace{-0.55cm}
\begin{adjustbox}{valign=t}
\begin{tabular}{c}
\end{tabular}
\end{adjustbox}
\begin{adjustbox}{valign=t}
\begin{tabular}{ccccccccccc}
\includegraphics[width=0.1065\linewidth]{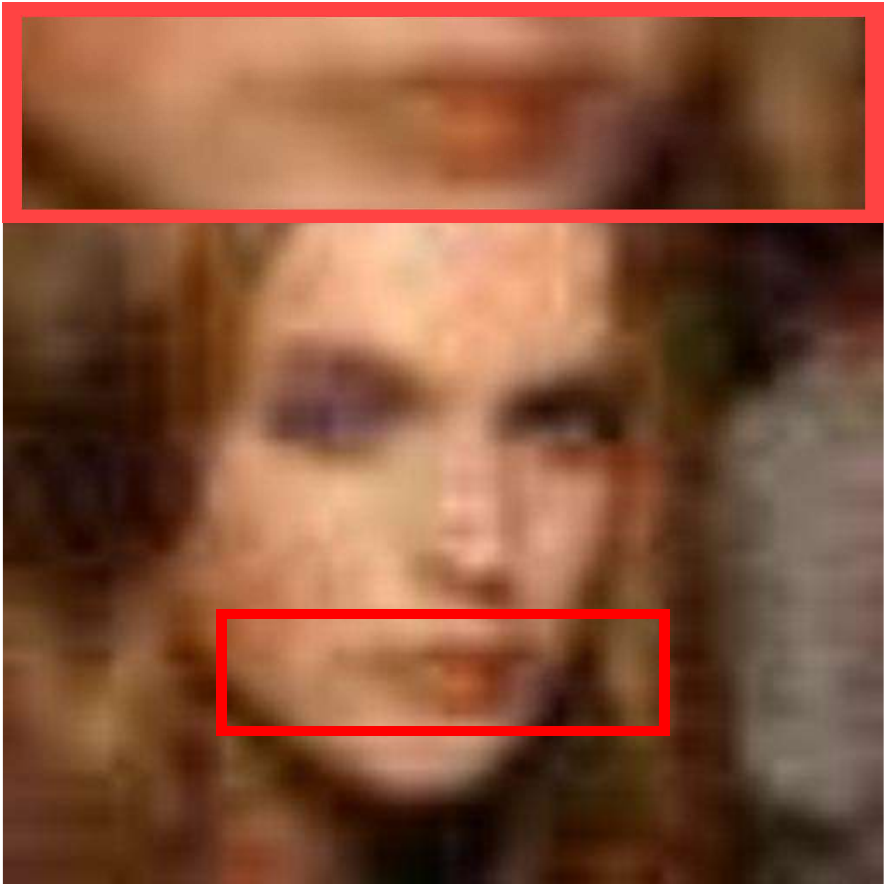} \hspace{-4mm} &
\includegraphics[width=0.1065\linewidth]{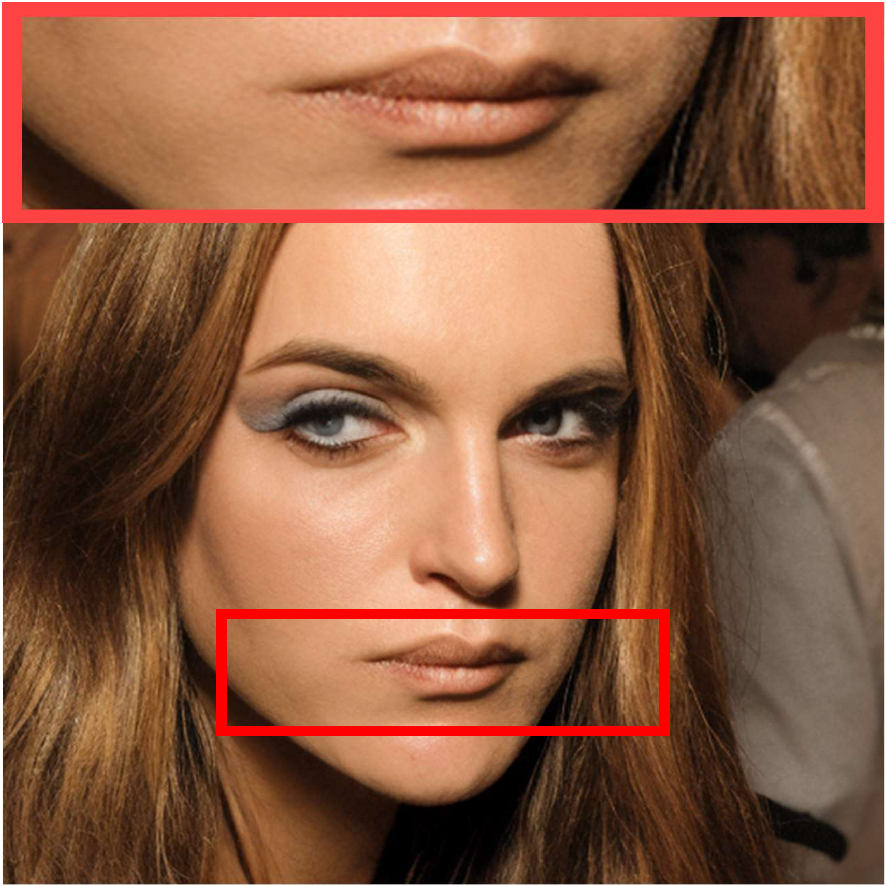} \hspace{-4mm} &
\includegraphics[width=0.1065\linewidth]{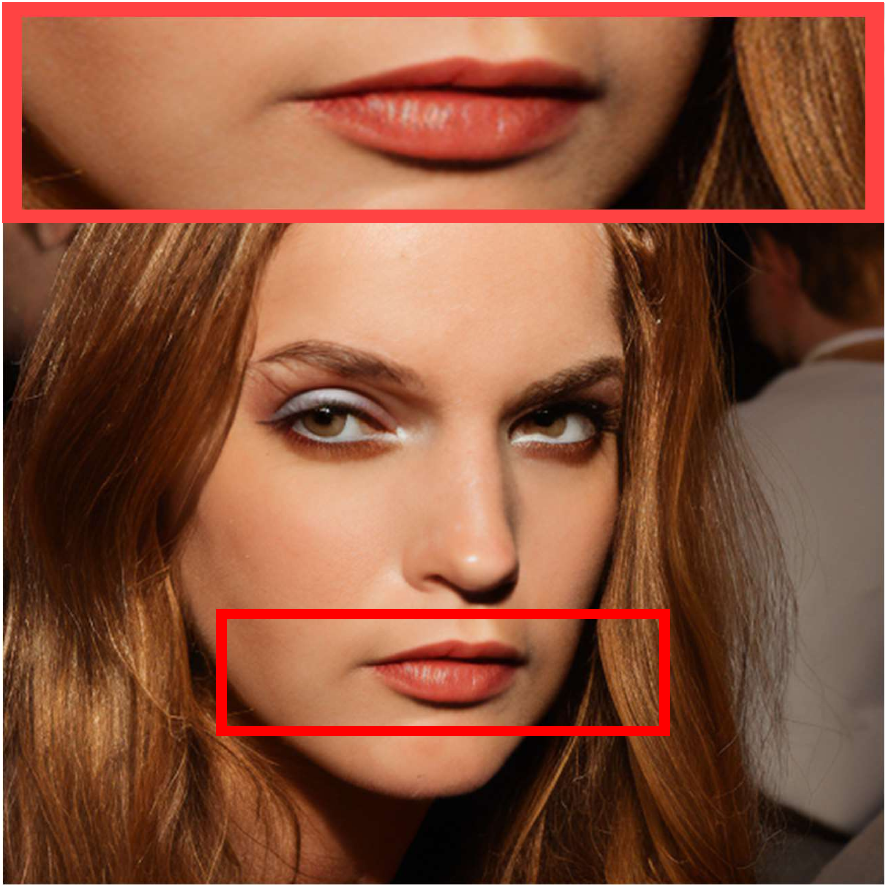} \hspace{-4mm} &
\includegraphics[width=0.1065\linewidth]{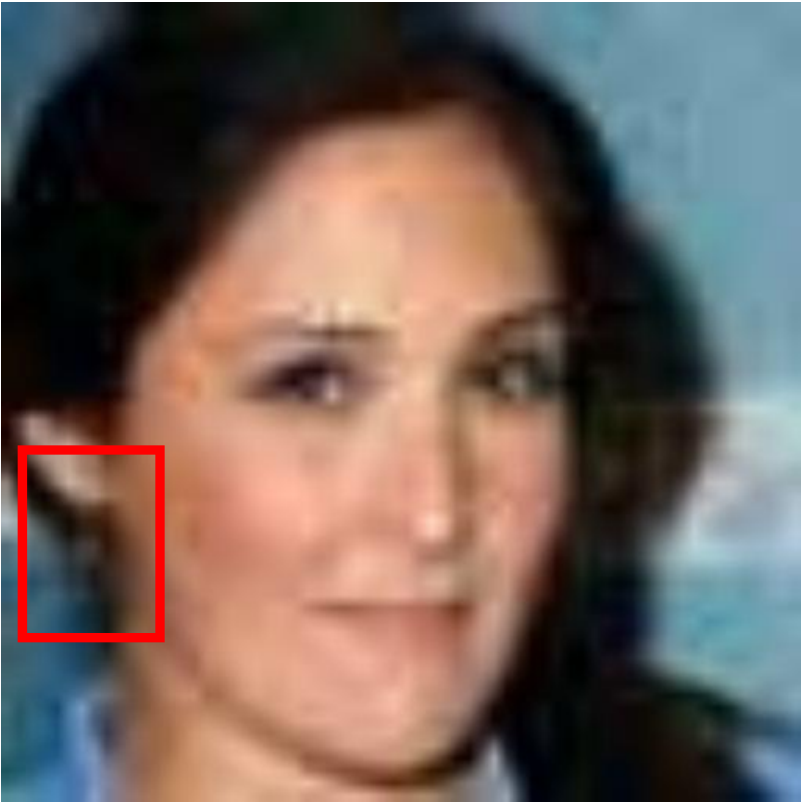}   \hspace{-4mm} &
\includegraphics[width=0.1065\linewidth]{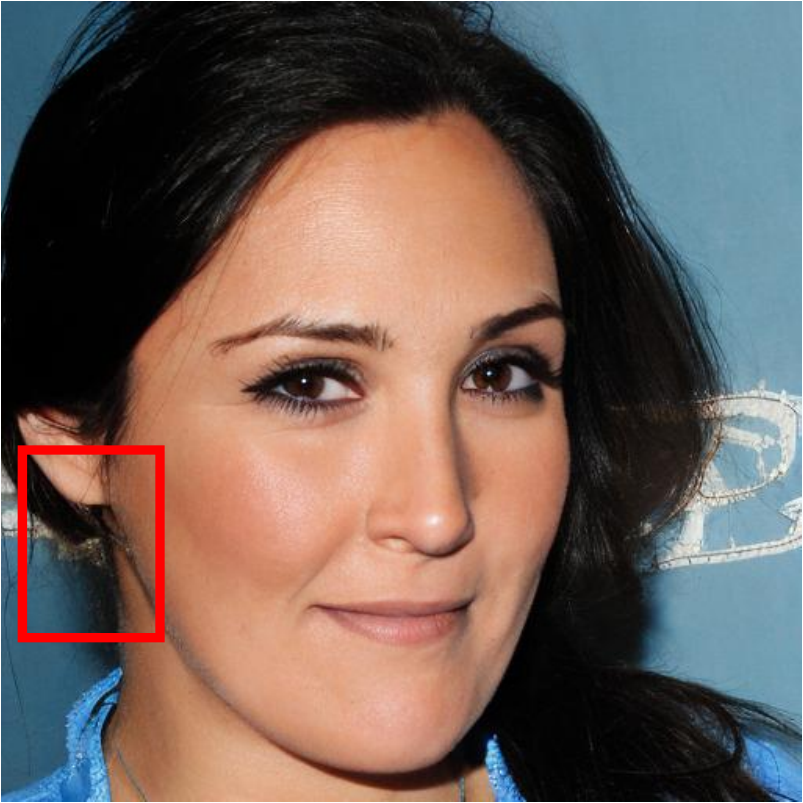} \hspace{-4mm} &
\includegraphics[width=0.1065\linewidth]{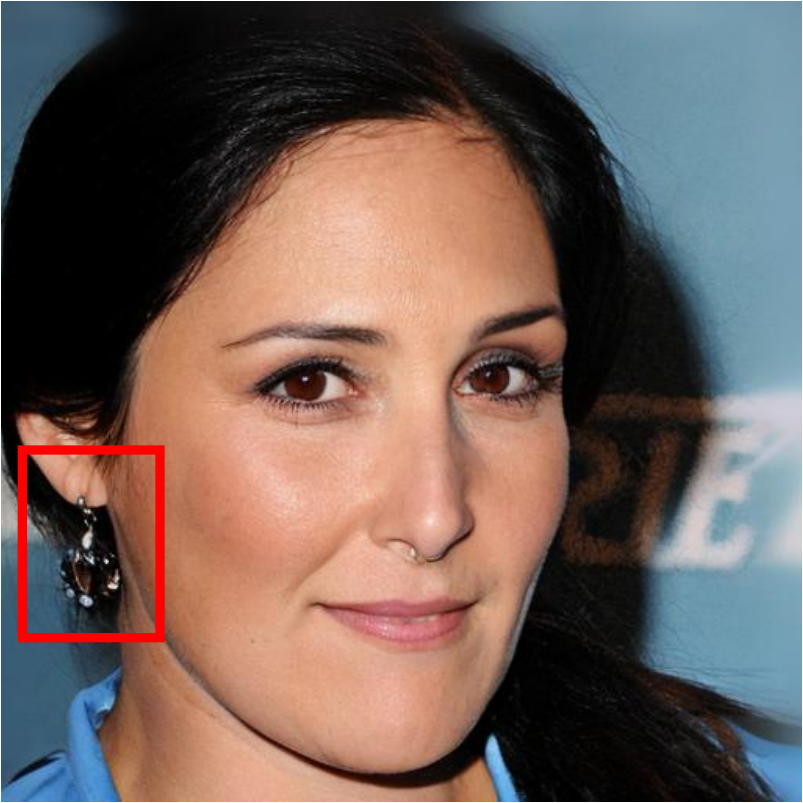}  \hspace{-4.1mm} &
\includegraphics[width=0.1065\linewidth]{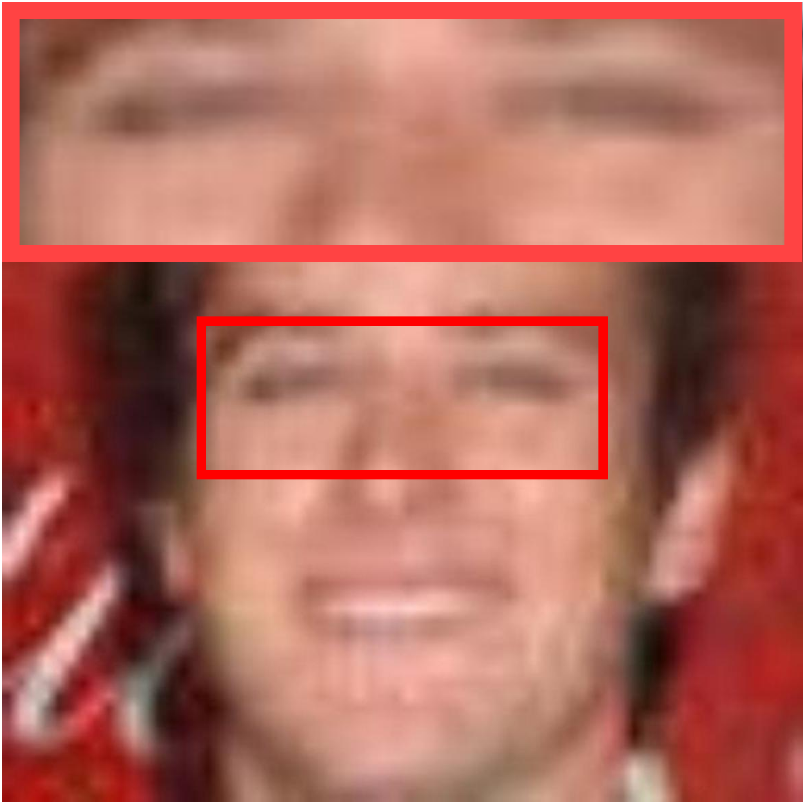}   \hspace{-4.1mm} &
\includegraphics[width=0.1065\linewidth]{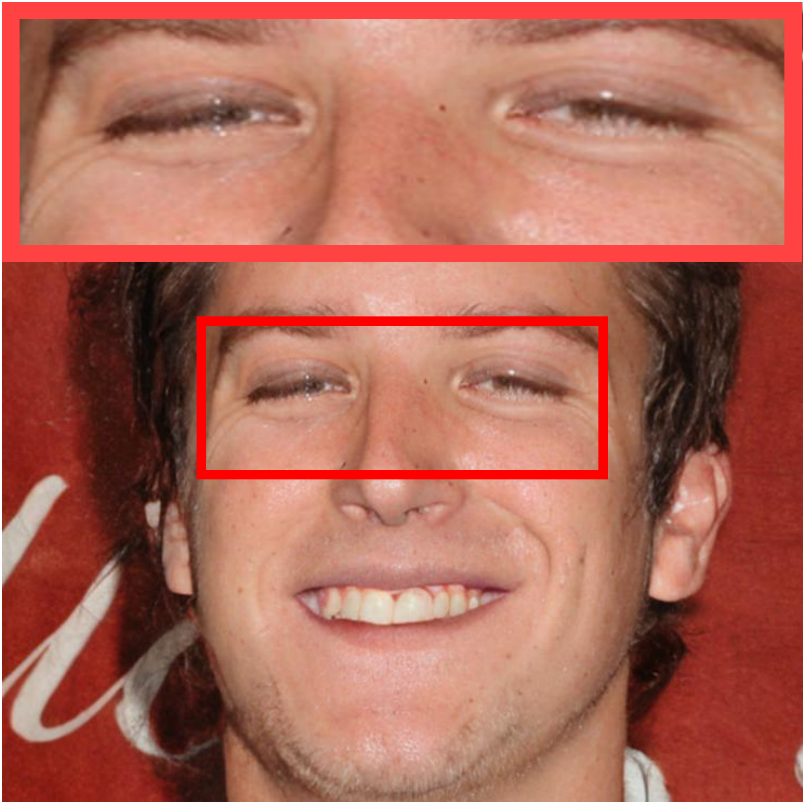}  
\hspace{-4mm} &
\includegraphics[width=0.1065\linewidth]{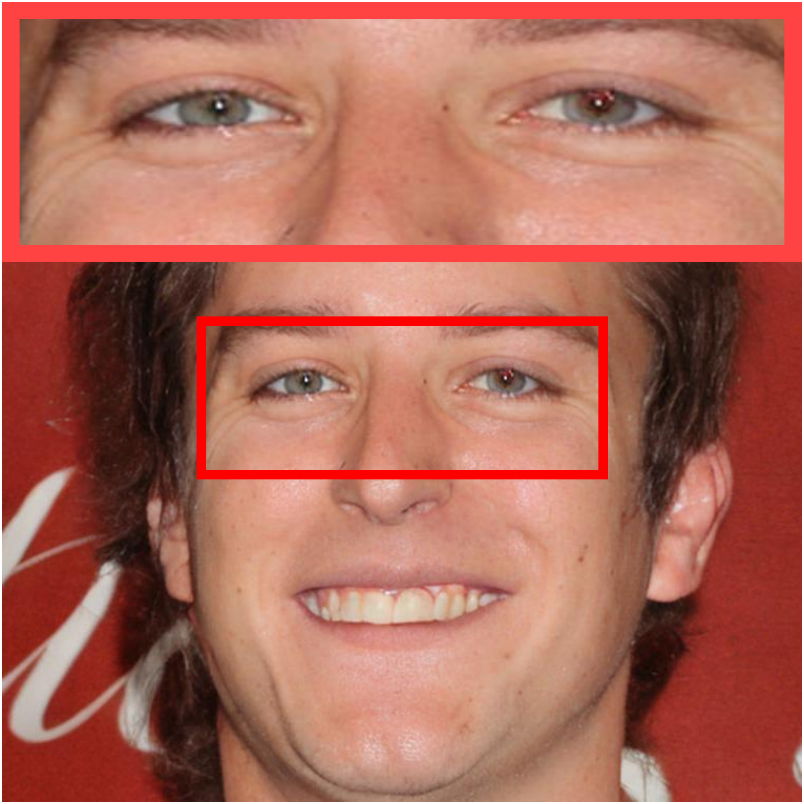}  
\\
Input \hspace{-4mm} &
No Prompt \hspace{-4mm} &
\makecell{Add: \emph{`lipstick'}} \hspace{-4mm} &
Input \hspace{-4mm} &
No Prompt \hspace{-4mm} &
\makecell{Add: \emph{`earrings'}} \hspace{-4mm} &
Input \hspace{-4mm} &
No Prompt \hspace{-4mm} &
\makecell{Add:\emph{`big eyes'}} 
\\
\end{tabular}
\end{adjustbox}
\end{tabular}
\vspace{-4mm}
\caption{MGFR demonstrates capability of face image restoration facilitated by text prompts. It possesses the capacity to artificially modulate specific aspects of the restoration outcomes, such as determining the presence of accessories like lipstick or glasses (Cases 1 \& 2), and orchestrating the restoration process in alignment with facial attributes (Case 3).}\label{fig:control}
\vspace{-2mm}
\end{figure*}

\vspace{-2mm}

\section{Experiments}
\vspace{-1mm}
\subsection{Experimental Setting}
\vspace{-2mm}
\textbf{Datasets.} 
Our two-stage training method requires different datasets for training.
For the first stage, we mainly train the model's ability to restore HQ images and process text prompts.
Therefore, we need HQ images with text annotations for training.
We synthesize training image pairs using the FFHQ dataset \citep{karras2019style}.
FFHQ contains 70,000 high-resolution face images, and we resize these images to $512\times512$ for training.
In the second stage, in addition to requiring HQ face images to create training image pairs, we also need to assign HQ reference images with consistent identities but different details to each image.
Although there are some datasets proposed for reference face restoration \citep{liu2015deep,yi2014learning}, the resolution and quality of these datasets cannot meet the current requirements.
In this work, we collect a new dataset for referenced face restoration called Reface-HQ.
Reface-HQ contains 21,500 high-quality and diverse images of over 4800 identities.
Additional details of Reface-HQ can be found in the \cref{sec:data}.
To synthesize LQ images, we follow the degradation model and setting used in \citep{dr2}.
Our test data also involves multiple sources, including CelebA-Test \citep{liu2015deep}, Reface-Test, and real-world LQ images collected from the Internet.
Specifically, CelebA-Test contains 3,000 testing images from the CelebA-HQ dataset.
Reface-Test contains 1,300 images of 280 identities split from the proposed Reface-HQ dataset.
The LQ images for testing are synthesized within the same degradation range as the training setting.

\begin{wrapfigure}{r}{0.5\textwidth}
\centering
\captionsetup{font={small}, skip=8pt}
\vspace{-4mm}
\includegraphics[width=0.5\textwidth]{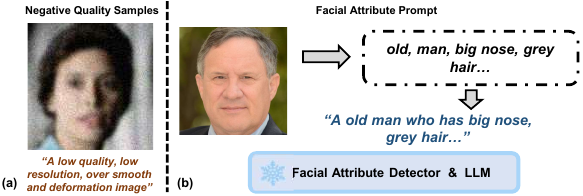}
\caption{\textbf{Training Data Composition.} Initially, negative quality samples are incorporated into the training to enhance the clarity and quality of the restored image. Furthermore, large language models, coupled with a facial attribute classifier, are employed to extract attribute texts for integration into the training.}
\vspace{-3.8mm}
\label{fig:data}
\end{wrapfigure}

\textbf{Attribute Prompt.}
\label{sec:att}
Text prompts are important for us to control face attributes and improve quality.
In our method, a total of three types of prompts are introduced. 
Two attribute prompts describe the face attributes, and the last one describes the negative quality of the image.
For attribute prompts, $pos$ contains positive descriptions of face attributes, while $na$ describes attributes that do not exist in this image to provide negative prompts of attributes.
To obtain these descriptions, we first use a pre-trained face attribute detector \citep{he2017adaptively} to extract the presence of each attribute in the face.
We considered 28 different attributes in this work.
For attributes that have high confidence to exist, we add them to the $pos$ positive attributes.
For the remaining attributes with low confidence, we classify them as $na$ negative attributes.
At this time, these attributes are still separate words.
We use a large language model to organize the separated words into natural language to facilitate the understanding of the CLIP text encoder.
Thus, each face image is associated with two attribute prompts detailing existing and non-existing attributes.
For the negative quality prompt, $nq$ involves ``low quality, low resolution, over-smoothed and distorted images'', as shown in \cref{fig:data} (a).
See \cref{sec:prompt} for comprehensive details on the training and inference procedures involving attribute prompts.

\textbf{Implementation.}
\label{sec:imp}
The training involved fine-tuning based on Stable Diffusion v2.1 \citep{rombach2022high}, with the control adapter structure adhering to \citep{zhang2023adding}. The Adam optimizer \citep{kingma2014adam} was employed, featuring a learning rate of $e^{-5}$. The initial training stage spanned 15 days, while the subsequent stage lasted 5 days, utilizing 4 Nvidia A100 GPUs with a batch size of 4. For testing purposes, the hyperparameters were set as $T=500$, $\lambda_{na}$ = 0.5 and $\lambda_{nq}$ = 0.5.

\textbf{Metrics.}
\label{met}
For quantitative comparison, followed by many previous works \citep{diffbir,yu2024scaling}, the selected metrics include full-reference metrics PSNR, SSIM, and LPIPS \citep{zhang2018unreasonable}, as well as non-reference metrics ManIQA \citep{yang2022maniqa}, ClipIQA \citep{wang2023exploring}, and MUSIQ \citep{ke2021musiq}. Furthermore, the Arcface identity distance \citep{deng2019arcface} (ID) is utilized to assess the similarity of identity information.

\begin{table*}
\captionsetup{font={small}, skip=14pt}
  \centering
\resizebox{\linewidth}{!}{
\begin{tabular}{c|cccc|cccc|cccc}
\toprule
 & \multicolumn{4}{c|}{Real-SR($\times$4)} & \multicolumn{4}{c|}{Real-SR($\times$8)} & \multicolumn{4}{c}{Real-SR($\times$16)} \\
\multirow{-2}{*}{Method} & LPIPS ↓& ManIQA & ClipIQA & MUSIQ & LPIPS ↓& ManIQA & ClipIQA & MUSIQ & LPIPS ↓& ManIQA & ClipIQA & MUSIQ \\ \midrule
PSFRGAN & 0.2938 & 0.5927 & 0.5702 & 73.39 & 0.3315 & 0.6015 & 0.5956 & 73.08 & 0.3788 & 0.5739 & 0.6274 & {\color[HTML]{3166FF} 71.76} \\
GPEN & 0.2828 & {\color[HTML]{3166FF} 0.6596} & 0.6430 & 69.25 & 0.3217 & {\color[HTML]{3166FF} 0.6754} & 0.6299 & 68.63 & 0.3831 & {\color[HTML]{3166FF} 0.6618} & 0.5897 & 66.61 \\
VQFR & 0.2951 & 0.2875 & 0.2490 & 62.95 & 0.3277 & 0.4163 & 0.2363 & 61.92 & {\color[HTML]{3166FF} 0.3761} & 0.6513 & 0.2148 & 60.49 \\
CodeFormer & 0.2927 & 0.5803 & 0.5179 & {\color[HTML]{3166FF} 75.47} & {\color[HTML]{3166FF} 0.3193} & 0.5970 & 0.6235 & {\color[HTML]{3166FF} 75.09} & {\color[HTML]{000000} 0.3821} & 0.5803 & 0.5877 & 70.85 \\
DR2 & 0.3264 & 0.5749 & 0.4441 & 63.43 & 0.3580 & 0.5246 & 0.4494 & 59.46 & 0.3796 & 0.5160 & 0.5035 & 70.31 \\
DiffBIR & {\color[HTML]{CB0000} 0.2611} & {\color[HTML]{000000} 0.6068} & {\color[HTML]{3166FF} 0.7681} & 74.27 & {\color[HTML]{CB0000} 0.3017} & 0.6058 & {\color[HTML]{3166FF} 0.7439} & 73.87 & 0.4238 & 0.5361 & {\color[HTML]{3166FF} 0.7164} & 67.41 \\
\add{BFRffusion} & 0.3258 & 0.5477 & 0.5572 & 45.32 & 0.3739 & 0.4404 & 0.5298 & 42.84 & 0.3735 & 0.4204 & 0.5098 & 43.16\\
Ours w/o Reference & {\color[HTML]{3166FF} 0.2925} & {\color[HTML]{CB0000} 0.6854} & {\color[HTML]{CB0000} 0.8244} & {\color[HTML]{CB0000} 76.22} & 0.3227 & {\color[HTML]{CB0000} 0.6776} & {\color[HTML]{CB0000} 0.8083} & {\color[HTML]{CB0000} 75.94} & {\color[HTML]{CB0000} 0.3760} & {\color[HTML]{CB0000} 0.6729} & {\color[HTML]{CB0000} 0.7944} & {\color[HTML]{CB0000} 75.76} \\ \bottomrule
\end{tabular}
}
\vspace{-2mm}
  \caption{\textbf{Quantitative Comparison in CelebA-Test.} Results in red and blue signify the highest and second highest, respectively. The $\downarrow$ indicates metrics whereby lower values constitute improved outcomes, with higher values preferred for all other metrics.}
  \label{tab1}
\vspace{-3mm}
\end{table*}

\subsection{Comparisons with State-of-the-art Methods}
\vspace{-2mm}
MGFR is qualitatively and quantitatively compared with state-of-the-art methods in FR. Notably, the model trained in the initial stage, which is a restoration model solely guided by attribute prompts, already achieves superior visual results. The non-reference prior-based methods selected include PSFRGAN \citep{chen2021progressive}, GPEN \citep{gpen}, VQFR \citep{gu2022vqfr}, CodeFormer \citep{coderformer}, DR2 \citep{dr2}, \add{BFRffusion} \citep{chen2024towards} and DiffBIR \citep{diffbir}, along with reference prior-based methods ASFFNet \citep{Li_2020_CVPR} and DMDNet \citep{9921338}. Particularly, to ensure contrastive fairness during the inference stage, the description text, containing restricted attributes, is obtained through low-resolution processing. In practical applications, however, users can freely set attribute prompts, enabling more precise and comprehensive guidance. For qualitative results comparing ASFFNet and DMDNet, please refer to the Appendix.

\begin{figure}[t]
\begin{minipage}[t]{0.49\textwidth}
\captionsetup{font={small}, skip=14pt}
\scriptsize
\centering
\begin{tabular}{ccc}
\hspace{-0.5cm}
\begin{adjustbox}{valign=t}
\begin{tabular}{c}
\end{tabular}
\end{adjustbox}
\begin{adjustbox}{valign=t}
\begin{tabular}{cccccc}
\includegraphics[width=0.24\linewidth]{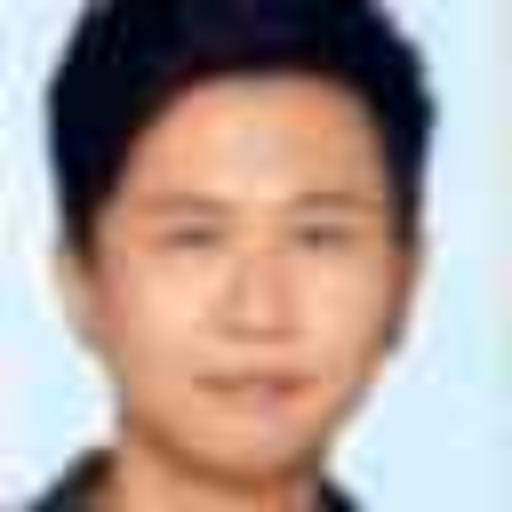} \hspace{-4mm} &
\includegraphics[width=0.24\linewidth]{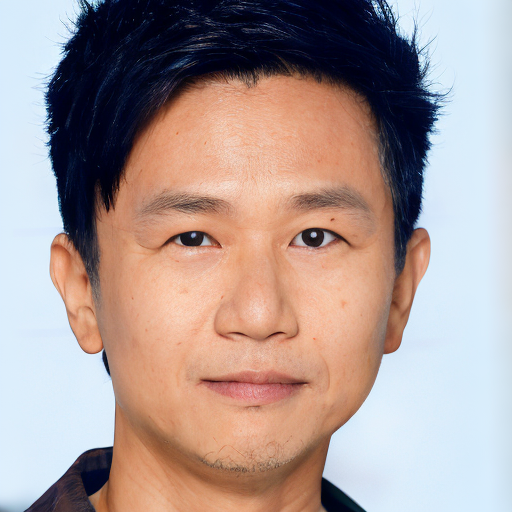} \hspace{-4mm} &
\includegraphics[width=0.24\linewidth]{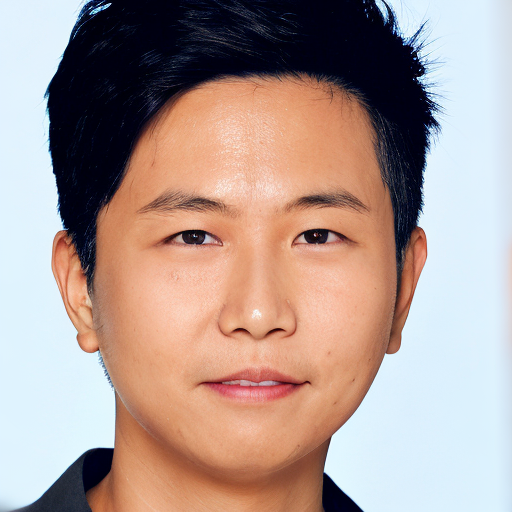} \hspace{-4mm} &
\includegraphics[width=0.24\linewidth]{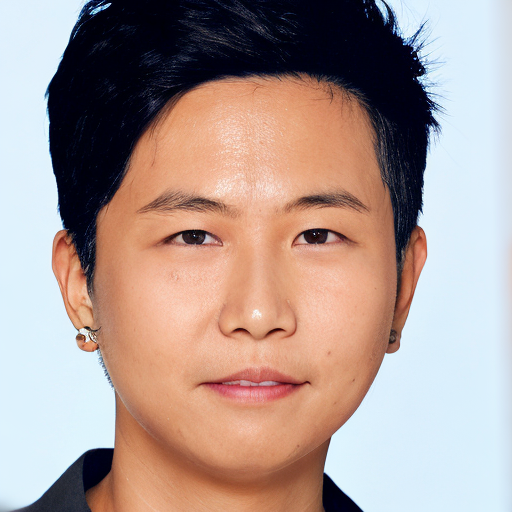} 
\\
Input \hspace{-4mm} &
No Prompt \hspace{-4mm} &
\emph{`No beard, young'} \hspace{-4mm} &
\makecell{\emph{`young, woman}\\ \emph{blonde hair'}} 

\\
\end{tabular}
\end{adjustbox}
\end{tabular}
\vspace{-5.mm}
\caption{Attribute prompts that manifestly contravene low-resolution inputs prove ineffectual and result in distortions and artifacts within the restored image.}
\label{fig:non-control}
\end{minipage}
\hfill
\begin{minipage}[t]{0.49\textwidth}
\captionsetup{font={small}, skip=16pt}
\scriptsize
\centering
\vspace{-12.mm}
\begin{tabular}{ccc}
\hspace{-0.5cm}
\begin{adjustbox}{valign=t}
\begin{tabular}{c}
\end{tabular}
\end{adjustbox}
\begin{adjustbox}{valign=t}
\begin{tabular}{cccccc}
\includegraphics[width=0.24\linewidth]{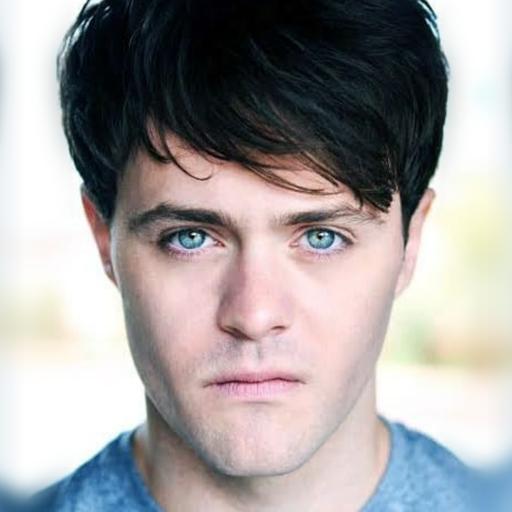} \hspace{-4mm} &
\includegraphics[width=0.24\linewidth]{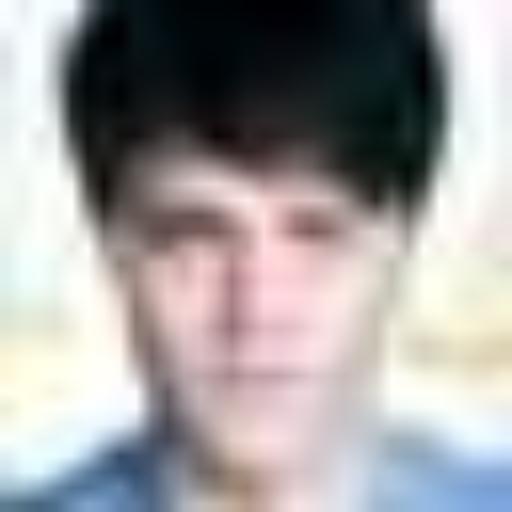} \hspace{-4mm} &
\includegraphics[width=0.24\linewidth]{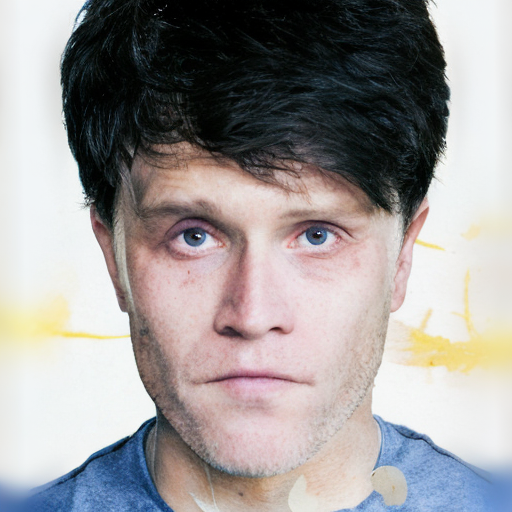} \hspace{-4mm} &
\includegraphics[width=0.24\linewidth]{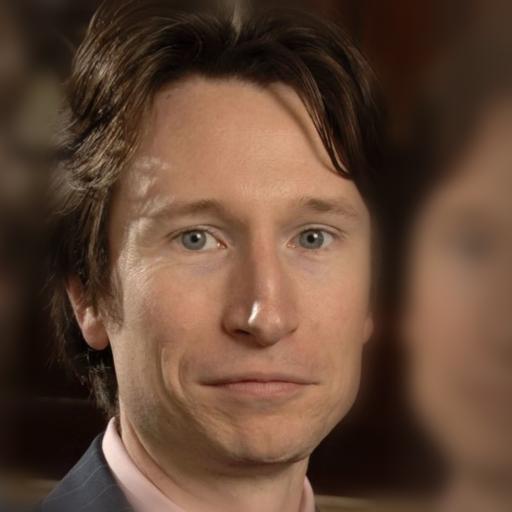} 
\\
Input \hspace{-4mm} &
High Blur \hspace{-4mm} &
Face Swap\hspace{-4mm} &
Reference 
\\
\end{tabular}
\end{adjustbox}
\end{tabular}
\vspace{-3mm}
\caption{Face Swapping: MGFR is capable of leveraging the reference map to alter the comprehensive components of the face.}
\label{fig:SWA}
\end{minipage}
\vspace{-2.5mm}
\end{figure}

\begin{figure*}[t]
\begin{minipage}{0.72\textwidth}
\captionsetup{font={small}, skip=2pt}
\scriptsize
\centering
\begin{tabular}{ccc}
\hspace{-0.5cm}
\begin{adjustbox}{valign=t}
\begin{tabular}{c}
\end{tabular}
\end{adjustbox}
\begin{adjustbox}{valign=t}
\begin{tabular}{cccccc}
\includegraphics[width=0.49\linewidth]{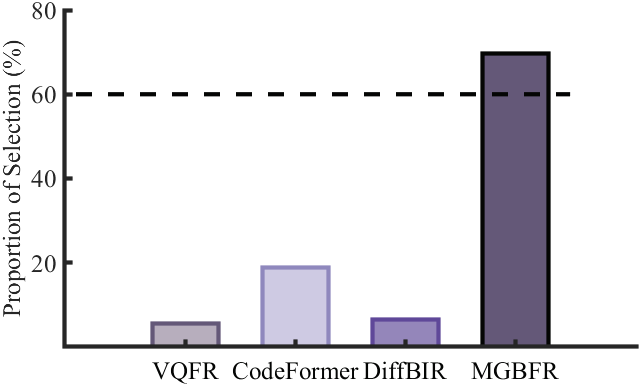} \hspace{-4mm} &
\includegraphics[width=0.49\linewidth]{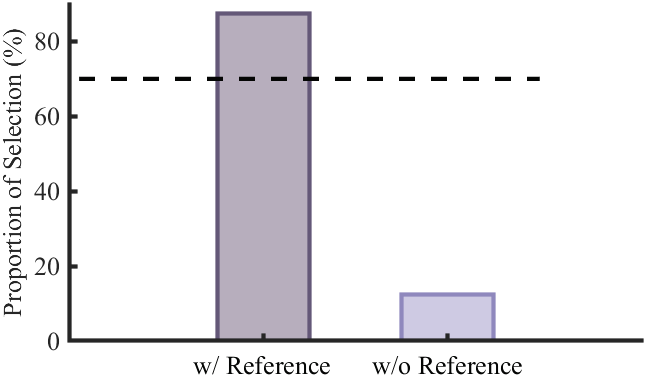} 
\\
(a) User Study on compared methods\hspace{-4mm} &
(b) User study on using Reference image\hspace{-4mm} &

\\
\end{tabular}
\end{adjustbox}
\end{tabular}
\label{fig:user}
\vspace{-2.mm}
\end{minipage}%
\hspace{0mm}
\begin{minipage}{0.26\textwidth}
\vspace{1mm}
\captionsetup{font={small}}
\caption{The results of our user study. We randomly select face images under multiple test datasets for user study. Our model achieves excellent recovery quality, which can be further enhanced with high-quality reference image and identity information guidance.}
\label{fig:real-world}
\end{minipage}
\vspace{-7mm}
\end{figure*}

\textbf{Comparison on Synthetic Degradations.}
Firstly, a quantitative comparison of our model without reference images on the synthetically degraded CelebA-Test dataset is conducted without reference image guidance. 
According to \cref{tab1}, our model achieves the best results on all non-reference metrics, indicative of the superior image quality of the results.
Due to space limitation, the values of SSIM and PSNR of \cref{tab1} are shown in \cref{more:tab1}.
Additionally, the method's limitations on full-reference metrics are also noted.
This phenomenon, preliminarily demonstrated by experiments in \citep{yu2024scaling,jinjin2020pipal}, necessitates a reevaluation of the reference value of indicators like PSNR, SSIM, LPIPS, and the proposal of more effective methods to assess advanced FR methods, particularly as quality improves.
More qualitative comparison results of our model can be found in \cref{sec:c_base}.
Subsequently, \cref{fig:result} and \cref{figt} (a) present a qualitative comparison of the MGFR method applied to the Reface-Test dataset.
Even in cases of severe degradation, our method successfully produces highly superior facial details guided by the reference image.
In addition, we provide a comparison between our model without reference images and MGFR, with a particular focus on FR tasks involving features like double eyelids, pupil color, and finer facial details, such as wrinkles and moles, which cannot be accurately captured without reference image guidance.
This further demonstrates the superiority of utilizing reference image guidance in the FR task.
Finally, \cref{tab2} offers quantitative comparison results, indicating that our method significantly surpasses other state-of-the-art methods in perceived quality. 

We also conducted a user study with a total of 40 participants, comparing MGFR to other approaches. Participants were asked to select the best quality recovery result from these test techniques for each pair of comparison images, or if no reference image was provided, the result that came closest to the Ground Truth. \cref{fig:user} presents the results, which demonstrate that our method outperforms the state-of-the-art methods in terms of recovery quality. Furthermore, the reconstruction effect can be further enhanced by using the reference image guidance.

\begin{wraptable}{r}{0.65\textwidth}
\centering
\captionsetup{font={small}, skip=8pt}
\vspace{-4mm}
\resizebox{\linewidth}{!}{
\begin{tabular}{@{}ccccccccc@{}}
\toprule
Degradation & Method & PSNR & SSIM & LPIPS ↓ & ManIQA & ClipIQA & MUSIQ & ID ↓ \\ \midrule
 & ASFFNet & 23.43 & 0.6811 & {\color[HTML]{CB0000} 0.2452} & {\color[HTML]{3166FF} 0.5685} & 0.6215 & 71.66 & 0.7053 \\
 & DMDNet & {\color[HTML]{000000} 23.85} & {\color[HTML]{CB0000} 0.7062} & {\color[HTML]{000000} 0.2667} & 0.5023 & 0.6023 & 72.31 & 0.6964 \\
 & DR2 & 23.58 & 0.6581 & {\color[HTML]{3166FF} 0.2532} & 0.5340 & 0.5956 & 69.00 & 0.7957 \\
 & CodeFormer & {\color[HTML]{3166FF} 23.88} & {\color[HTML]{3166FF} 0.6904} & {\color[HTML]{000000} 0.2912} & 0.4959 & 0.5823 & {\color[HTML]{3166FF} 74.80} & 0.6579 \\
 & DiffBIR & {\color[HTML]{CB0000} 24.12} & 0.6717 & {\color[HTML]{000000} 0.2785} & 0.5547 & {\color[HTML]{3166FF} 0.7474} & 73.73 & {\color[HTML]{3166FF} 0.6379} \\
\multirow{-6}{*}{$\times$8} & MGBFR(Ours) & 23.10 & 0.6248 & {\color[HTML]{000000} 0.2688} & {\color[HTML]{CB0000} 0.6535} & {\color[HTML]{CB0000} 0.8147} & {\color[HTML]{CB0000} 75.51} & {\color[HTML]{CB0000} 0.5166} \\ \midrule
 & ASFFNet & 21.70 & 0.6472 & {\color[HTML]{3166FF} 0.3013} & {\color[HTML]{3166FF} 0.5803} & 0.6221 & {\color[HTML]{3166FF} 71.57} & 0.9361 \\
 & DMDNet & {\color[HTML]{CB0000} 22.37} & {\color[HTML]{CB0000} 0.6761} & 0.3179 & 0.4579 & 0.4727 & 67.27 & 0.9270 \\
 & DR2 & {\color[HTML]{3166FF} 22.28} & {\color[HTML]{3166FF} 0.6720} & 0.3269 & 0.5233 & 0.5693 & 66.39 & 0.8676 \\
 & CodeFormer & 21.88 & 0.6124 & 0.3400 & 0.5547 & 0.5855 & 71.30 & {\color[HTML]{3166FF} 0.8658} \\
 & DiffBIR & 21.51 & 0.5939 & 0.3944 & 0.4937 & {\color[HTML]{3166FF} 0.7144} & 67.42 & 0.8876 \\
\multirow{-6}{*}{$\times$16} & MGBFR(Ours) & {\color[HTML]{000000} 21.75} & {\color[HTML]{000000} 0.6033} & {\color[HTML]{CB0000} 0.2989} & {\color[HTML]{CB0000} 0.6524} & {\color[HTML]{CB0000} 0.8046} & {\color[HTML]{CB0000} 75.06} & {\color[HTML]{CB0000} 0.7401} \\ \bottomrule
\end{tabular}

}
\caption{\textbf{Quantitative Comparison in Reface-Test.} Quantitative comparison of guided recovery results based on reference images. DR2, CodeFormer, and DiffBIR do not use reference images.}
\label{tab2}
\vspace{-5mm}
\end{wraptable}

\textbf{Comparison on Real-world Degradations.}
Additionally, our method was tested on real-world LQ images, which involved collecting degraded face images of publicly available images alongside reference images. The qualitative results, presented in \cref{fig:real-world}, demonstrate that the resulting images possess realistic visual effects with minimal facial illusions. More results are presented in \cref{sec:c_base}.

\subsection{Controlling Restoration with Attributes Prompts}

Our method facilitates targeted image restoration guided by attribute prompts.
As illustrated in \cref{fig:control}, the comparison between the first and second cases reveals that the integration of supplementary attribute prompts facilitates the manipulation of subtle facial attributes absent in the original image.
This includes the addition of glasses, earrings, and accessories.
In scenarios of severe degradation, exemplified by the third case, reconstructing facial features like eyes poses a significant challenge without external prompts. More results are shown in \cref{sec:more}. 

However, it is imperative to acknowledge that attribute prompts do not invariably yield efficacy.
As demonstrated in \cref{fig:non-control}, our model is capable of control tuning through attribute prompts.
However, prompts that starkly contradict LQ inputs, like ``blonde hair'', are found to be ineffective.
This ensures the model's adherence to the provided LQ inputs. Furthermore, as illustrated by the input of an LQ male face in \cref{fig:non-control}, when the input attribute is ``Female'', the model subtly incorporates the attribute label ``Female'' into the image. This is achieved through modifications like the addition of an earring and the removal of the beard while remaining faithful to the LQ input. Such modifications further underscore the efficacy of attribute text in guiding the restoration process. 
This outcome is not unexpected. On the contrary, excessive control capability might lead to a reduction in the restoration effectiveness, countering the fundamental intent of image reconstruction efforts and thereby demonstrating the robustness of the proposed method.

\begin{figure*}

\hspace{-5.5mm} 
\begin{minipage}{0.55\textwidth}
\centering
\scriptsize
\begin{tabular}{ccc}
\begin{adjustbox}{valign=t}
\begin{tabular}{c}
\end{tabular}
\end{adjustbox}
\begin{adjustbox}{valign=t}
\begin{tabular}{ccccc}
\includegraphics[width=0.24\linewidth]{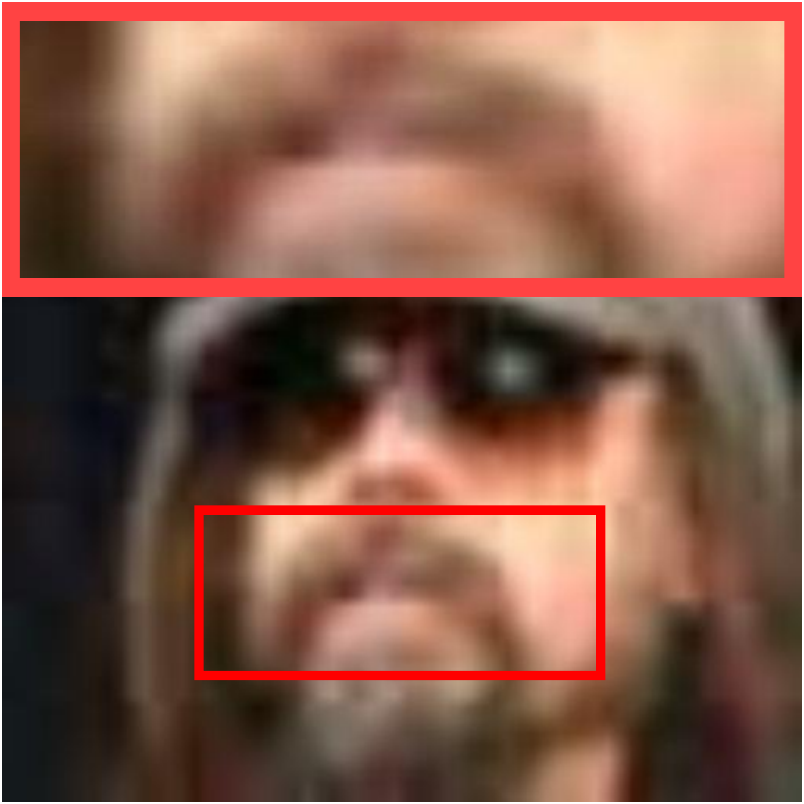} \hspace{-4mm} &
\includegraphics[width=0.24\linewidth]{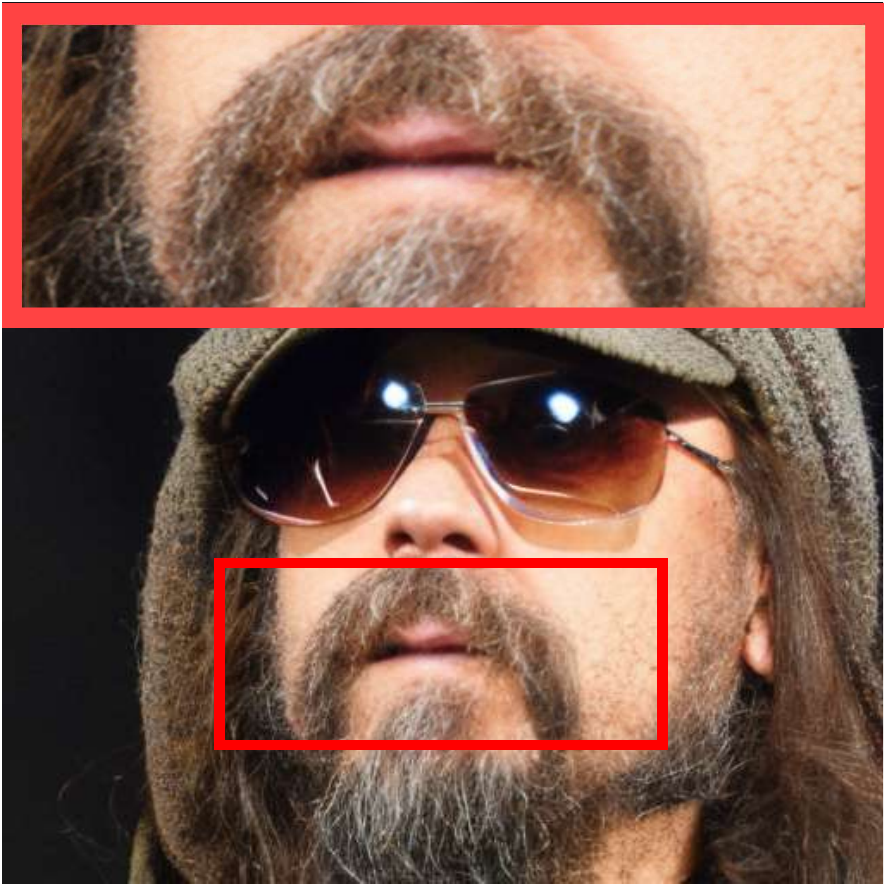} \hspace{-4mm} &
\includegraphics[width=0.24\linewidth]{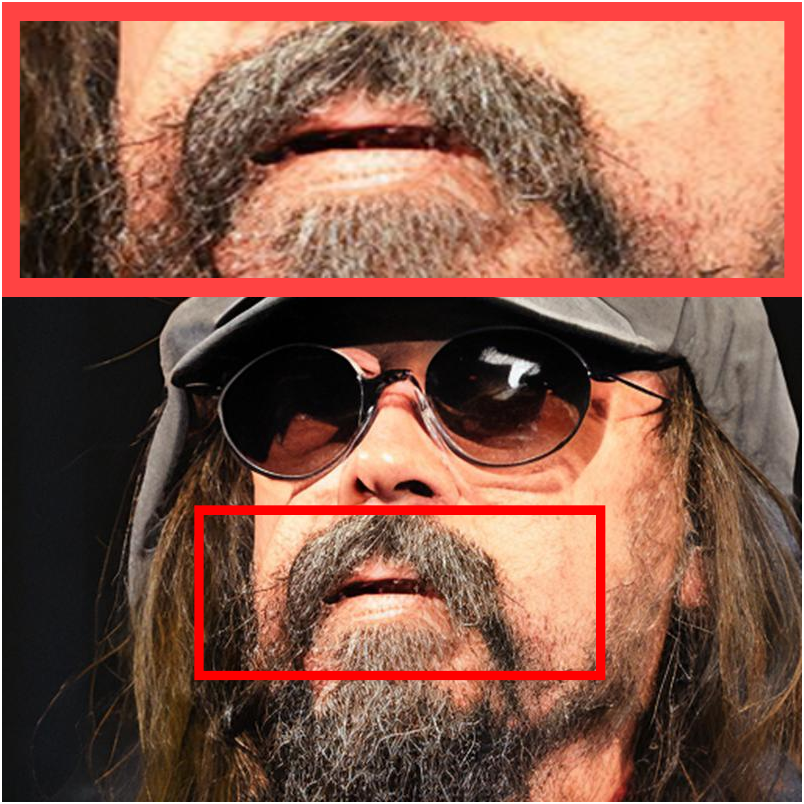} \hspace{-4mm} &
\includegraphics[width=0.24\linewidth]{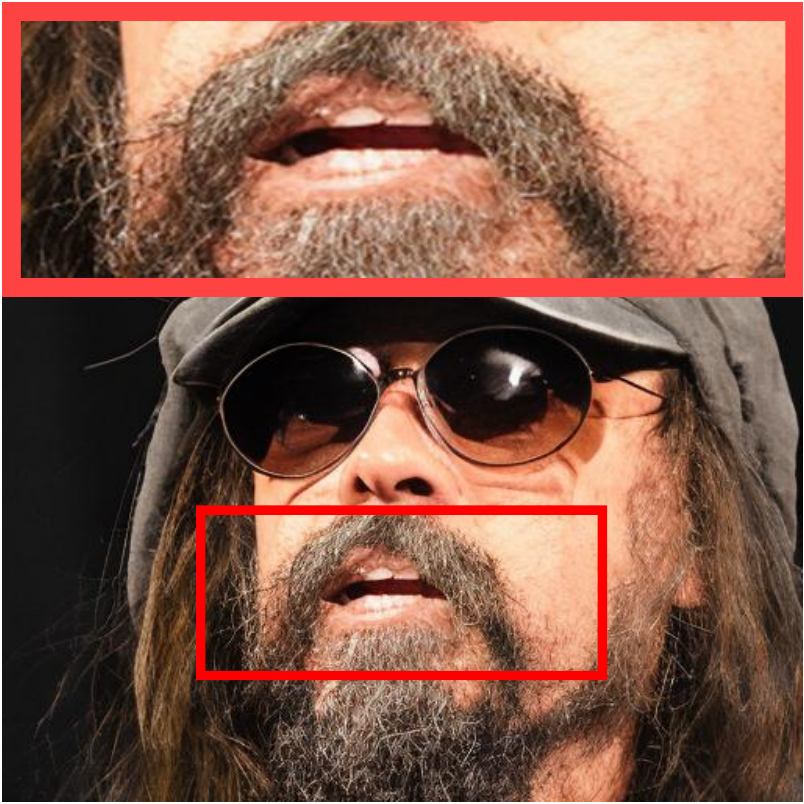} \hspace{-4mm} &
\includegraphics[width=0.24\linewidth]{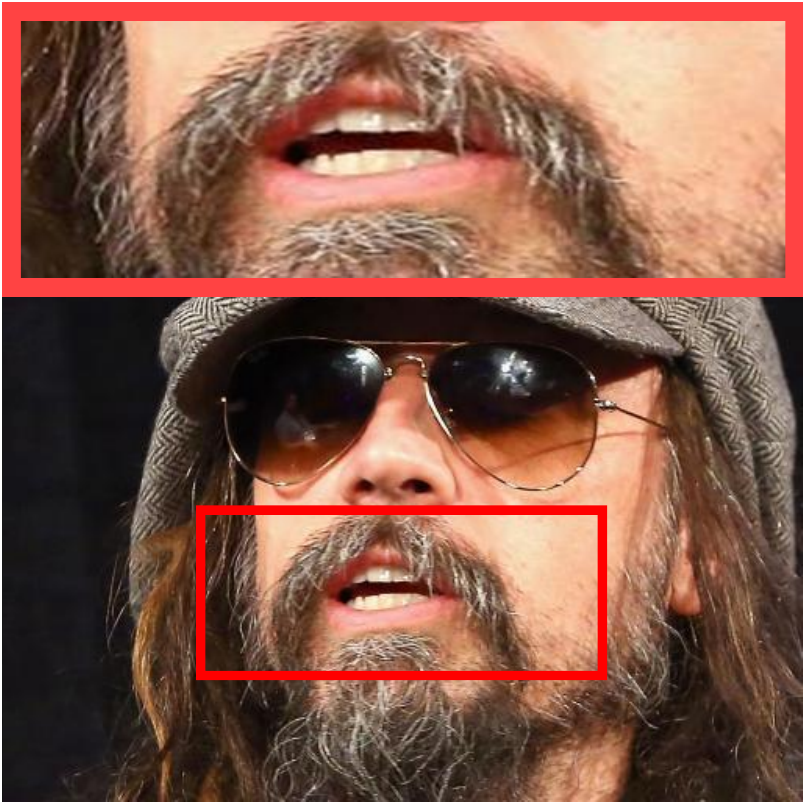} %
\\
Input \hspace{-4mm} &
w/o  Negative Prompt\hspace{-4mm} &
\makecell{Use Negative \\ Quality Prompt} \hspace{-4mm} &
\makecell{Use Negative \\ Attribute Prompt} \hspace{-4mm} &
GT
\\
\end{tabular}
\end{adjustbox}
\hspace{10mm} 
\end{tabular}
\hspace{10mm} 
\end{minipage}%
\hspace{28mm}
\begin{minipage}{0.28\textwidth}
\vspace{-2.5mm}
\captionsetup{font={small}}
\caption{
Negative quality prompts engender restoration outcomes characterized by high definition, whereas negative attribute prompts yield results with enhanced detail.}
\label{fig:a_prompt}
\end{minipage}
\vspace{-7mm}
\end{figure*}
\vspace{-2.mm}

\subsection{Ablation Study}
\vspace{-2.mm}

\begin{table}[t]
\vspace{2.5mm}
\begin{minipage}[t]{0.49\textwidth}
\centering
\captionsetup{font={small}, skip=8pt}
\vspace{0mm}
\resizebox{\linewidth}{!}{
\begin{tabular}{@{}ccccccc@{}}
\toprule
Real-SR(×8) & SSIM & PSNR & LPIPS & ManIQA & ClipIQA & MUSIQ \\ \midrule
w/o Link-UL & 0.6372 & 22.17 & 0.3275 & 0.5931 & 0.7082 & 71.33 \\
\rowcolor[HTML]{FFFDFA} 
{\color[HTML]{333333} w/o Link-LR} & {\color[HTML]{333333} 0.6659} & {\color[HTML]{333333} 23.86} & {\color[HTML]{333333} 0.2873} & {\color[HTML]{333333} 0.6152} & {\color[HTML]{333333} 0.6645} & {\color[HTML]{333333} 65.70} \\
MGBFR(Ours) & \multicolumn{1}{l}{\cellcolor[HTML]{FFFDFA}{\color[HTML]{000000} 0.6248}} & \multicolumn{1}{l}{\cellcolor[HTML]{FFFDFA}{\color[HTML]{000000} 23.10}} & \multicolumn{1}{l}{\cellcolor[HTML]{FFFDFA}{\color[HTML]{9A0000} 0.2688}} & \multicolumn{1}{l}{\cellcolor[HTML]{FFFDFA}{\color[HTML]{CB0000} 0.6535}} & \multicolumn{1}{l}{\cellcolor[HTML]{FFFDFA}{\color[HTML]{CB0000} 0.8147}} & \multicolumn{1}{l}{\cellcolor[HTML]{FFFDFA}{\color[HTML]{CB0000} 75.51}} \\ \bottomrule
\end{tabular}

}
\caption{Ablation study of additional information exchange in the MGBFR model. 'w/o Link-LR' means that the upward flow of information from LCA to RCA is removed.}
\label{tab10}
\end{minipage}
\hfill
\begin{minipage}[t]{0.49\textwidth}
 \centering
\captionsetup{font={small}, skip=8pt}
\vspace{0mm}
\resizebox{\linewidth}{!}{
\begin{tabular}{@{}ccc|cccccc@{}}
\toprule
\multicolumn{3}{c|}{Prompts} &  &  &  &  &  &  \\
$pos$ & $nq$ & $na$ & \multirow{-2}{*}{LPIPS ↓} & \multirow{-2}{*}{SSIM} & \multirow{-2}{*}{PSNR} & \multirow{-2}{*}{ManIQA} & \multirow{-2}{*}{ClipIQA} & \multirow{-2}{*}{MUSIQ} \\ \midrule
{} & {} & {} & 0.3264 & 0.6858 & 25.15 & 0.4782 & 0.2568 & 49.97 \\
{\checkmark} & {} & {} & 0.2690 & 0.6484 & 24.43 & 0.6441 & 0.7008 & 73.26 \\
{\checkmark} & {\checkmark} & {} & 0.2930 & 0.6066 & 23.27 & 0.6656 & 0.7999 & 75.34 \\
{\checkmark} & {} & {\checkmark} & 0.2702 & 0.6511 & 24.49 & 0.6437 & 0.7029 & 73.11 \\
{\checkmark} & {\checkmark} & {\checkmark} & 0.3227 & 0.5904 & 22.34 & {\color[HTML]{CB0000} 0.6776} & \color[HTML]{CB0000} 0.8083 & {\color[HTML]{CB0000} 75.94} \\ \bottomrule
\end{tabular}

}
\caption{Ablation study of attribute prompts and negative prompts}
\label{tab3}
\end{minipage}
\vspace{-6mm}
\end{table}

\textbf{Attribute Prompt and Negative Samples.}
\cref{fig:a_prompt} displays qualitative results under various settings, aligning with the strategies outlined in \cref{sec:Negative}. It can be seen that incorporating a negative quality prompt significantly enhances restoration quality, while the addition of a negative attribute prompt yields images with finer details. Quantitative results under various settings are also presented in \cref{tab3}. Adding either positive attribute prompts or negative quality prompts is observed to improve the perceived quality of the images significantly. Utilizing both types of prompts in conjunction with the negative attribute prompt achieves the most favourable perceived effect. The impact of hyperparameters on the results was also explored, revealing that settings of $\lambda_{na}=0.5$ and $\lambda_{nq}= 0.5$ yield the best perceptual outcomes, balancing sharpness and definition. Please refer to \cref{sec:Appen_prompt} for detailed qualitative results in different hyperparameters.

\textbf{Face Swapping.}
MGFR can facilitate face-swapping operations involving the processing of highly degraded LQ images to obscure identities, as shown in \cref{fig:SWA}. Face images and identity information from different identities are utilized as guides to achieve face swapping and identity replacement. Besides proposing an additional application for the model, this experiment further illustrates the method's efficacy in utilizing identity information and reference images for guidance.

\textbf{Additional Information Exchange.}
Unlike \citep{zhang2023adding}, we have integrated an additional information flow exchange link (Link-UL) from the U-net model to LCA, and a bidirectional information flow link (Link-LR) between LCA and RCA. \cref{tab10} displays the quantitative test results for the presence of the aforementioned information exchange links. Notably, `w/o Link-UL' refers to results obtained with a single information flow from LCA to U-net model. It is evident that additional information flow exchanges result in improved perceived quality.

\textbf{Arcface Identity Embedding.}
Our model is able to leverage identity information to guide the image restoration process, aiming to mitigate the deficit in facial identity information substantially.
As demonstrated in \cref{fig:ID}, our model employs the identity encoding formulated by the identity information extractor to mitigate the deficiency of facial identity information in the restored image. After losing arcface identity embedding, the recovered results still have high quality but there is a false illusion of face identity information.

\begin{figure}[t]
\captionsetup{font={small}, skip=14pt}
\scriptsize
\centering
\begin{tabular}{ccc}
\hspace{-0.5cm}
\begin{adjustbox}{valign=t}
\begin{tabular}{c}
\end{tabular}
\end{adjustbox}
\begin{adjustbox}{valign=t}
\begin{tabular}{ccccccccc}
\includegraphics[width=0.12\linewidth]{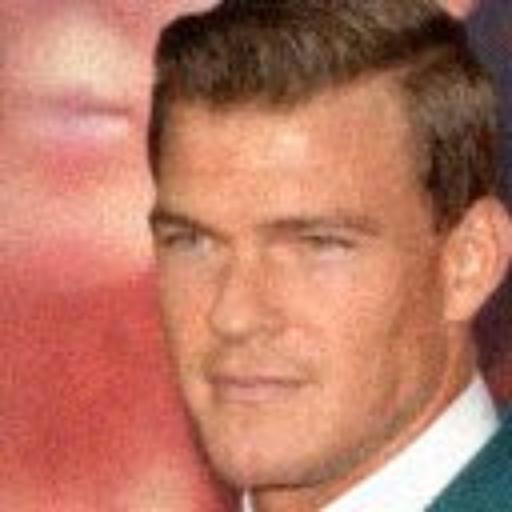} \hspace{-4mm} &
\includegraphics[width=0.12\linewidth]{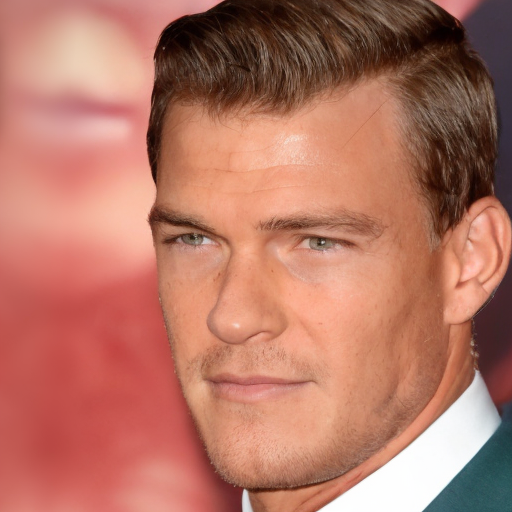} \hspace{-4mm} &
\includegraphics[width=0.122\linewidth]{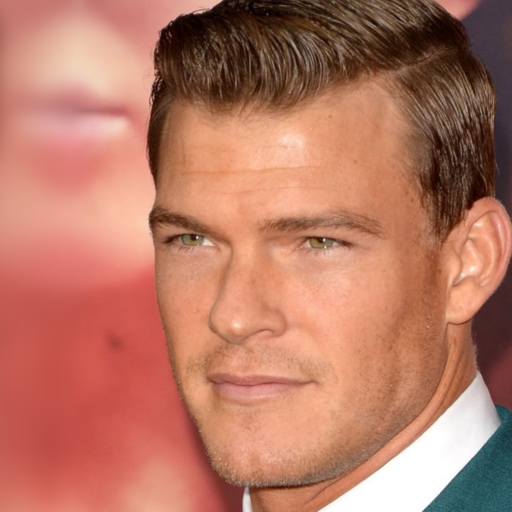} \hspace{-4mm} &
\includegraphics[width=0.12\linewidth]{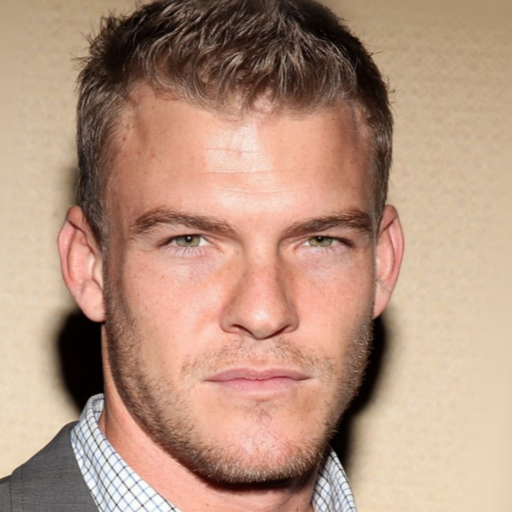} \hspace{-4mm} &
\tikz[baseline,overlay] \draw[dashed, thick] (0,0.12\linewidth) -- (0,-\tabcolsep); 
\includegraphics[width=0.12\linewidth]{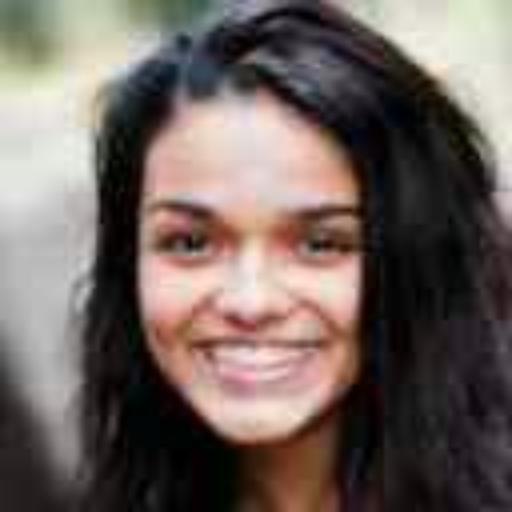} \hspace{-4mm} &
\includegraphics[width=0.12\linewidth]{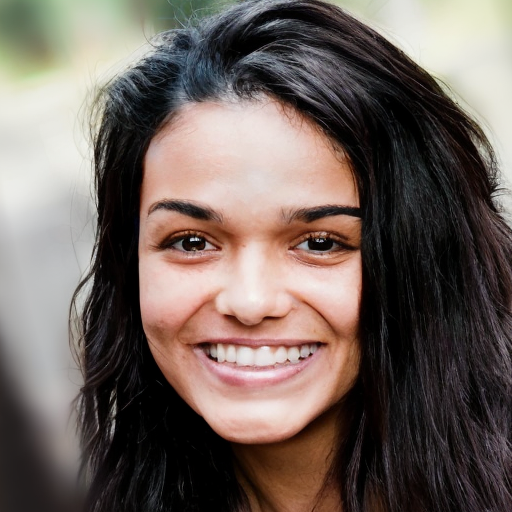} \hspace{-4mm} &
\includegraphics[width=0.12\linewidth]{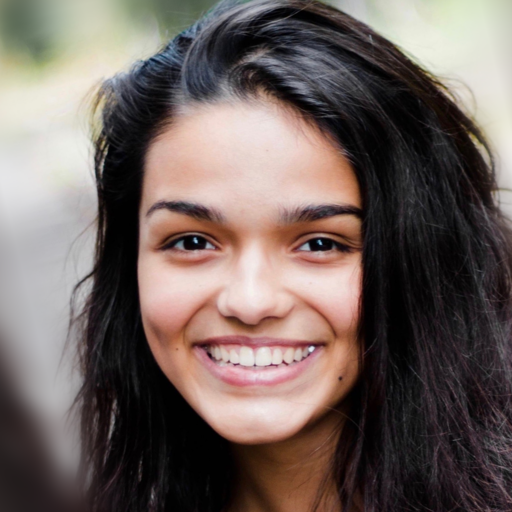} 
\hspace{-4mm} &
\includegraphics[width=0.12\linewidth]{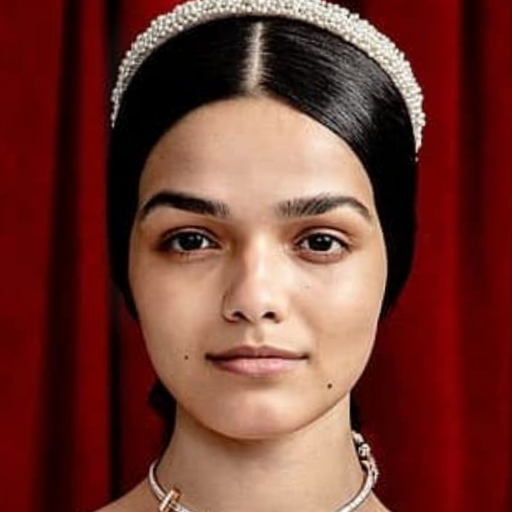} \hspace{-4mm} &
\\
\includegraphics[width=0.12\linewidth]{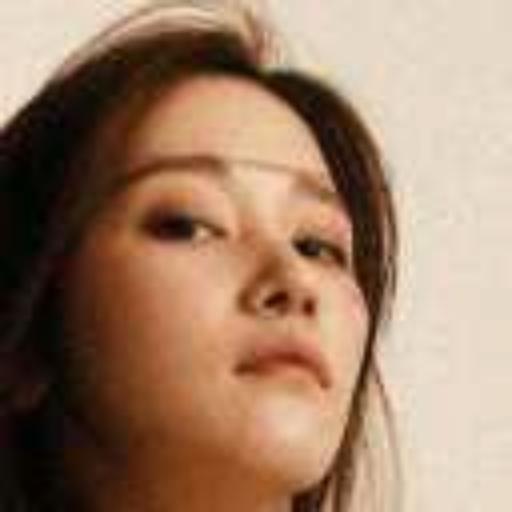} \hspace{-4mm} &
\includegraphics[width=0.12\linewidth]{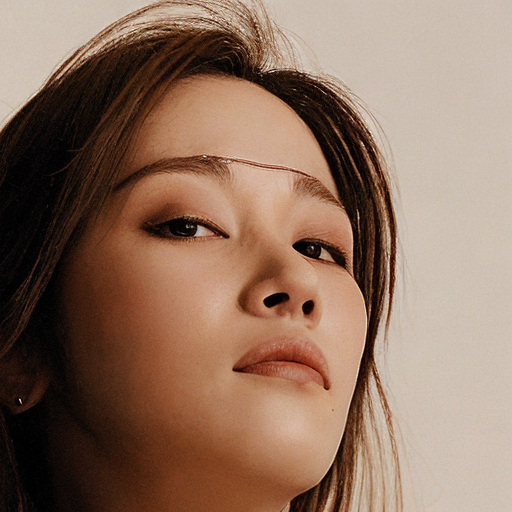} \hspace{-4mm} &
\includegraphics[width=0.122\linewidth]{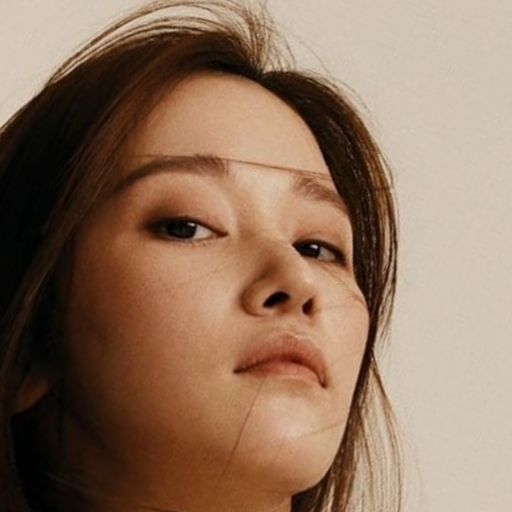} \hspace{-4mm} &
\includegraphics[width=0.12\linewidth]{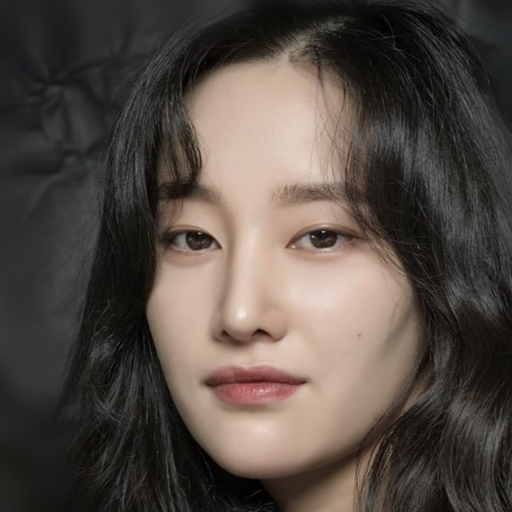} \hspace{-4mm} &
\tikz[baseline,overlay] \draw[dashed, thick] (0,0.12\linewidth) -- (0,-\tabcolsep); 
\includegraphics[width=0.12\linewidth]{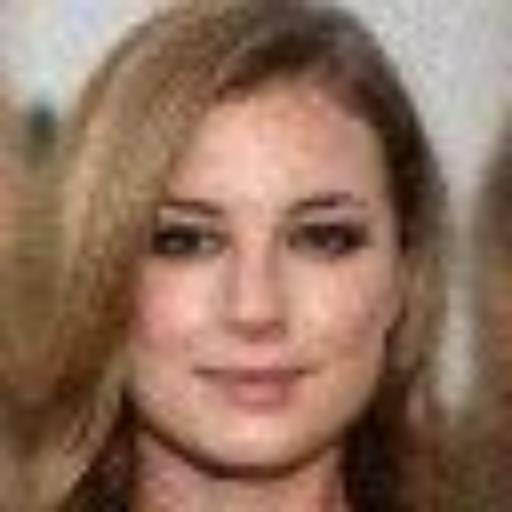} \hspace{-4mm} &
\includegraphics[width=0.12\linewidth]{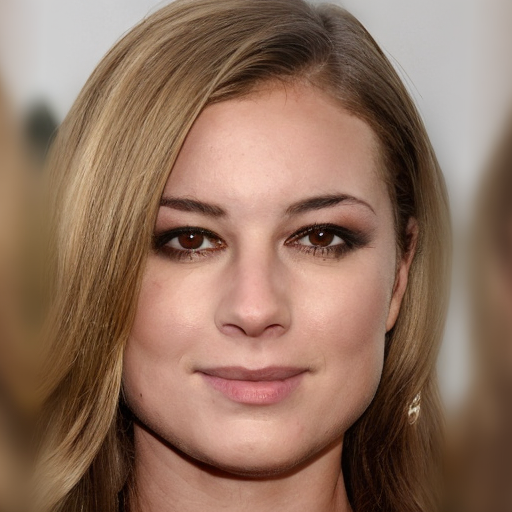} \hspace{-4mm} &
\includegraphics[width=0.12\linewidth]{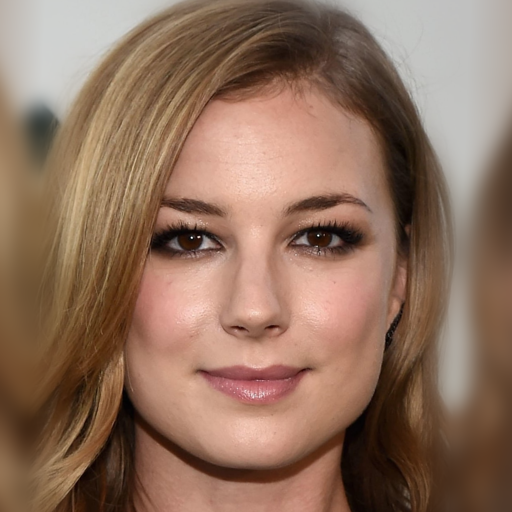} 
\hspace{-4mm} &
\includegraphics[width=0.12\linewidth]{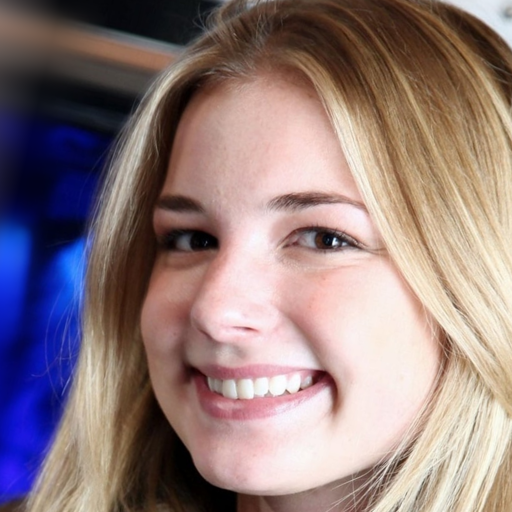} \hspace{-4mm} &
\\
\textbf{(a)} \hspace{3mm} LQ \hspace{0mm} &
Ours \hspace{-4mm} &
GT \hspace{-4mm} &
Reference \hspace{-4mm} &

\textbf{(b)} \hspace{3mm} LQ \hspace{0mm} &
Ours \hspace{-4mm} &
GT \hspace{-4mm} &
Reference \hspace{-4mm} &
\vspace{1mm}
\end{tabular}
\end{adjustbox}
\hspace{-0.55cm}
\end{tabular}
\vspace{-5mm}
\caption{\textbf{Face restoration results with reference images with different expressions and poses.} When there are differences in pose and expression between the reference image and the low-quality input image, our model can still achieve a good restoration effect without the generation of artifacts.}
\label{fig-ab}
\vspace{-2mm}
\end{figure}

\begin{figure}[t]
\captionsetup{font={small}, skip=14pt}
\scriptsize
\centering
\begin{tabular}{ccc}
\hspace{-0.5cm}
\begin{adjustbox}{valign=t}
\begin{tabular}{c}
\end{tabular}
\end{adjustbox}
\begin{adjustbox}{valign=t}
\begin{tabular}{ccccccccc}
\includegraphics[width=0.12\linewidth]{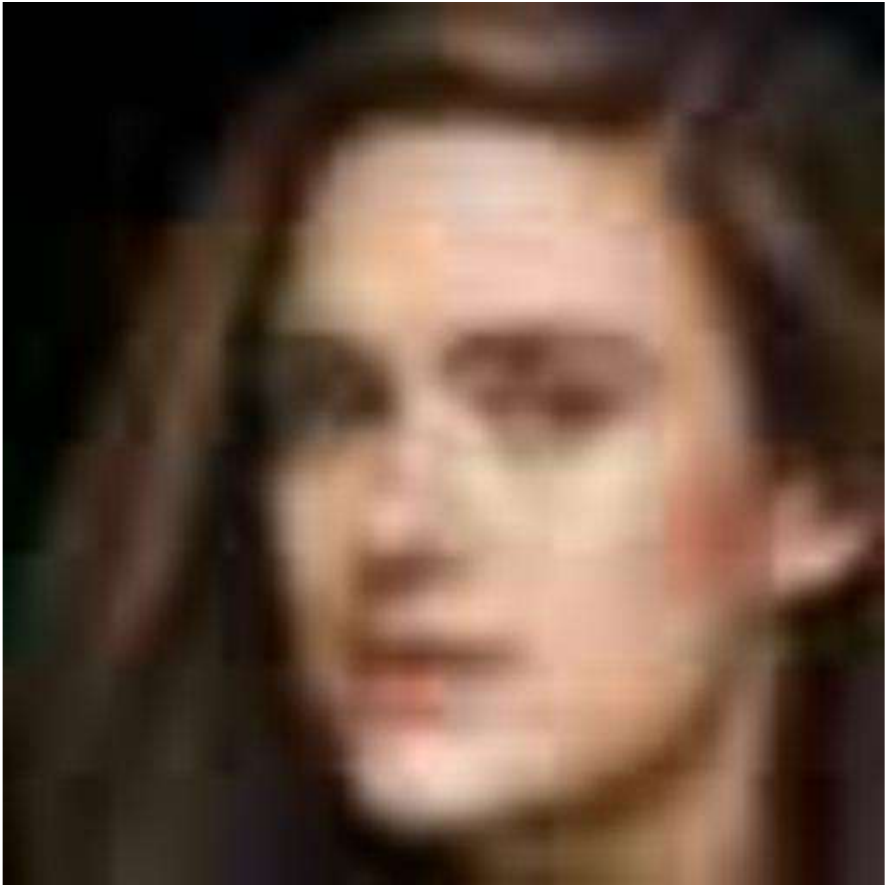} \hspace{-4mm} &
\includegraphics[width=0.12\linewidth]{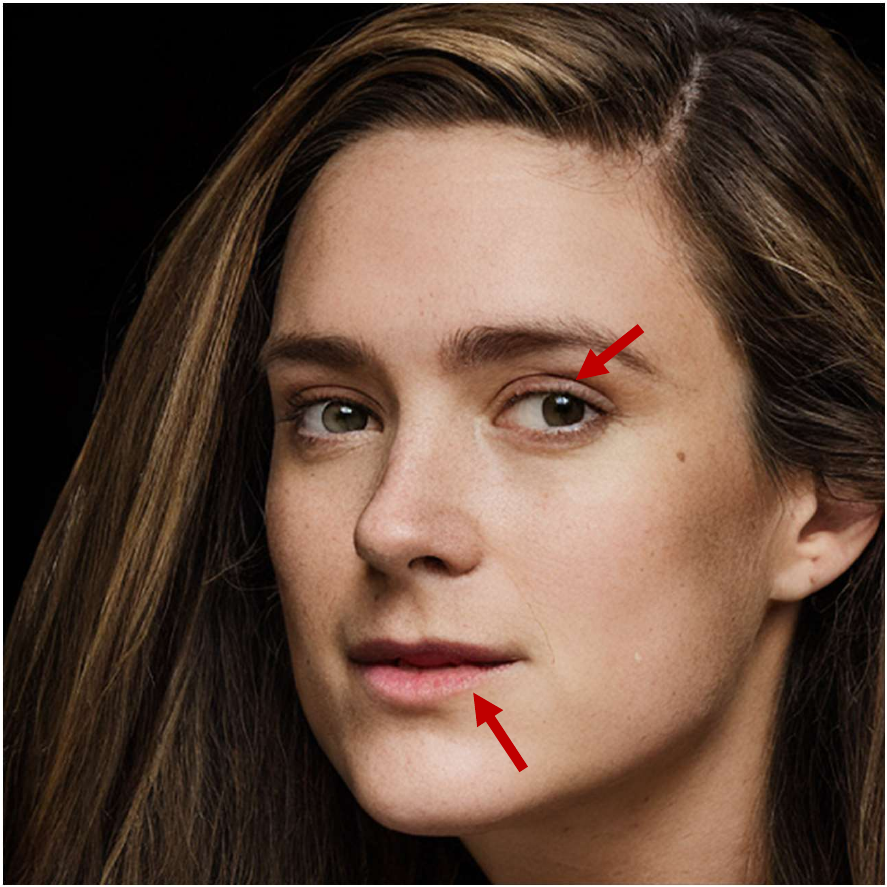} \hspace{-4mm} &
\includegraphics[width=0.12\linewidth]{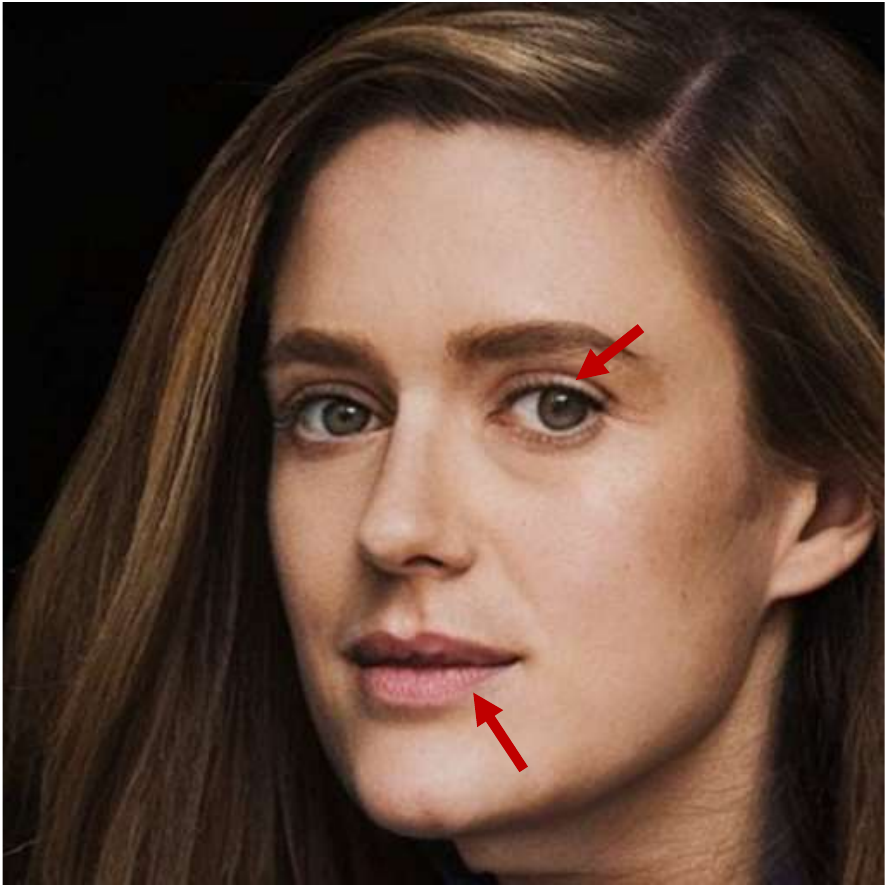} \hspace{-4mm} &
\includegraphics[width=0.12\linewidth]{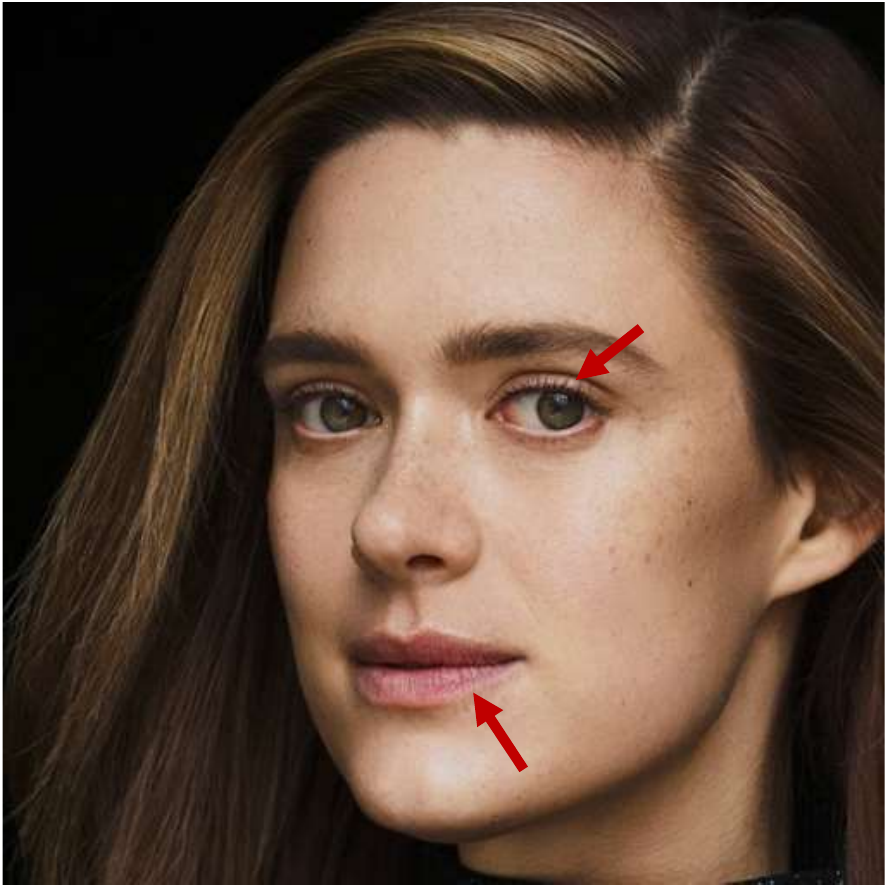} \hspace{-4mm} &
\includegraphics[width=0.12\linewidth]{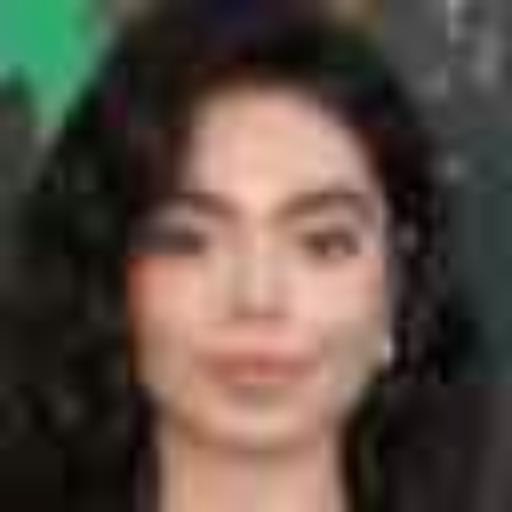} \hspace{-4mm} &
\includegraphics[width=0.12\linewidth]{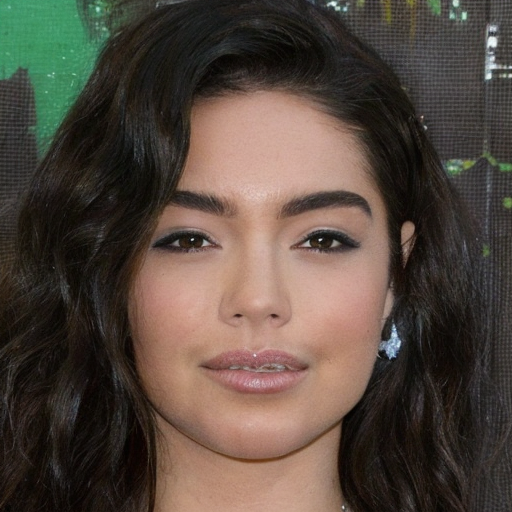} \hspace{-4mm} &
\includegraphics[width=0.12\linewidth]{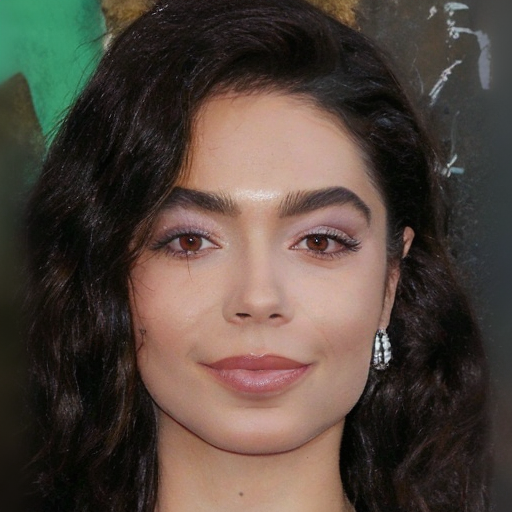} 
\hspace{-4mm} &
\includegraphics[width=0.12\linewidth]{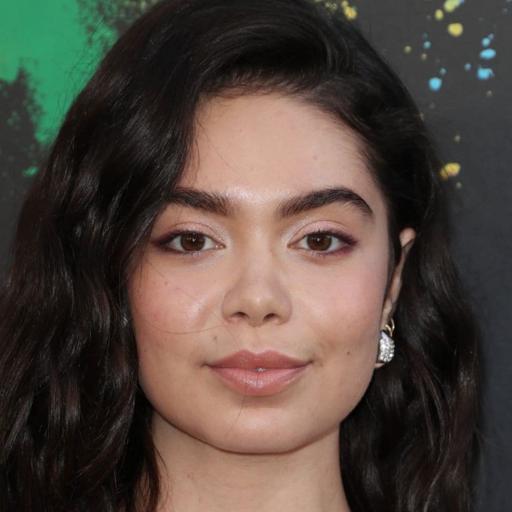} \hspace{-4mm} &
\\
\textbf{(a)} \hspace{3mm} LQ \hspace{0mm} &
Ours w/o ID\hspace{-4mm} &
Ours w/ ID\hspace{-4mm} &
GT \hspace{-4mm} &

\textbf{(b)} \hspace{3mm} LQ \hspace{0mm} &
Ours w/o ID \hspace{-4mm} &
Ours w/ ID \hspace{-4mm} &
GT \hspace{-4mm} &
\vspace{1mm}
\end{tabular}
\end{adjustbox}
\hspace{-0.55cm}
\end{tabular}
\vspace{-5mm}
\caption{\textbf{Ablation experiments about arcface identity embedding (ID).} The additional identity embedding can greatly reduce the false illusion of identity information in the recovery results.}
\label{fig:ID}
\vspace{-6mm}
\end{figure}

\textbf{Different expressions and poses reference images.}
In previous studies, reference image-based face restoration has been widely explored, but its efficacy is constrained by the need for strict alignment between the reference image and the low-quality (LQ) input. As shown in \cref{fig:B_4}, the recovery results of ASFFNet and MDMNet exhibit severe distortions when the reference image and the LQ input are slightly misaligned. However, our method completely resolves this issue, as it imposes no strict requirements on the expression, pose, or other variations of the reference image. As demonstrated in \cref{fig-ab}, even when there are discrepancies between the reference image and the LQ input, such as face orientation, labeling, makeup, or pose, our model consistently achieves high-quality restoration without any artifacts.

\vspace{-2mm}
\subsection{Limitations and Discussion}
\label{sec:limit}
\vspace{-2mm}
Although this represents an initial foray into attribute text-guided face image restoration, the flexibility of its text input is somewhat constrained due to the nature of the training samples. The model struggles to fully comprehend freely composed attribute description sentences, tending instead to rely on attribute labels embedded within fixed template text prompts, which limits its applicability in broader contexts. Furthermore, when users input attribute labels unseen during training, 
these do not effectively guide the recovery process. These limitations highlight the importance and necessity of utilizing high-quality data on a larger scale. 

Moreover, we believe that excessive flexibility in controlling facial attributes poses a risk of model misuse. Our model demonstrates robust recovery even in the absence of attribute hints. Therefore, for the first time, we incorporate additional multimodal information to enhance recovery quality without increasing task complexity. The reference image serves solely to enhance facial details and does not alter the recovered results. In summary, we remain committed to the core objective of the recovery task: ensuring that the output remains faithful to the low-quality input. Furthermore, leveraging multiple reference images enables more precise detail recovery, while the exploration of utilizing multiple low-quality reference images for collaborative guidance is left for future research.

\vspace{-2mm}
\section{Conclusion}
\vspace{-2mm}
We introduce MGFR as a pioneering method in real-world face restoration at the cutting edge of face image restoration technology, capable of using multi-modal information for guidance to achieve realistic visual effects. Simultaneously, MGFR extends the possibilities of face restoration by controlling text prompts with attributes. The proposed Reface-HQ dataset also offers significant potential for advancing the development of face restoration models based on reference images. As the first multi-modal face image restoration model, MGFR establishes a new benchmark for future technological advancements.

\section*{Acknowledge}
This work was supported by the National Key Research and Development Program of China (2024YFB2907500).

\bibliography{iclr2025_conference}

\begin{thebibliography}{60}
\providecommand{\natexlab}[1]{#1}
\providecommand{\url}[1]{\texttt{#1}}
\expandafter\ifx\csname urlstyle\endcsname\relax
  \providecommand{\doi}[1]{doi: #1}\else
  \providecommand{\doi}{doi: \begingroup \urlstyle{rm}\Url}\fi

\bibitem[Boutros et~al.(2023)Boutros, Grebe, Kuijper, and Damer]{Boutros2023IDiffFace}
Fadi Boutros, Jonas~Henry Grebe, Arjan Kuijper, and Naser Damer.
\newblock Idiff-face: Synthetic-based face recognition through fizzy identity-conditioned diffusion models.
\newblock In \emph{Proceedings of the IEEE/CVF International Conference on Computer Vision (ICCV)}, October 2023.

\bibitem[Cao et~al.(2018)Cao, Shen, Xie, Parkhi, and Zisserman]{cao2018vggface2}
Qiong Cao, Li~Shen, Weidi Xie, Omkar~M Parkhi, and Andrew Zisserman.
\newblock Vggface2: A dataset for recognising faces across pose and age.
\newblock In \emph{2018 13th IEEE international conference on automatic face \& gesture recognition (FG 2018)}, pp.\  67--74. IEEE, 2018.

\bibitem[Chan et~al.(2021)Chan, Wang, Xu, Gu, and Loy]{chan2021glean}
Kelvin~CK Chan, Xintao Wang, Xiangyu Xu, Jinwei Gu, and Chen~Change Loy.
\newblock Glean: Generative latent bank for large-factor image super-resolution.
\newblock In \emph{Proceedings of the IEEE/CVF conference on computer vision and pattern recognition}, pp.\  14245--14254, 2021.

\bibitem[Chen et~al.(2021)Chen, Li, Yang, Lin, Zhang, and Wong]{chen2021progressive}
Chaofeng Chen, Xiaoming Li, Lingbo Yang, Xianhui Lin, Lei Zhang, and Kwan-Yee~K Wong.
\newblock Progressive semantic-aware style transformation for blind face restoration.
\newblock In \emph{Proceedings of the IEEE/CVF conference on computer vision and pattern recognition}, pp.\  11896--11905, 2021.

\bibitem[Chen et~al.(2024)Chen, Tan, Wang, Zhang, Luo, and Cao]{chen2024towards}
Xiaoxu Chen, Jingfan Tan, Tao Wang, Kaihao Zhang, Wenhan Luo, and Xiaochun Cao.
\newblock Towards real-world blind face restoration with generative diffusion prior.
\newblock \emph{IEEE Transactions on Circuits and Systems for Video Technology}, 2024.

\bibitem[Chen et~al.(2018)Chen, Tai, Liu, Shen, and Yang]{chen2018fsrnet}
Yu~Chen, Ying Tai, Xiaoming Liu, Chunhua Shen, and Jian Yang.
\newblock Fsrnet: End-to-end learning face super-resolution with facial priors.
\newblock In \emph{Proceedings of the IEEE conference on computer vision and pattern recognition}, pp.\  2492--2501, 2018.

\bibitem[Chen et~al.(2023{\natexlab{a}})Chen, Zhang, Gu, Yuan, Kong, Chen, and Yang]{chen2023image}
Zheng Chen, Yulun Zhang, Jinjin Gu, Xin Yuan, Linghe Kong, Guihai Chen, and Xiaokang Yang.
\newblock Image super-resolution with text prompt diffusion.
\newblock \emph{arXiv preprint arXiv:2311.14282}, 2023{\natexlab{a}}.

\bibitem[Chen et~al.(2023{\natexlab{b}})Chen, Zhang, Liu, Xia, Gu, Kong, and Yuan]{chen2023hierarchical}
Zheng Chen, Yulun Zhang, Ding Liu, Bin Xia, Jinjin Gu, Linghe Kong, and Xin Yuan.
\newblock Hierarchical integration diffusion model for realistic image deblurring.
\newblock \emph{arXiv preprint arXiv:2305.12966}, 2023{\natexlab{b}}.

\bibitem[Deng et~al.(2019)Deng, Guo, Xue, and Zafeiriou]{deng2019arcface}
Jiankang Deng, Jia Guo, Niannan Xue, and Stefanos Zafeiriou.
\newblock Arcface: Additive angular margin loss for deep face recognition.
\newblock In \emph{Proceedings of the IEEE/CVF conference on computer vision and pattern recognition}, pp.\  4690--4699, 2019.

\bibitem[Dogan et~al.(2019)Dogan, Gu, and Timofte]{Dogan_2019_CVPR_Workshops}
Berk Dogan, Shuhang Gu, and Radu Timofte.
\newblock Exemplar guided face image super-resolution without facial landmarks.
\newblock In \emph{Proceedings of the IEEE/CVF Conference on Computer Vision and Pattern Recognition (CVPR) Workshops}, June 2019.

\bibitem[Dong et~al.(2014)Dong, Loy, He, and Tang]{dong2014learning}
Chao Dong, Chen~Change Loy, Kaiming He, and Xiaoou Tang.
\newblock Learning a deep convolutional network for image super-resolution.
\newblock In \emph{Computer Vision--ECCV 2014: 13th European Conference, Zurich, Switzerland, September 6-12, 2014, Proceedings, Part IV 13}, pp.\  184--199. Springer, 2014.

\bibitem[Dong et~al.(2015)Dong, Deng, Loy, and Tang]{dong2015compression}
Chao Dong, Yubin Deng, Chen~Change Loy, and Xiaoou Tang.
\newblock Compression artifacts reduction by a deep convolutional network.
\newblock In \emph{Proceedings of the IEEE international conference on computer vision}, pp.\  576--584, 2015.

\bibitem[Gu et~al.(2020)Gu, Shen, and Zhou]{gu2020image}
Jinjin Gu, Yujun Shen, and Bolei Zhou.
\newblock Image processing using multi-code gan prior.
\newblock In \emph{Proceedings of the IEEE/CVF conference on computer vision and pattern recognition}, pp.\  3012--3021, 2020.

\bibitem[Gu et~al.(2022)Gu, Wang, Xie, Dong, Li, Shan, and Cheng]{gu2022vqfr}
Yuchao Gu, Xintao Wang, Liangbin Xie, Chao Dong, Gen Li, Ying Shan, and Ming-Ming Cheng.
\newblock Vqfr: Blind face restoration with vector-quantized dictionary and parallel decoder.
\newblock In \emph{European Conference on Computer Vision}, pp.\  126--143. Springer, 2022.

\bibitem[He et~al.(2017)He, Wang, Fu, Feng, Jiang, and Xue]{he2017adaptively}
Keke He, Zhanxiong Wang, Yanwei Fu, Rui Feng, Yu-Gang Jiang, and Xiangyang Xue.
\newblock Adaptively weighted multi-task deep network for person attribute classification.
\newblock In \emph{Proceedings of the 25th ACM international conference on Multimedia}, pp.\  1636--1644, 2017.

\bibitem[Ho \& Salimans(2022)Ho and Salimans]{ho2022classifier}
Jonathan Ho and Tim Salimans.
\newblock Classifier-free diffusion guidance.
\newblock \emph{arXiv preprint arXiv:2207.12598}, 2022.

\bibitem[Hu et~al.(2023)Hu, Wang, and Zhang]{hu2023dear}
Yujie Hu, Yinhuai Wang, and Jian Zhang.
\newblock Dear-gan: Degradation-aware face restoration with gan prior.
\newblock \emph{IEEE Transactions on Circuits and Systems for Video Technology}, 2023.

\bibitem[Jinjin et~al.(2020)Jinjin, Haoming, Haoyu, Xiaoxing, Ren, and Chao]{jinjin2020pipal}
Gu~Jinjin, Cai Haoming, Chen Haoyu, Ye~Xiaoxing, Jimmy~S Ren, and Dong Chao.
\newblock Pipal: a large-scale image quality assessment dataset for perceptual image restoration.
\newblock In \emph{Computer Vision--ECCV 2020: 16th European Conference, Glasgow, UK, August 23--28, 2020, Proceedings, Part XI 16}, pp.\  633--651. Springer, 2020.

\bibitem[Karras et~al.(2019{\natexlab{a}})Karras, Laine, and Aila]{ffhq}
Tero Karras, Samuli Laine, and Timo Aila.
\newblock A style-based generator architecture for generative adversarial networks.
\newblock In \emph{Proceedings of the IEEE/CVF conference on computer vision and pattern recognition}, pp.\  4401--4410, 2019{\natexlab{a}}.

\bibitem[Karras et~al.(2019{\natexlab{b}})Karras, Laine, and Aila]{karras2019style}
Tero Karras, Samuli Laine, and Timo Aila.
\newblock A style-based generator architecture for generative adversarial networks.
\newblock In \emph{Proceedings of the IEEE/CVF conference on computer vision and pattern recognition}, pp.\  4401--4410, 2019{\natexlab{b}}.

\bibitem[Karras et~al.(2020)Karras, Laine, Aittala, Hellsten, Lehtinen, and Aila]{karras2020analyzing}
Tero Karras, Samuli Laine, Miika Aittala, Janne Hellsten, Jaakko Lehtinen, and Timo Aila.
\newblock Analyzing and improving the image quality of stylegan.
\newblock In \emph{Proceedings of the IEEE/CVF conference on computer vision and pattern recognition}, pp.\  8110--8119, 2020.

\bibitem[Ke et~al.(2021)Ke, Wang, Wang, Milanfar, and Yang]{ke2021musiq}
Junjie Ke, Qifei Wang, Yilin Wang, Peyman Milanfar, and Feng Yang.
\newblock Musiq: Multi-scale image quality transformer.
\newblock In \emph{Proceedings of the IEEE/CVF International Conference on Computer Vision}, pp.\  5148--5157, 2021.

\bibitem[Kingma \& Ba(2014)Kingma and Ba]{kingma2014adam}
Diederik~P Kingma and Jimmy Ba.
\newblock Adam: A method for stochastic optimization.
\newblock \emph{arXiv preprint arXiv:1412.6980}, 2014.

\bibitem[Kupyn et~al.(2018)Kupyn, Budzan, Mykhailych, Mishkin, and Matas]{kupyn2018deblurgan}
Orest Kupyn, Volodymyr Budzan, Mykola Mykhailych, Dmytro Mishkin, and Ji{\v{r}}{\'\i} Matas.
\newblock Deblurgan: Blind motion deblurring using conditional adversarial networks.
\newblock In \emph{Proceedings of the IEEE conference on computer vision and pattern recognition}, pp.\  8183--8192, 2018.

\bibitem[Li et~al.(2018{\natexlab{a}})Li, Liu, Ye, Zuo, Lin, and Yang]{Li_2018_ECCV}
Xiaoming Li, Ming Liu, Yuting Ye, Wangmeng Zuo, Liang Lin, and Ruigang Yang.
\newblock Learning warped guidance for blind face restoration.
\newblock In \emph{Proceedings of the European Conference on Computer Vision (ECCV)}, September 2018{\natexlab{a}}.

\bibitem[Li et~al.(2018{\natexlab{b}})Li, Liu, Ye, Zuo, Lin, and Yang]{li2018learning}
Xiaoming Li, Ming Liu, Yuting Ye, Wangmeng Zuo, Liang Lin, and Ruigang Yang.
\newblock Learning warped guidance for blind face restoration.
\newblock In \emph{Proceedings of the European conference on computer vision (ECCV)}, pp.\  272--289, 2018{\natexlab{b}}.

\bibitem[Li et~al.(2020{\natexlab{a}})Li, Chen, Zhou, Lin, Zuo, and Zhang]{li2020blind}
Xiaoming Li, Chaofeng Chen, Shangchen Zhou, Xianhui Lin, Wangmeng Zuo, and Lei Zhang.
\newblock Blind face restoration via deep multi-scale component dictionaries.
\newblock In \emph{European conference on computer vision}, pp.\  399--415. Springer, 2020{\natexlab{a}}.

\bibitem[Li et~al.(2020{\natexlab{b}})Li, Li, Ren, Zhang, Wang, and Zuo]{Li_2020_CVPR}
Xiaoming Li, Wenyu Li, Dongwei Ren, Hongzhi Zhang, Meng Wang, and Wangmeng Zuo.
\newblock Enhanced blind face restoration with multi-exemplar images and adaptive spatial feature fusion.
\newblock In \emph{Proceedings of the IEEE/CVF Conference on Computer Vision and Pattern Recognition (CVPR)}, June 2020{\natexlab{b}}.

\bibitem[Li et~al.(2020{\natexlab{c}})Li, Li, Ren, Zhang, Wang, and Zuo]{li2020enhanced}
Xiaoming Li, Wenyu Li, Dongwei Ren, Hongzhi Zhang, Meng Wang, and Wangmeng Zuo.
\newblock Enhanced blind face restoration with multi-exemplar images and adaptive spatial feature fusion.
\newblock In \emph{Proceedings of the IEEE/CVF Conference on Computer Vision and Pattern Recognition}, pp.\  2706--2715, 2020{\natexlab{c}}.

\bibitem[Li et~al.(2022)Li, Zhang, Zhou, Zhang, and Zuo]{li2022learning}
Xiaoming Li, Shiguang Zhang, Shangchen Zhou, Lei Zhang, and Wangmeng Zuo.
\newblock Learning dual memory dictionaries for blind face restoration.
\newblock \emph{IEEE Transactions on Pattern Analysis and Machine Intelligence}, 45\penalty0 (5):\penalty0 5904--5917, 2022.

\bibitem[Li et~al.(2023)Li, Zhang, Zhou, Zhang, and Zuo]{9921338}
Xiaoming Li, Shiguang Zhang, Shangchen Zhou, Lei Zhang, and Wangmeng Zuo.
\newblock Learning dual memory dictionaries for blind face restoration.
\newblock \emph{IEEE Transactions on Pattern Analysis and Machine Intelligence}, 45\penalty0 (5):\penalty0 5904--5917, 2023.
\newblock \doi{10.1109/TPAMI.2022.3215251}.

\bibitem[Lim et~al.(2017)Lim, Son, Kim, Nah, and Mu~Lee]{lim2017enhanced}
Bee Lim, Sanghyun Son, Heewon Kim, Seungjun Nah, and Kyoung Mu~Lee.
\newblock Enhanced deep residual networks for single image super-resolution.
\newblock In \emph{Proceedings of the IEEE conference on computer vision and pattern recognition workshops}, pp.\  136--144, 2017.

\bibitem[Lin et~al.(2023)Lin, He, Chen, Lyu, Fei, Dai, Ouyang, Qiao, and Dong]{diffbir}
Xinqi Lin, Jingwen He, Ziyan Chen, Zhaoyang Lyu, Ben Fei, Bo~Dai, Wanli Ouyang, Yu~Qiao, and Chao Dong.
\newblock Diffbir: Towards blind image restoration with generative diffusion prior.
\newblock \emph{arXiv preprint arXiv:2308.15070}, 2023.

\bibitem[Liu et~al.(2015)Liu, Luo, Wang, and Tang]{liu2015deep}
Ziwei Liu, Ping Luo, Xiaogang Wang, and Xiaoou Tang.
\newblock Deep learning face attributes in the wild.
\newblock In \emph{Proceedings of the IEEE international conference on computer vision}, pp.\  3730--3738, 2015.

\bibitem[Menon et~al.(2020)Menon, Damian, Hu, Ravi, and Rudin]{menon2020pulse}
Sachit Menon, Alexandru Damian, Shijia Hu, Nikhil Ravi, and Cynthia Rudin.
\newblock Pulse: Self-supervised photo upsampling via latent space exploration of generative models.
\newblock In \emph{Proceedings of the ieee/cvf conference on computer vision and pattern recognition}, pp.\  2437--2445, 2020.

\bibitem[Mou et~al.(2023)Mou, Wang, Xie, Wu, Zhang, Qi, Shan, and Qie]{mou2023t2i}
Chong Mou, Xintao Wang, Liangbin Xie, Yanze Wu, Jian Zhang, Zhongang Qi, Ying Shan, and Xiaohu Qie.
\newblock T2i-adapter: Learning adapters to dig out more controllable ability for text-to-image diffusion models.
\newblock \emph{arXiv preprint arXiv:2302.08453}, 2023.

\bibitem[Nitzan et~al.(2022)Nitzan, Aberman, He, Liba, Yarom, Gandelsman, Mosseri, Pritch, and Cohen-or]{nitzan2022mystyle}
Yotam Nitzan, Kfir Aberman, Qiurui He, Orly Liba, Michal Yarom, Yossi Gandelsman, Inbar Mosseri, Yael Pritch, and Daniel Cohen-or.
\newblock Mystyle: A personalized generative prior, 2022.

\bibitem[Radford et~al.(2021)Radford, Kim, Hallacy, Ramesh, Goh, Agarwal, Sastry, Askell, Mishkin, Clark, et~al.]{radford2021learning}
Alec Radford, Jong~Wook Kim, Chris Hallacy, Aditya Ramesh, Gabriel Goh, Sandhini Agarwal, Girish Sastry, Amanda Askell, Pamela Mishkin, Jack Clark, et~al.
\newblock Learning transferable visual models from natural language supervision.
\newblock In \emph{International conference on machine learning}, pp.\  8748--8763. PMLR, 2021.

\bibitem[Rombach et~al.(2022)Rombach, Blattmann, Lorenz, Esser, and Ommer]{rombach2022high}
Robin Rombach, Andreas Blattmann, Dominik Lorenz, Patrick Esser, and Bj{\"o}rn Ommer.
\newblock High-resolution image synthesis with latent diffusion models.
\newblock In \emph{Proceedings of the IEEE/CVF conference on computer vision and pattern recognition}, pp.\  10684--10695, 2022.

\bibitem[Ronneberger et~al.(2015)Ronneberger, Fischer, and Brox]{ronneberger2015u}
Olaf Ronneberger, Philipp Fischer, and Thomas Brox.
\newblock U-net: Convolutional networks for biomedical image segmentation.
\newblock In \emph{Medical Image Computing and Computer-Assisted Intervention--MICCAI 2015: 18th International Conference, Munich, Germany, October 5-9, 2015, Proceedings, Part III 18}, pp.\  234--241. Springer, 2015.

\bibitem[Shen et~al.(2020)Shen, Gu, Tang, and Zhou]{shen2020interpreting}
Yujun Shen, Jinjin Gu, Xiaoou Tang, and Bolei Zhou.
\newblock Interpreting the latent space of gans for semantic face editing.
\newblock In \emph{Proceedings of the IEEE/CVF conference on computer vision and pattern recognition}, pp.\  9243--9252, 2020.

\bibitem[Shen et~al.(2018)Shen, Lai, Xu, Kautz, and Yang]{shen2018deep}
Ziyi Shen, Wei-Sheng Lai, Tingfa Xu, Jan Kautz, and Ming-Hsuan Yang.
\newblock Deep semantic face deblurring.
\newblock In \emph{Proceedings of the IEEE conference on computer vision and pattern recognition}, pp.\  8260--8269, 2018.

\bibitem[Teng et~al.(2022)Teng, Yu, and Wu]{teng2022blind}
Zi~Teng, Xiaosheng Yu, and Chengdong Wu.
\newblock Blind face restoration via multi-prior collaboration and adaptive feature fusion.
\newblock \emph{Frontiers in Neurorobotics}, 16:\penalty0 797231, 2022.

\bibitem[Wang et~al.(2023{\natexlab{a}})Wang, Chan, and Loy]{wang2023exploring}
Jianyi Wang, Kelvin~CK Chan, and Chen~Change Loy.
\newblock Exploring clip for assessing the look and feel of images.
\newblock In \emph{Proceedings of the AAAI Conference on Artificial Intelligence}, volume~37, pp.\  2555--2563, 2023{\natexlab{a}}.

\bibitem[Wang et~al.(2021{\natexlab{a}})Wang, Li, Zhang, and Shan]{gfp}
Xintao Wang, Yu~Li, Honglun Zhang, and Ying Shan.
\newblock Towards real-world blind face restoration with generative facial prior.
\newblock In \emph{Proceedings of the IEEE/CVF conference on computer vision and pattern recognition}, pp.\  9168--9178, 2021{\natexlab{a}}.

\bibitem[Wang et~al.(2021{\natexlab{b}})Wang, Li, Zhang, and Shan]{wang2021towards}
Xintao Wang, Yu~Li, Honglun Zhang, and Ying Shan.
\newblock Towards real-world blind face restoration with generative facial prior.
\newblock In \emph{Proceedings of the IEEE/CVF conference on computer vision and pattern recognition}, pp.\  9168--9178, 2021{\natexlab{b}}.

\bibitem[Wang et~al.(2023{\natexlab{b}})Wang, Zhang, Zhang, Zheng, Zhou, Zhang, and Wang]{dr2}
Zhixin Wang, Ziying Zhang, Xiaoyun Zhang, Huangjie Zheng, Mingyuan Zhou, Ya~Zhang, and Yanfeng Wang.
\newblock Dr2: Diffusion-based robust degradation remover for blind face restoration.
\newblock In \emph{Proceedings of the IEEE/CVF Conference on Computer Vision and Pattern Recognition}, pp.\  1704--1713, 2023{\natexlab{b}}.

\bibitem[Wang et~al.(2022)Wang, Zhang, Chen, Wang, and Luo]{wang2022restoreformer}
Zhouxia Wang, Jiawei Zhang, Runjian Chen, Wenping Wang, and Ping Luo.
\newblock Restoreformer: High-quality blind face restoration from undegraded key-value pairs.
\newblock In \emph{Proceedings of the IEEE/CVF Conference on Computer Vision and Pattern Recognition}, pp.\  17512--17521, 2022.

\bibitem[Yang et~al.(2022)Yang, Wu, Shi, Lao, Gong, Cao, Wang, and Yang]{yang2022maniqa}
Sidi Yang, Tianhe Wu, Shuwei Shi, Shanshan Lao, Yuan Gong, Mingdeng Cao, Jiahao Wang, and Yujiu Yang.
\newblock Maniqa: Multi-dimension attention network for no-reference image quality assessment.
\newblock In \emph{Proceedings of the IEEE/CVF Conference on Computer Vision and Pattern Recognition}, pp.\  1191--1200, 2022.

\bibitem[Yang et~al.(2021{\natexlab{a}})Yang, Ren, Xie, and Zhang]{gpen}
Tao Yang, Peiran Ren, Xuansong Xie, and Lei Zhang.
\newblock Gan prior embedded network for blind face restoration in the wild.
\newblock In \emph{Proceedings of the IEEE/CVF Conference on Computer Vision and Pattern Recognition}, pp.\  672--681, 2021{\natexlab{a}}.

\bibitem[Yang et~al.(2021{\natexlab{b}})Yang, Ren, Xie, and Zhang]{yang2021gan}
Tao Yang, Peiran Ren, Xuansong Xie, and Lei Zhang.
\newblock Gan prior embedded network for blind face restoration in the wild.
\newblock In \emph{Proceedings of the IEEE/CVF Conference on Computer Vision and Pattern Recognition}, pp.\  672--681, 2021{\natexlab{b}}.

\bibitem[Yi et~al.(2014)Yi, Lei, Liao, and Li]{yi2014learning}
Dong Yi, Zhen Lei, Shengcai Liao, and Stan~Z Li.
\newblock Learning face representation from scratch.
\newblock \emph{arXiv preprint arXiv:1411.7923}, 2014.

\bibitem[Yu et~al.(2024)Yu, Gu, Li, Hu, Kong, Wang, He, Qiao, and Dong]{yu2024scaling}
Fanghua Yu, Jinjin Gu, Zheyuan Li, Jinfan Hu, Xiangtao Kong, Xintao Wang, Jingwen He, Yu~Qiao, and Chao Dong.
\newblock Scaling up to excellence: Practicing model scaling for photo-realistic image restoration in the wild.
\newblock \emph{arXiv preprint arXiv:2401.13627}, 2024.

\bibitem[Yu et~al.(2018)Yu, Fernando, Ghanem, Porikli, and Hartley]{yu2018face}
Xin Yu, Basura Fernando, Bernard Ghanem, Fatih Porikli, and Richard Hartley.
\newblock Face super-resolution guided by facial component heatmaps.
\newblock In \emph{Proceedings of the European conference on computer vision (ECCV)}, pp.\  217--233, 2018.

\bibitem[Zavadski et~al.(2023)Zavadski, Feiden, and Rother]{zavadski2023controlnet}
Denis Zavadski, Johann-Friedrich Feiden, and Carsten Rother.
\newblock Controlnet-xs: Designing an efficient and effective architecture for controlling text-to-image diffusion models.
\newblock \emph{arXiv preprint arXiv:2312.06573}, 2023.

\bibitem[Zhang et~al.(2017)Zhang, Zuo, Chen, Meng, and Zhang]{zhang2017beyond}
Kai Zhang, Wangmeng Zuo, Yunjin Chen, Deyu Meng, and Lei Zhang.
\newblock Beyond a gaussian denoiser: Residual learning of deep cnn for image denoising.
\newblock \emph{IEEE transactions on image processing}, 26\penalty0 (7):\penalty0 3142--3155, 2017.

\bibitem[Zhang et~al.(2023)Zhang, Rao, and Agrawala]{zhang2023adding}
Lvmin Zhang, Anyi Rao, and Maneesh Agrawala.
\newblock Adding conditional control to text-to-image diffusion models, 2023.

\bibitem[Zhang et~al.(2018)Zhang, Isola, Efros, Shechtman, and Wang]{zhang2018unreasonable}
Richard Zhang, Phillip Isola, Alexei~A Efros, Eli Shechtman, and Oliver Wang.
\newblock The unreasonable effectiveness of deep features as a perceptual metric.
\newblock In \emph{Proceedings of the IEEE conference on computer vision and pattern recognition}, pp.\  586--595, 2018.

\bibitem[Zhou et~al.(2022)Zhou, Chan, Li, and Loy]{coderformer}
Shangchen Zhou, Kelvin Chan, Chongyi Li, and Chen~Change Loy.
\newblock Towards robust blind face restoration with codebook lookup transformer.
\newblock \emph{Advances in Neural Information Processing Systems}, 35:\penalty0 30599--30611, 2022.

\bibitem[Zhu et~al.(2022)Zhu, Zhu, Chu, Zhang, Ji, Wang, and Tai]{zhu2022blind}
Feida Zhu, Junwei Zhu, Wenqing Chu, Xinyi Zhang, Xiaozhong Ji, Chengjie Wang, and Ying Tai.
\newblock Blind face restoration via integrating face shape and generative priors.
\newblock In \emph{Proceedings of the IEEE/CVF Conference on Computer Vision and Pattern Recognition}, pp.\  7662--7671, 2022.

\end{thebibliography}
\bibliographystyle{iclr2025_conference}

\appendix

\newpage
\appendix
\section*{Appendix}
\section{Reface-HQ Dataset}
\label{sec:data}
\begin{wraptable}{r}{0.55\textwidth}
\centering
\captionsetup{font={small}, skip=8pt}
\vspace{-2mm}
\resizebox{\linewidth}{!}{
\begin{tabular}{@{}ccccc@{}}
\toprule
\textbf{Dataset} & \textbf{Number of ID} & \textbf{Image} & \textbf{Size} & \textbf{Synthesized} \\ \midrule
CASIA-WebFace & 10575 & 494414 & 256$\times$256 & \XSolidBrush \\
Celeba & 10,177 & 202599 & 178$\times$218 & \XSolidBrush \\
IDiff-Face & - & - & 128$\times$128 & \Checkmark \\ \midrule
VggFace2 - HQ & 1200 & 24000 & 512$\times$512 & \XSolidBrush \\
CelebRef-HQ & 1000 & 10000 & 512$\times$512 & \XSolidBrush \\
Reface-HQ & {\color[HTML]{000000} 4800} & {\color[HTML]{000000} 21500} & 512$\times$512 & \XSolidBrush\\ \bottomrule
\end{tabular}
}
\caption{Datasets Comparison.}
\label{tab4}
\vspace{-4mm}
\end{wraptable}
The bulk of prior reference-based face restoration methodologies commonly focus on training and testing with $256 \times 256$ images. 
This is primarily due to the limitations of existing datasets such as CelebA \cite{liu2015deep}, VggFace2 \cite{cao2018vggface2}, and CASIA-WebFace \cite{yi2014learning}, which offer reference images mainly for face or attribute recognition but do not include high-quality images suitable for training at higher resolutions, like $512 \times 512$ or $1024 \times 1024$, thereby limiting their practical applications. 
Additionally, high-definition datasets recently introduced, such as CelebRef-HQ \cite{li2022learning} and VggFace2-HQ, face challenges in maximizing the potential of models due to their limited number of images and narrow range of identities. 

\begin{figure}[H]
  \captionsetup{font={small}, skip=6pt}
  \includegraphics[width=1\textwidth]{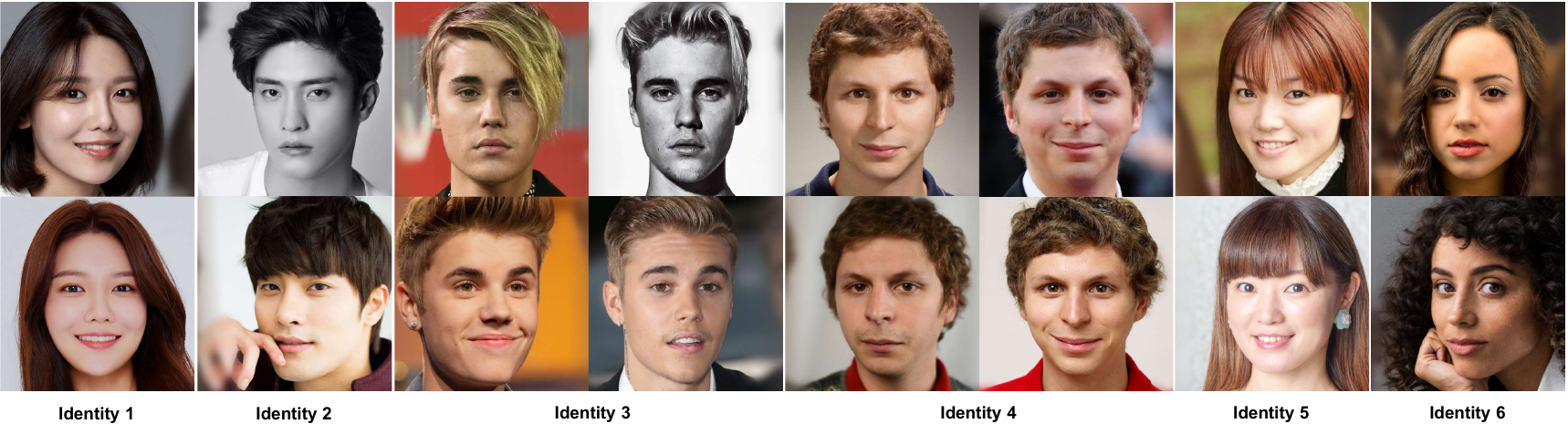}
  \caption{Demonstration of the Reface-HQ dataset.}
  \label{reface}
\end{figure}

To address this challenge, we have created a new \textbf{real-world} dataset named Reface-HQ, as shown in \cref{reface}.
The Reface-HQ dataset encompasses high-definition facial images of celebrities, which have been collected from the Internet. 
Initially, images with inadequate resolution (minimum 512), low quality and outliers lacking facial features were eliminated. 
Subsequently, identities represented by fewer than two images were excluded, and face image crop alignment was conducted. 
Each identity was also manually inspected to eliminate discrepancies in age and makeup. 
Additionally, to enhance the fairness and inclusiveness of the algorithm, we meticulously review the dataset to ensure it includes samples from all races and skin colors. We strive to ensure the diversity of the training data, thereby minimizing algorithmic bias and discrimination, and further enhancing the algorithm's fairness and inclusiveness.
In summary, Reface-HQ encompasses 4,800 identities, totaling 21500 images with a resolution of 512, subsequently partitioned into three segments: 4520 identities for the training set and 280 for the Reface-Test. The comparison of datasets available for special face restoration tasks is shown in \cref{tab4}. IDiff-Face \cite{Boutros2023IDiffFace} is a composite dataset with an indefinite number of images.

\subsection{Ablation Experiment}

For diffusion models and adapter structures, both the quality and quantity of training data are critical factors affecting the model's final performance. \cref{tab5} presents the quantitative comparison results of our proposed model under various training data volumes. It is evident that the model's performance significantly decreases with only 10,000 training data samples.
\begin{table}[h!]
\centering
\captionsetup{font={small}, skip=8pt}

\resizebox{0.75\linewidth}{!}{
\begin{tabular}{@{}ccccccc@{}}
\toprule
\textbf{Real-SR(×8)} & \textbf{SSIM} & \textbf{PSNR} & \textbf{LPIPS} & \textbf{ManIQA} & \textbf{ClipIQA} & \textbf{MUSIQ} \\ \midrule
{\color[HTML]{333333} 10K Training Samples} & {\color[HTML]{333333} 0.6254} & {\color[HTML]{333333} 23.46} & {\color[HTML]{333333} 0.2535} & {\color[HTML]{333333} 0.6388} & {\color[HTML]{333333} 0.7424} & {\color[HTML]{333333} 72.23} \\
{\color[HTML]{333333} 20K Training Samples} & {\color[HTML]{333333} 0.6248} & {\color[HTML]{333333} 23.10} & {\color[HTML]{333333} 0.2688} & {\color[HTML]{CB0000} 0.6535} & {\color[HTML]{CB0000} 0.8147} & {\color[HTML]{CB0000} 75.51} \\ \bottomrule
\end{tabular}
}
\caption{Ablation Experiment about training.}
\vspace{-6mm}
\label{tab5}
\end{table}

\clearpage
\section{Attribute Prompt}
\label{sec:prompt}
\begin{figure}[h]
  \captionsetup{font={small}, skip=6pt}
  \includegraphics[width=1\textwidth]{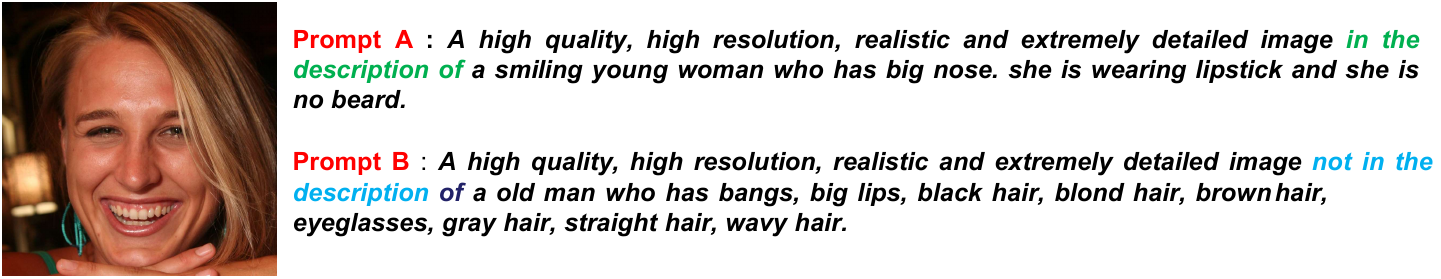}
  \caption{Attribute prompts composition in training.}
  \label{A:Prompt}
\end{figure}

\begin{wraptable}{r}{0.55\textwidth}
\centering
\captionsetup{font={small}, skip=8pt}
\vspace{-3mm}
\resizebox{\linewidth}{!}{
\begin{tabular}{@{}ccccc@{}}
\toprule
\textbf{Row 1} & \textbf{Row 2} & \textbf{Row 3} & \textbf{Row 4}\\ 
\midrule
Black Hair & Blond Hair & Blurry & Brown Hair\\ 
- & Eyeglasses & Gray Hair & Heavy Makeup \\ 
Mouth Slightly Open & Mustache & Big Eyes & No Beard \\
\midrule
Receding Hairline & Sideburns & Smiling & Straight Hair  \\
Wearing Earrings & Wearing Hat & Male & Wearing Necklace \\
 Big Nose  & - & Wearing Lipstick & Young \\
 Wavy Hair & Big Lips & Bald & Bangs \\
\bottomrule
\end{tabular}
}
\caption{Face Attribute.}
\label{tab-a}
\vspace{-4mm}
\end{wraptable}
This section provides a supplementary note on the attribute text prompts utilized in MGFR. For the training data, attribute labels are first extracted from the FFHQ or Reface-HQ dataset's face images using a facial attribute classifier. 
The 28 types of attributes included are listed in \cref{tab-a}, while labels with binomial characteristics (such as Male and Female, no beard and beard, etc.) are not repetitively shown. 
Regarding the classification threshold, attributes with a probability greater than 0.6 are considered positive, those with a probability less than 0.4 as negative, and the rest as uncertain in describing facial features. 
LLM is utilized to embed the attribute labels into a descriptive sentence template, thereby enhancing the model's understanding. To augment the model's grasp of negative attribute descriptions, two sentences of prompt text are provided for each image, as illustrated in \cref{A:Prompt}. Both descriptions offer a positive portrayal of the face, with Prompt B specifically focusing on the negative attributes.

In the inference stage, following the approach detailed in \cref{sec:Negative}, we apply positive attribute prompts ($pos$), negative quality prompts ($nq$), and negative attribute prompts ($na$) in each iteration.
For example, in restoring a LQ image, if it is assumed to contain attributes like 'smiling, man, black hair, eyeglasses,' the corresponding text for image restoration can be generated as follows:

\begin{itemize}
    \item \textbf{Positive Prompt:} A high quality, high resolution, realistic and extremely detailed image in the description of a smiling man who has black hair and eyeglasses.
    \item \textbf{Negative Attribute Prompt:} A high quality, high resolution, realistic and extremely detailed image not in the description of a smiling man who has black hair and eyeglasses.
    \item \textbf{Negative Quality Prompt:} A low quality, low resolution, over smooth and deformation image.
\end{itemize}

The underlying premise is to prevent our model from generating low-quality images and images with mismatched facial attributes. Extensive experiments demonstrate the effectiveness of our proposed attribute prompts.

\clearpage
\section{More Qualitative Comparisons For Our Model without Reference Images.}
\label{sec:c_base}
\begin{wraptable}{r}{0.4\textwidth}
\centering
\captionsetup{font={small}, skip=8pt}
\vspace{-4mm}
\resizebox{\linewidth}{!}{
\begin{tabular}{@{}cccc@{}}
\toprule
\textbf{WebPhoto-Test} & \textbf{ManIQA} & \textbf{ClipIQA} & \textbf{MUSIQ} \\ \midrule
DR2 & 0.4868 & 0.6184 & 64.36 \\
DiffBIR & 0.4068 & 0.6858 & 55.73 \\
Ours & \textbf{0.5901} & \textbf{0.8397} & \textbf{72.52} \\ \bottomrule
\end{tabular}
}
\caption{Quantitative comparison with other diffusion model-based methods on real-world
degradations in WebPhoto-Test.}
\label{tab11}
\vspace{-5mm}
\end{wraptable}
This section presents qualitative comparisons experimental results of our model without reference images, focusing on attribute text-guided face recovery. 
Importantly, for a fair comparison, the attribute during the inference phase are derived from the LQ input, which means the model's maximum potential is not fully realized. 
We assert that in practical scenarios, users will be able to supply more precise attribute text for enhanced recovery guidance. 
Although, our model demonstrates the most superior visual effects and details when compared to other state-of-the-art methods.

\begin{figure*}[h]
\captionsetup{font={small}, skip=12pt}
\scriptsize
\begin{tabular}{ccc}
\hspace{-0.55cm}
\begin{adjustbox}{valign=t}
\begin{tabular}{c}
\end{tabular}
\end{adjustbox}
\begin{adjustbox}{valign=t}
\begin{tabular}{ccccccccc}
\includegraphics[width=0.109\linewidth]{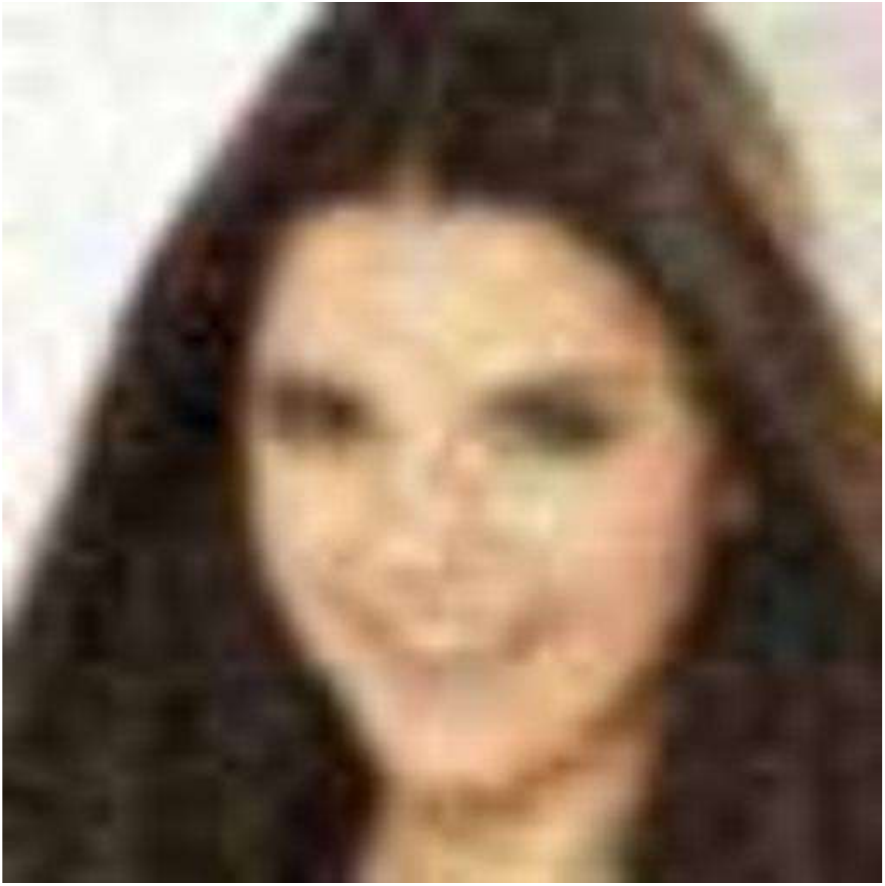} \hspace{-4.6mm} &
\includegraphics[width=0.1092\linewidth]{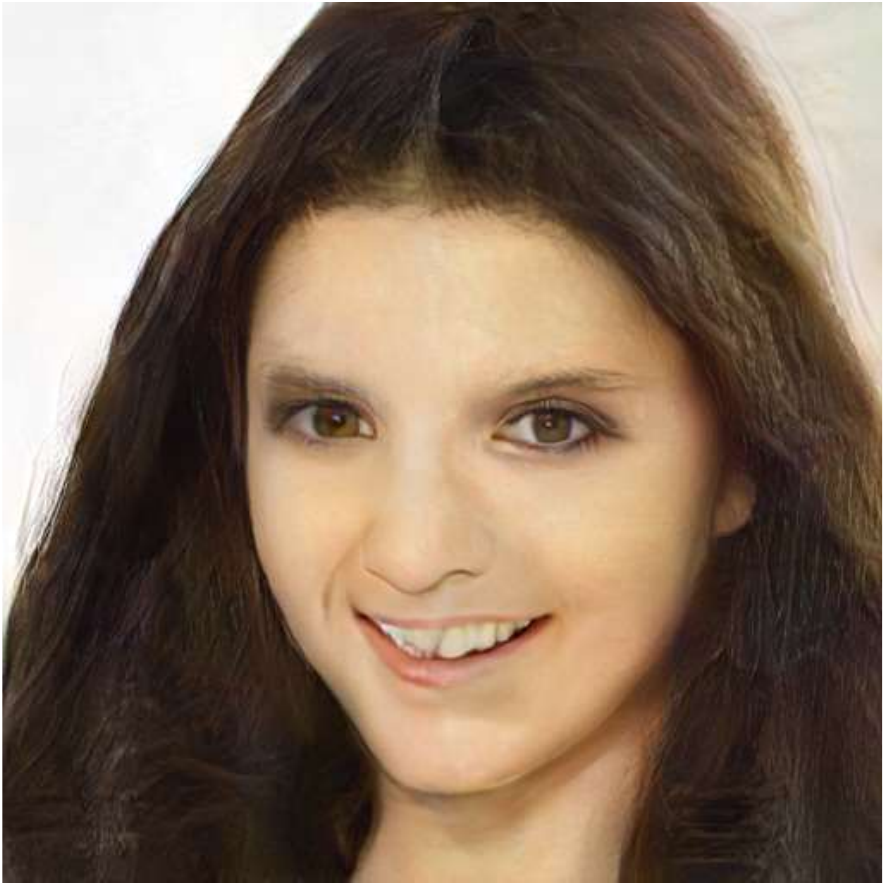} \hspace{-4.6mm} &
\includegraphics[width=0.1092\linewidth]{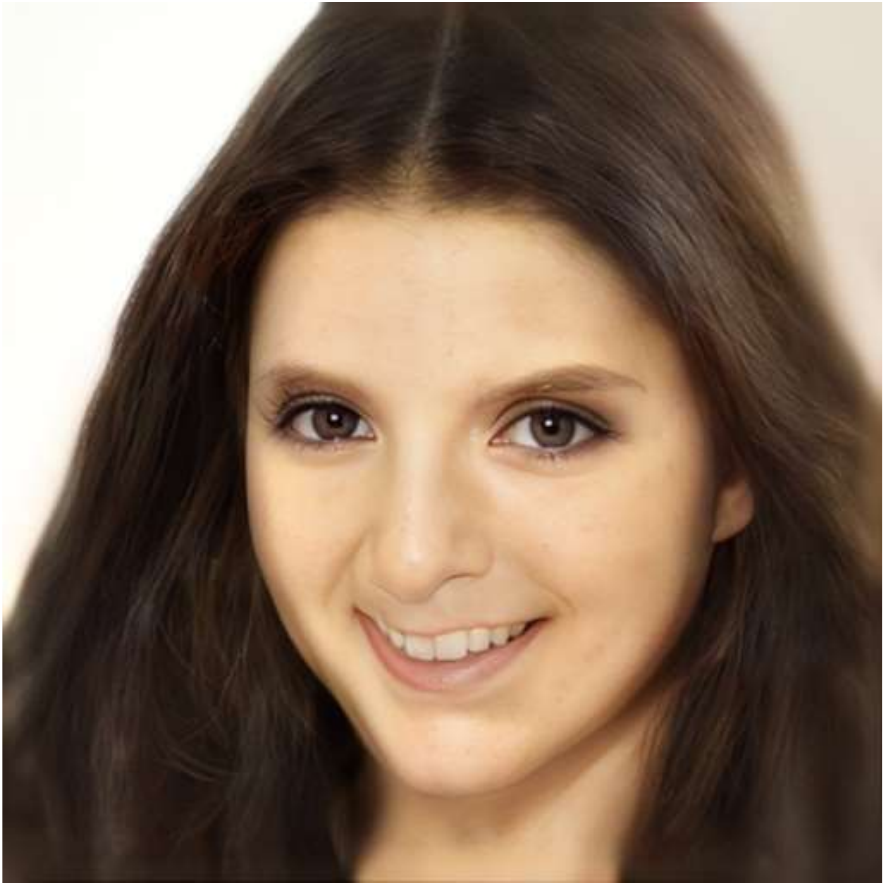} \hspace{-4.6mm} &
\includegraphics[width=0.1092\linewidth]{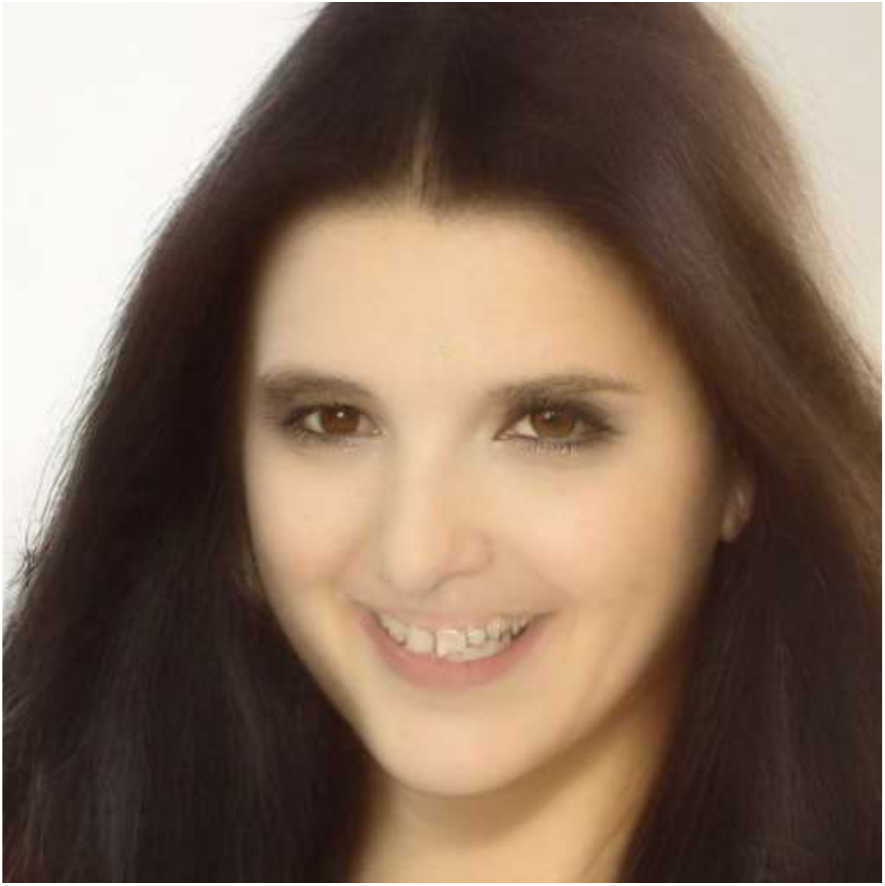} \hspace{-4.6mm} &
\includegraphics[width=0.1092\linewidth]{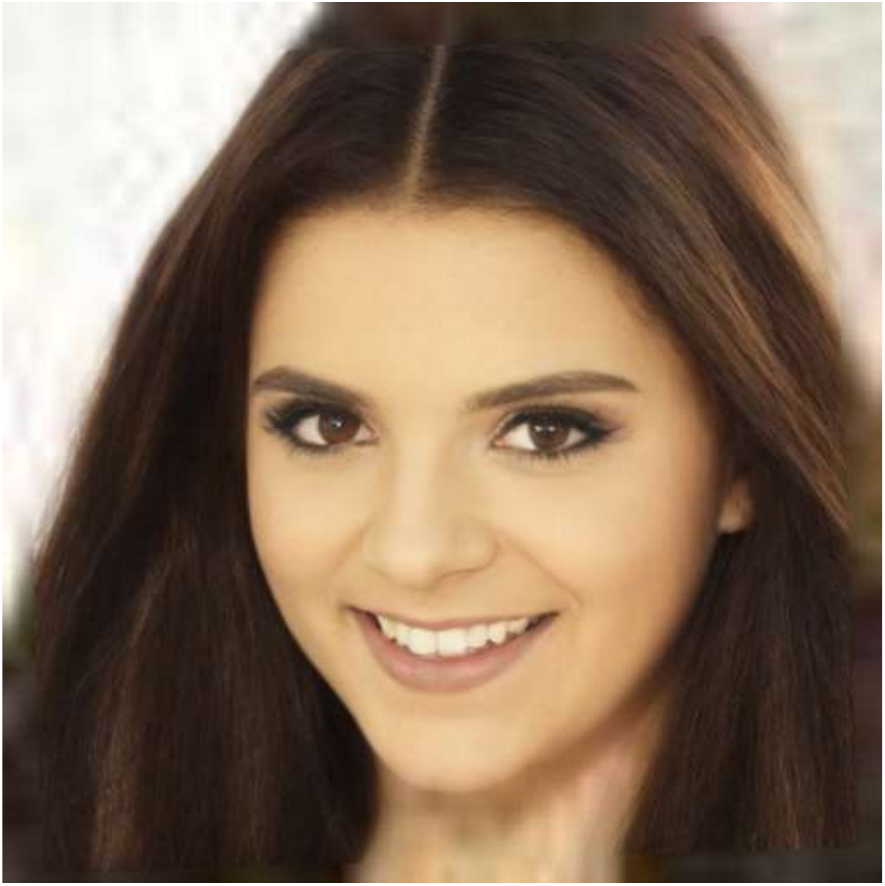} \hspace{-4.6mm} &
\includegraphics[width=0.1092\linewidth]{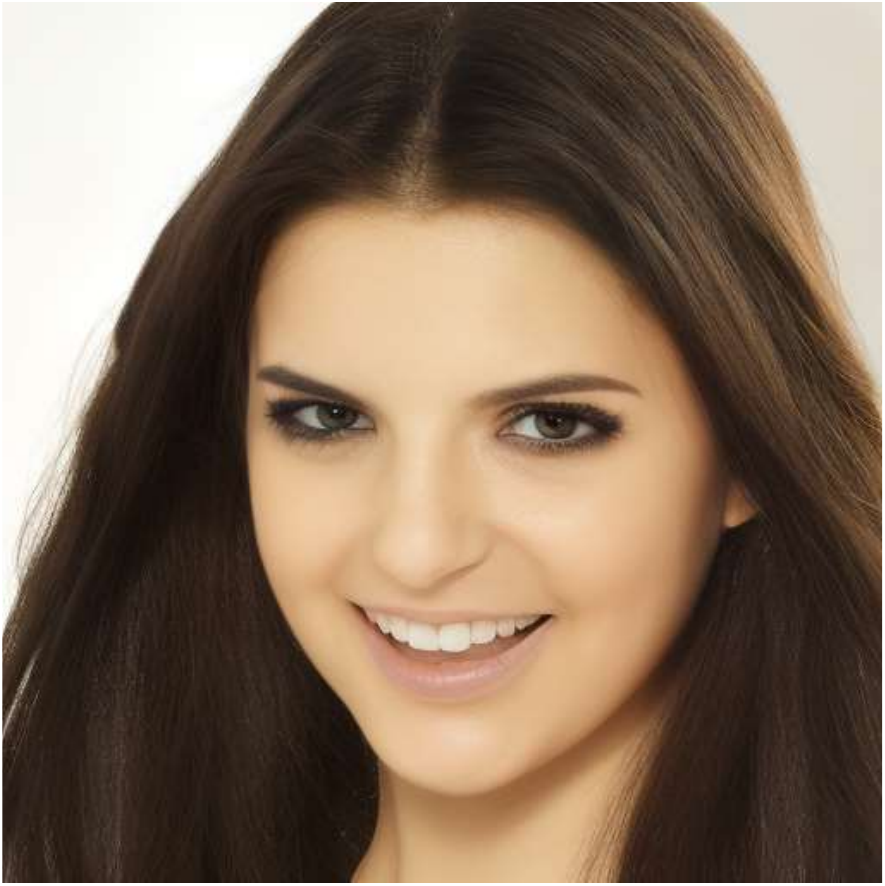} \hspace{-4.6mm} &
\includegraphics[width=0.1092\linewidth]{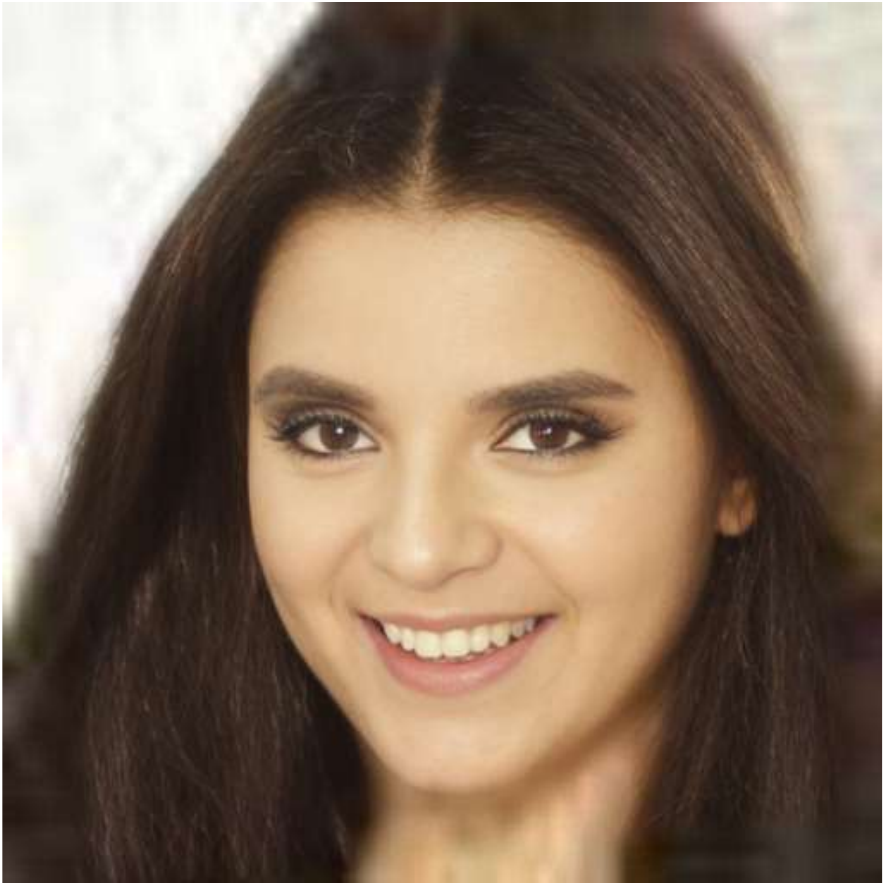} \hspace{-4.6mm} &
\includegraphics[width=0.1092\linewidth]{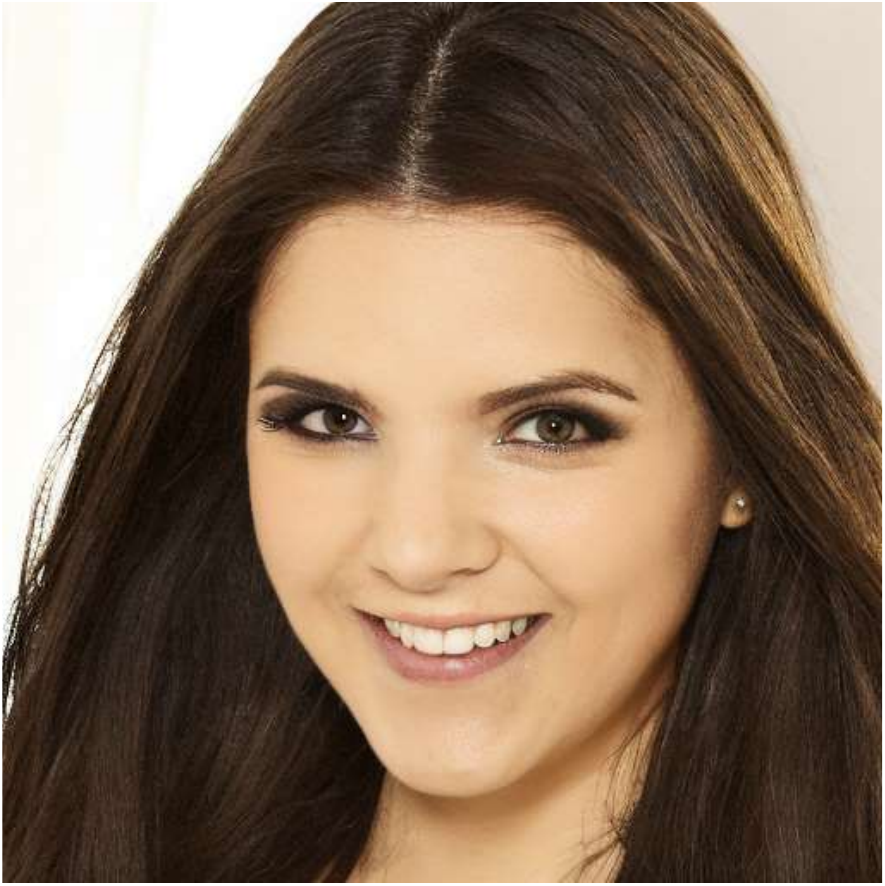} \hspace{-4.6mm} &
\includegraphics[width=0.1092\linewidth]
{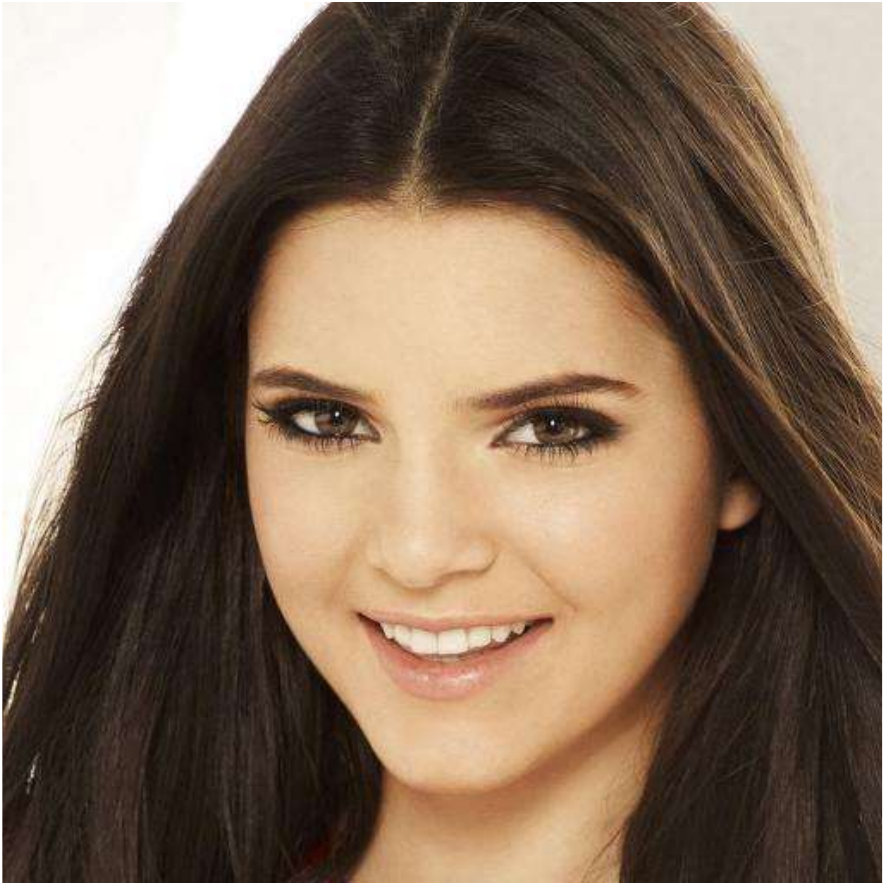}
\end{tabular}
\end{adjustbox}
\vspace{0.05mm}
\\
\hspace{-0.55cm}
\begin{adjustbox}{valign=t}
\begin{tabular}{c}
\end{tabular}
\end{adjustbox}
\begin{adjustbox}{valign=t}
\begin{tabular}{ccccccccc}
\includegraphics[width=0.109\linewidth]
{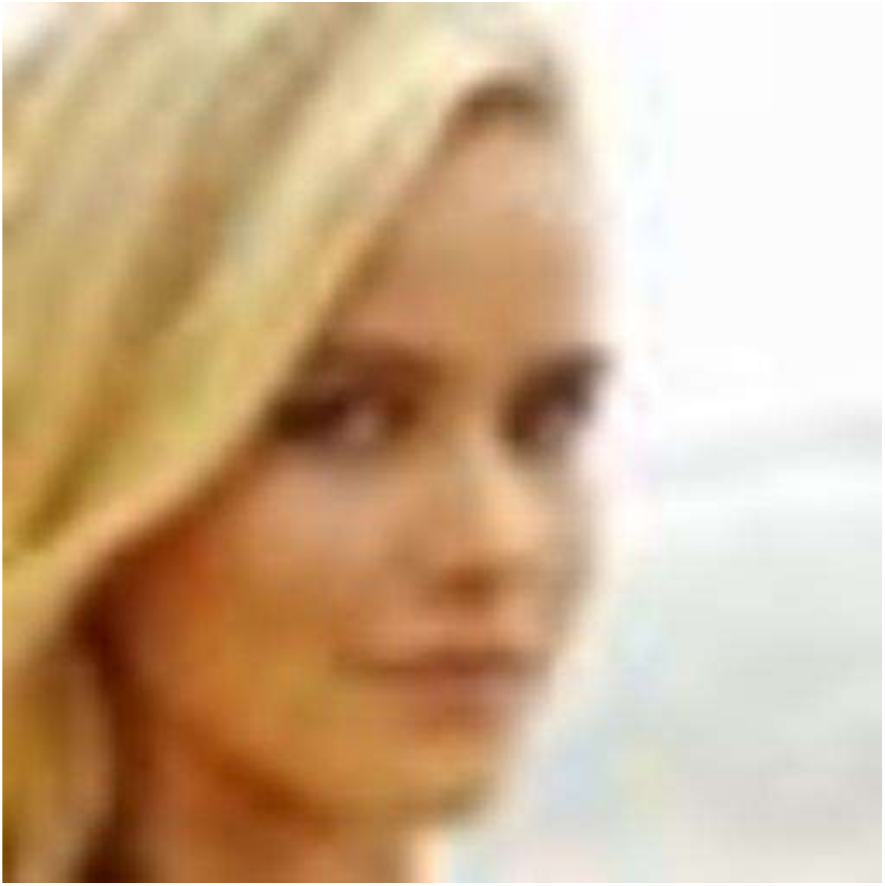} \hspace{-4.6mm} &
\includegraphics[width=0.1092\linewidth]{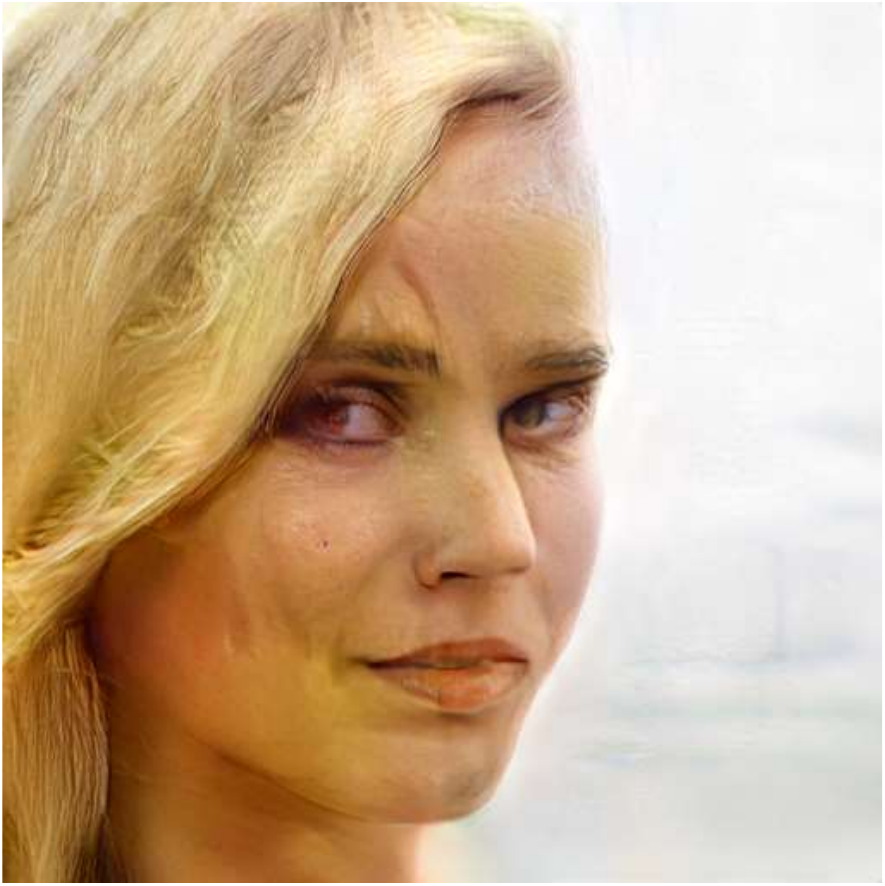} \hspace{-4.6mm} &
\includegraphics[width=0.1092\linewidth]{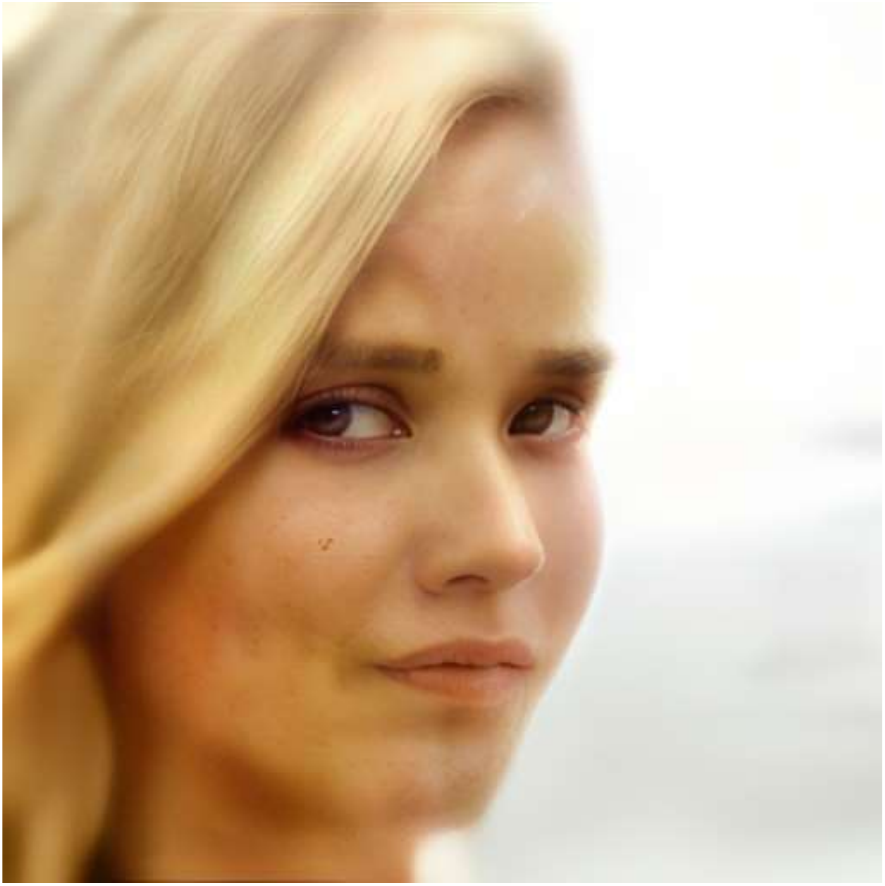} \hspace{-4.6mm} &
\includegraphics[width=0.1092\linewidth]{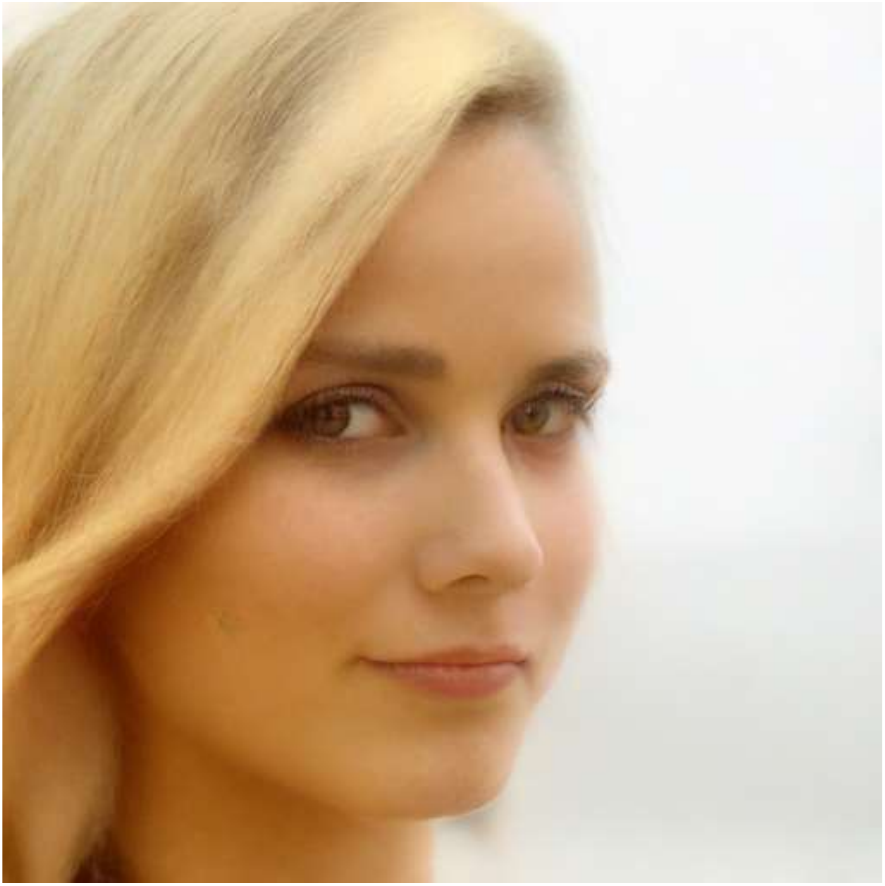} \hspace{-4.6mm} &
\includegraphics[width=0.1092\linewidth]{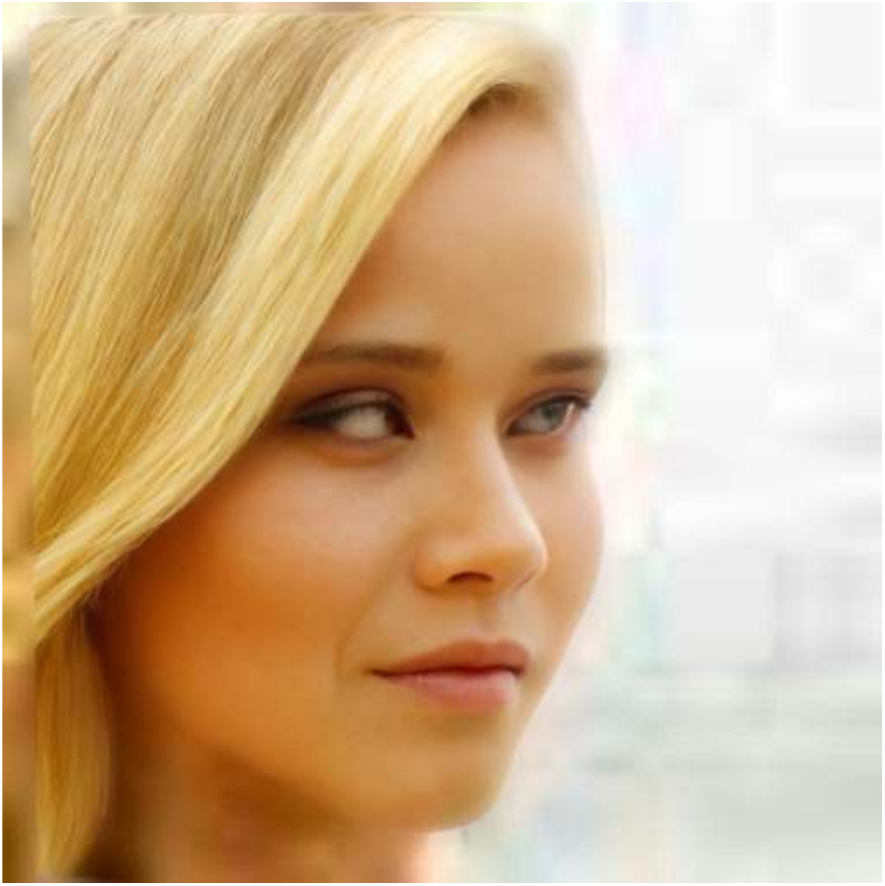} \hspace{-4.6mm} &
\includegraphics[width=0.1092\linewidth]{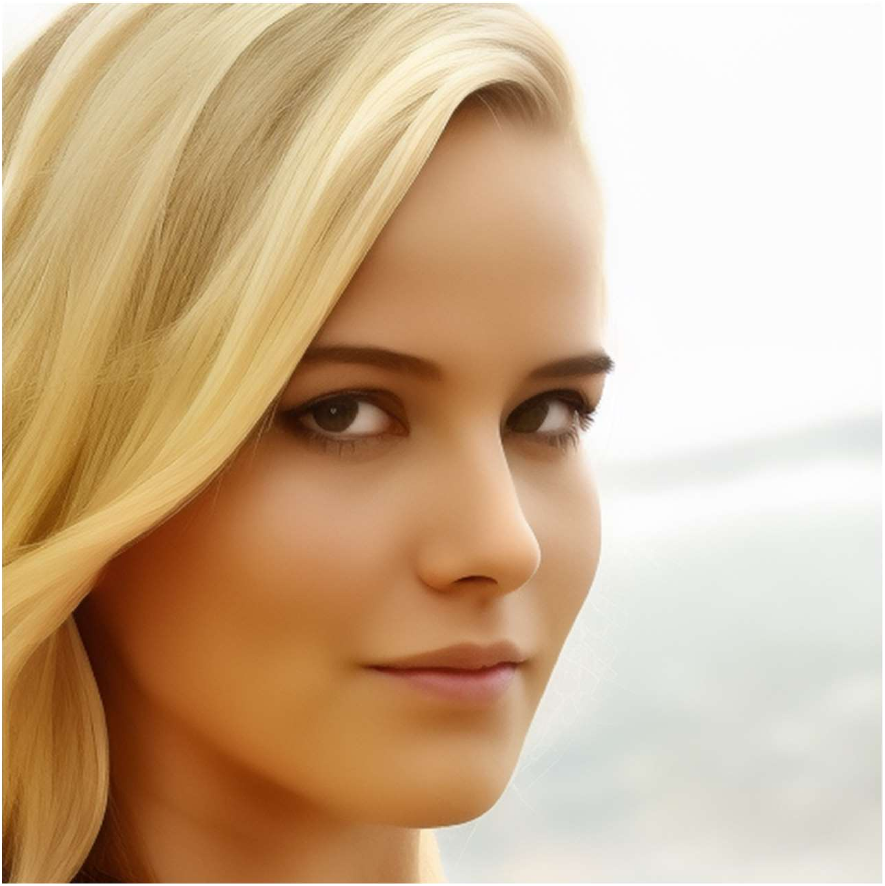} \hspace{-4.6mm} &
\includegraphics[width=0.1092\linewidth]{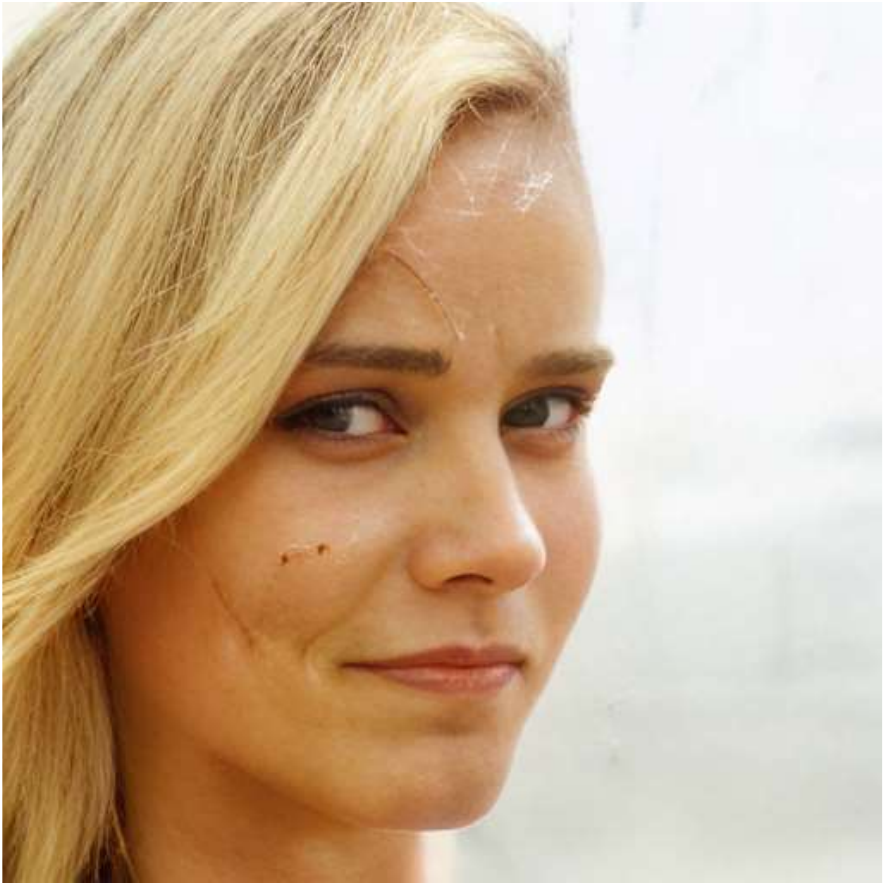} \hspace{-4.6mm} &
\includegraphics[width=0.1092\linewidth]{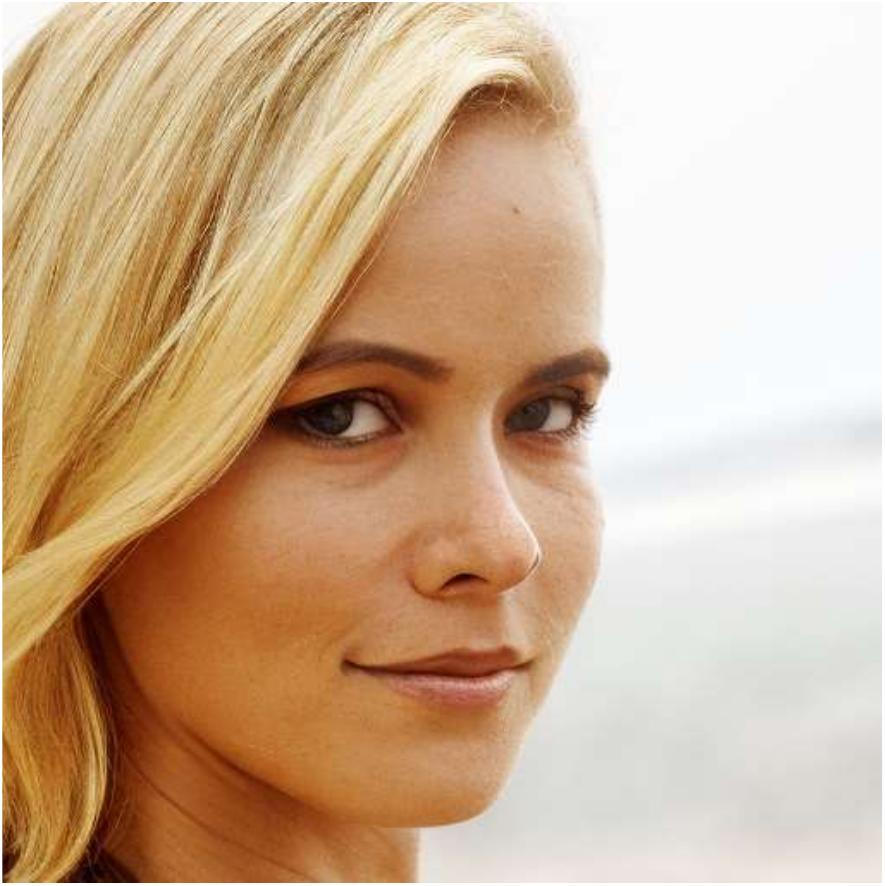} \hspace{-4.6mm} &
\includegraphics[width=0.1092\linewidth]
{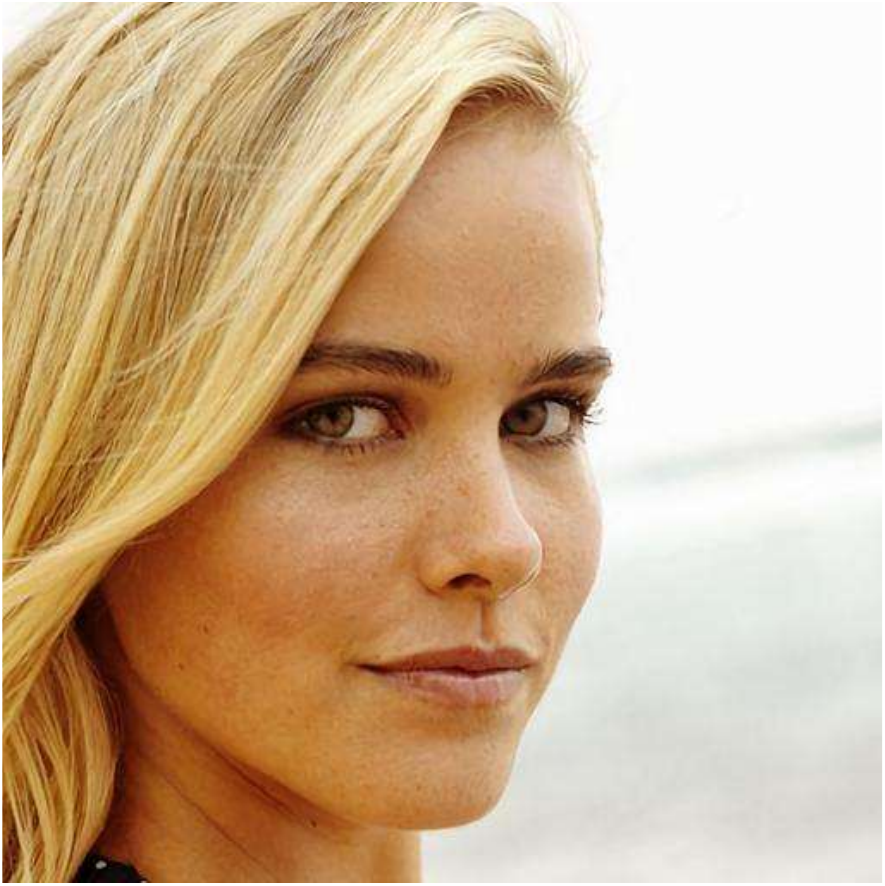}
\vspace{-0mm}
\\
LR \hspace{-4.6mm} &
PSFRGAN \hspace{-4.6mm} &
GPEN \hspace{-4.6mm} &
DR2 \hspace{-4.6mm} &
VQFR \hspace{-4.6mm} &
DiffBIR \hspace{-4.6mm} &
CodeFormer \hspace{-4.6mm} &
Ours w/o Ref. \hspace{-4.6mm} &
GT
\\
\end{tabular}
\end{adjustbox}
\end{tabular}
\vspace{-5.mm}
\caption{More qualitative comparisons for our text-guided baseline model on synthetic dataset under mild degradation in CelebA-Test dataset. Zoom in for best view.}
\label{fig:B_1}
\vspace{-2.mm}
\end{figure*}

\begin{figure*}[h]
\captionsetup{font={small}, skip=12pt}
\scriptsize
\begin{tabular}{ccc}
\hspace{-0.5cm}
\begin{adjustbox}{valign=t}
\begin{tabular}{c}
\end{tabular}
\end{adjustbox}
\begin{adjustbox}{valign=t}
\begin{tabular}{cccccccc}
\includegraphics[width=0.12\linewidth]{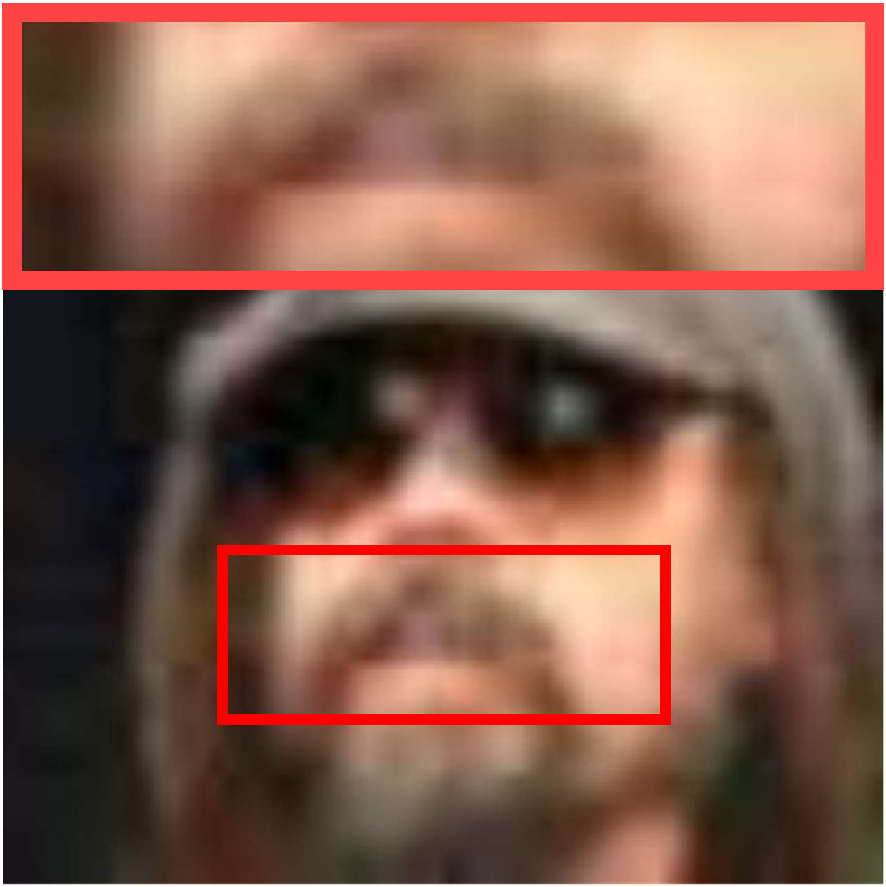} \hspace{-4mm} &
\includegraphics[width=0.12\linewidth]{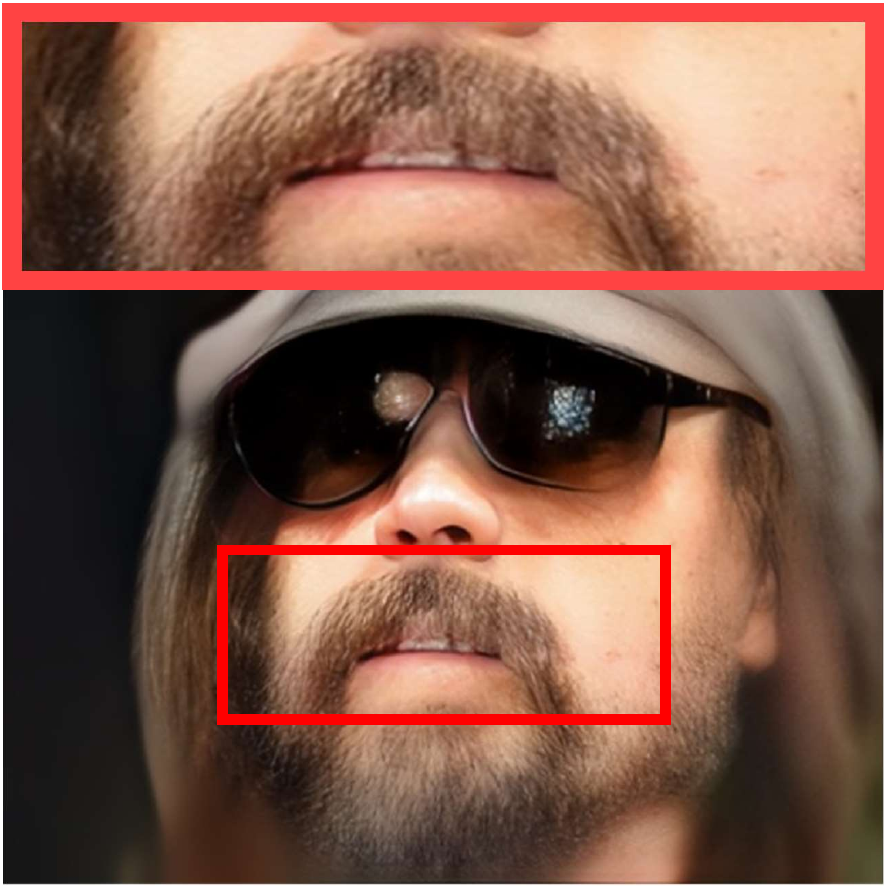}   \hspace{-4mm} &
\includegraphics[width=0.12\linewidth]{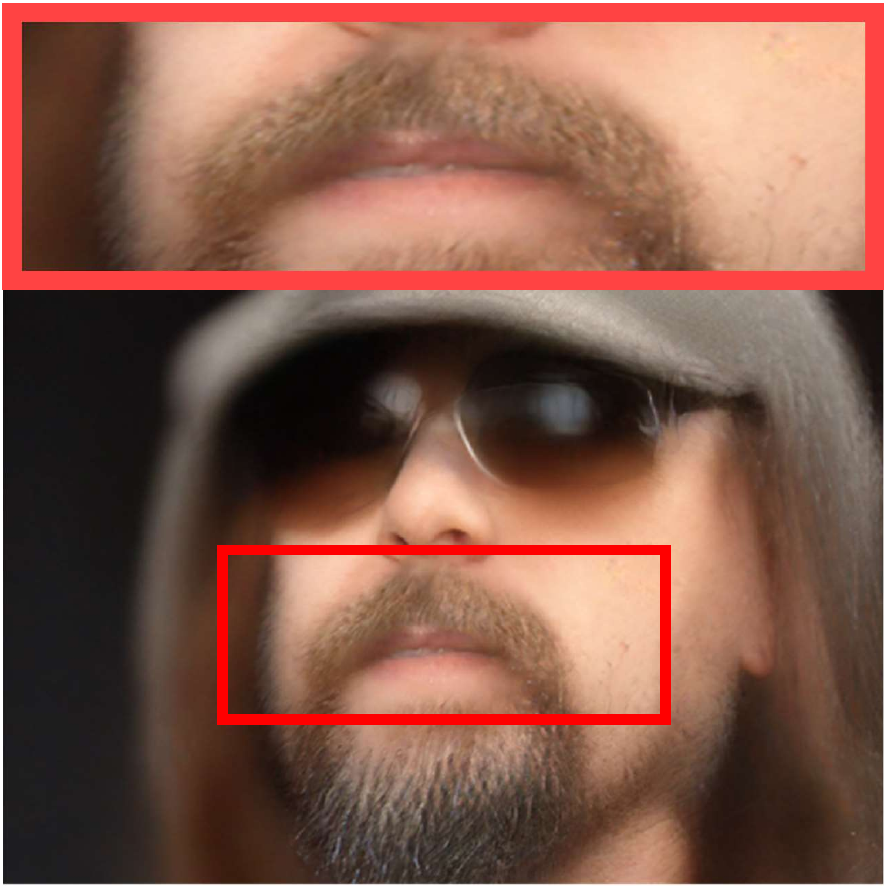}  \hspace{-4mm} &
\includegraphics[width=0.12\linewidth]{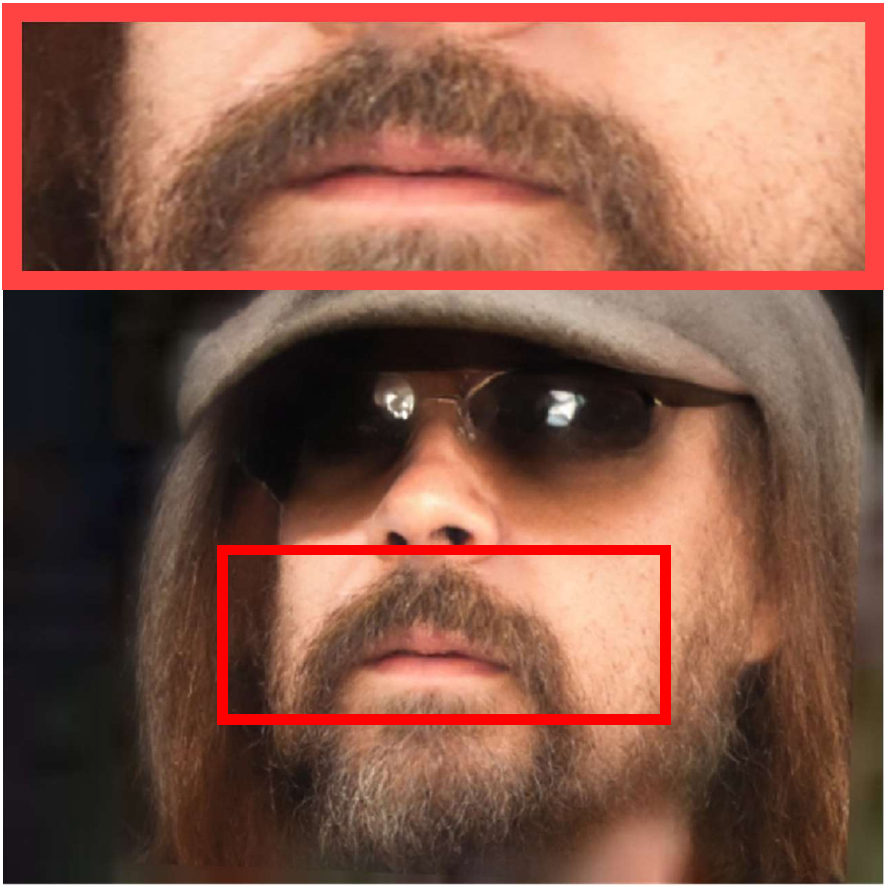}  \hspace{-4mm} &
\includegraphics[width=0.12\linewidth]{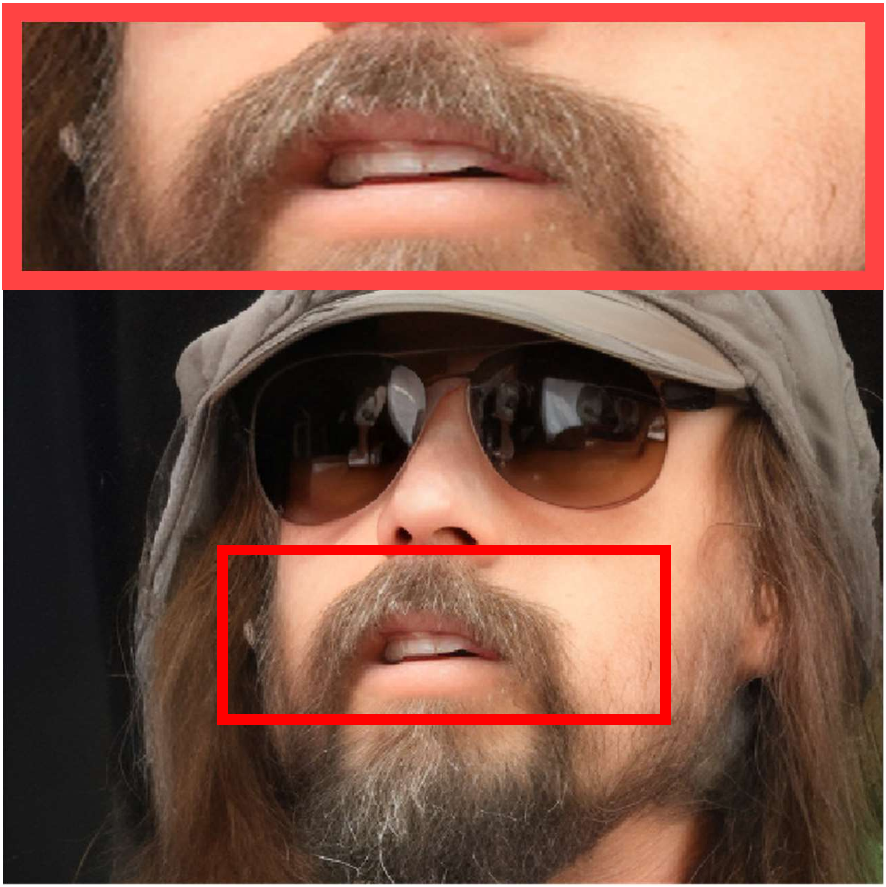}  \hspace{-4mm} &
\includegraphics[width=0.12\linewidth]{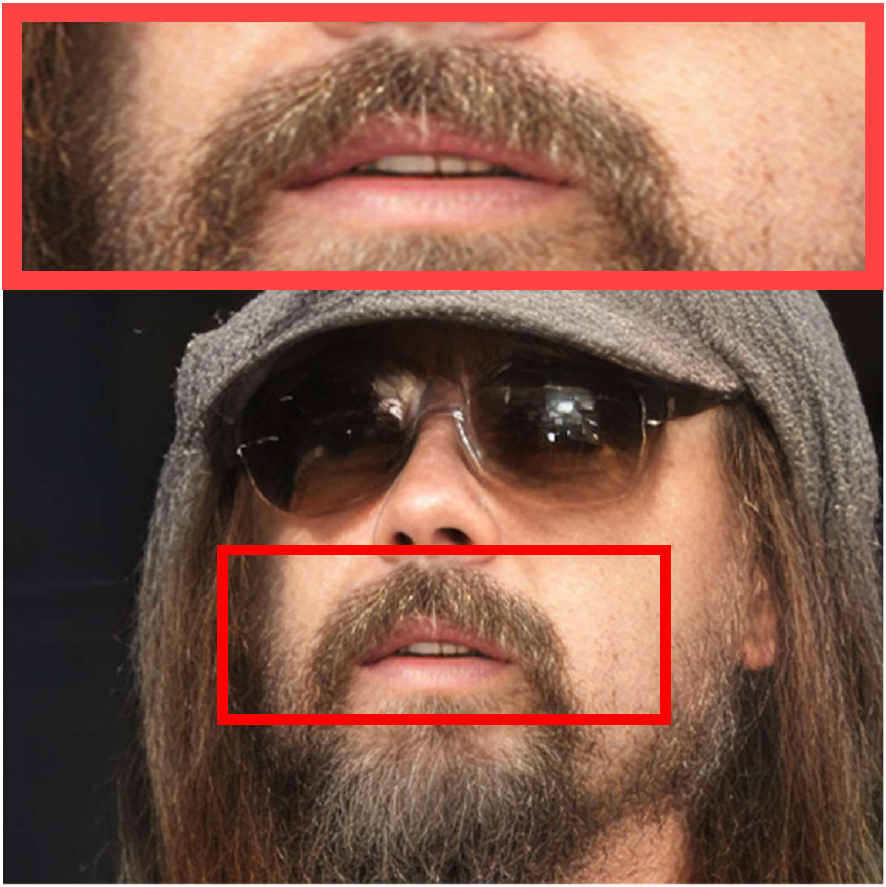}  \hspace{-4mm} &
\includegraphics[width=0.12\linewidth]{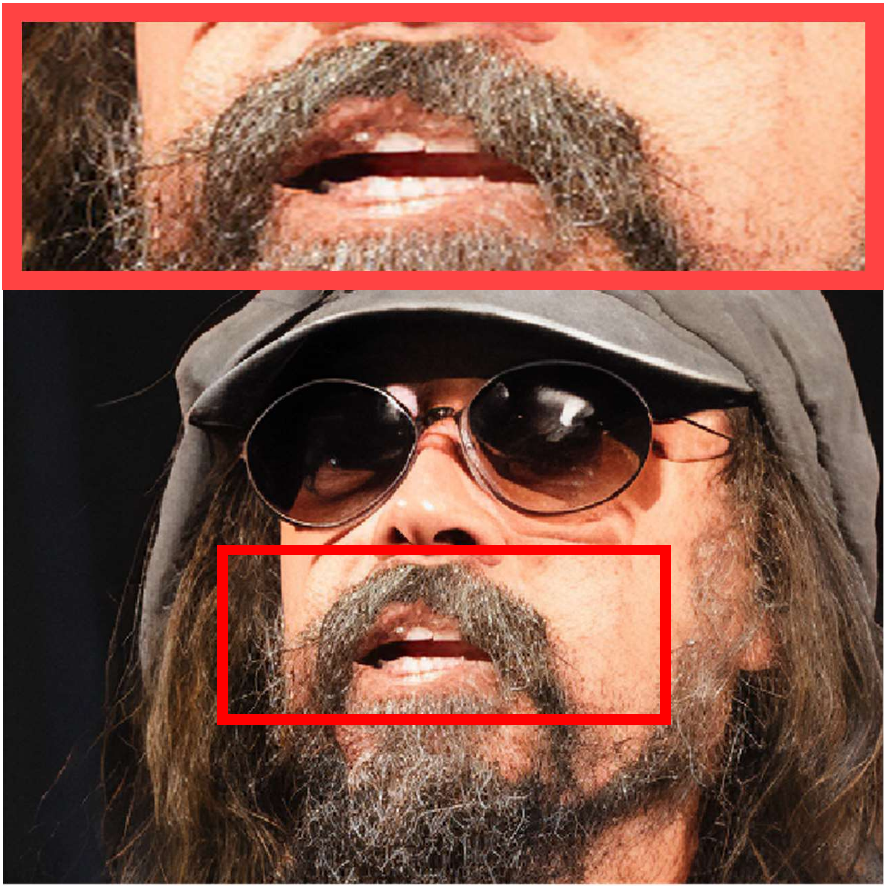}  \hspace{-4mm} &
\includegraphics[width=0.12\linewidth]{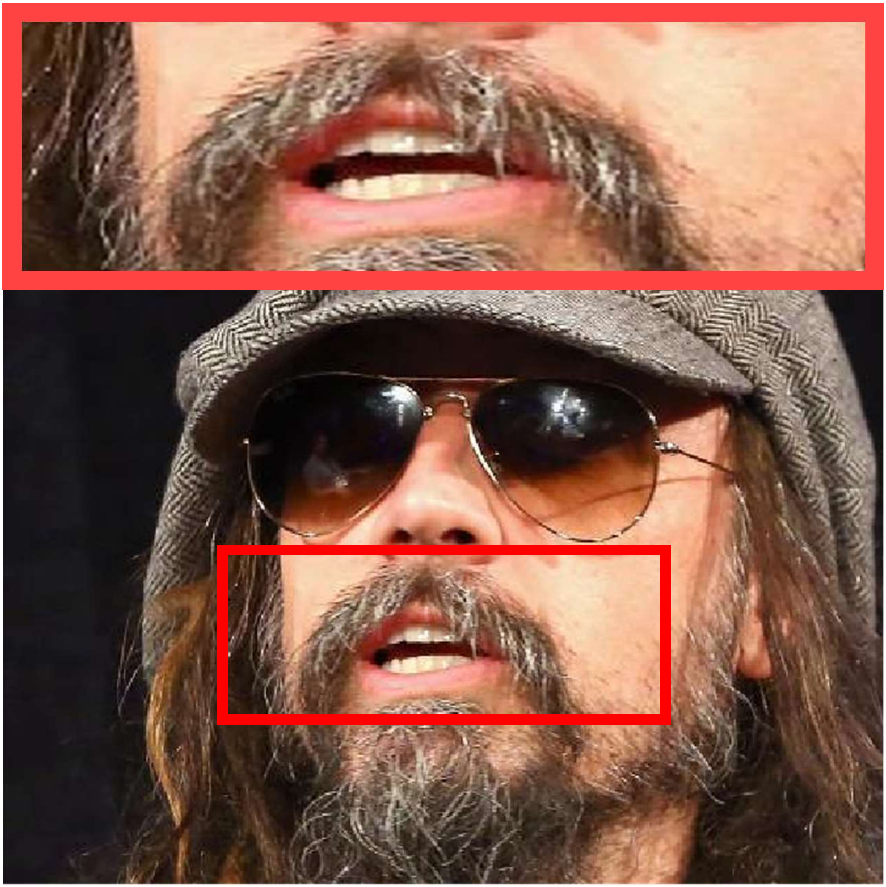}   
\end{tabular}
\end{adjustbox}
\vspace{0.1mm}
\\
\hspace{-0.55cm}
\begin{adjustbox}{valign=t}
\begin{tabular}{c}
\end{tabular}
\end{adjustbox}
\begin{adjustbox}{valign=t}
\begin{tabular}{cccccccc}
\includegraphics[width=0.12\linewidth]{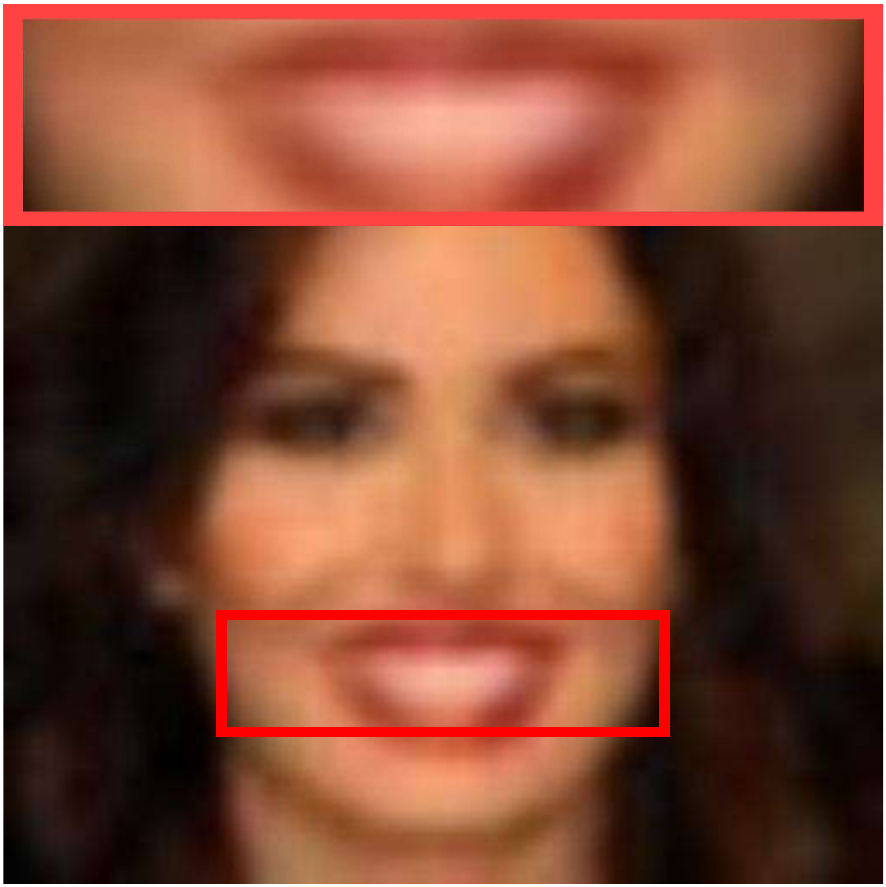} \hspace{-4mm} &
\includegraphics[width=0.12\linewidth]{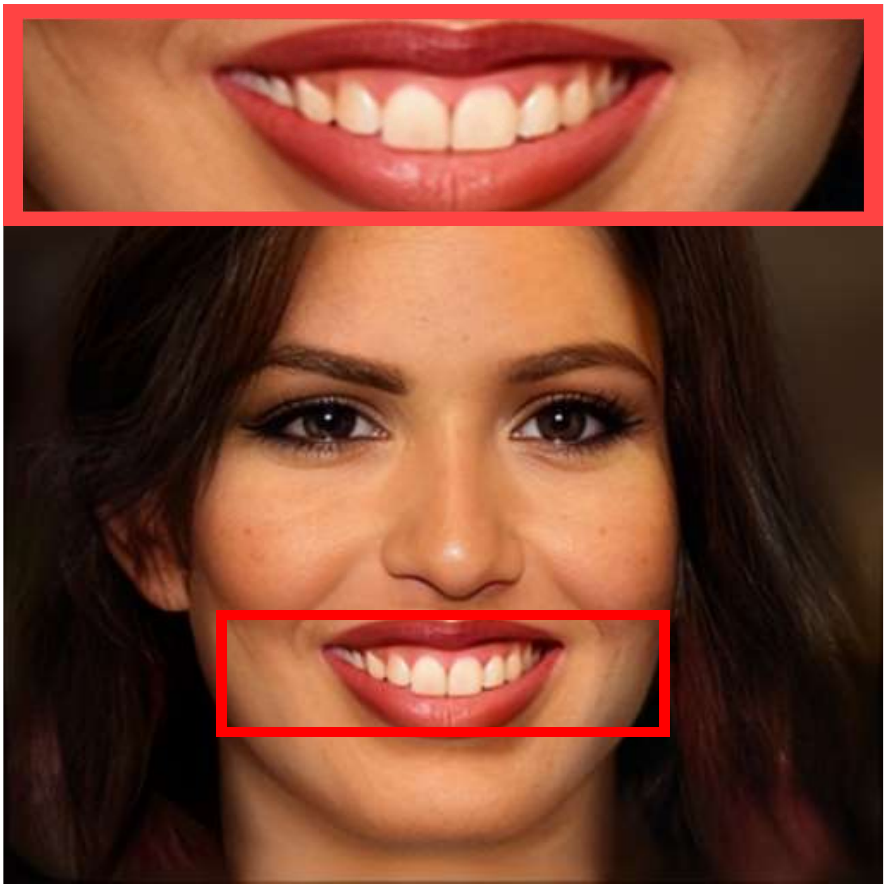}   \hspace{-4mm} &
\includegraphics[width=0.12\linewidth]{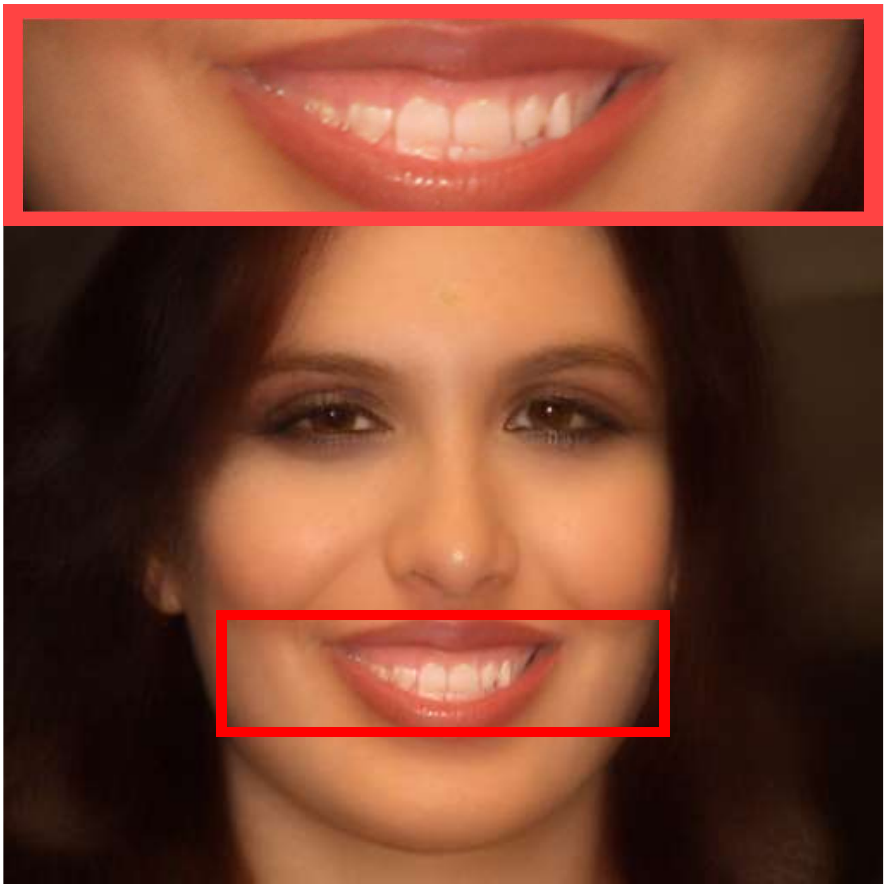}  \hspace{-4mm} &
\includegraphics[width=0.12\linewidth]{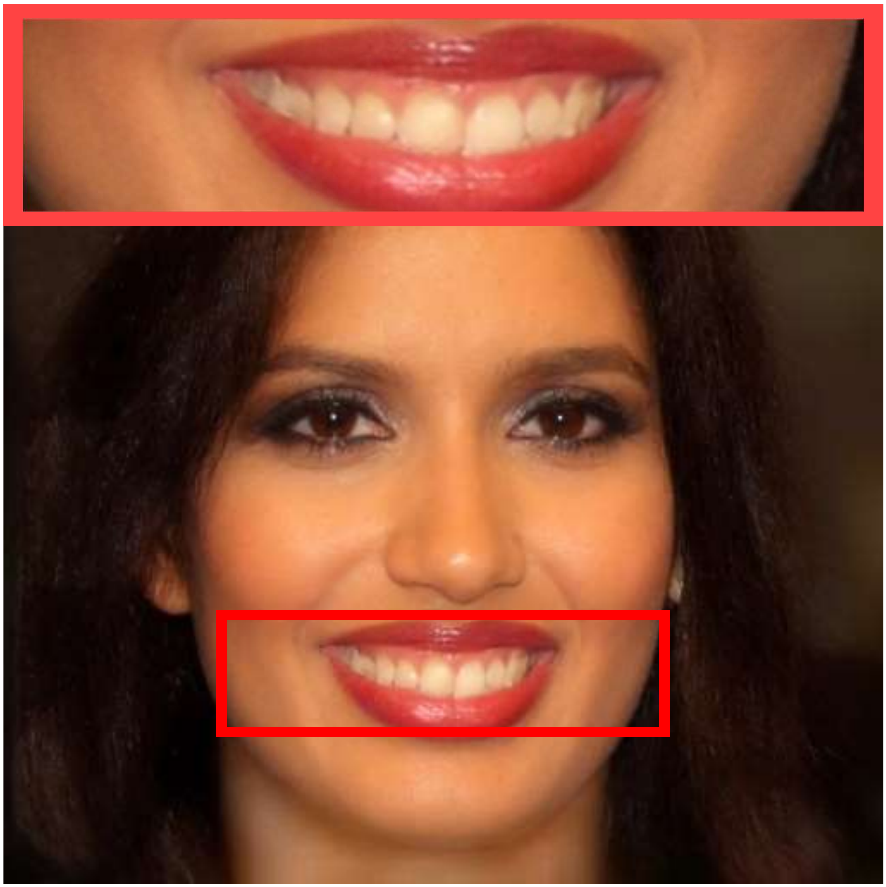}  \hspace{-4mm} &
\includegraphics[width=0.12\linewidth]{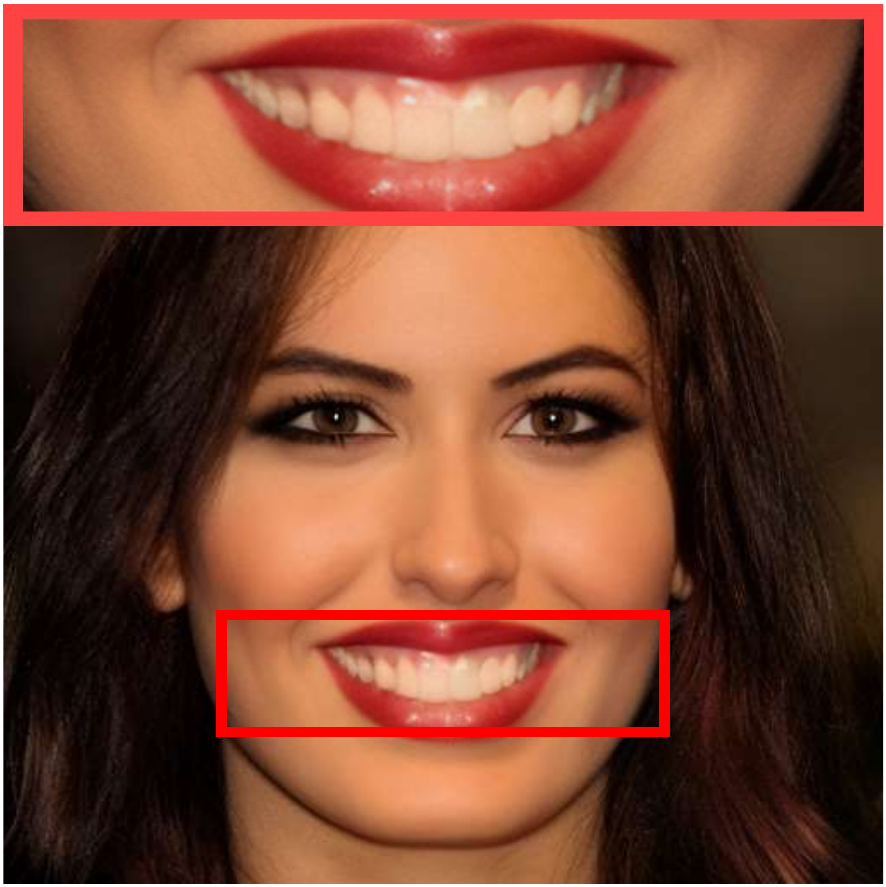}  \hspace{-4mm} &
\includegraphics[width=0.12\linewidth]{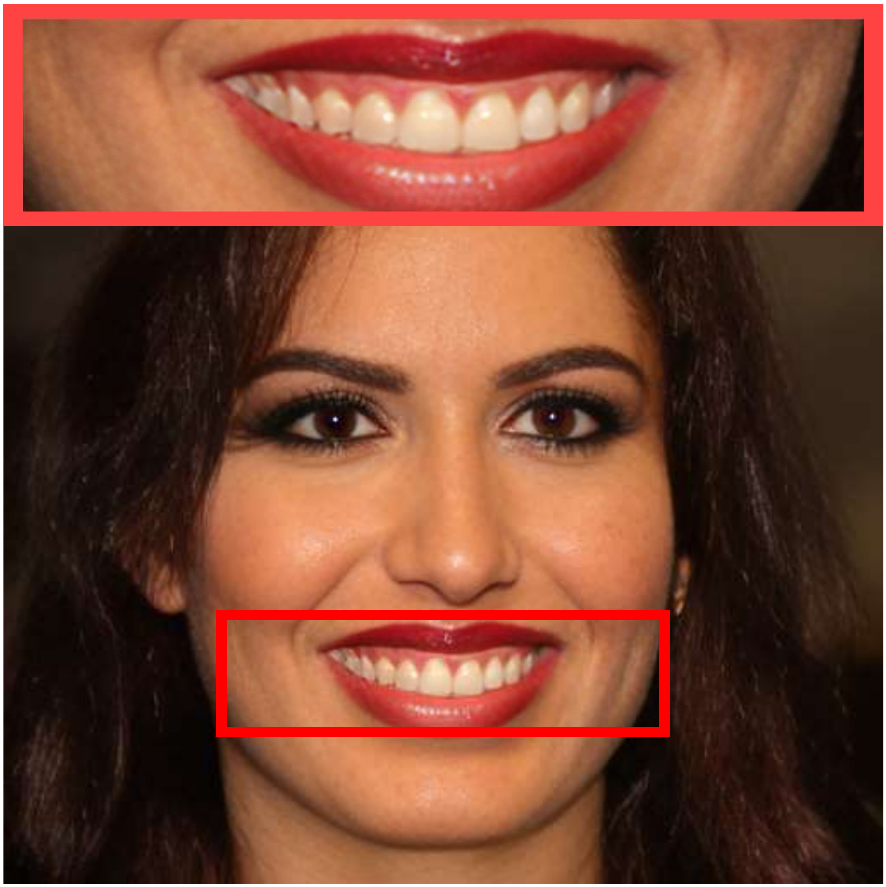}  \hspace{-4mm} &
\includegraphics[width=0.12\linewidth]{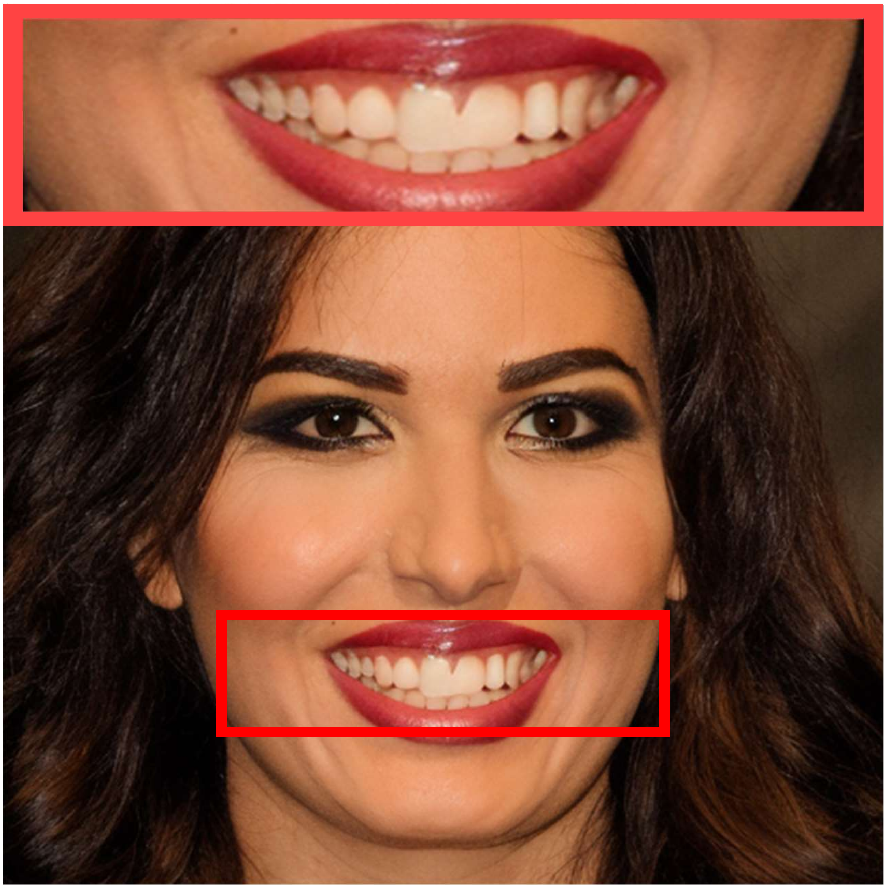}  \hspace{-4mm} &
\includegraphics[width=0.12\linewidth]{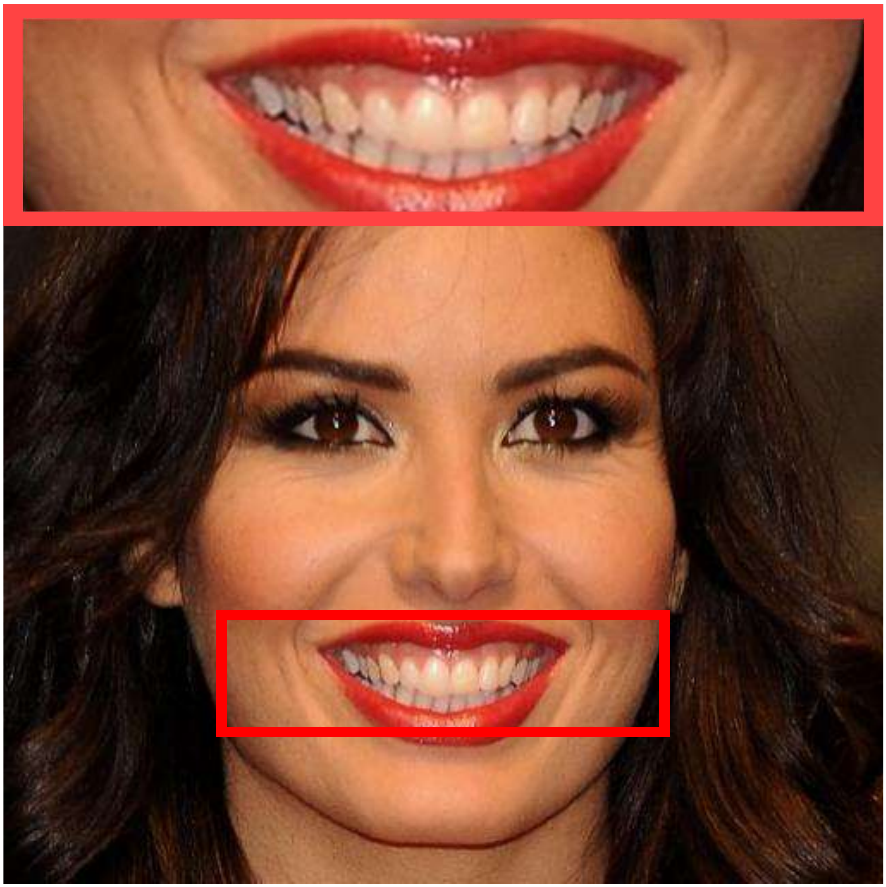}   
\\
LR \hspace{-4mm} &
GPEN \hspace{-4mm} &
DR2 \hspace{-4mm} &
VQFR \hspace{-4mm} &
DiffBIR \hspace{-4mm} &
CodeFormer \hspace{-4mm} &
Ours w/o Ref. \hspace{-4mm} &
GT
\\
\end{tabular}
\end{adjustbox}
\end{tabular}
\vspace{-5.mm}
\caption{More qualitative comparisons for our text-guided baseline model on synthetic dataset under moderate degradation in CelebA-Test dataset. Zoom in for best view.}
\label{fig:B_2}
\vspace{-2.mm}
\end{figure*}

\begin{figure*}[h]
\captionsetup{font={small}, skip=12pt}
\scriptsize
\begin{tabular}{ccc}
\hspace{-0.5cm}
\begin{adjustbox}{valign=t}
\begin{tabular}{c}
\end{tabular}
\end{adjustbox}
\begin{adjustbox}{valign=t}
\begin{tabular}{cccccccc}
\includegraphics[width=0.12\linewidth]{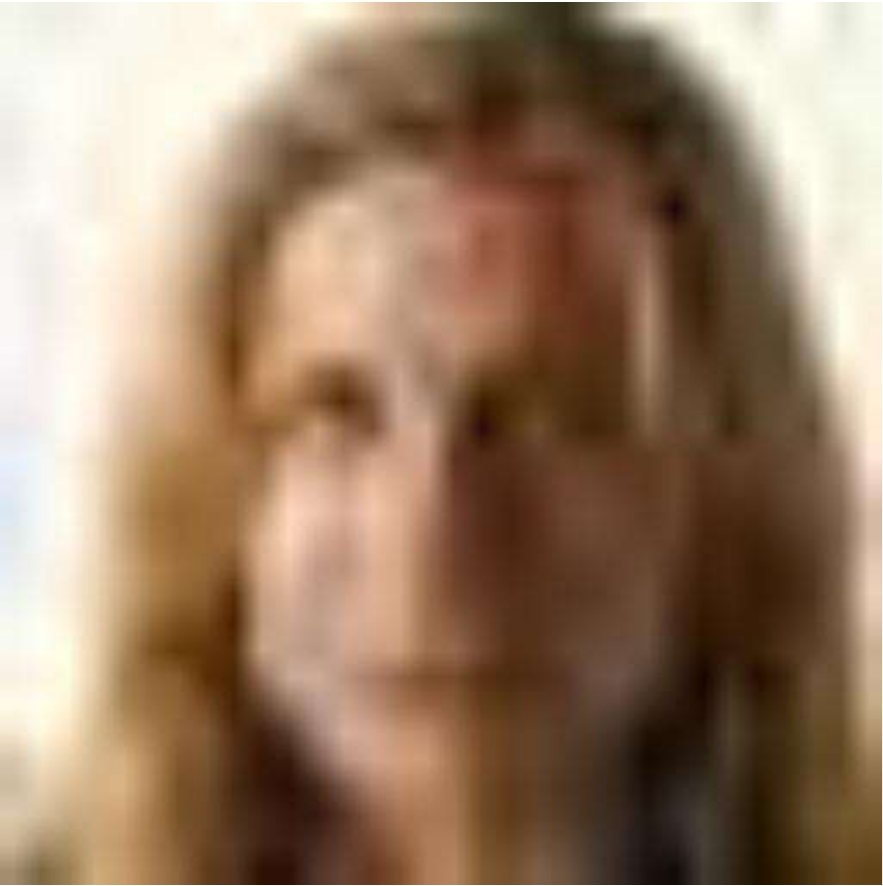} \hspace{-4mm} &
\includegraphics[width=0.12\linewidth]{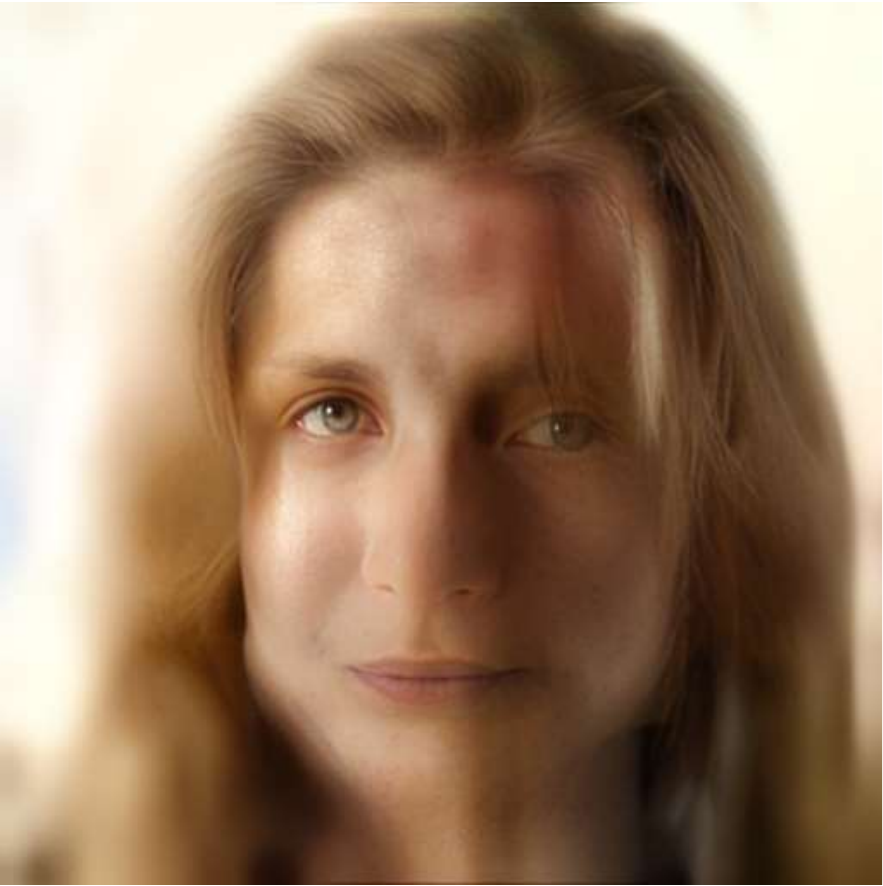}  \hspace{-4mm} &
\includegraphics[width=0.12\linewidth]{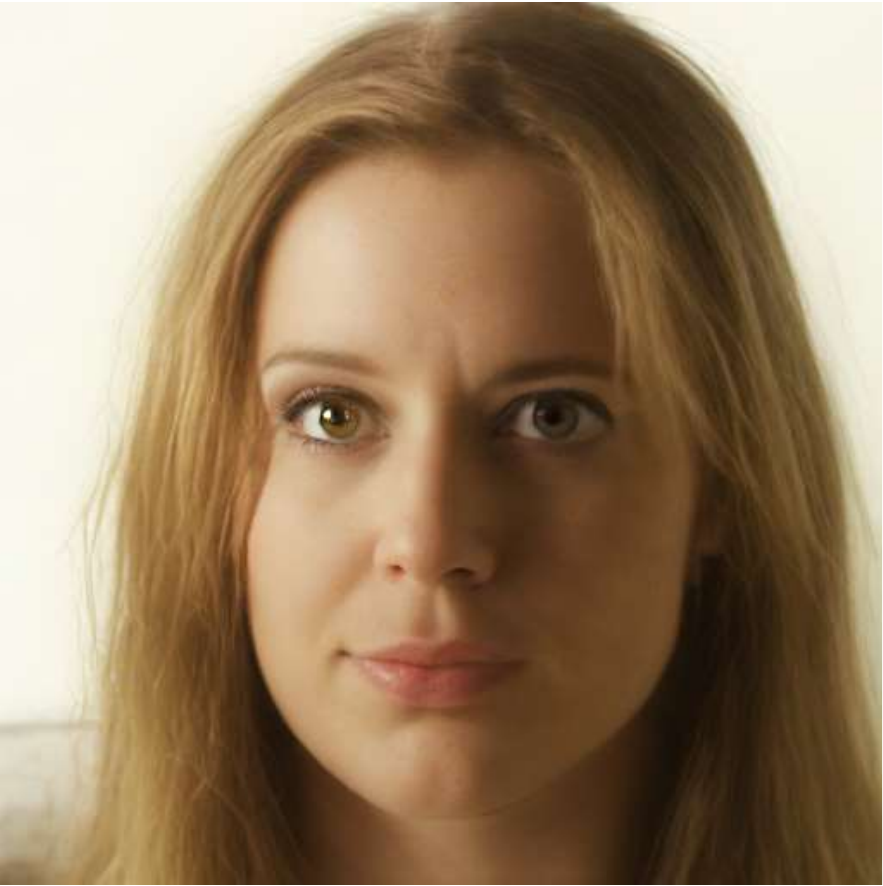}  \hspace{-4mm} &
\includegraphics[width=0.12\linewidth]{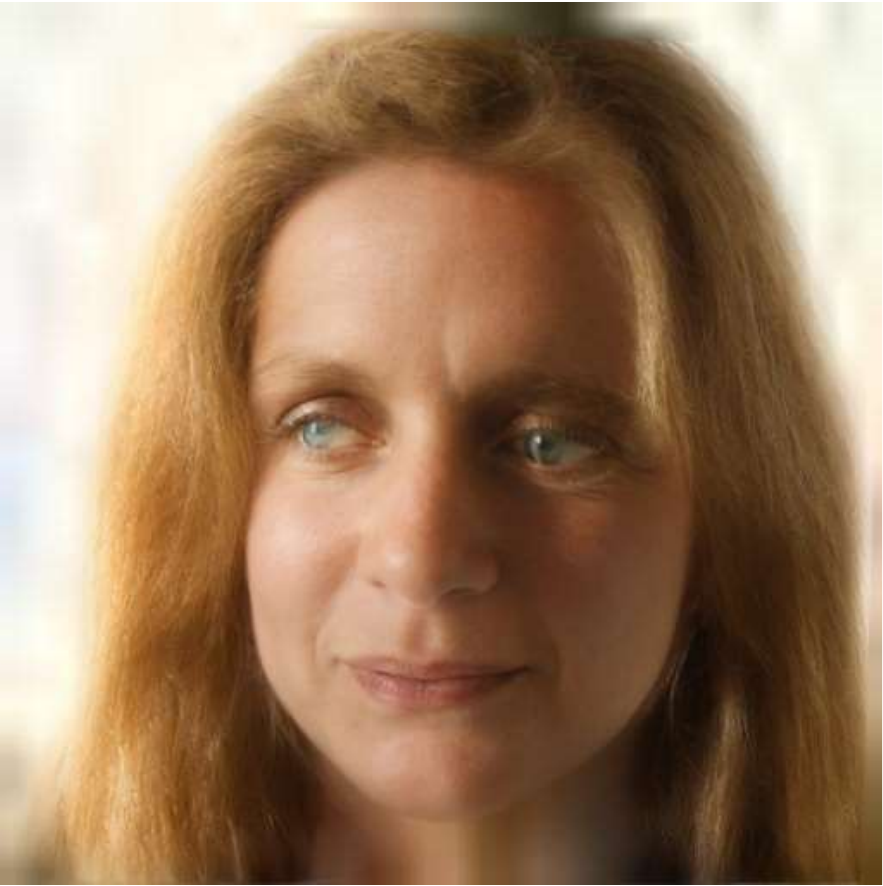} \hspace{-4mm} &
\includegraphics[width=0.12\linewidth]{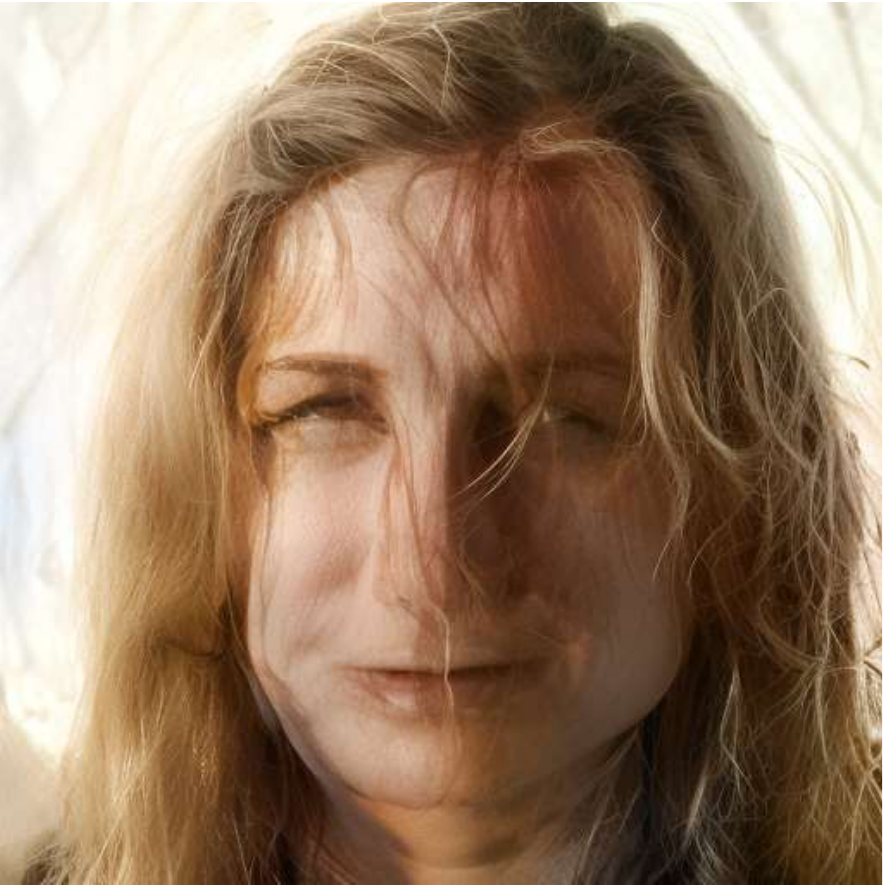} \hspace{-4mm} &
\includegraphics[width=0.12\linewidth]{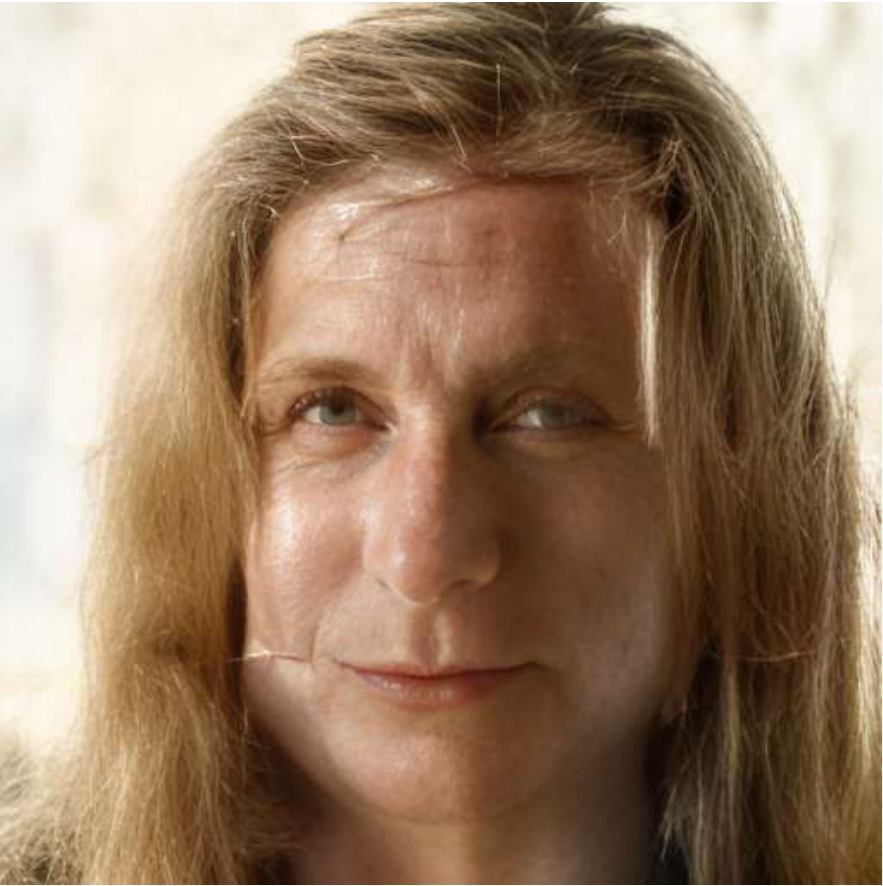}
\hspace{-4mm} &
\includegraphics[width=0.12\linewidth]{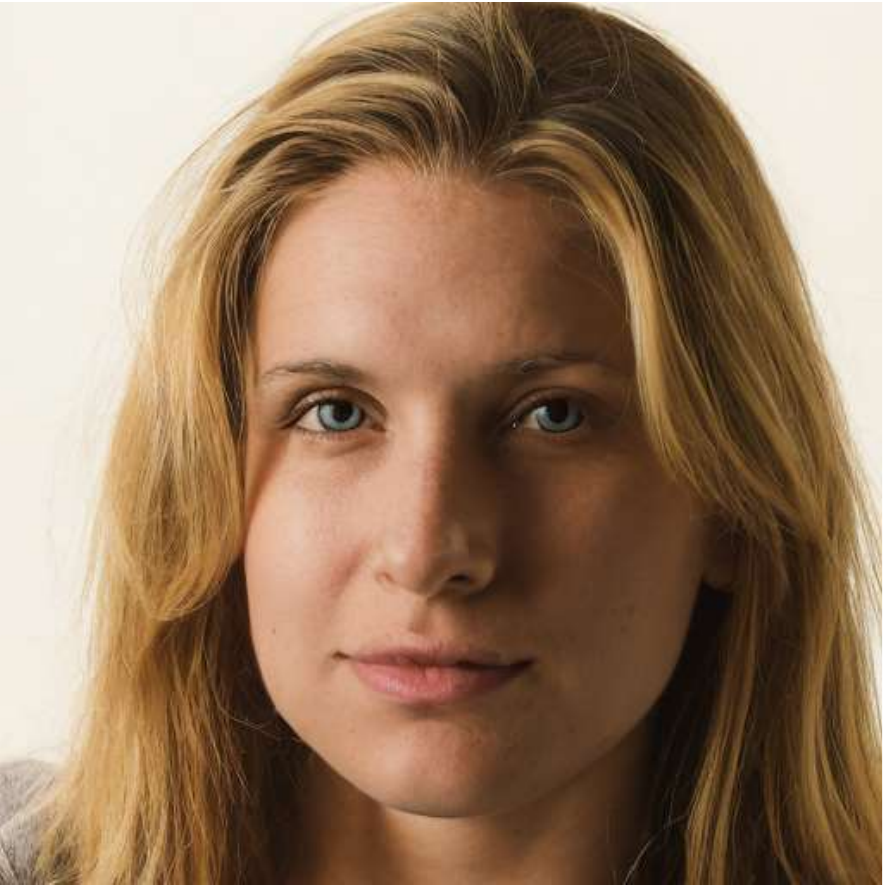} \hspace{-4mm} &
\includegraphics[width=0.12\linewidth]{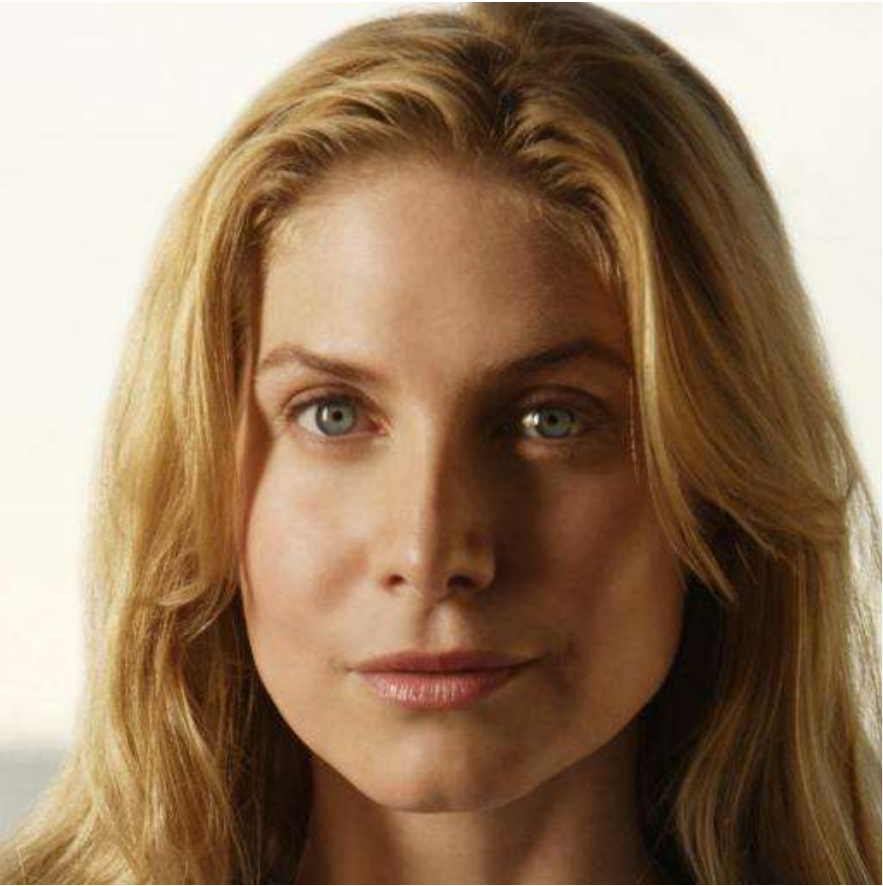}   
\end{tabular}
\end{adjustbox}
\vspace{0.1mm}
\\
\hspace{-0.55cm}
\begin{adjustbox}{valign=t}
\begin{tabular}{c}
\end{tabular}
\end{adjustbox}
\begin{adjustbox}{valign=t}
\begin{tabular}{cccccccc}
\includegraphics[width=0.12\linewidth]{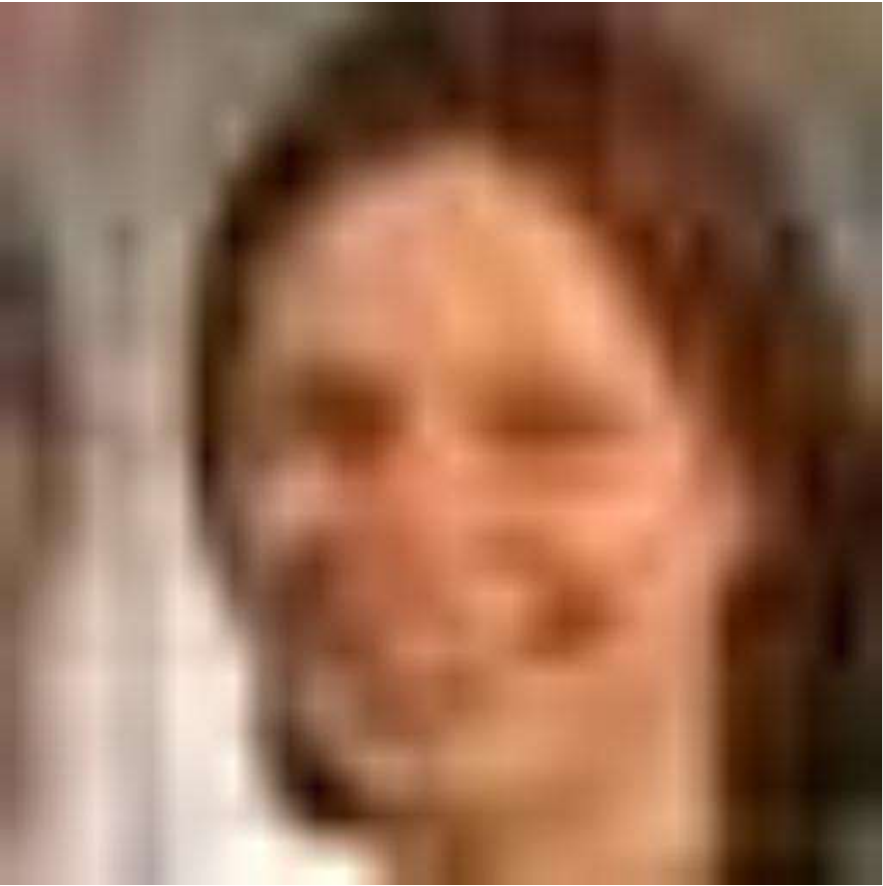} \hspace{-4mm} &
\includegraphics[width=0.12\linewidth]{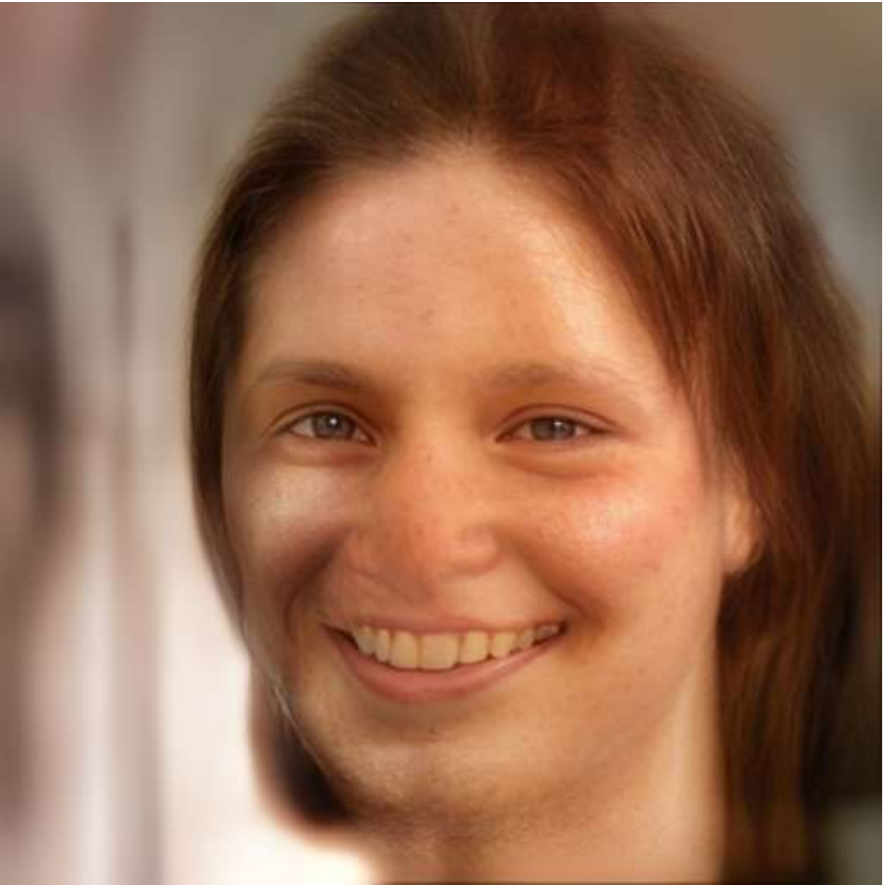}  \hspace{-4mm} &
\includegraphics[width=0.12\linewidth]{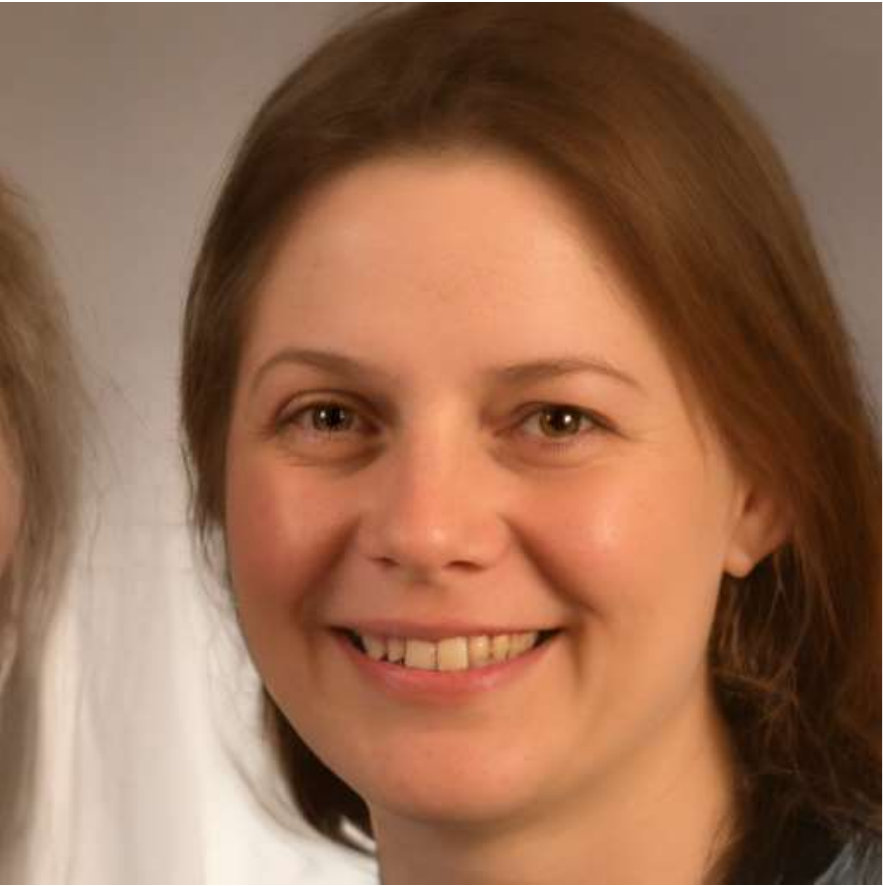}  \hspace{-4mm} &
\includegraphics[width=0.12\linewidth]{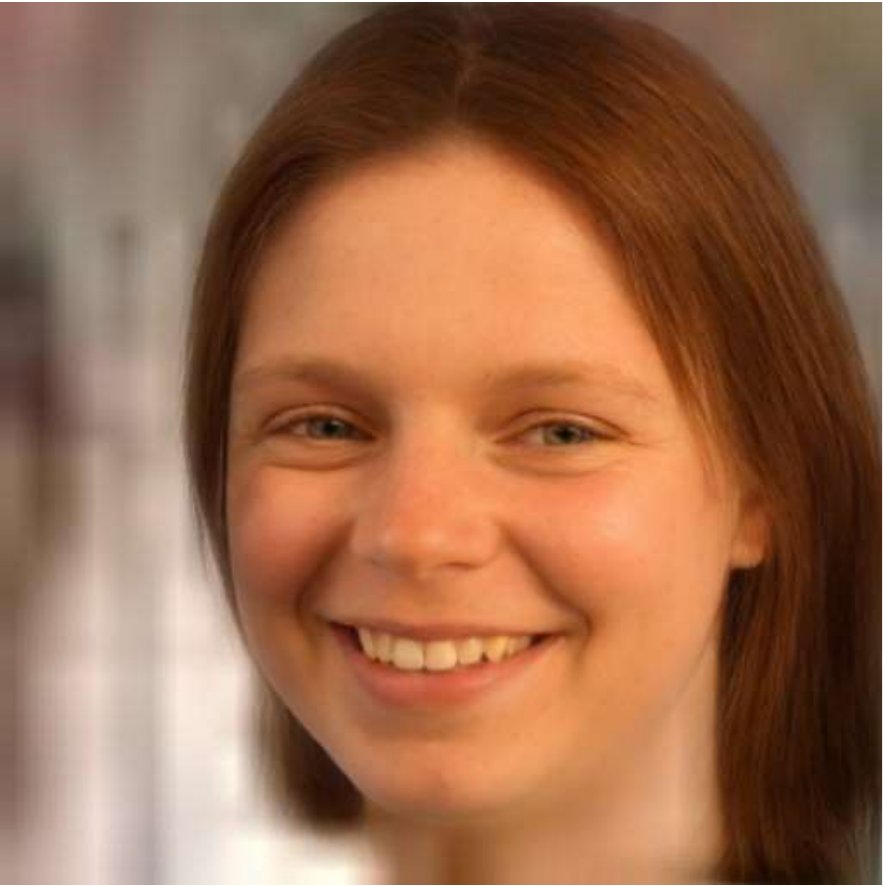} \hspace{-4mm} &
\includegraphics[width=0.12\linewidth]{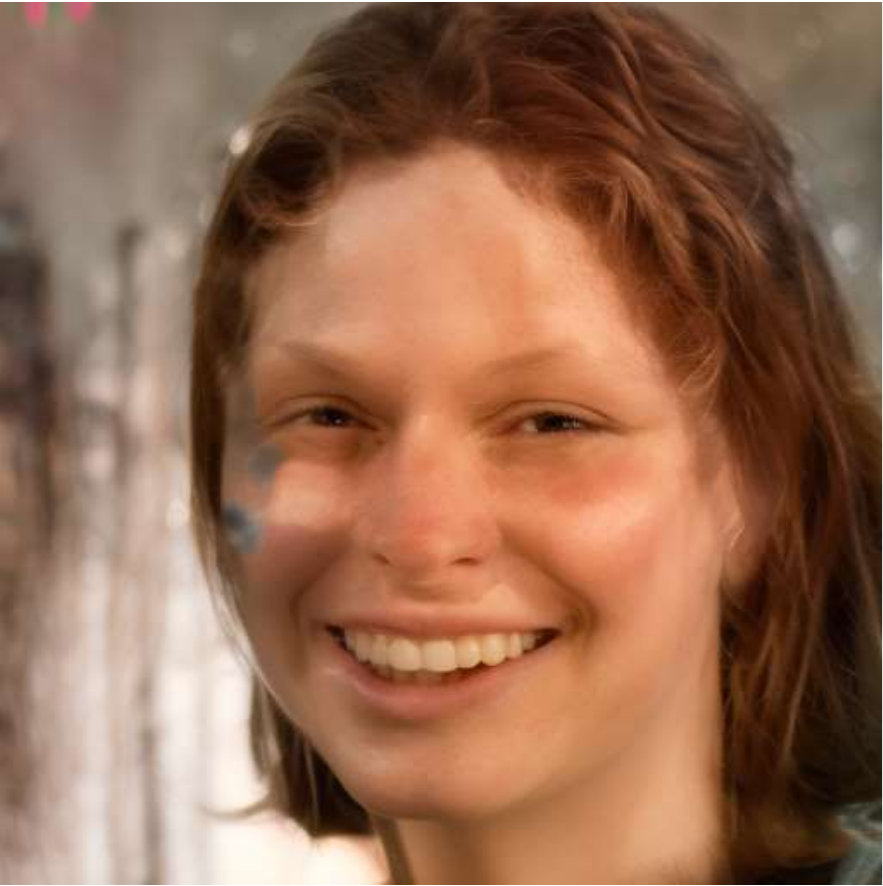} \hspace{-4mm} &
\includegraphics[width=0.12\linewidth]{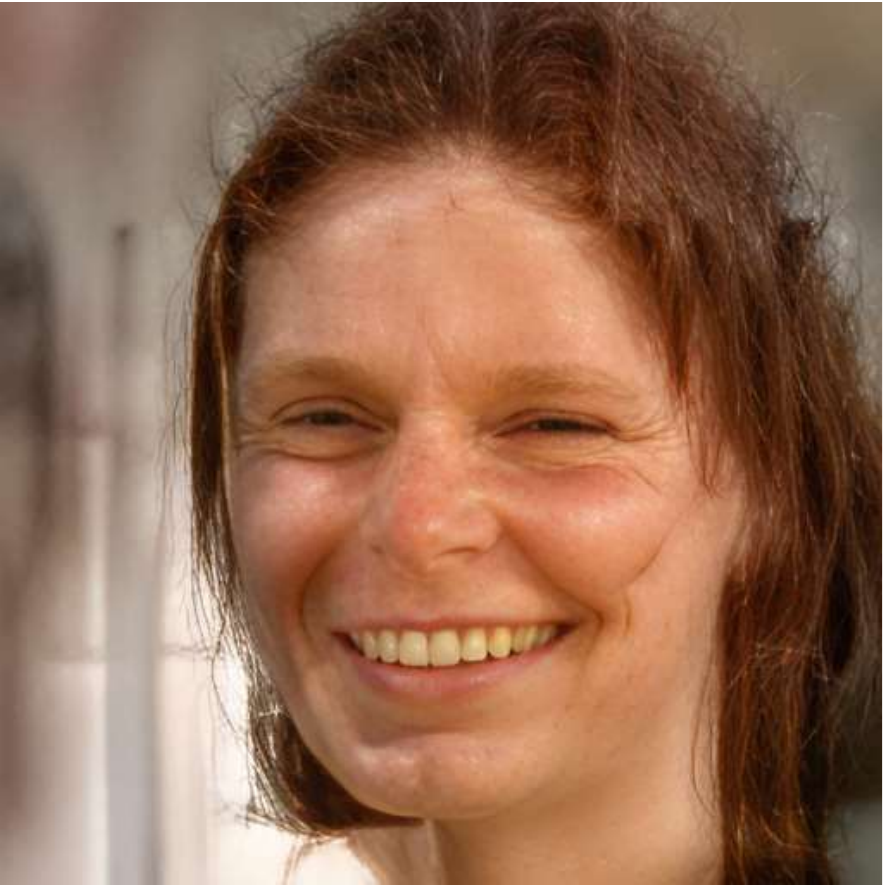}
\hspace{-4mm} &
\includegraphics[width=0.12\linewidth]{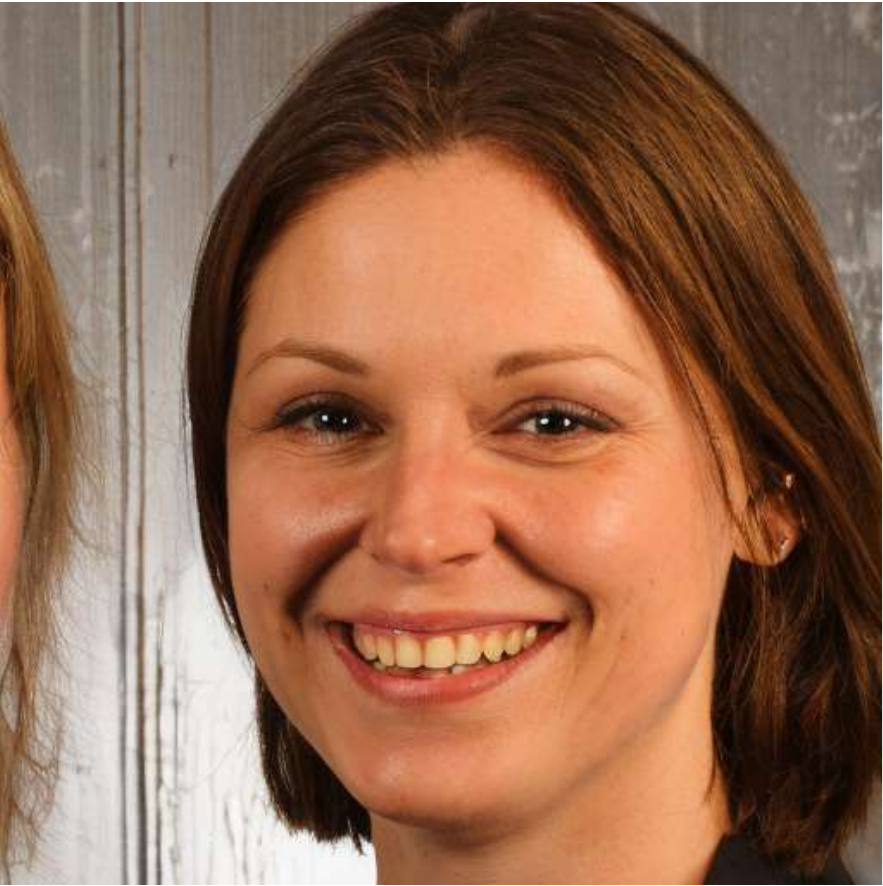} \hspace{-4mm} &
\includegraphics[width=0.12\linewidth]{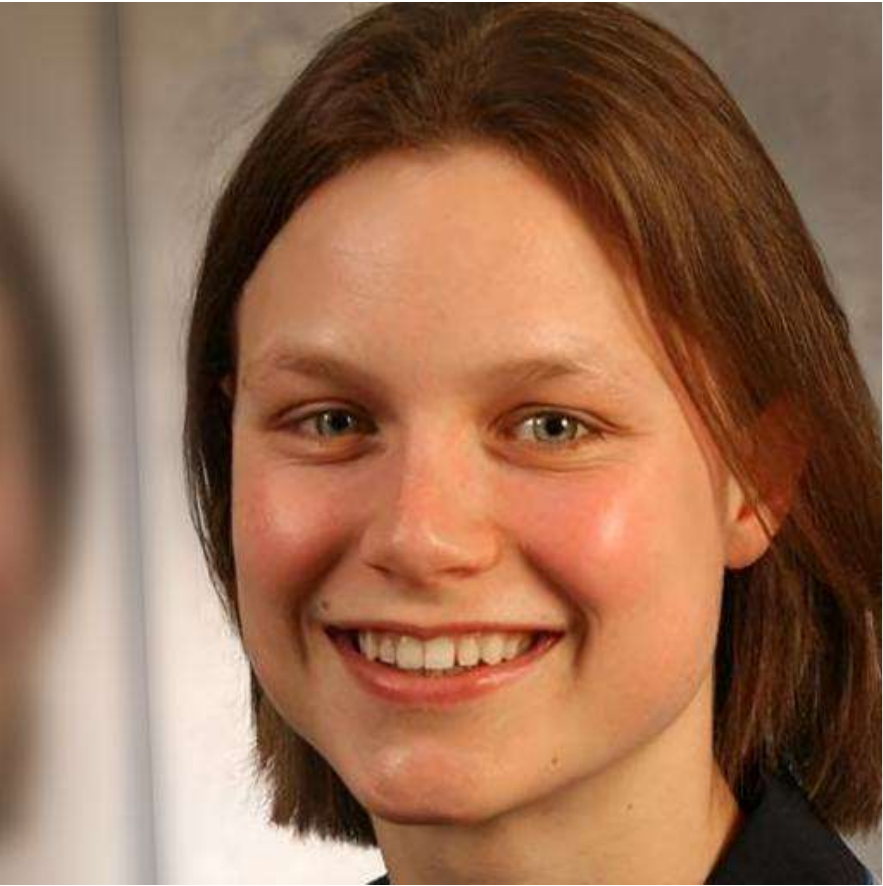}    
\\
LR \hspace{-4mm} &
GPEN \hspace{-4mm} &
DR2 \hspace{-4mm} &
VQFR \hspace{-4mm} &
DiffBIR \hspace{-4mm} &
CodeFormer \hspace{-4mm} &
Ours w/o Ref. \hspace{-4mm} &
GT
\\
\end{tabular}
\end{adjustbox}
\end{tabular}
\vspace{-5.mm}
\caption{More qualitative comparisons for our text-guided baseline model on synthetic dataset under severe degradation in CelebA-Test dataset. Zoom in for best view.}
\label{fig:B_3}
\vspace{-2.mm}
\end{figure*}

\begin{figure*}[t]
\captionsetup{font={small}, skip=12pt}
\scriptsize
\begin{tabular}{ccc}
\hspace{-0.5cm}
\begin{adjustbox}{valign=t}
\begin{tabular}{c}
\end{tabular}
\end{adjustbox}
\begin{adjustbox}{valign=t}
\begin{tabular}{cccccccc}
\includegraphics[width=0.12\linewidth]{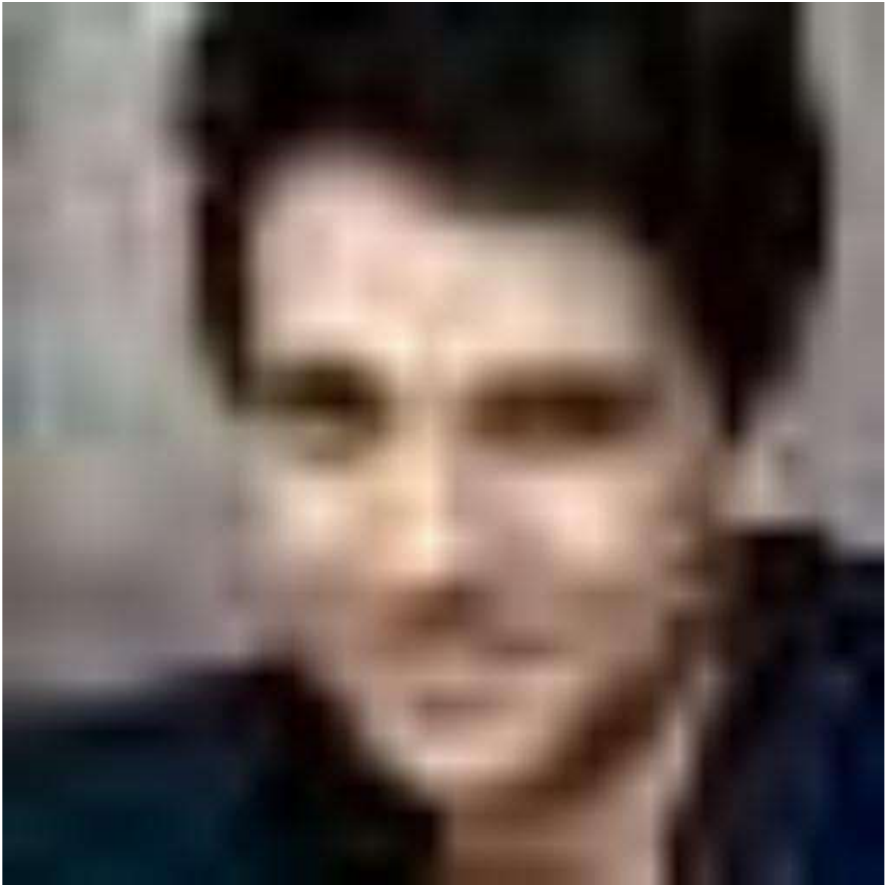} \hspace{-4mm} &
\includegraphics[width=0.12\linewidth]{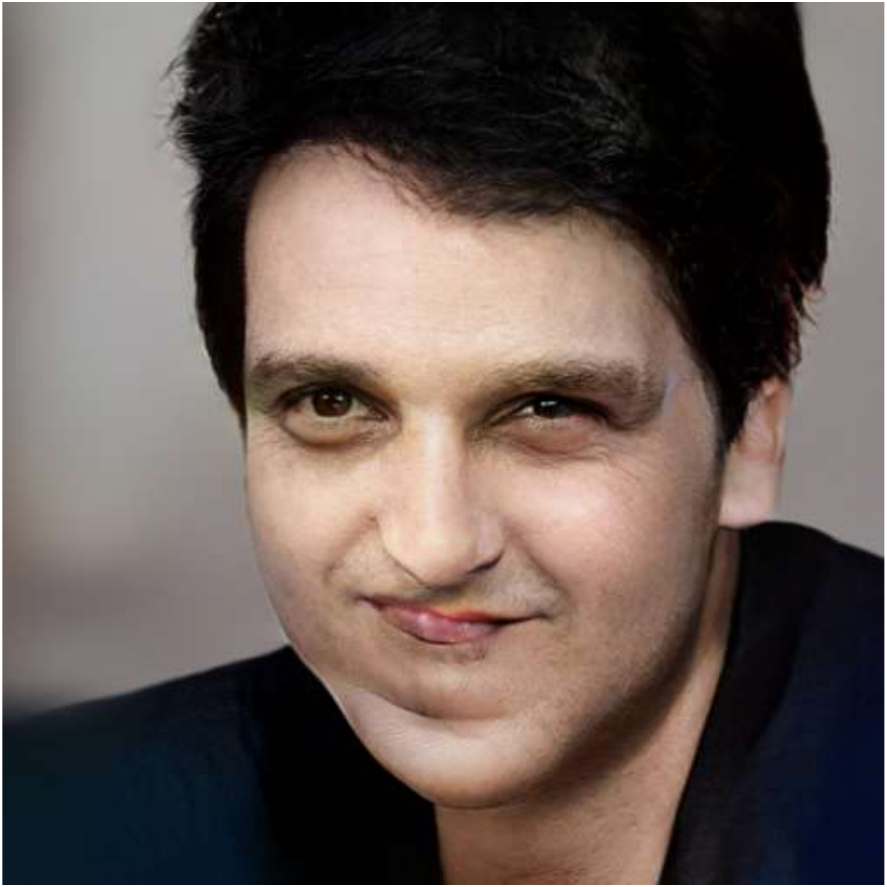}   \hspace{-4mm} &
\includegraphics[width=0.12\linewidth]{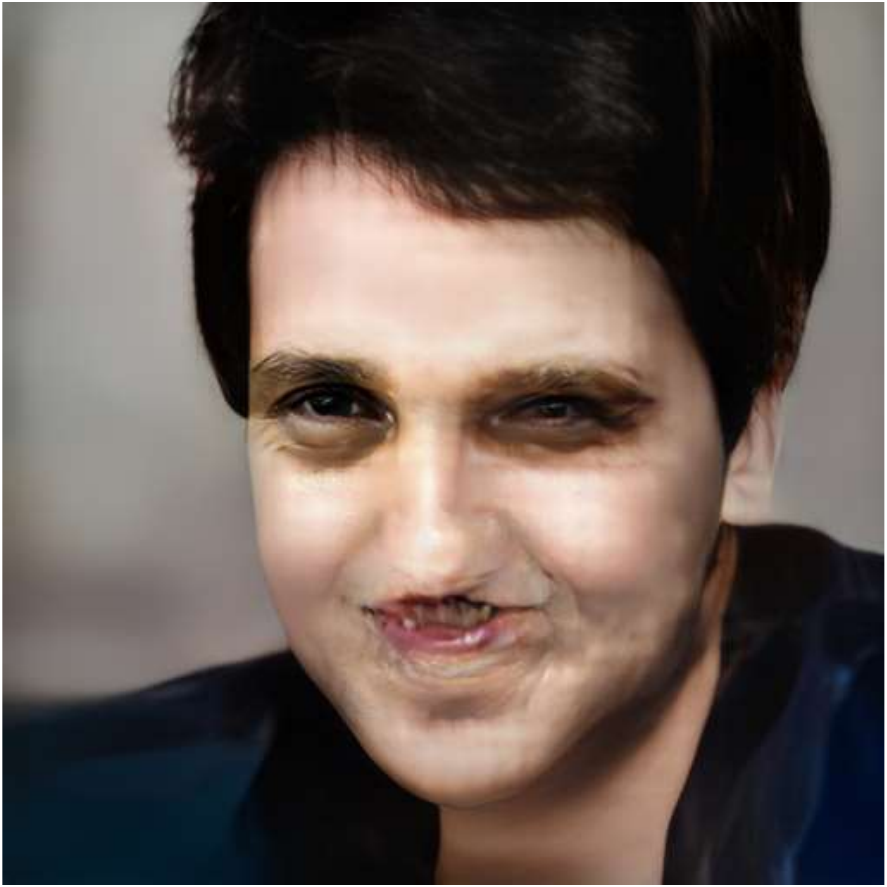}  \hspace{-4mm} &
\includegraphics[width=0.12\linewidth]{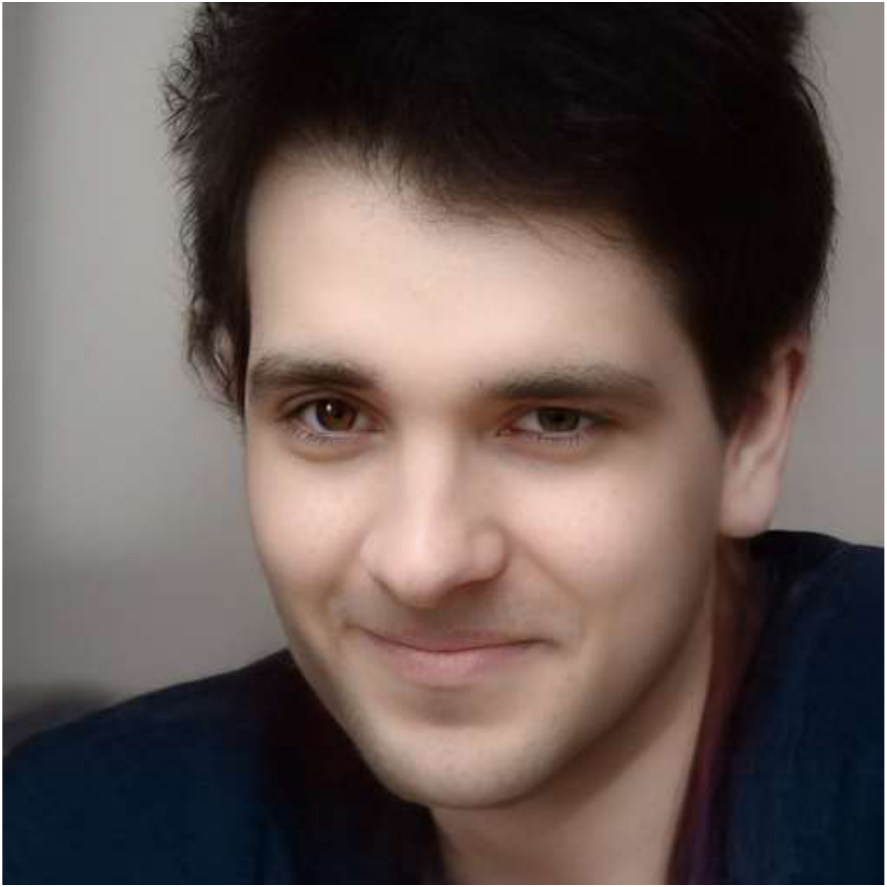}  \hspace{-4mm} &
\includegraphics[width=0.12\linewidth]{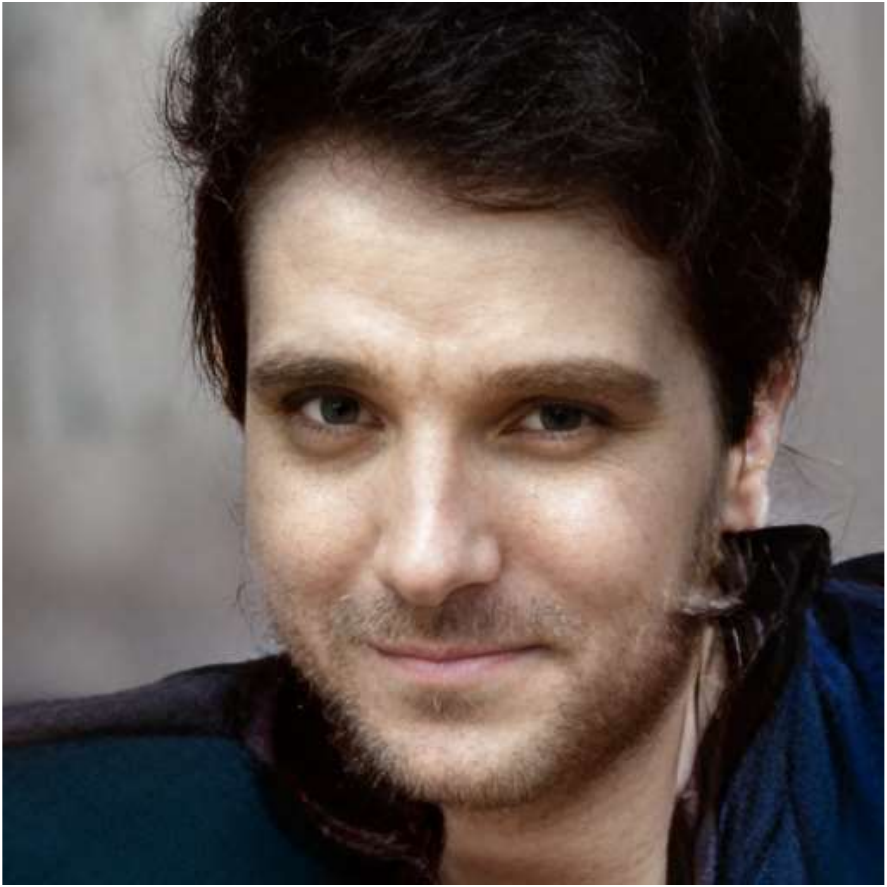}  \hspace{-4mm} &
\includegraphics[width=0.12\linewidth]{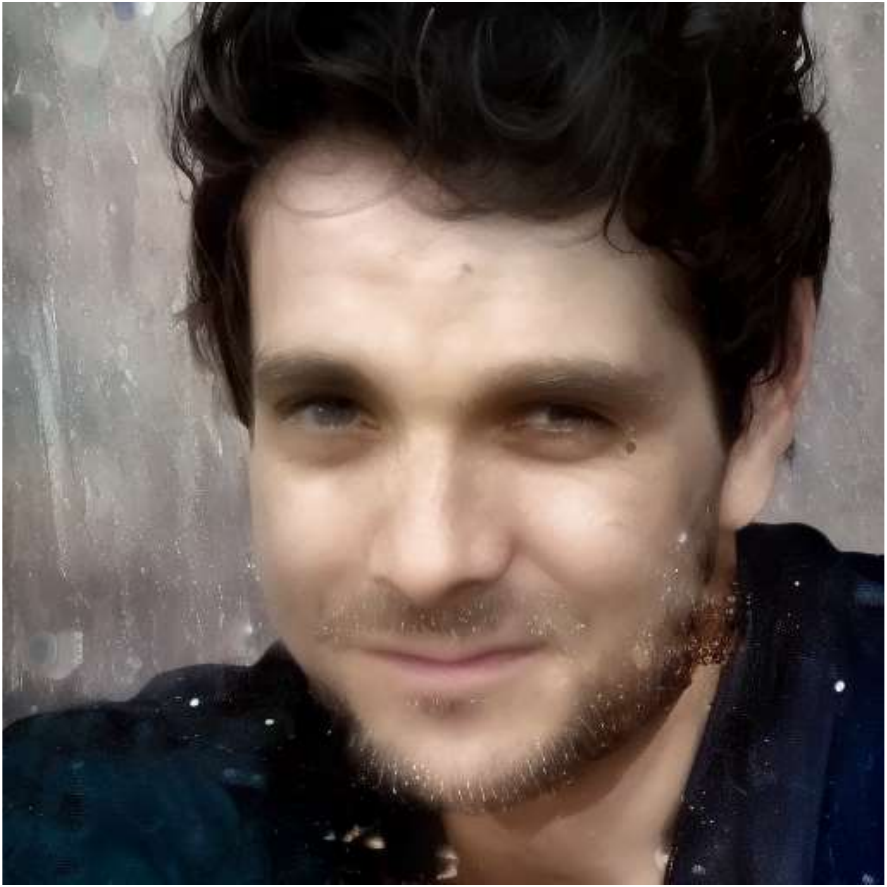}  \hspace{-4mm} &
\includegraphics[width=0.12\linewidth]{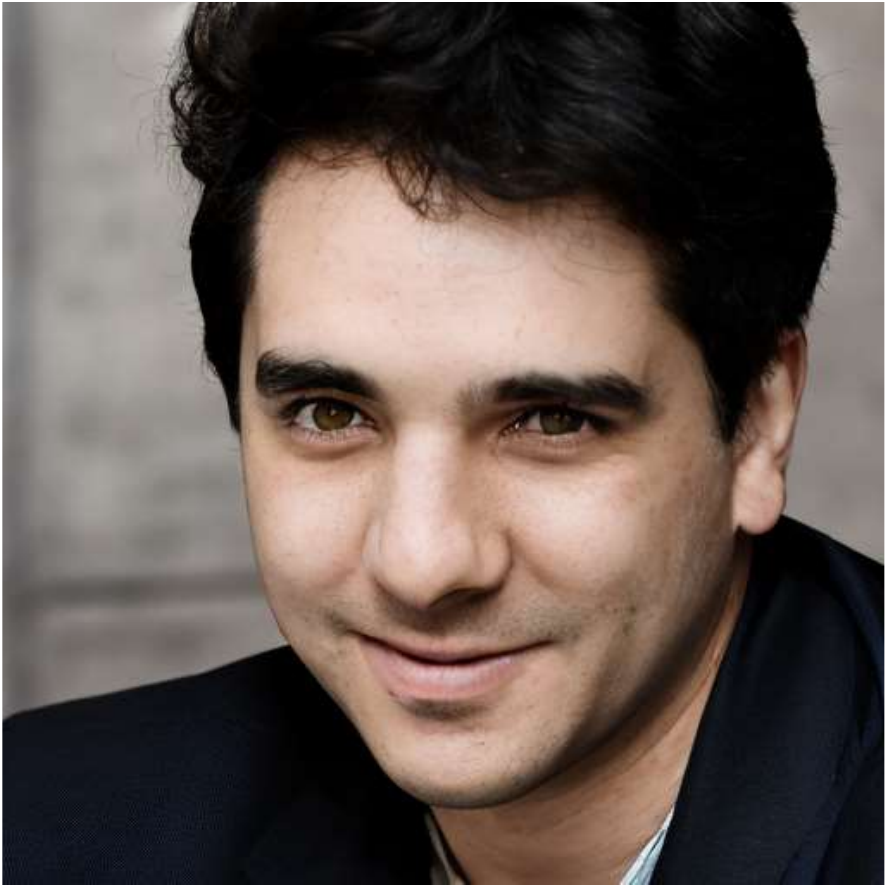}  \hspace{-4mm} &
\includegraphics[width=0.12\linewidth]{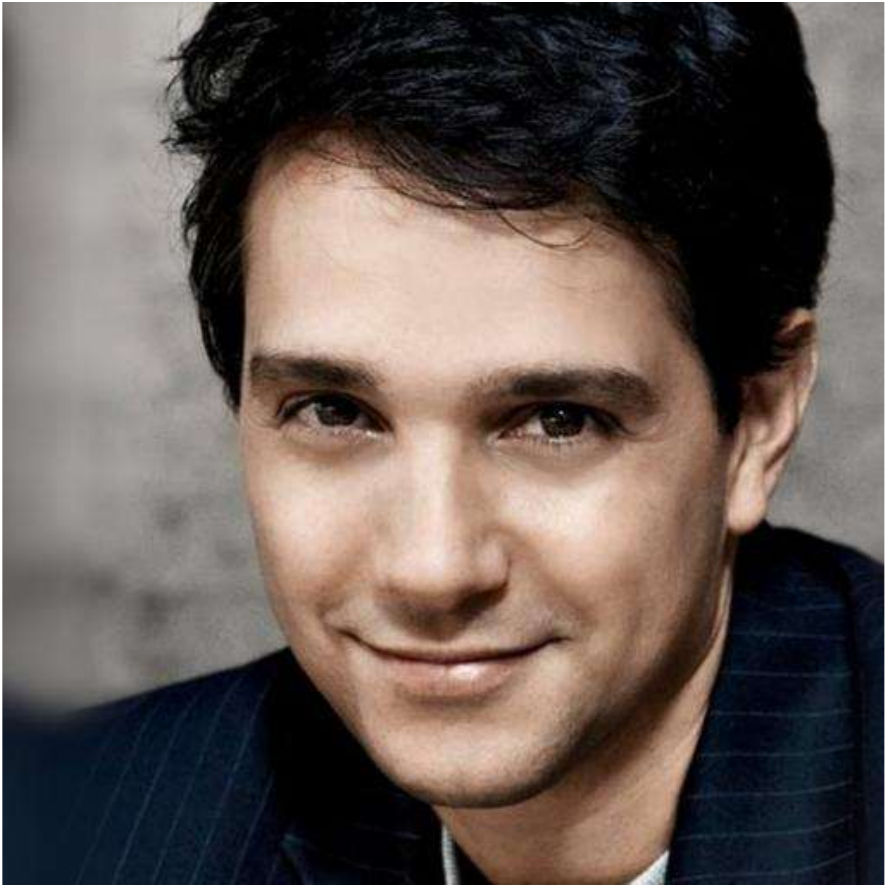} 
\end{tabular}
\end{adjustbox}
\vspace{0.1mm}
\\
\hspace{-0.55cm}
\begin{adjustbox}{valign=t}
\begin{tabular}{c}
\end{tabular}
\end{adjustbox}
\begin{adjustbox}{valign=t}
\begin{tabular}{cccccccc}
\includegraphics[width=0.12\linewidth]{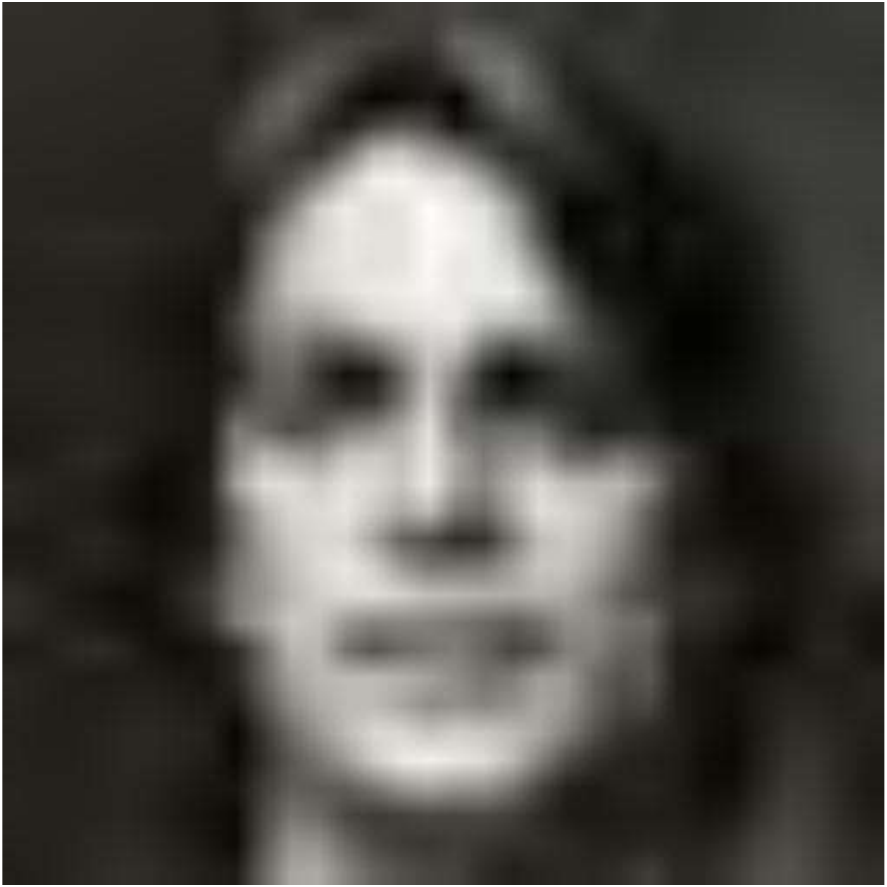} \hspace{-4mm} &
\includegraphics[width=0.12\linewidth]{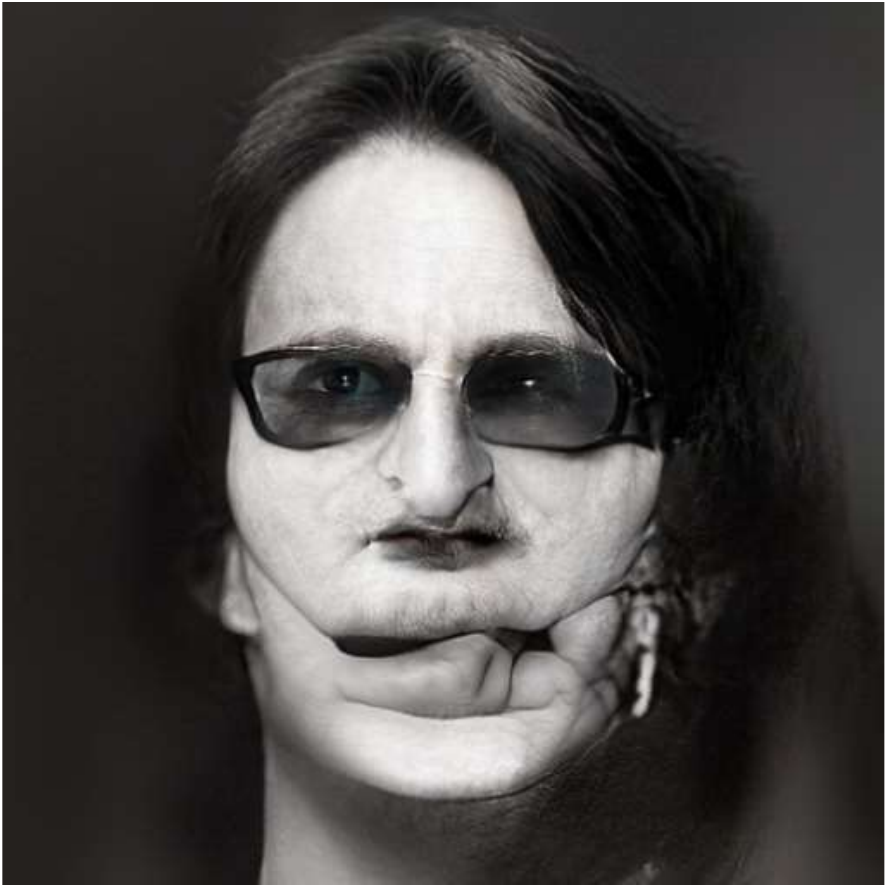}   \hspace{-4mm} &
\includegraphics[width=0.12\linewidth]{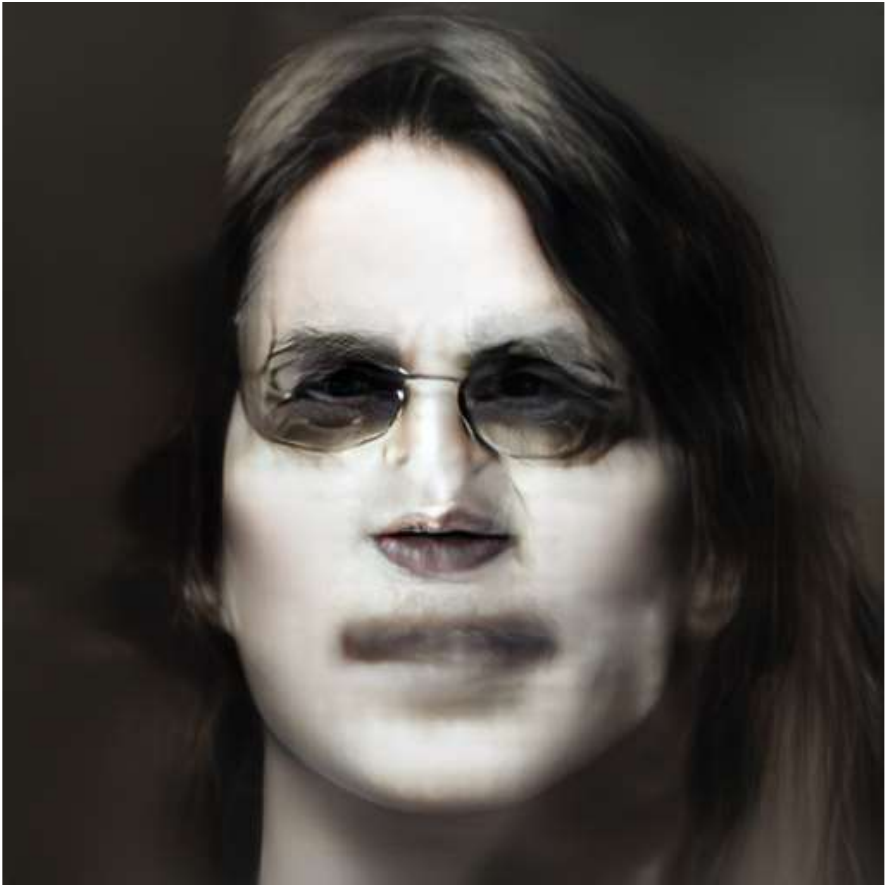}  \hspace{-4mm} &
\includegraphics[width=0.12\linewidth]{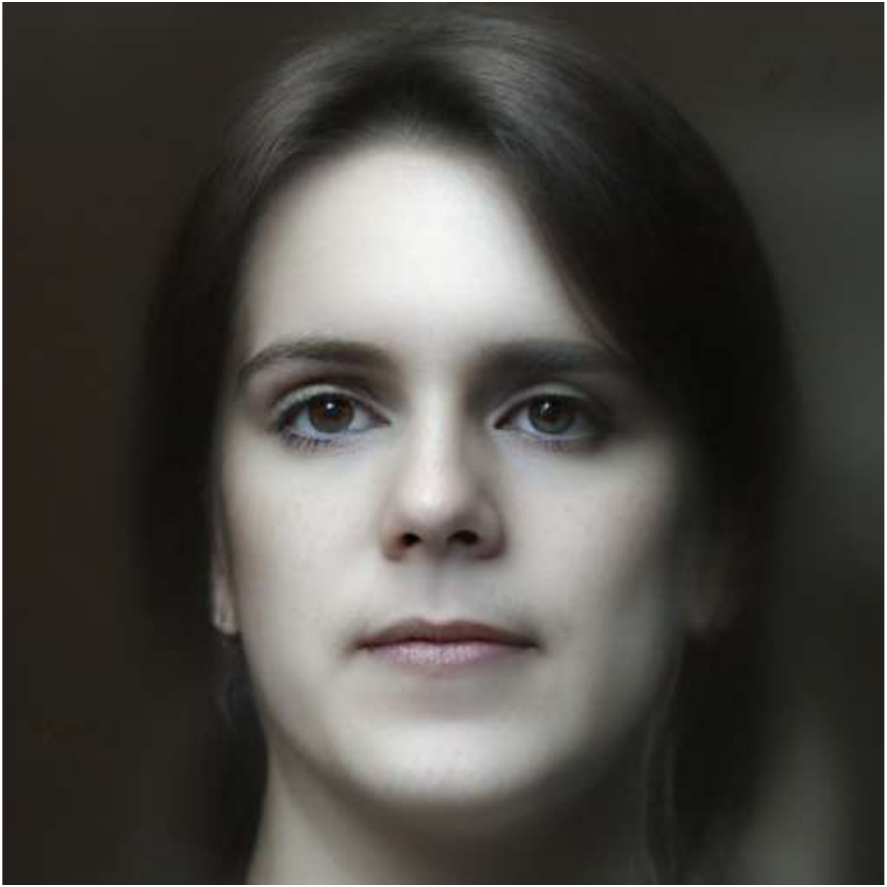}  \hspace{-4mm} &
\includegraphics[width=0.12\linewidth]{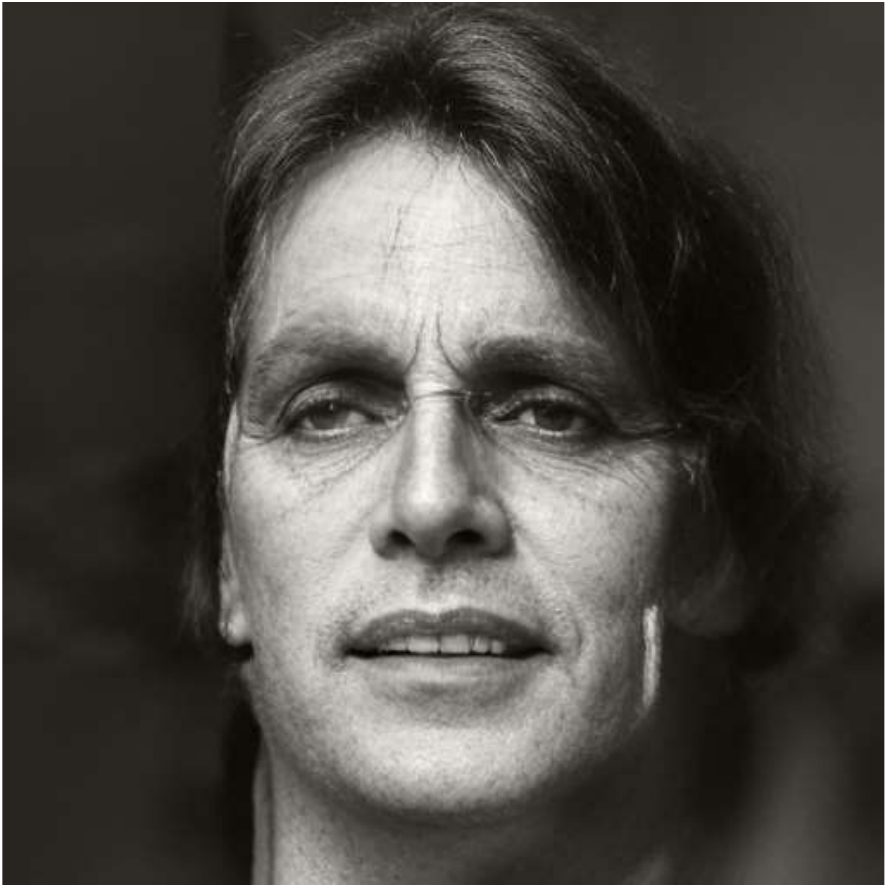}  \hspace{-4mm} &
\includegraphics[width=0.12\linewidth]{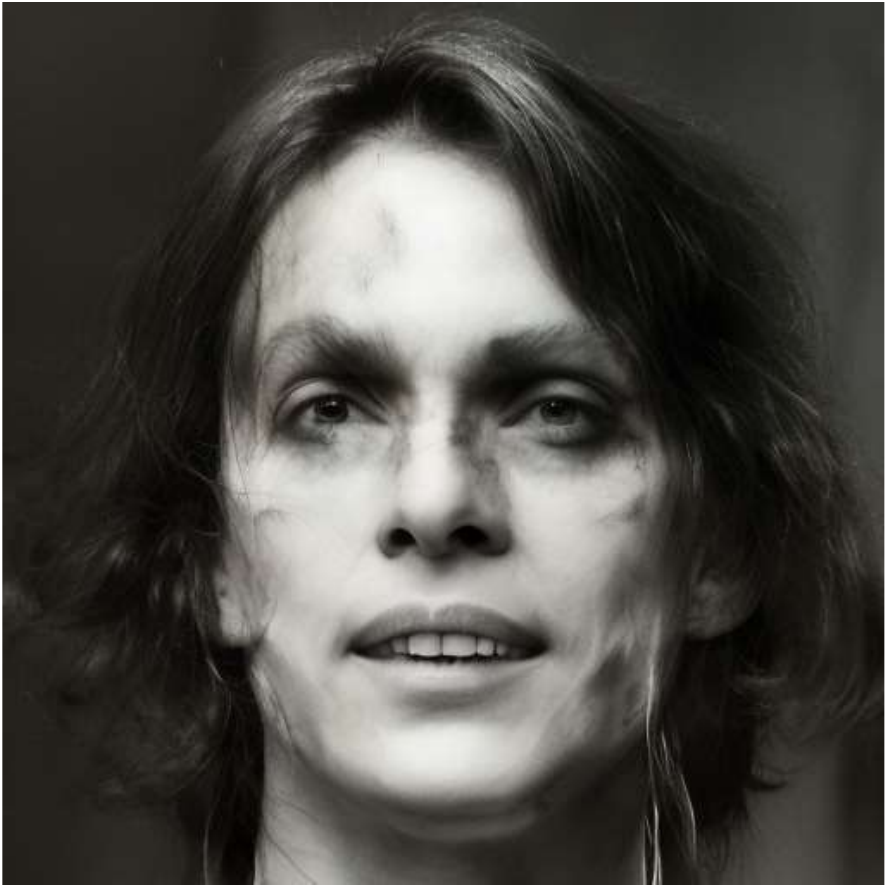}  \hspace{-4mm} &
\includegraphics[width=0.12\linewidth]{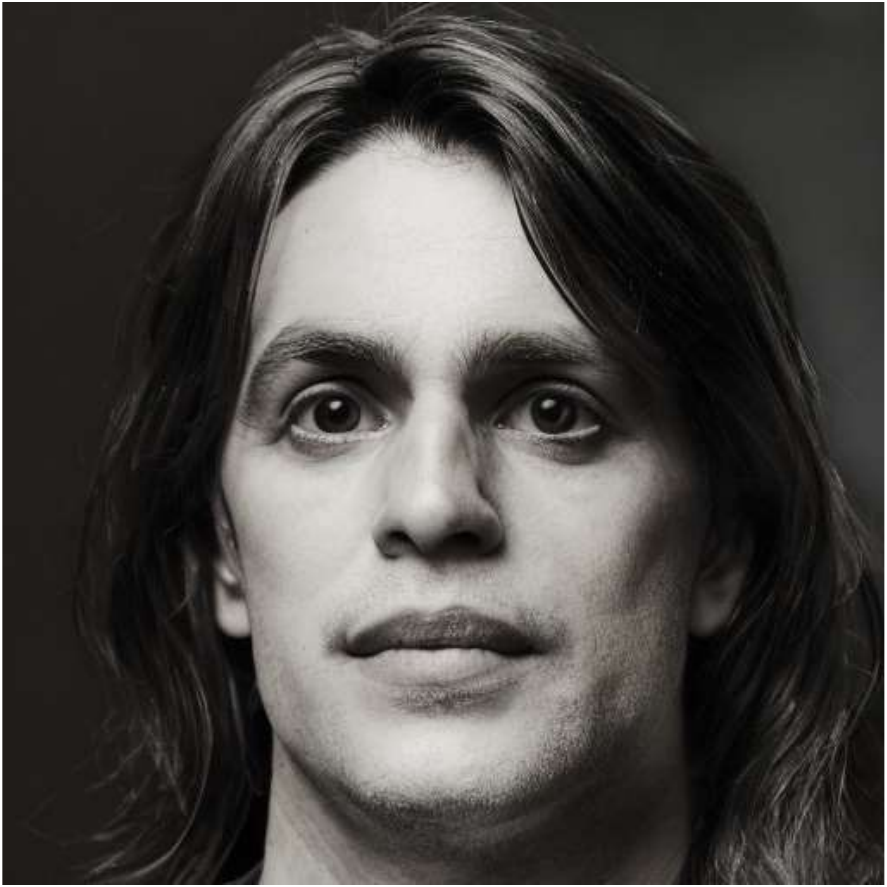}  \hspace{-4mm} &
\includegraphics[width=0.12\linewidth]{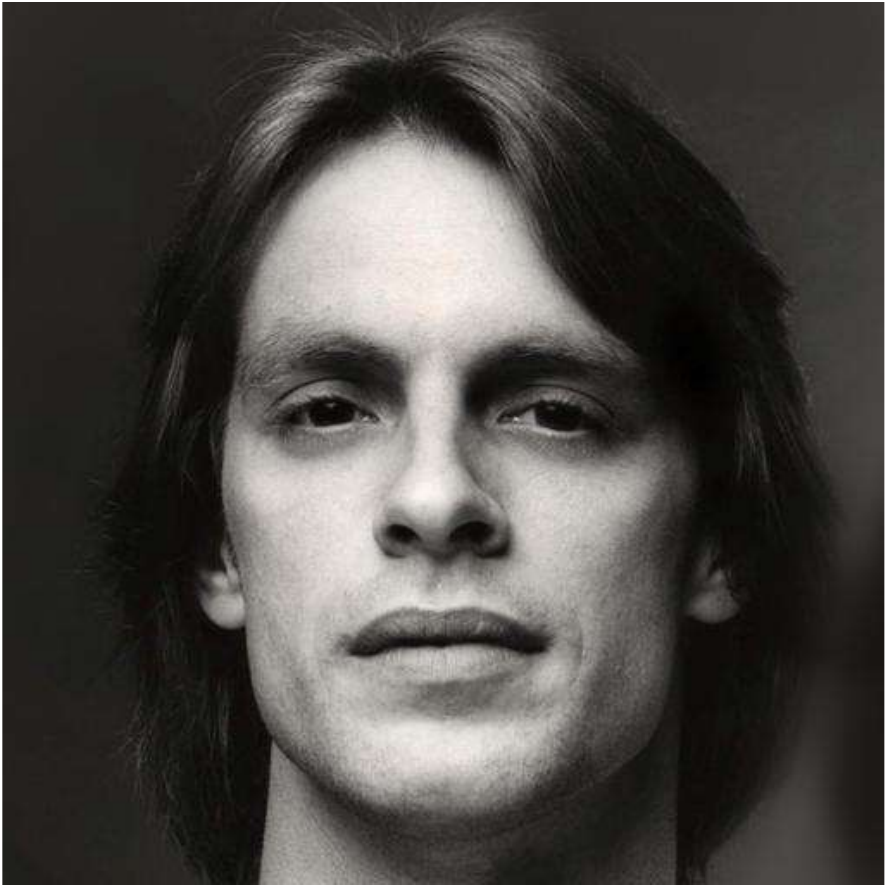} 
\\
LR \hspace{-4mm} &
ASFFNet \hspace{-4mm} &
MDMNet \hspace{-4mm} &
DR2 \hspace{-4mm} &
CodeFormer \hspace{-4mm} &
DiffBIR \hspace{-4mm} &
Ours w/o Ref. \hspace{-4mm} &
GT 
\\
\end{tabular}
\end{adjustbox}
\end{tabular}
\vspace{-5.mm}
\caption{More qualitative comparisons for our text-guided baseline model on synthetic dataset under severe degradation in Reface-Test dataset. Zoom in for best view.}
\label{fig:B_4}
\vspace{-2.mm}
\end{figure*}

\begin{figure*}[h]
\captionsetup{font={small}, skip=14pt}
\scriptsize
\begin{tabular}{ccc}
\hspace{-0.55cm}
\begin{adjustbox}{valign=t}
\begin{tabular}{c}
\end{tabular}
\end{adjustbox}
\begin{adjustbox}{valign=t}
\begin{tabular}{cccccccc}
\includegraphics[width=0.1933\linewidth]{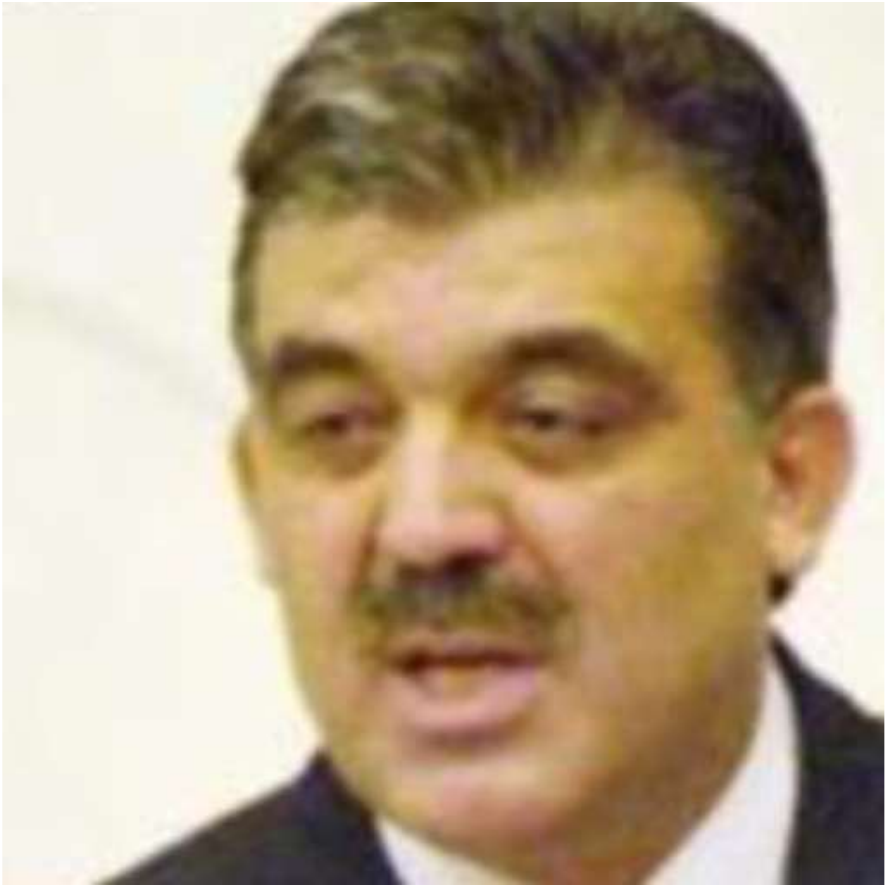} \hspace{-4mm} &
\includegraphics[width=0.1933\linewidth]{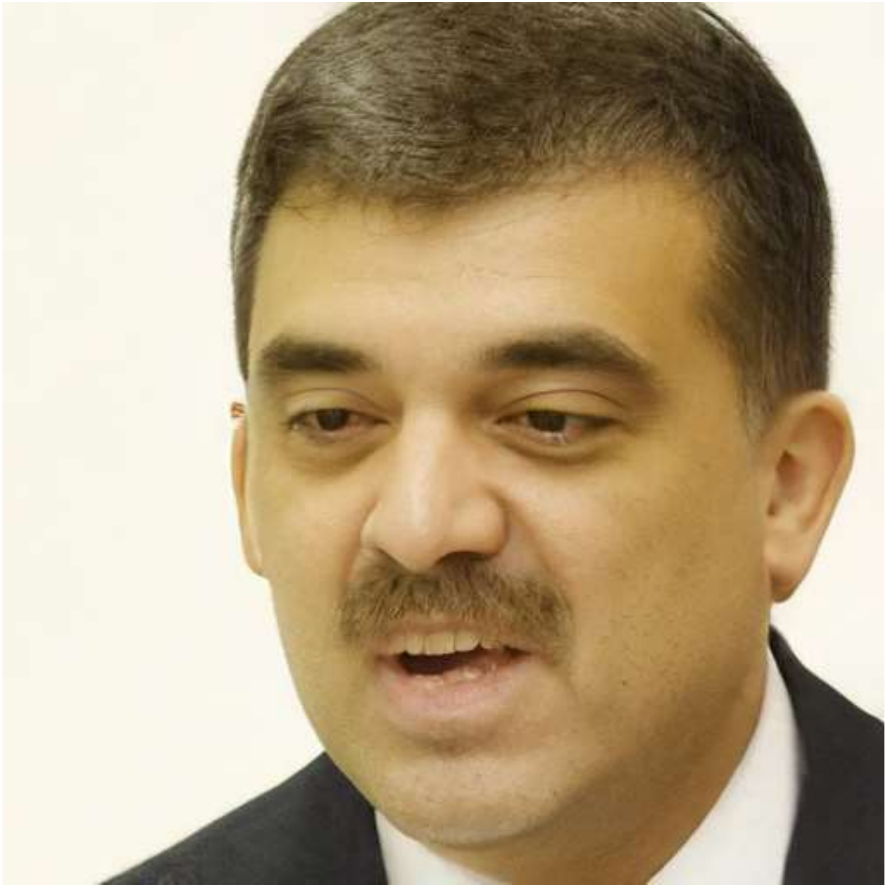}    \hspace{-4mm} &
\includegraphics[width=0.1933\linewidth]{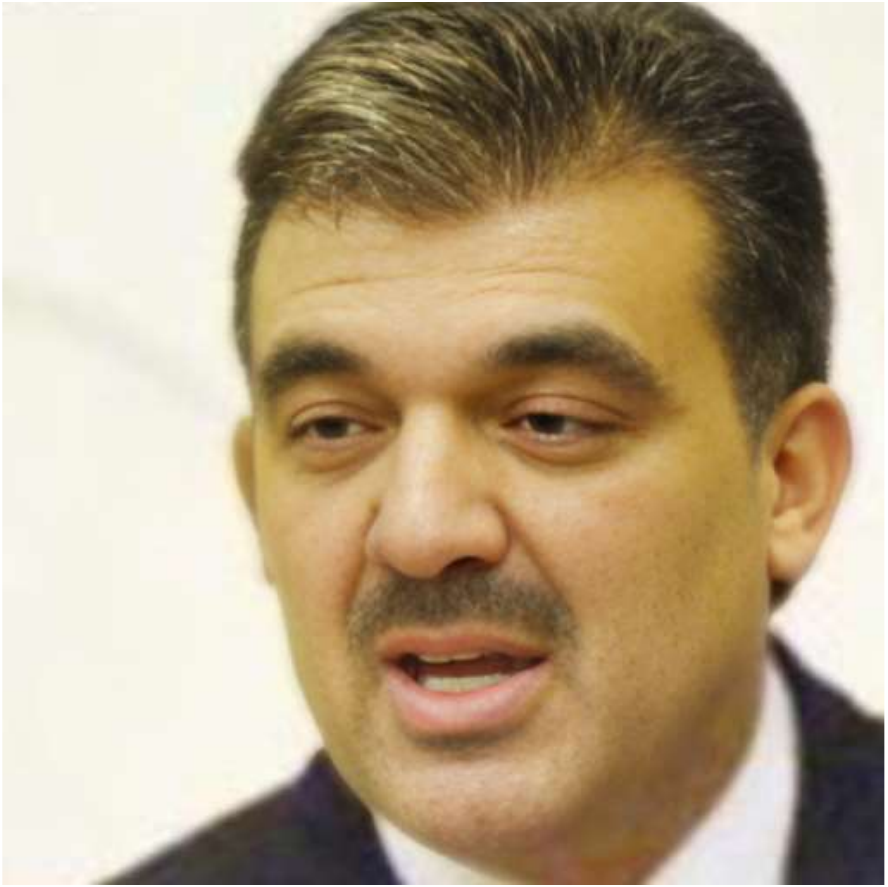}  \hspace{-4mm} &
\includegraphics[width=0.1933\linewidth]{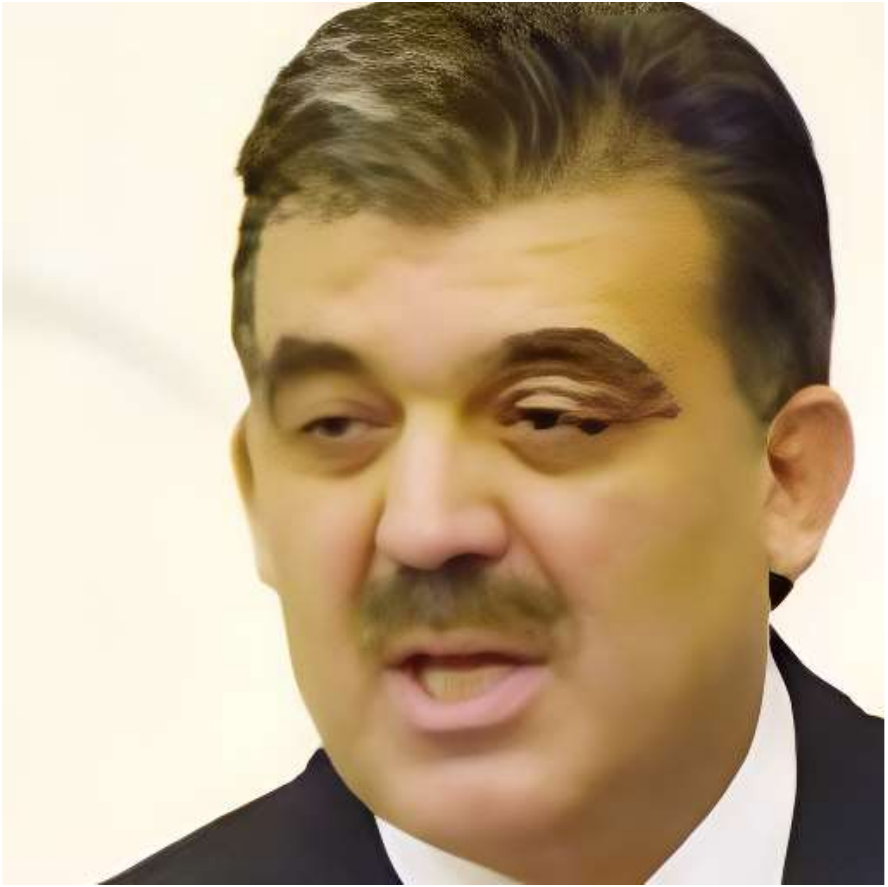}  \hspace{-4mm} &
\includegraphics[width=0.1933\linewidth]{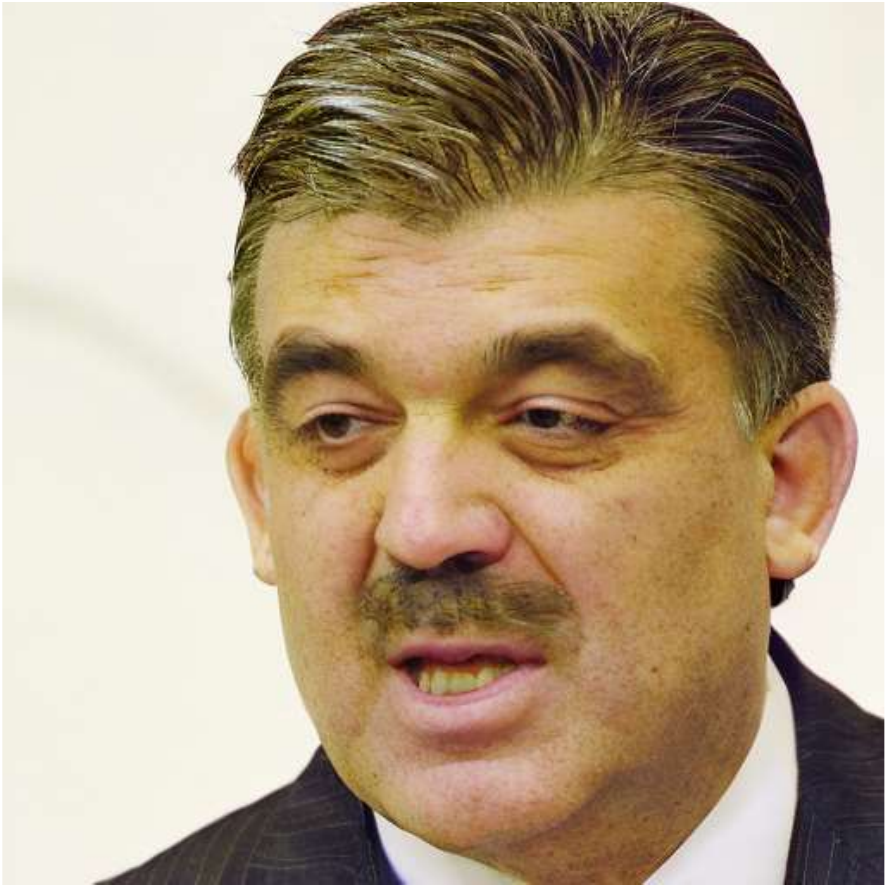} 
\end{tabular}
\end{adjustbox}
\vspace{0.1mm}
\\
\hspace{-0.55cm}
\begin{adjustbox}{valign=t}
\begin{tabular}{c}
\end{tabular}
\end{adjustbox}
\begin{adjustbox}{valign=t}
\begin{tabular}{cccccccc}
\includegraphics[width=0.1933\linewidth]{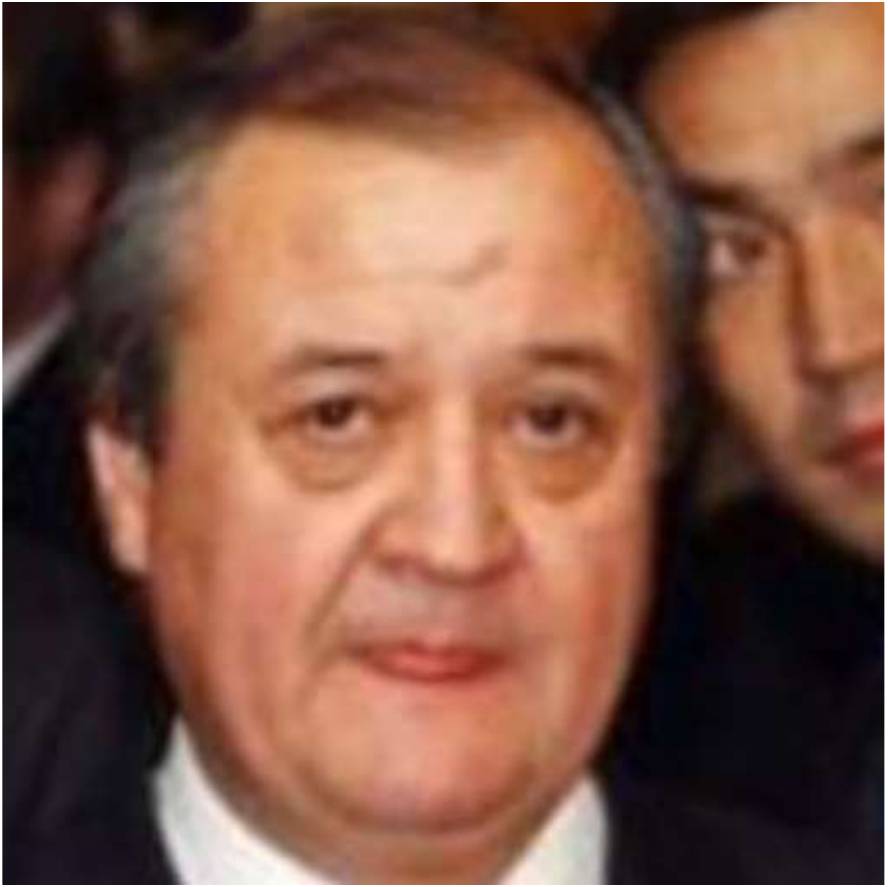} \hspace{-4mm} &
\includegraphics[width=0.1933\linewidth]{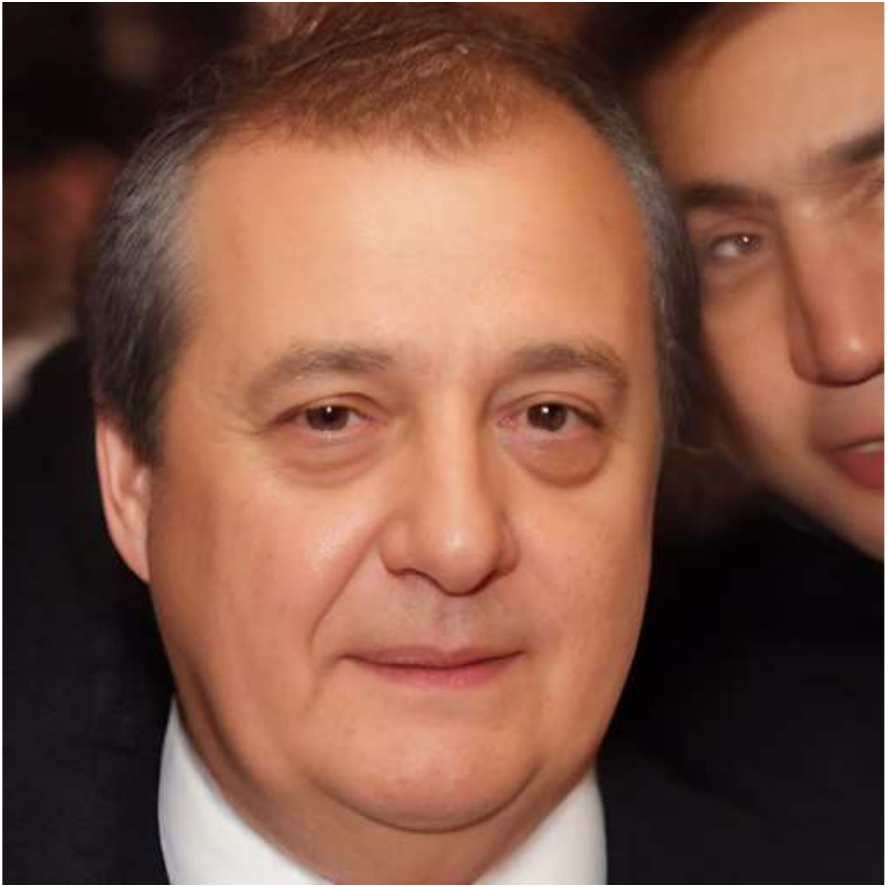}    \hspace{-4mm} &
\includegraphics[width=0.1933\linewidth]{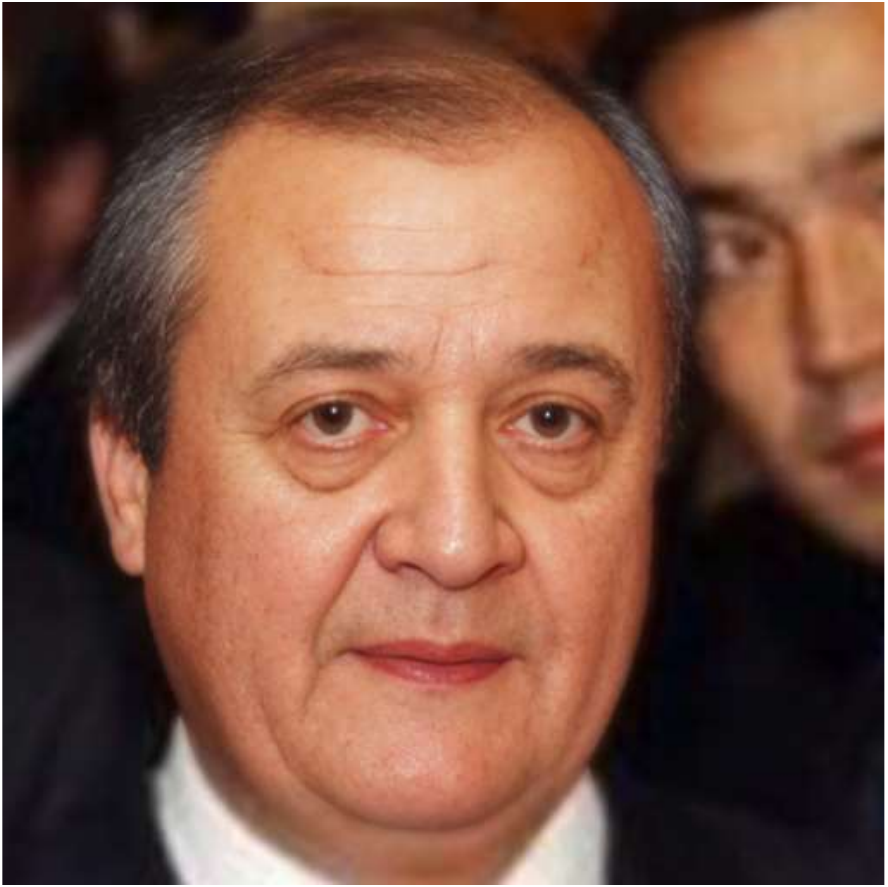}  \hspace{-4mm} &
\includegraphics[width=0.1933\linewidth]{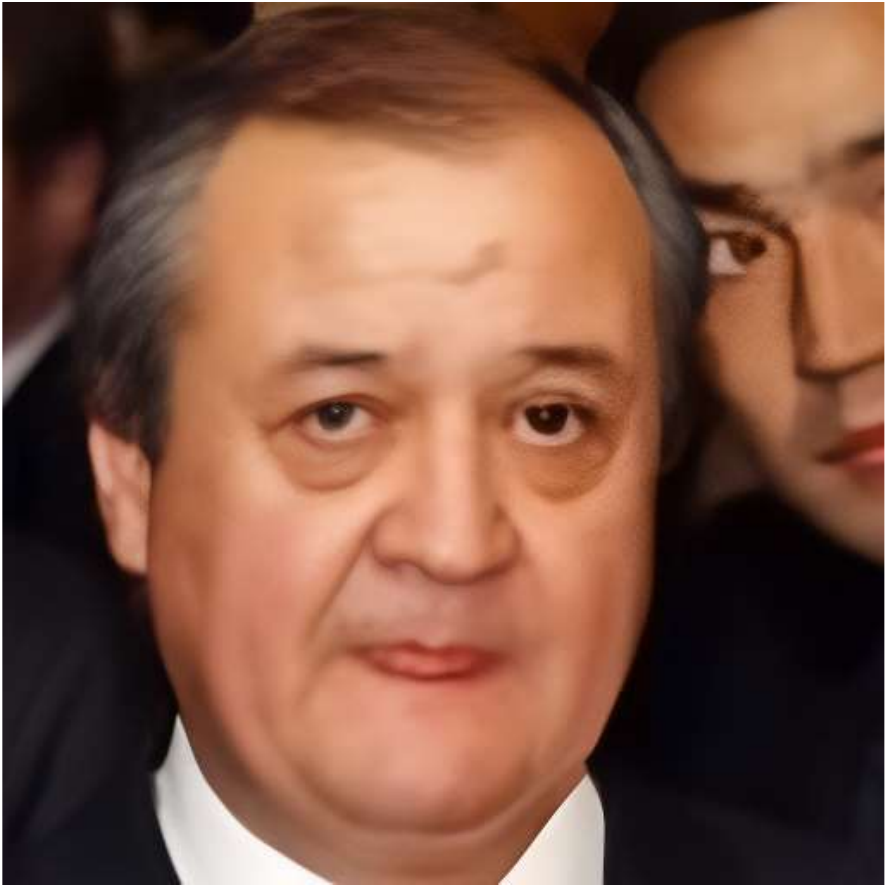}  \hspace{-4mm} &
\includegraphics[width=0.1933\linewidth]{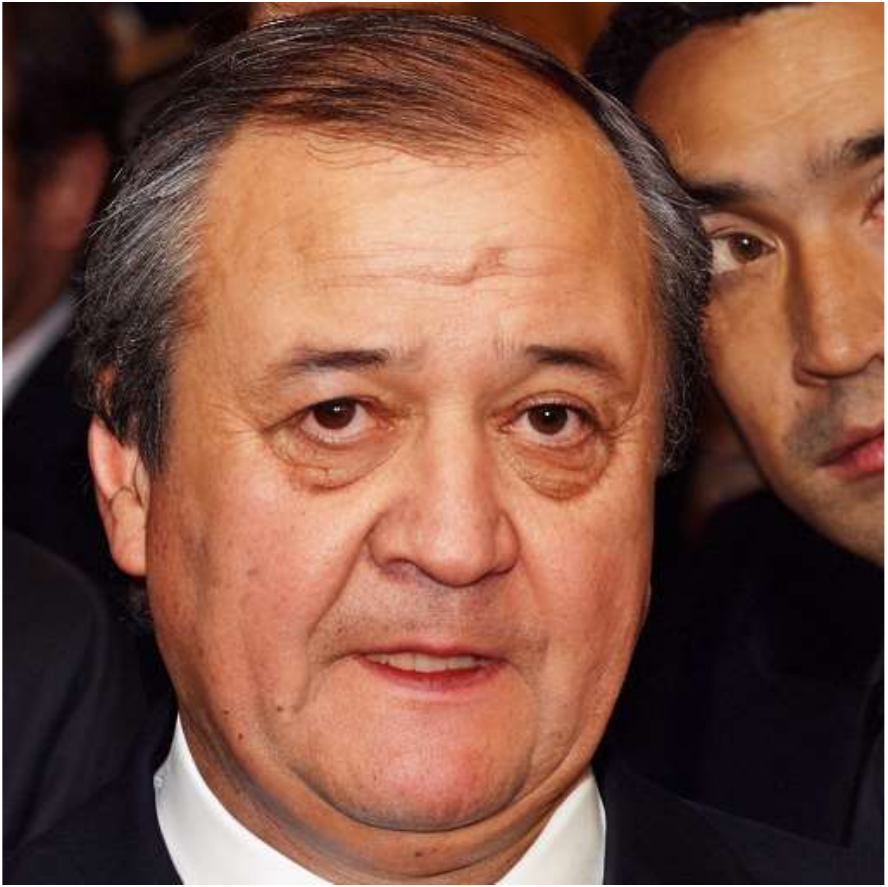} 
\\
LR \hspace{-4mm} &
DR2 \hspace{-4mm} &
CodeFormer \hspace{-4mm} &
DiffBIR \hspace{-4mm} &
Ours w/o Reference

\\
\end{tabular}
\end{adjustbox}
\end{tabular}
\vspace{-5.mm}
\caption{More qualitative comparisons for our model without reference images on real world images. Zoom in for best view.}
\label{fig:B_real}
\vspace{-2.mm}
\end{figure*}

\cref{fig:B_1} and \cref{fig:B_2} display the qualitative comparison results of our model against other advanced models under conditions of mild and moderate degradation of LQ input, respectively. 
It is evident that the previous methods exhibit severe facial illusion, whereas our model attains the best visual outcomes. Notably, as shown in \cref{fig:B_3} and \cref{fig:B_4}, our model demonstrates a remarkable ability to recover severely degraded input images with high quality and fidelity. Finally, \cref{fig:B_real} shows the effect of restoration on \textbf{real-world} LQ inputs and \cref{tab11} presents the quantitative comparison results between our model and the principal comparison methods using real-world LQ inputs from the WebPhoto-Test dataset.

\subsection{Additional Supplement to Table 1}
\label{more:tab1}
Due to space limitation, for the quantitative comparison results of our model without reference images, \cref{tab1} does not show the numerical values of PSNR and SSIM, and we supplement them in \cref{tabA:1}, \cref{tabA:2} and \cref{tabA:3}.

\begin{table*}[h]
\captionsetup{font={small}, skip=14pt}
  \centering
\resizebox{0.7\linewidth}{!}{
\begin{tabular}{c|cccccc}
\hline
 & \multicolumn{6}{c}{\textbf{Real-SR($\times$4)}} \\
\multirow{-2}{*}{\textbf{Method}} & \textbf{LPIPS} & \textbf{PSNR} & \textbf{SSIM} & \textbf{ManIQA} & \textbf{ClipIQA} & \textbf{MUSIQ} \\ \midrule
PSFRGAN & 0.2938 & 23.72 & 0.6522 & 0.5927 & 0.5702 & 73.39 \\
GPEN & 0.2828 & {\color[HTML]{3166FF} 24.78} & {\color[HTML]{CB0000} 0.7056} & {\color[HTML]{3166FF} 0.6596} & 0.6430 & 69.25 \\
VQFR & 0.2951 & 23.81 & 0.6878 & 0.2875 & 0.2490 & 62.95 \\
CodeFormer & 0.2927 & 24.56 & {\color[HTML]{000000} 0.6809} & 0.5803 & 0.5179 & {\color[HTML]{3166FF} 75.47} \\
DR2 & 0.3264 & 23.74 & {\color[HTML]{000000} 0.6827} & 0.5749 & 0.4441 & 63.43 \\
DiffBIR & {\color[HTML]{CB0000} 0.2611} & {\color[HTML]{000000} 24.49} & 0.6778 & {\color[HTML]{000000} 0.6068} & {\color[HTML]{3166FF} 0.7681} & 74.27 \\
\add{BFRffusion} & 0.3258 & {\color[HTML]{CB0000} 24.87} & {\color[HTML]{3166FF} 0.7014} & 0.5477 & 0.5572 & 45.32\\
MGFR(Ours) & {\color[HTML]{3166FF} 0.2925} & 23.25 & {\color[HTML]{343434} 0.6104} & {\color[HTML]{CB0000} 0.6854} & {\color[HTML]{CB0000} 0.8244} & {\color[HTML]{CB0000} 76.22} \\ \hline
\end{tabular}
}
\vspace{-2mm}
  \caption{\textbf{Quantitative Comparison in CelebA-Test.} Results in red and blue signify the highest and second highest, respectively. The $\downarrow$ indicates metrics whereby lower values constitute improved outcomes, with higher values preferred for all other metrics.}
  \label{tabA:1}
\vspace{-3mm}
\end{table*}

\begin{table*}[h]
\captionsetup{font={small}, skip=14pt}
  \centering
\resizebox{0.7\linewidth}{!}{
\begin{tabular}{c|cccccc}
\hline
 & \multicolumn{6}{c}{\textbf{Real-SR($\times$8)}} \\
\multirow{-2}{*}{\textbf{Method}} & \textbf{LPIPS} & \textbf{PSNR} & \textbf{SSIM} & \textbf{ManIQA} & \textbf{ClipIQA} & \textbf{MUSIQ} \\ \midrule
PSFRGAN & 0.3315 & 22.85 & 0.6232 & 0.6015 & 0.5956 & 73.08 \\
GPEN & 0.3217 & {\color[HTML]{CB0000} 23.88} & {\color[HTML]{CB0000} 0.6822} & {\color[HTML]{3166FF} 0.6754} & 0.6299 & 68.63 \\
VQFR & 0.3277 & 23.16 & 0.6683 & 0.4163 & 0.2363 & 61.92 \\
CodeFormer & {\color[HTML]{3166FF} 0.3193} & 21.81 & 0.5799 & 0.5970 & 0.6235 & {\color[HTML]{3166FF} 75.09} \\
DR2 & 0.3580 & {\color[HTML]{000000} 23.26} & {\color[HTML]{3166FF} 0.6725} & 0.5246 & 0.4494 & 59.46 \\
DiffBIR & {\color[HTML]{CB0000} 0.3017} & 23.47 & {\color[HTML]{000000} 0.6442} & 0.6058 & {\color[HTML]{3166FF} 0.7439} & 73.87 \\
\add{BFRffusion} & 0.3739 & {\color[HTML]{3166FF} 23.72} & 0.6718 & 0.4404 & 0.5298 & 42.84\\
MGFR(Ours) & 0.3227 & 22.34 & 0.5904 & {\color[HTML]{CB0000} 0.6776} & {\color[HTML]{CB0000} 0.8083} & {\color[HTML]{CB0000} 75.94} \\ \hline
\end{tabular}
}
\vspace{-2mm}
  \caption{\textbf{Quantitative Comparison in CelebA-Test.} Results in red and blue signify the highest and second highest, respectively. The $\downarrow$ indicates metrics whereby lower values constitute improved outcomes, with higher values preferred for all other metrics.}
  \label{tabA:2}
\vspace{-3mm}
\end{table*}

\begin{table*}[h]
\captionsetup{font={small}, skip=14pt}
  \centering
\resizebox{0.7\linewidth}{!}{
\begin{tabular}{c|cccccc}
\hline
 & \multicolumn{6}{c}{\textbf{Real-SR($\times$16)}} \\
\multirow{-2}{*}{\textbf{Method}} & \textbf{LPIPS} & \textbf{PSNR} & \textbf{SSIM} & \textbf{ManIQA} & \textbf{ClipIQA} & \textbf{MUSIQ} \\ \midrule
PSFRGAN & 0.3788 & 21.27 & 0.5899 & 0.5739 & 0.6274 & {\color[HTML]{3166FF} 71.76} \\
GPEN & 0.3831 & {\color[HTML]{CB0000} 22.22} & {\color[HTML]{CB0000} 0.6541} & {\color[HTML]{3166FF} 0.6618} & 0.5897 & 66.61 \\
VQFR & {\color[HTML]{3166FF} 0.3761} & {\color[HTML]{3166FF} 21.72} & {\color[HTML]{3166FF} 0.6413} & 0.6513 & 0.2148 & 60.49 \\
CodeFormer & {\color[HTML]{000000} 0.3821} & {\color[HTML]{000000} 21.19} & {\color[HTML]{000000} 0.5717} & 0.5803 & 0.5877 & 70.85 \\
DR2 & 0.3796 & 21.06 & 0.6225 & 0.5160 & 0.5035 & 70.31 \\
DiffBIR & 0.4238 & {\color[HTML]{000000} 21.21} & {\color[HTML]{000000} 0.5654} & 0.5361 & {\color[HTML]{3166FF} 0.7164} & 67.41 \\
\add{BFRffusion} & 0.3735 & {\color[HTML]{3166FF} 23.67} & 0.6716 & 0.4204 & 0.5098 & 43.16\\
MGFR(Ours) & {\color[HTML]{CB0000} 0.3760} & 20.54 & 0.5452 & {\color[HTML]{CB0000} 0.6729} & {\color[HTML]{CB0000} 0.7944} & {\color[HTML]{CB0000} 75.76} \\ \hline
\end{tabular}
}
\vspace{-2mm}
  \caption{\textbf{Quantitative Comparison in CelebA-Test.} Results in red and blue signify the highest and second highest, respectively. The $\downarrow$ indicates metrics whereby lower values constitute improved outcomes, with higher values preferred for all other metrics.}
  \label{tabA:3}
\vspace{-3mm}
\end{table*}

\add{\section{Training and Inference Consuming Analysis}}
\label{sec:memory}
\add{In terms of training consumption, the proposed MGFR model employs a two-stage training strategy for the dual-control adapter, leading to a moderate increase in training cost. However, for the diffusion-based image restoration model, this additional training time remains relatively short. Nonetheless, this investment is justified, as the proposed MGFR model demonstrates excellent recovery performance. Additionally, the dual-control adapter's specialized design enables superior restoration results depend on the guidance of multimodal information. Our experiments (\cref{fig:one-1}) confirm that employing a single traditional adapter structure for multimodal input often results in redundancy between the reference image and the low-quality input, as well as color inconsistencies in the recovered output. This observation, however, does not preclude further exploration in this area. Our future work will focus on employing a specially designed single-transformer adapter to replace the dual-control adapter, aiming to reduce the model's complexity.}

\add{In addition, \cref{tab7} presents the average inference time, memory consumption, parameter count, and FLOPs statistics. Notably, the CFG strategy is compatible with all LDM-based recovery models. Results are presented separately to reflect the CFG strategy's influence during inference. Without the CFG strategy, our model exhibits slightly higher time and memory consumption compared to DiffBIR \cite{diffbir}. DR2 \cite{dr2} and BFRfusion \cite{chen2024towards} exhibit faster inference times; however, their recovery performance is suboptimal. Furthermore, SUPIR's large-scale model design results in significantly higher training and testing costs compared to other methods, including MGFR. However, MGFR outperforms SUPIR \cite{yu2024scaling} on the face image restoration task while incurring lower costs (see \cref{sec:supir}). It should be noted that efficiency is not the primary focus of this work. Moreover, we believe that the development of efficient lightweight models is grounded in the superior performance of large-scale models. Our future iterations will explore model compression techniques, such as quantization and pruning, to enhance inference speed and reduce parameter counts while maintaining MGFR's superior performance on the face recovery task.}

\begin{table}[h]
\centering
\captionsetup{font={small}, skip=8pt}
\resizebox{0.78\linewidth}{!}{
\begin{tabular}{@{}ccccc@{}}
\toprule
\textbf{Method} & \textbf{Average time (s)} & \textbf{Memory consuming (M)} & \add{\textbf{\#Params (M)}} & \add{\textbf{FLOPs (G)}}\\ \midrule
DiffBIR & 5.1 & 11260 & 1716.7 & 897.5\\
DR2 & 2.6 & 3144 & 93.56 & 388.94  \\
\add{BFRffusion} & 3.2 & 8338 & 1197.4 & 784.5 \\
\add{SUPIR} & 47.6 & 54318 & 3870.0 &  11950\\
Ours(w/o CFG) & 6.9 & 15351 & 2029.3 & 890.5\\
Ours (w/ CFG) & 12.5 & 15351 & 2029.3 & 2672.4\\ \bottomrule
\end{tabular}
}
\caption{Inference consuming compared with other diffusion model-based methods.}
\label{tab7}
\vspace{-1mm}
\end{table}

\begin{figure}[h]
\captionsetup{font={small}, skip=8pt}
\scriptsize
\centering
\begin{tabular}{ccc}
\hspace{-0.51cm}
\begin{adjustbox}{valign=t}
\begin{tabular}{c}
\end{tabular}
\end{adjustbox}
\begin{adjustbox}{valign=t}
\begin{tabular}{cccccc}
\includegraphics[width=0.163\linewidth]{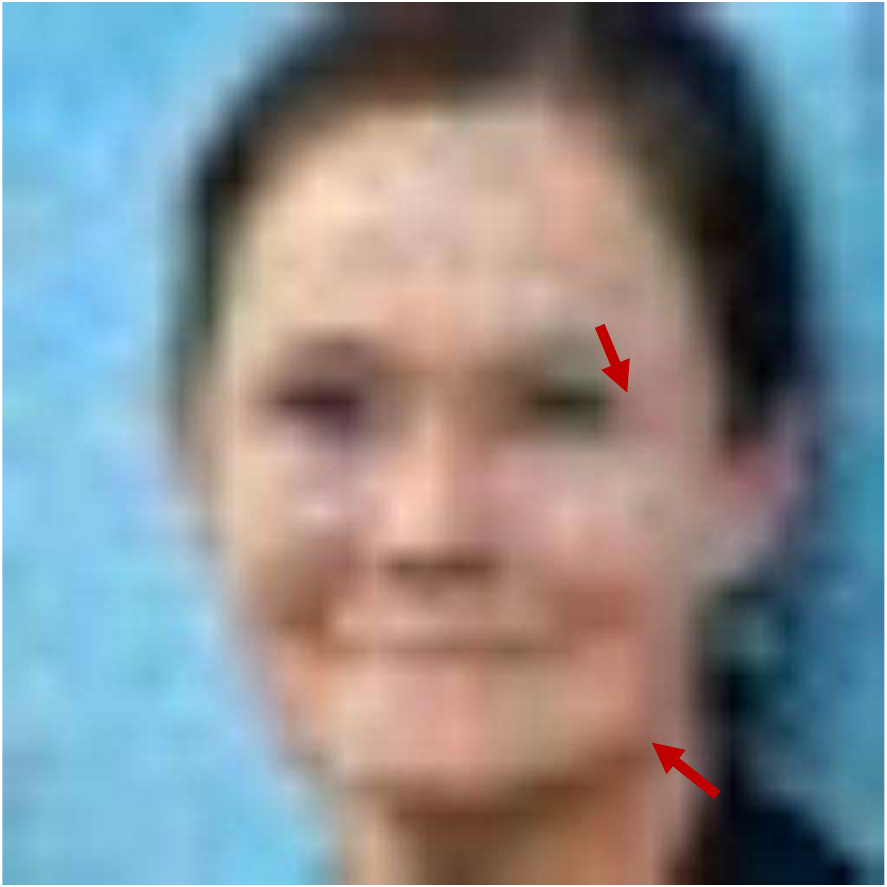} \hspace{-4mm} &
\includegraphics[width=0.163\linewidth]{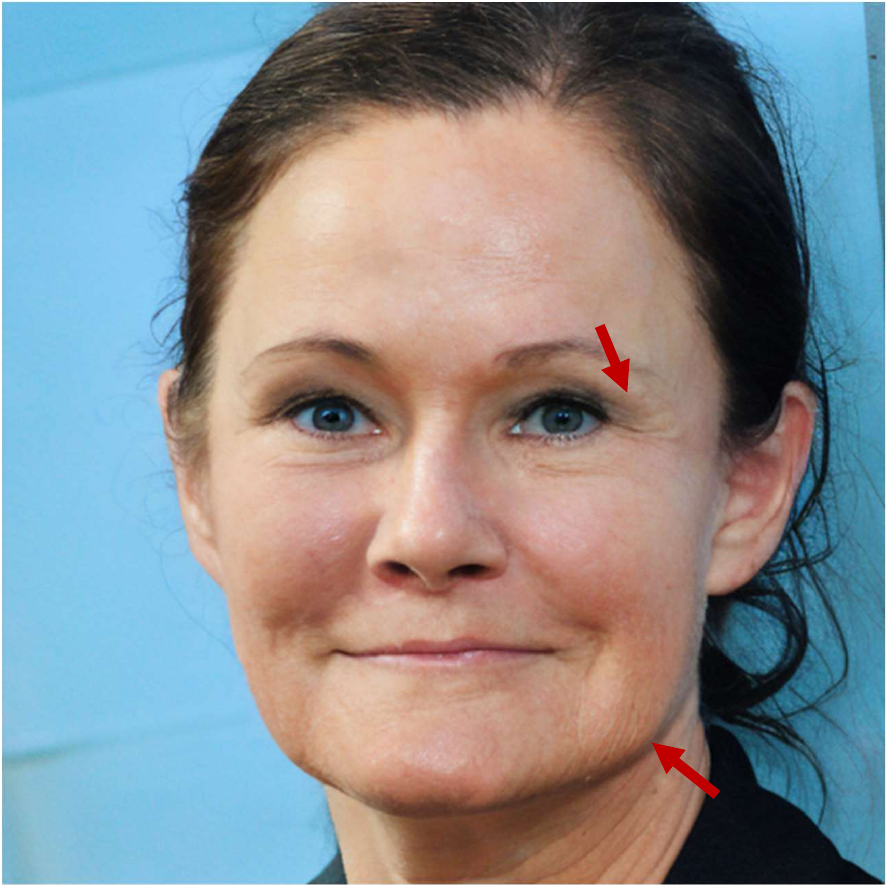} \hspace{-4mm} &
\includegraphics[width=0.163\linewidth]{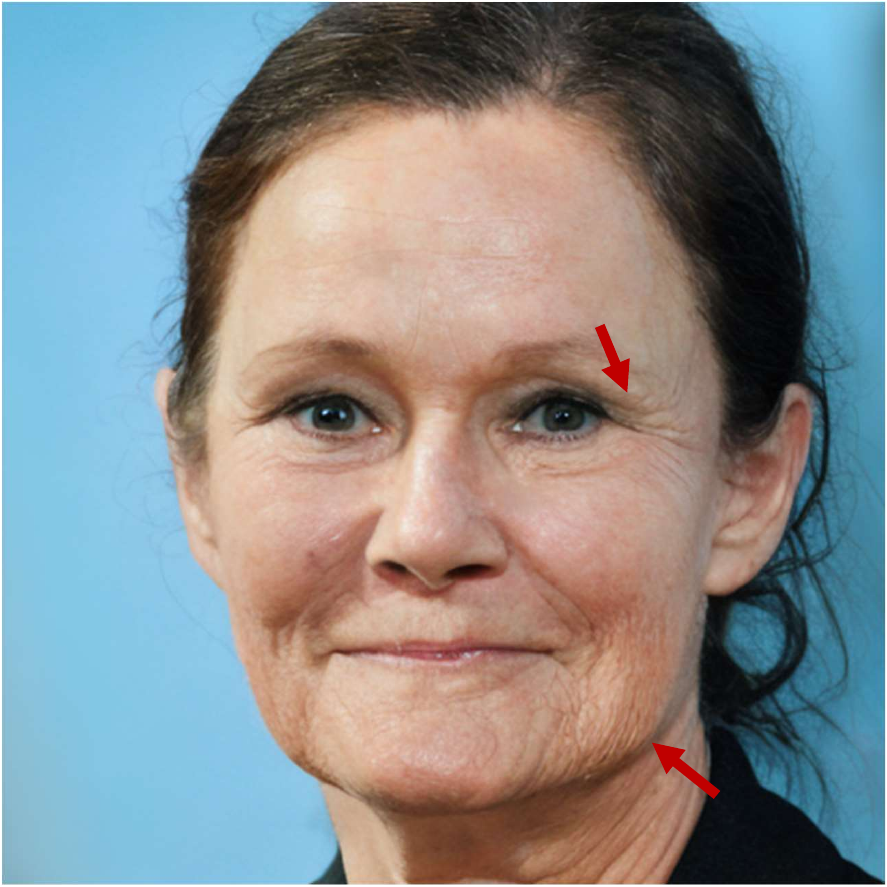} \hspace{-4mm} &
\includegraphics[width=0.163\linewidth]{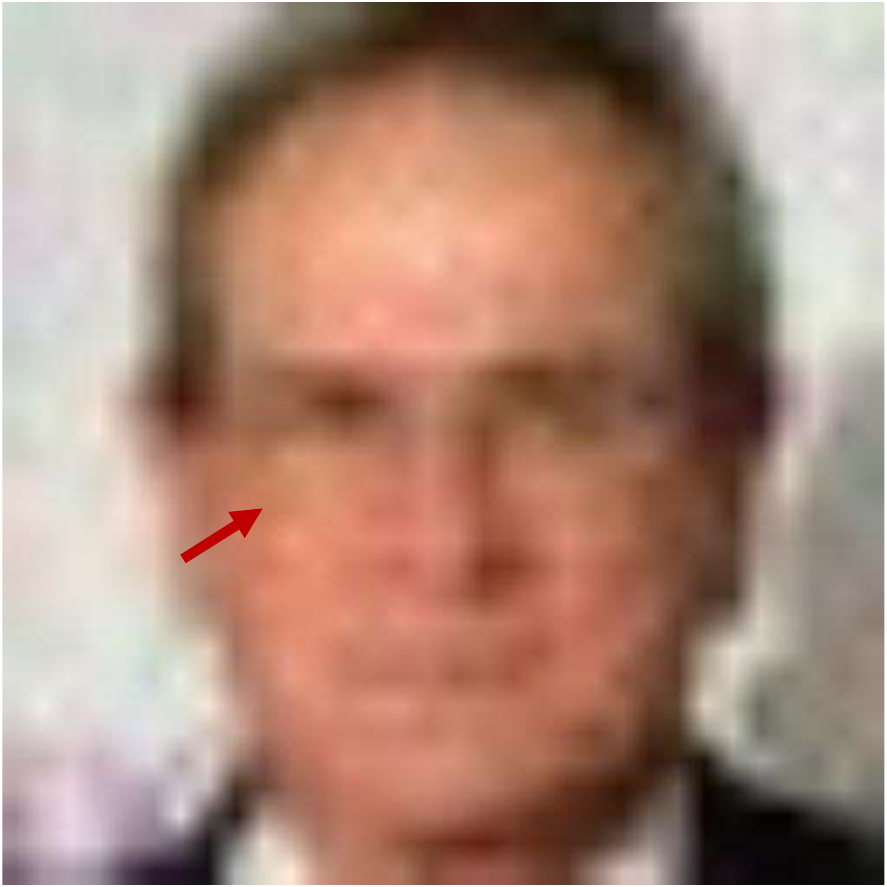} \hspace{-4mm} &
\includegraphics[width=0.163\linewidth]{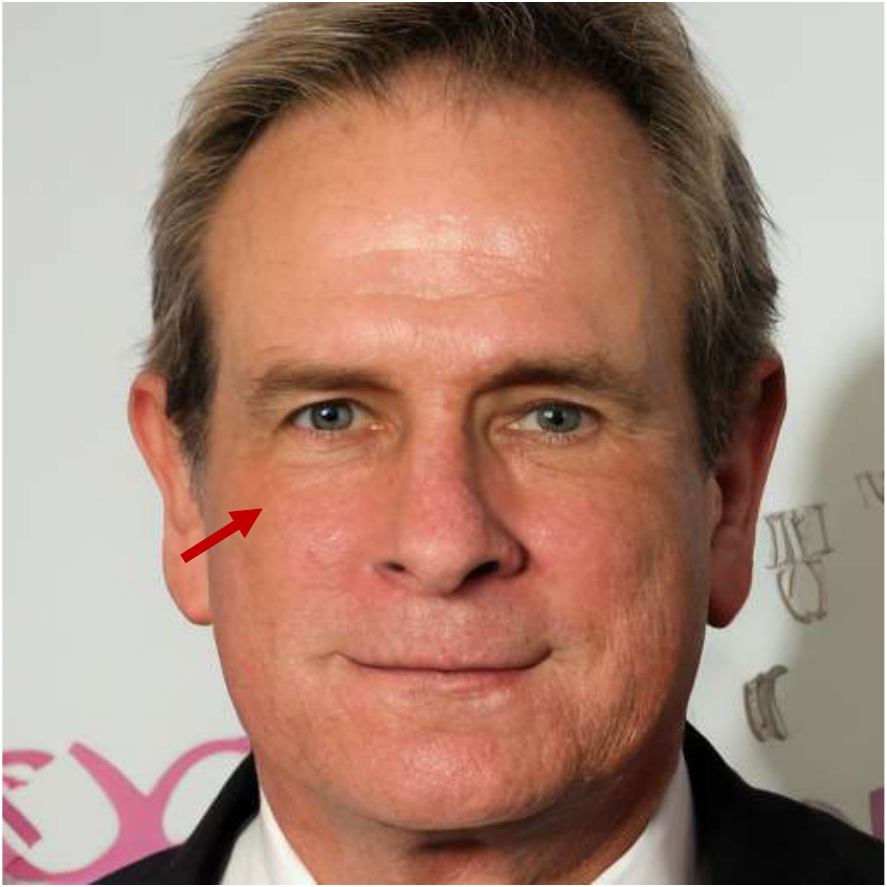} \hspace{-4mm} &
\includegraphics[width=0.163\linewidth]{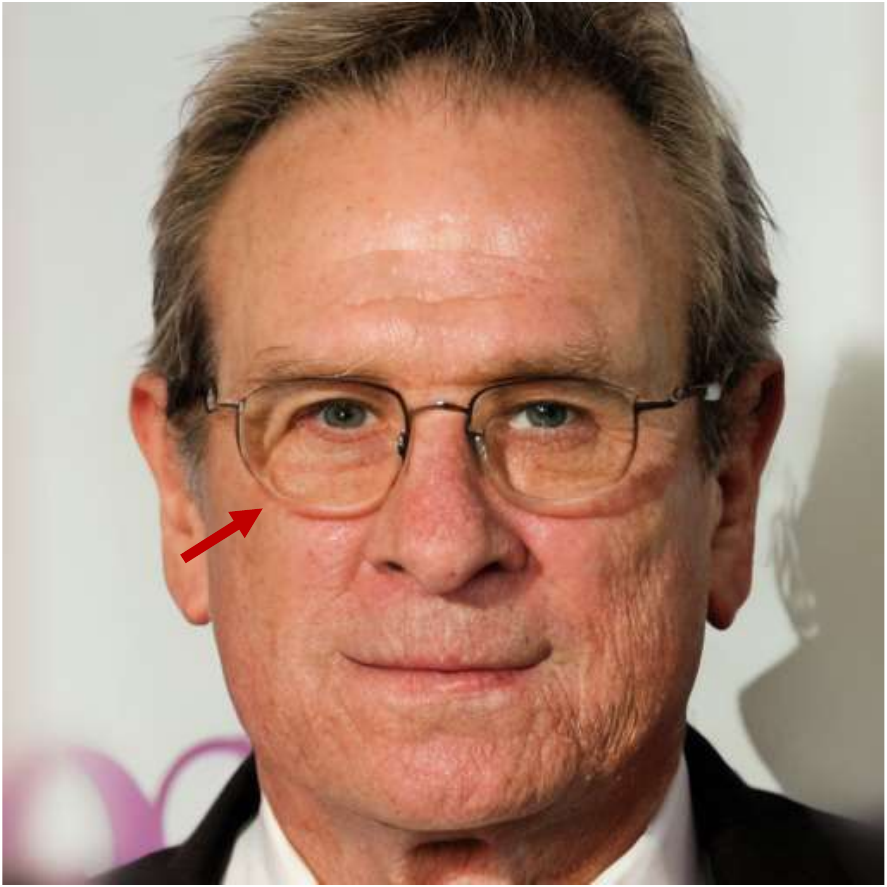}
\\
Input \hspace{-4mm} &
No Prompt \hspace{-4mm} &
\emph{`old, woman'}\hspace{-4mm} &
Input \hspace{-4mm} &
No Prompt \hspace{-4mm} &
\emph{`eyeglasses'} 
\\
\end{tabular}
\end{adjustbox}
\\
\hspace{-0.5cm}
\begin{adjustbox}{valign=t}
\begin{tabular}{cccccc}
\includegraphics[width=0.163\linewidth]{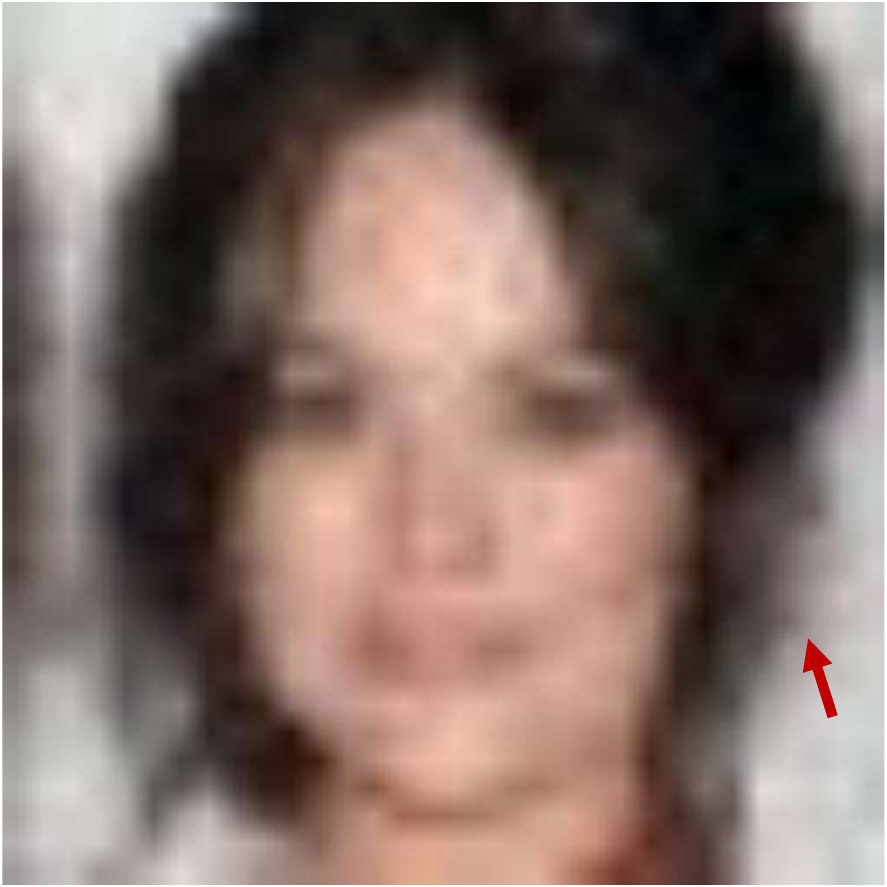} \hspace{-4mm} &
\includegraphics[width=0.163\linewidth]{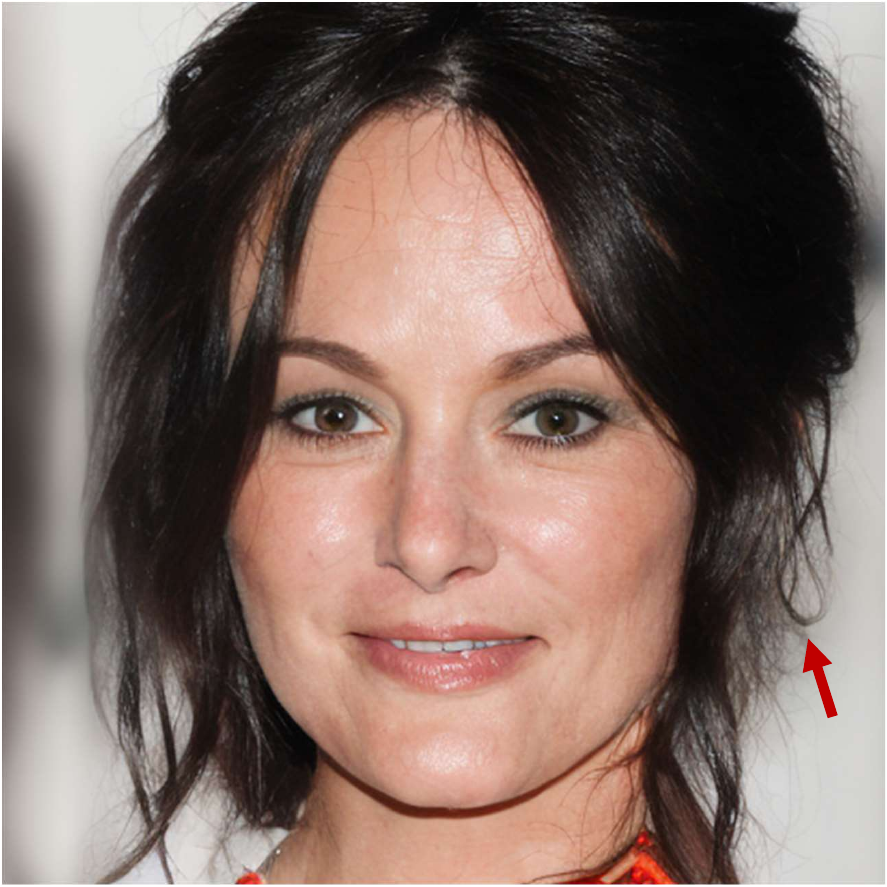} \hspace{-4mm} &
\includegraphics[width=0.163\linewidth]{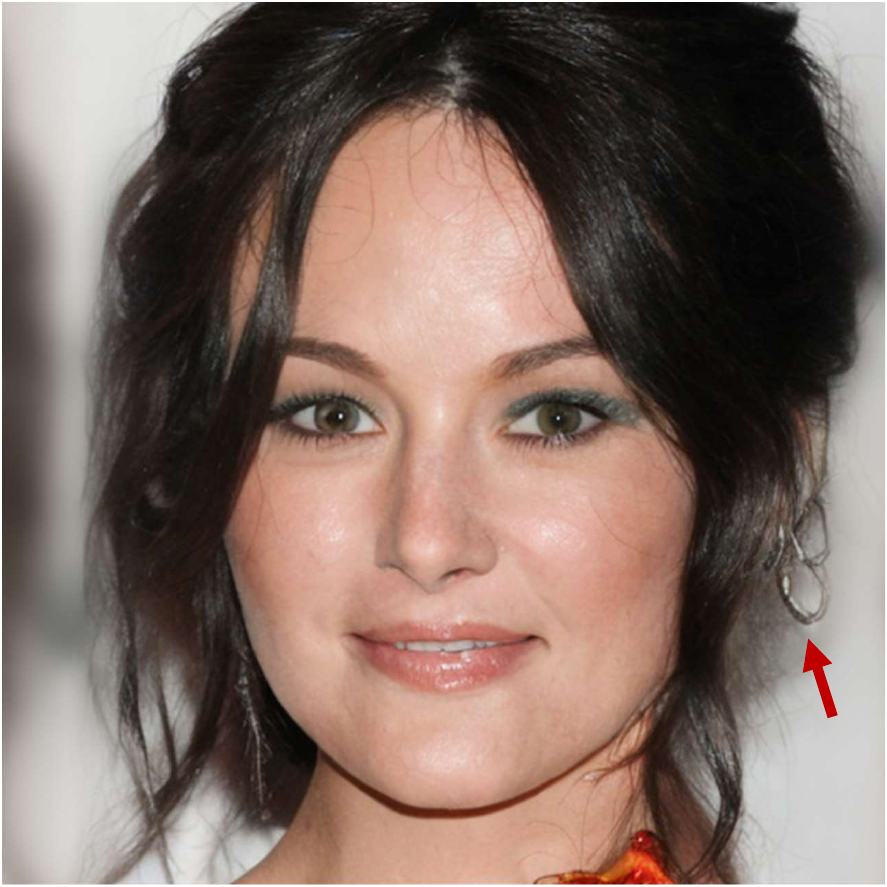} \hspace{-4mm} &
\includegraphics[width=0.163\linewidth]{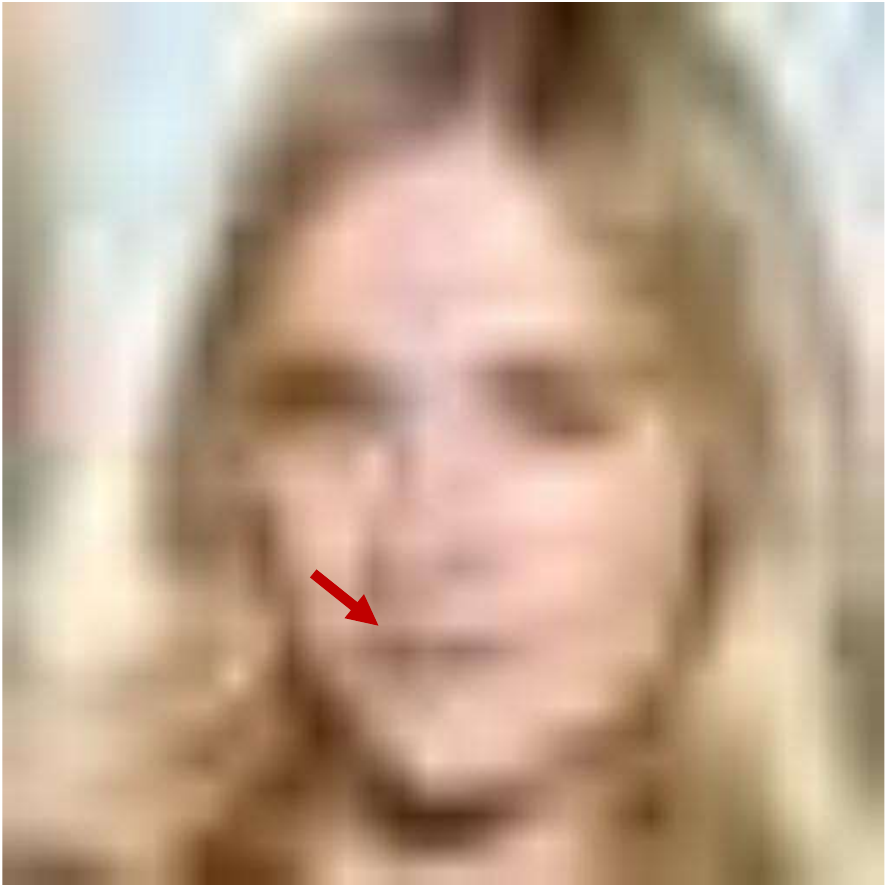} \hspace{-4mm} &
\includegraphics[width=0.163\linewidth]{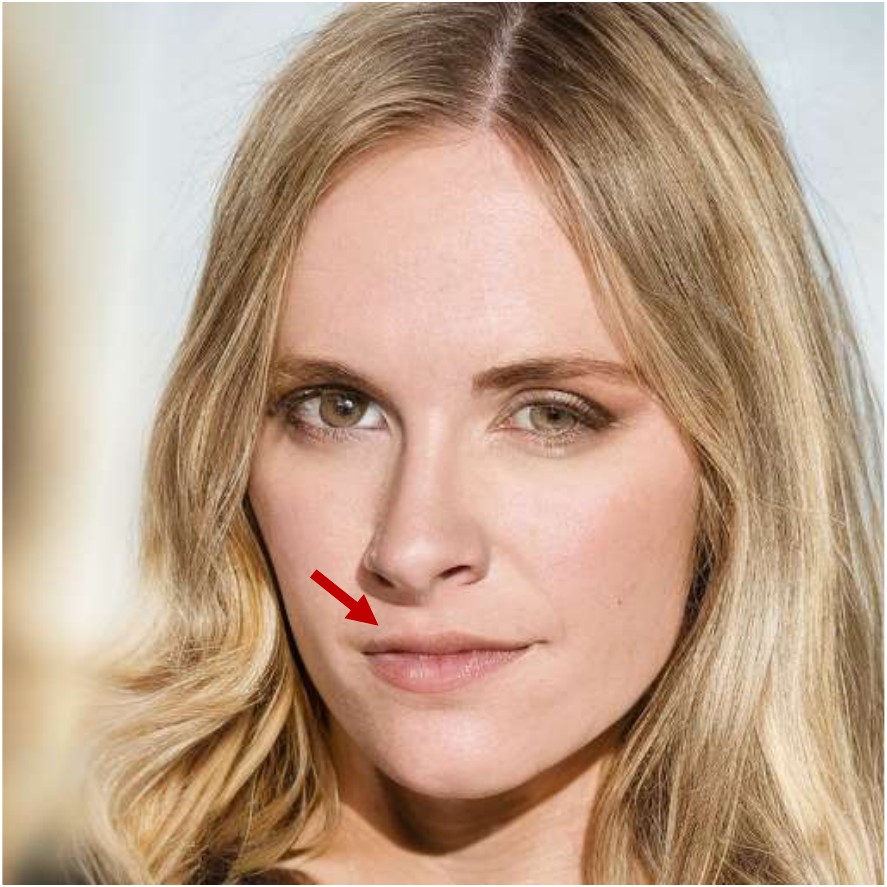} \hspace{-4mm} &
\includegraphics[width=0.163\linewidth]{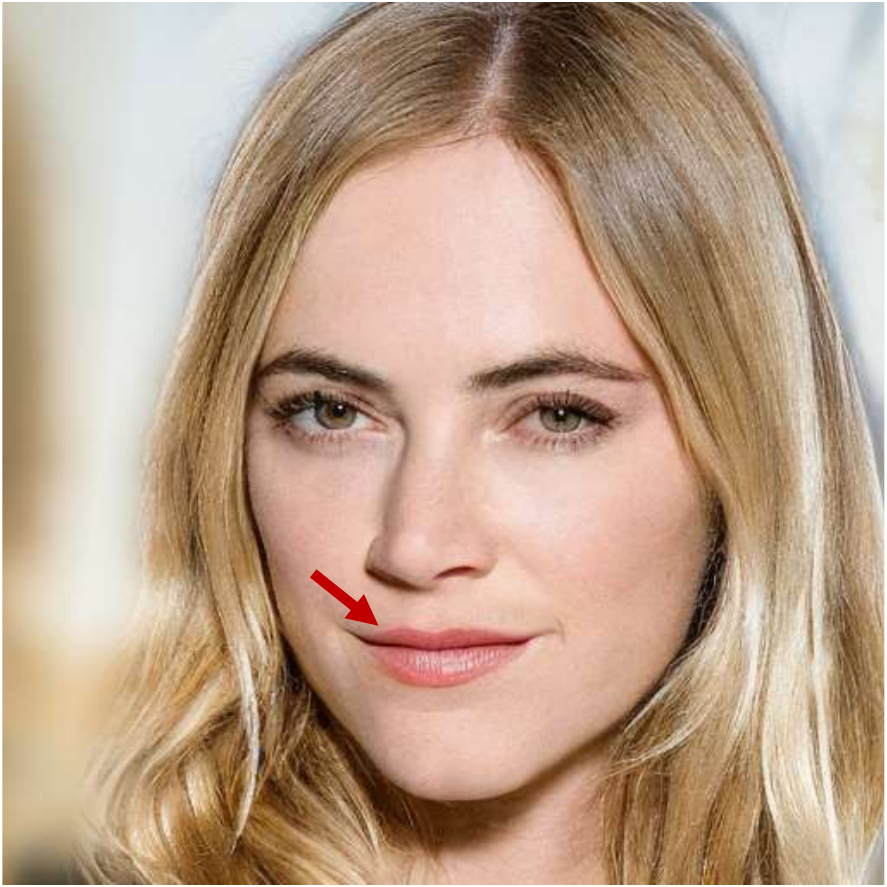}
\\
Input \hspace{-4mm} &
No Prompt \hspace{-4mm} &
\emph{`earrings'}\hspace{-4mm} &
Input \hspace{-4mm} &
No Prompt \hspace{-4mm} &
\emph{`lipstick'} 
\\
\end{tabular}
\end{adjustbox}
\\
\hspace{-0.5cm}
\begin{adjustbox}{valign=t}
\begin{tabular}{cccccc}
\includegraphics[width=0.163\linewidth]{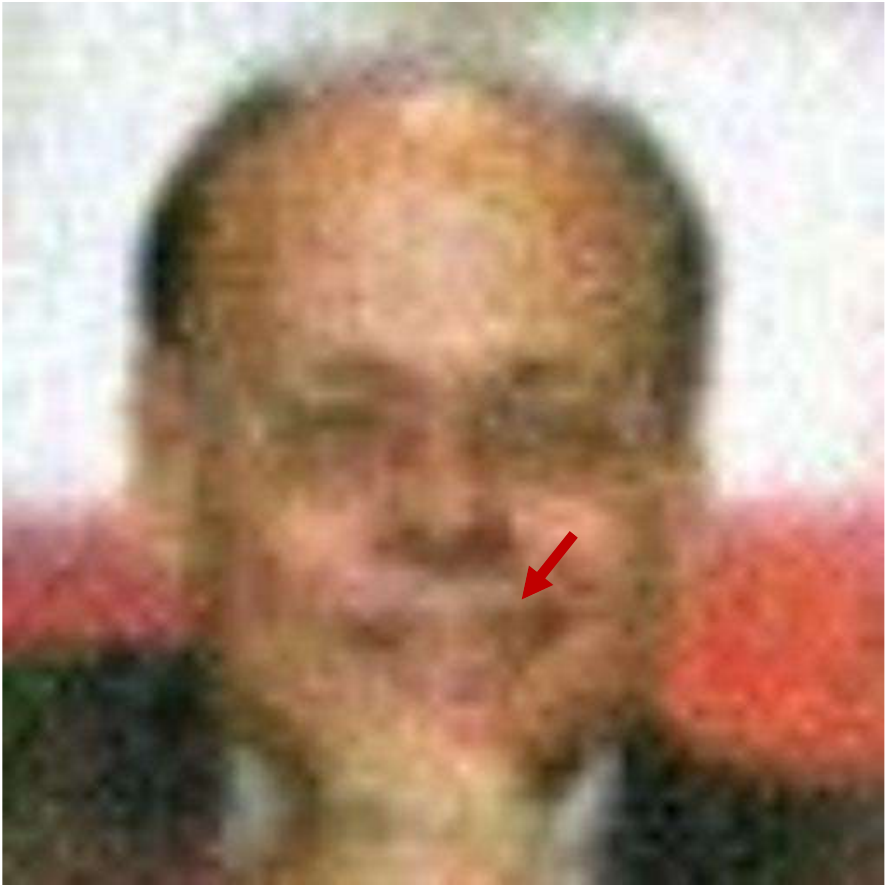} \hspace{-4mm} &
\includegraphics[width=0.163\linewidth]{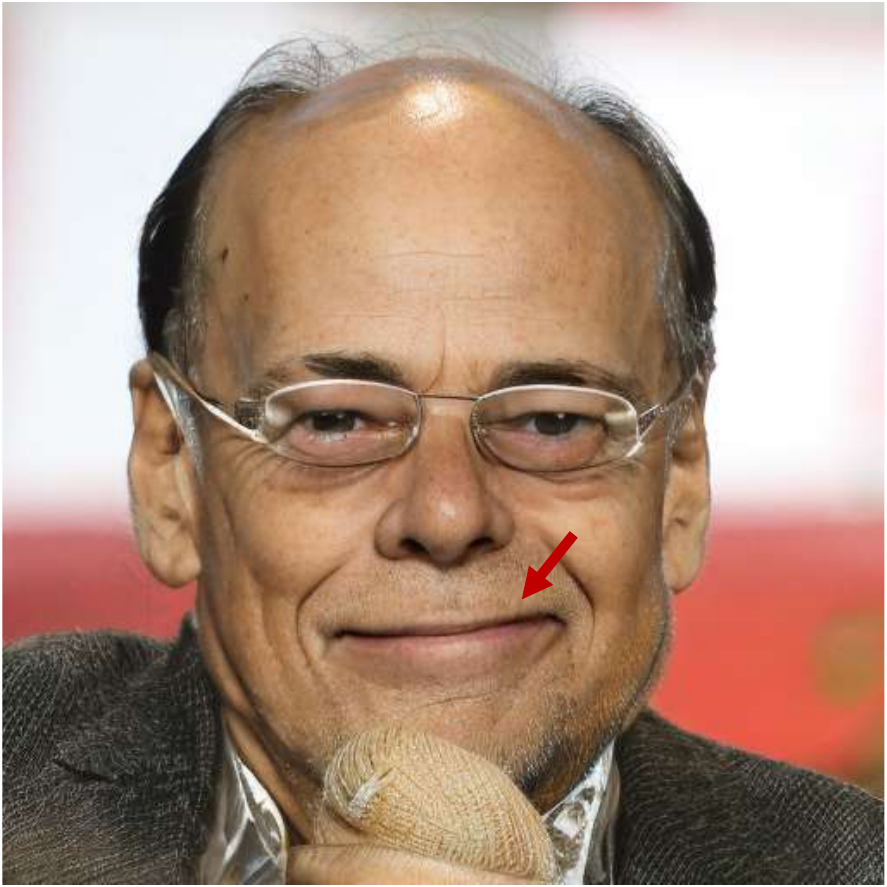} \hspace{-4mm} &
\includegraphics[width=0.163\linewidth]{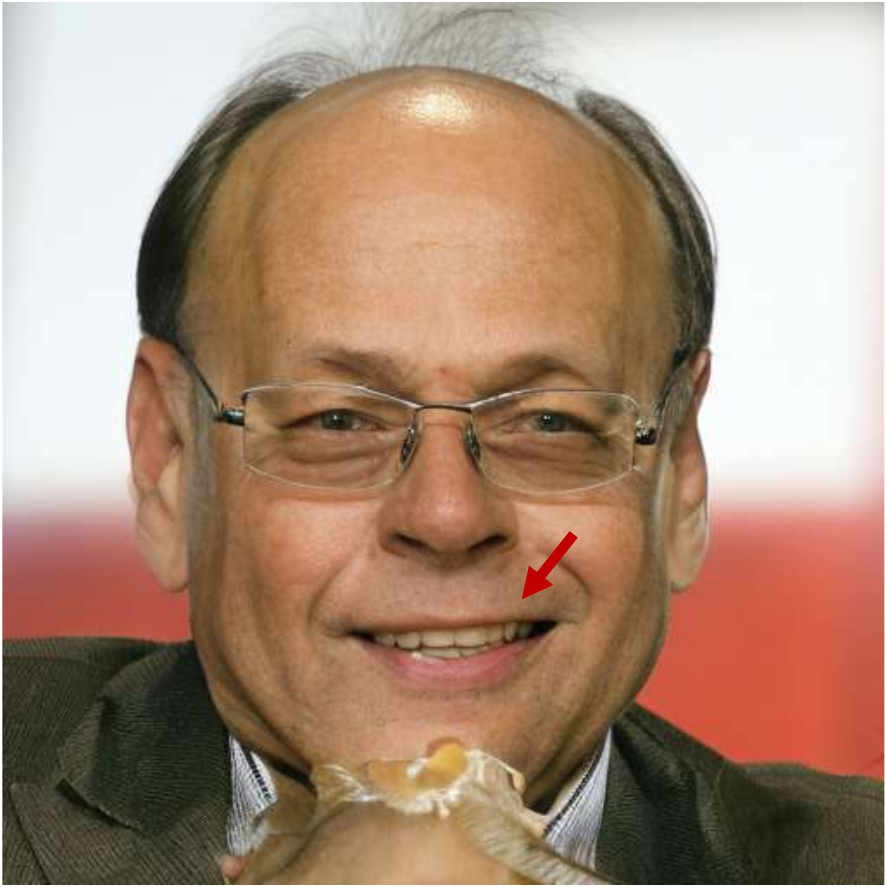} \hspace{-4mm} &
\includegraphics[width=0.163\linewidth]{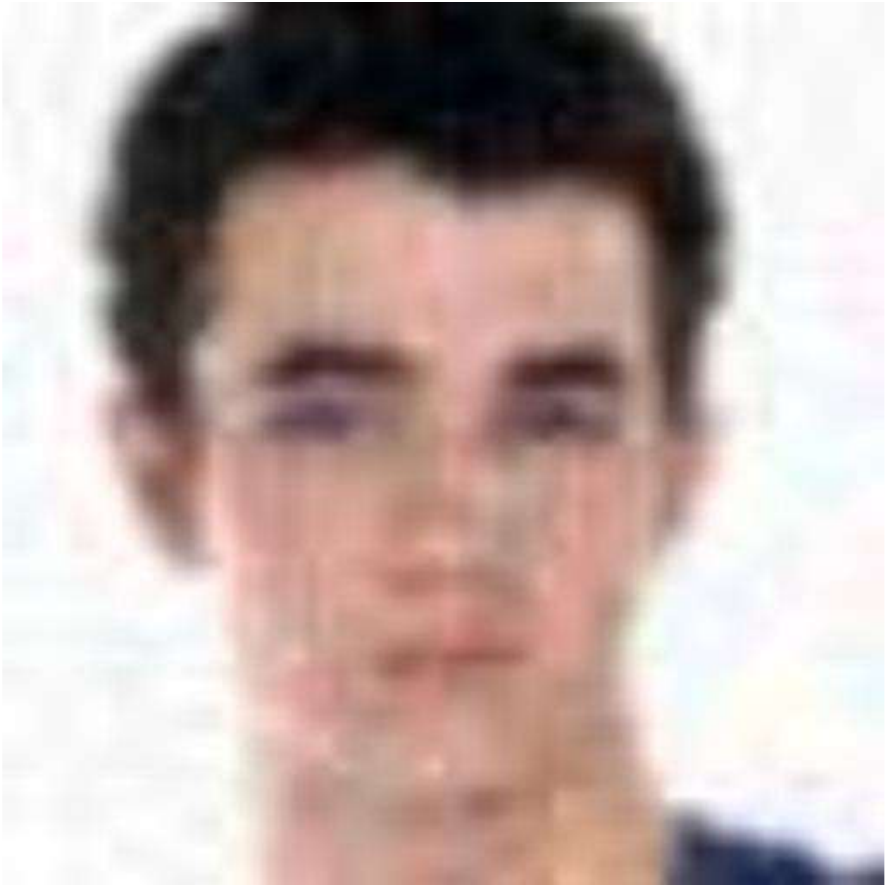} \hspace{-4mm} &
\includegraphics[width=0.163\linewidth]{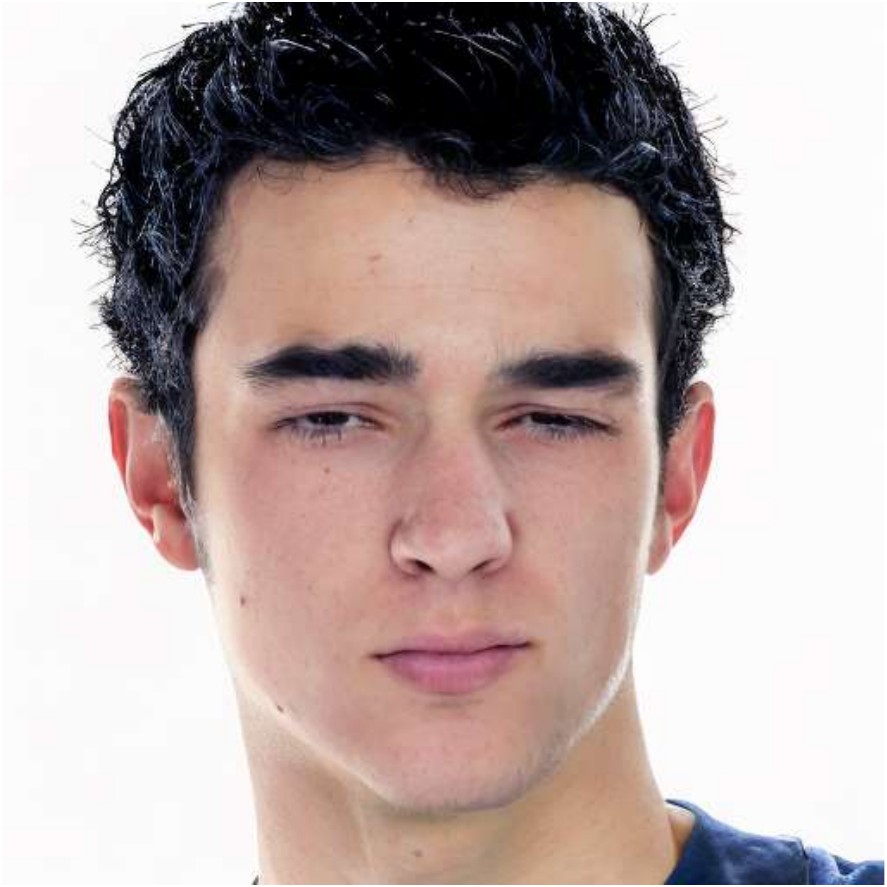} \hspace{-4mm} &
\includegraphics[width=0.163\linewidth]{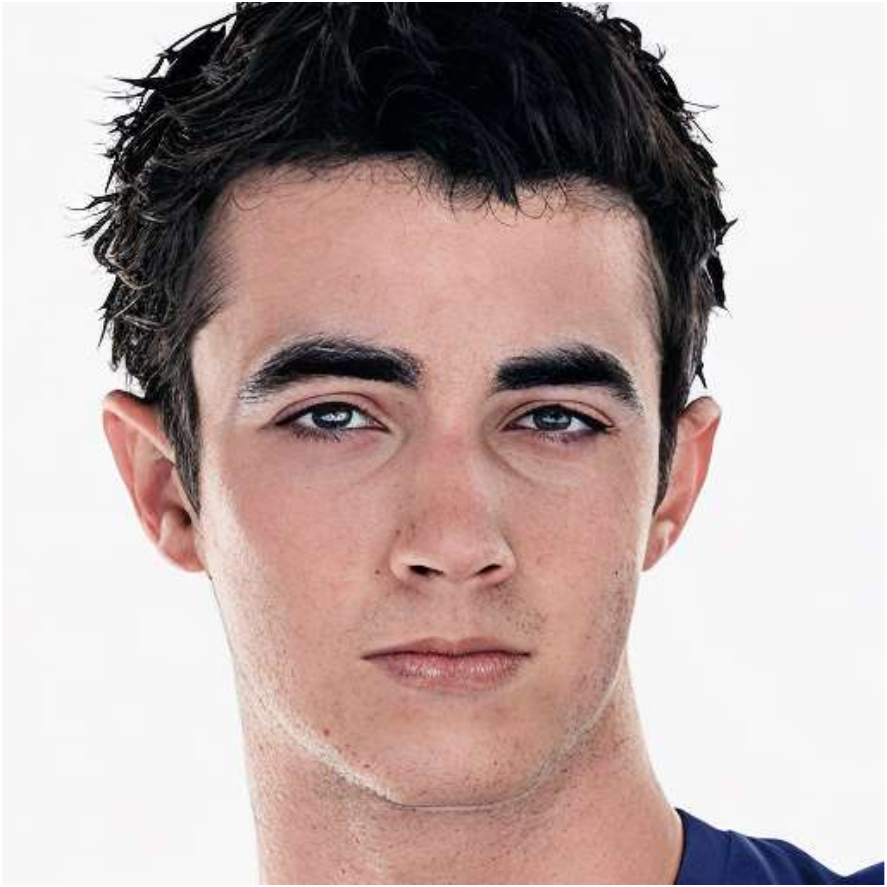}
\\
Input \hspace{-4mm} &
No Prompt \hspace{-4mm} &
\emph{`mouth slightly open'}\hspace{-4mm} &
Input \hspace{-4mm} &
No Prompt \hspace{-4mm} &
\emph{`big eyes'} 
\\
\end{tabular}
\end{adjustbox}
\end{tabular}
\caption{Influences of attribute prompts.}
\label{fig:D_control}
\vspace{-4.mm}
\end{figure}

\section{Controlling with Attributes Prompts}
\subsection{Controlling Restoration}
\label{sec:more}
Our model facilitates guidance through user-defined attribute prompts during testing. 
\cref{fig:D_control} exemplifies this with a demonstration of attribute prompt-controlled recovery. Notably, 'No Prompt' refers to the initial prompt input, 'A high quality, high resolution, realistic, and extremely detailed image.' As illustrates, users can employ prompts like 'old' to define the approximate age in the restored image, or 'eyeglasses' and 'earrings' to add accessories to the image. Furthermore, users can provide additional attribute prompts to refine unsatisfactory results. For instance, 'lipstick' can be used to add lipstick, or 'mouth slightly open' to adjust the mouth's appearance. More significantly, severe illusions, particularly in the eye area, are common in previous methods due to insufficient information in LQ inputs. This observation underscores the importance of attribute prompts in our method, as using 'big eyes' leads to more realistic eye effects. Therefore, we posit that attribute text holds potential as a versatile tool for controlling face recovery.

\subsection{Sensitivity Analysis}
\label{sec:sen}
\begin{figure*}[h]
\captionsetup{font={small}, skip=8pt}
\scriptsize
\centering
\begin{tabular}{ccc}
\hspace{-0.5cm}
\\
\hspace{-0.55cm}
\begin{adjustbox}{valign=t}
\begin{tabular}{c}
\end{tabular}
\end{adjustbox}
\begin{adjustbox}{valign=t}
\begin{tabular}{ccccccccccc}
\includegraphics[width=0.1092\linewidth]{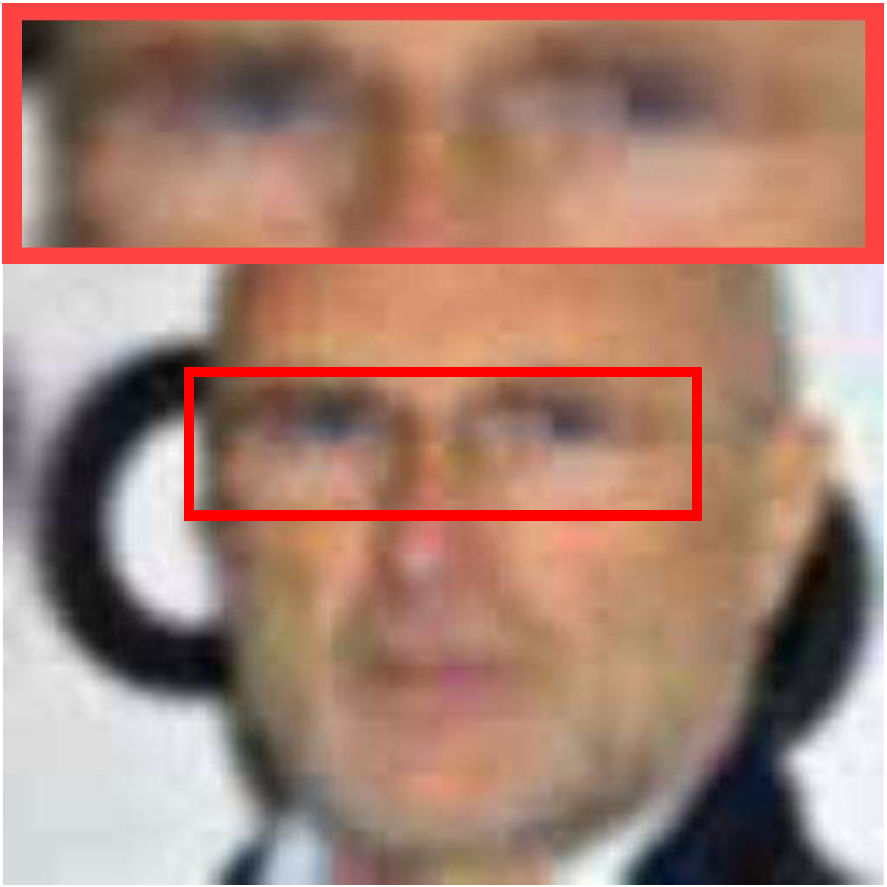} \hspace{-4.8mm} &
\includegraphics[width=0.1092\linewidth]{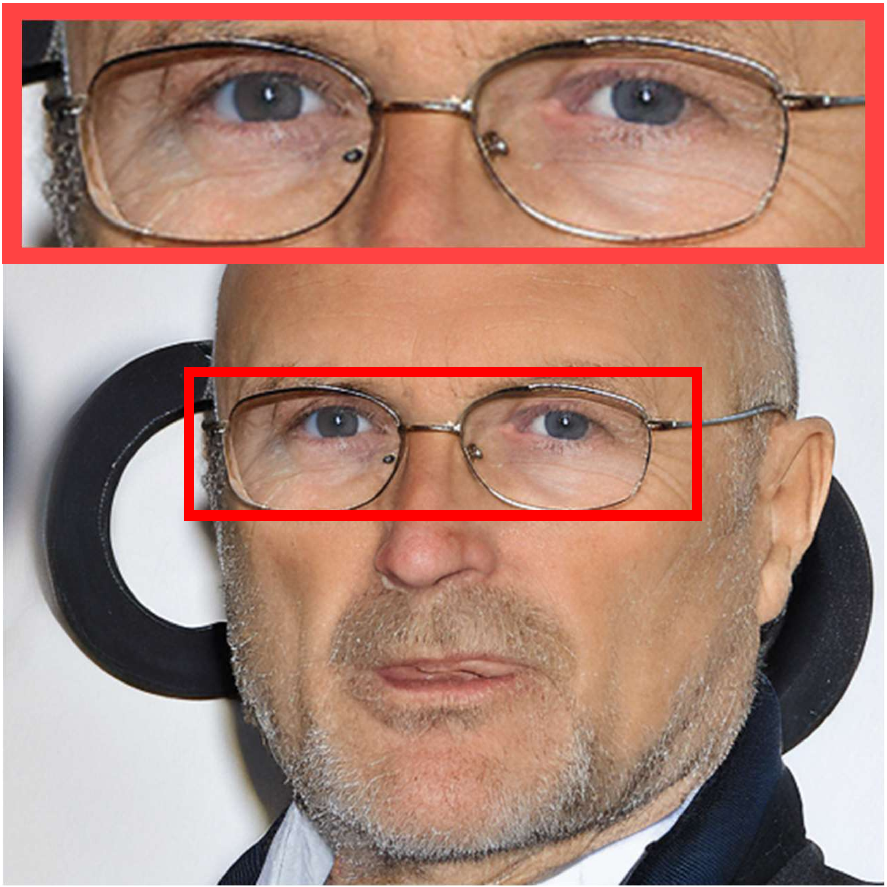}  \hspace{-4.6mm} &
\includegraphics[width=0.1092\linewidth]{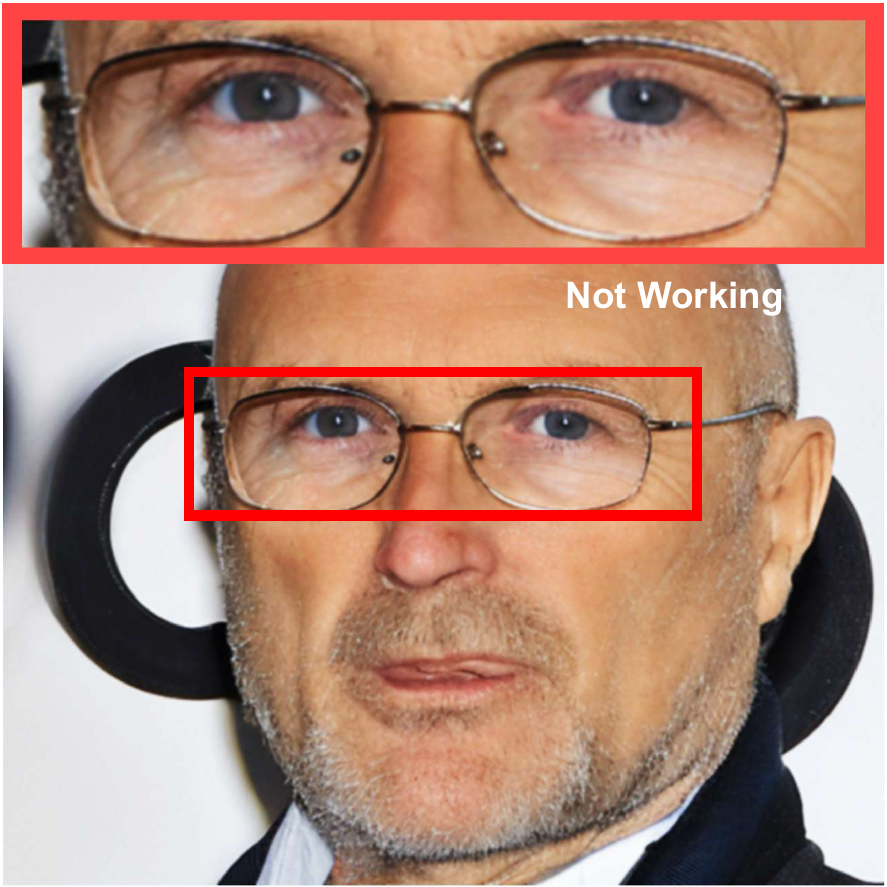}  \hspace{-4.6mm} &
\includegraphics[width=0.1092\linewidth]{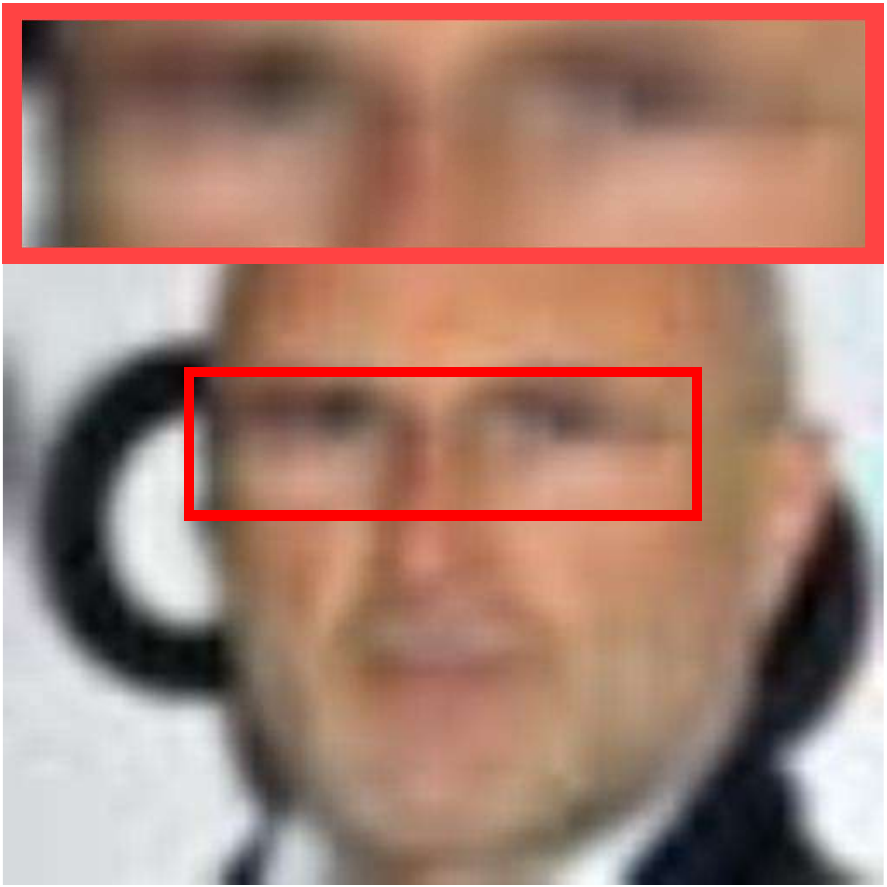}    \hspace{-4.6mm} &
\includegraphics[width=0.1092\linewidth]{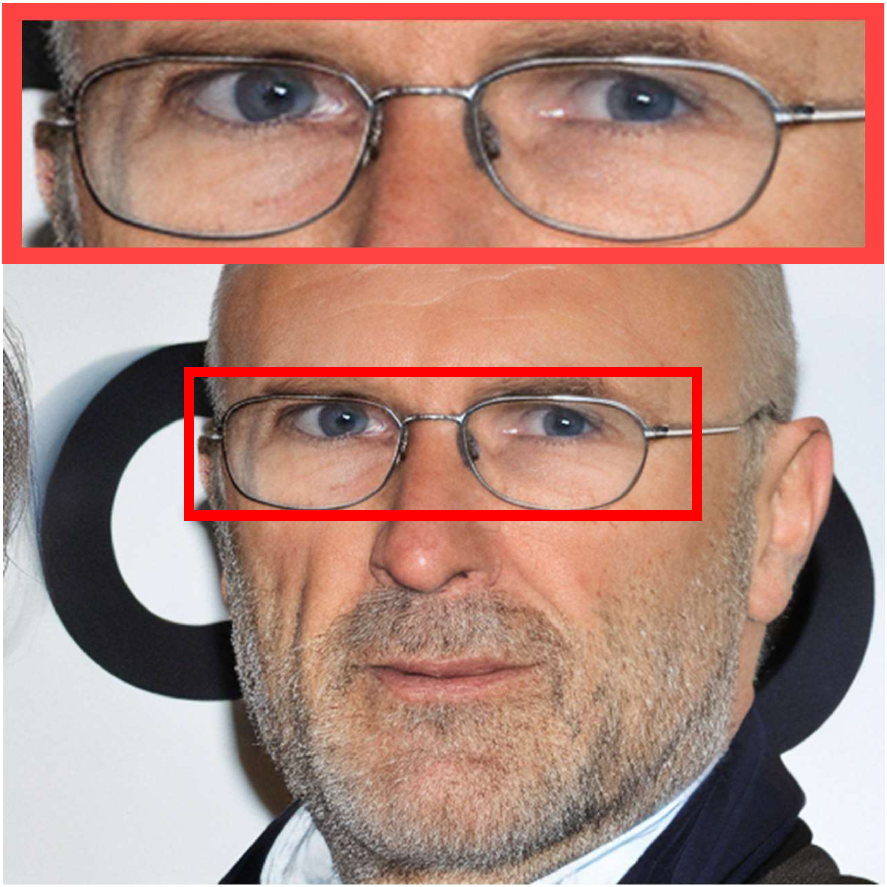}  \hspace{-4.6mm} &
\includegraphics[width=0.1092\linewidth]{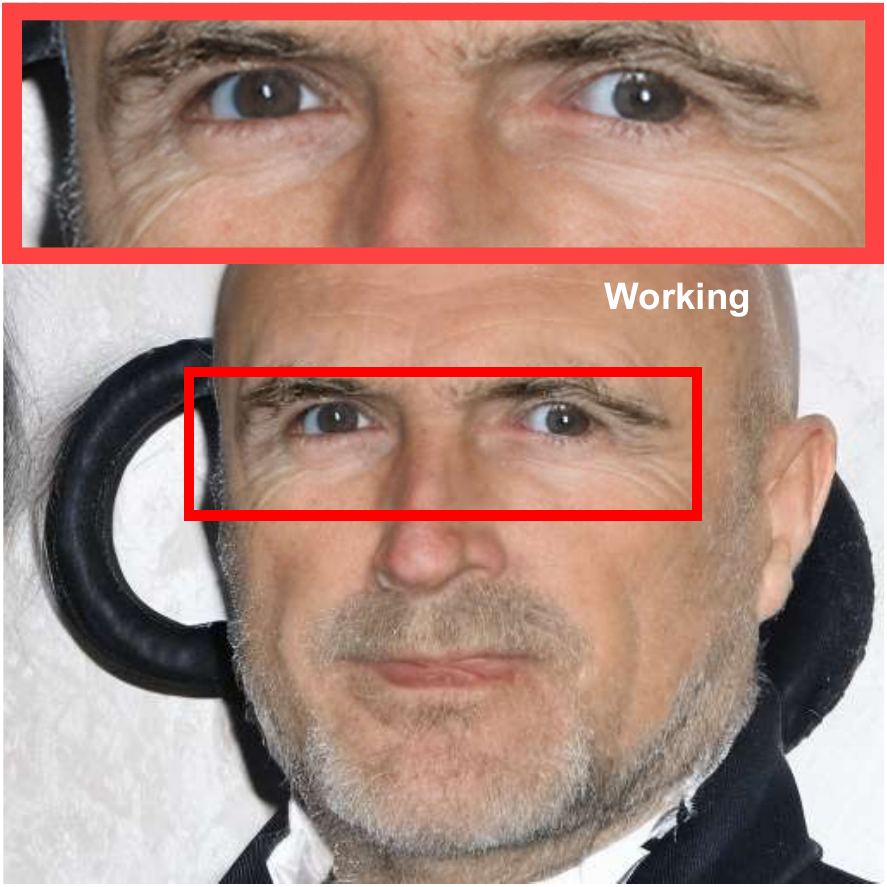}   \hspace{-4.6mm} &
\includegraphics[width=0.1092\linewidth]{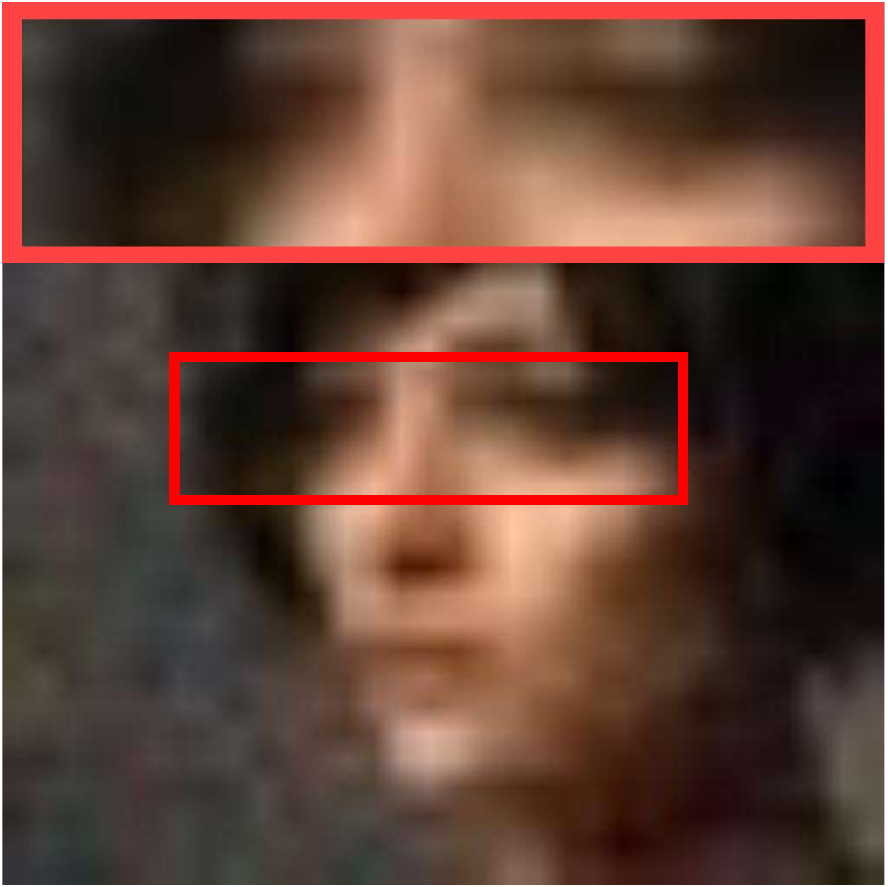}  \hspace{-4.6mm} &
\includegraphics[width=0.1092\linewidth]{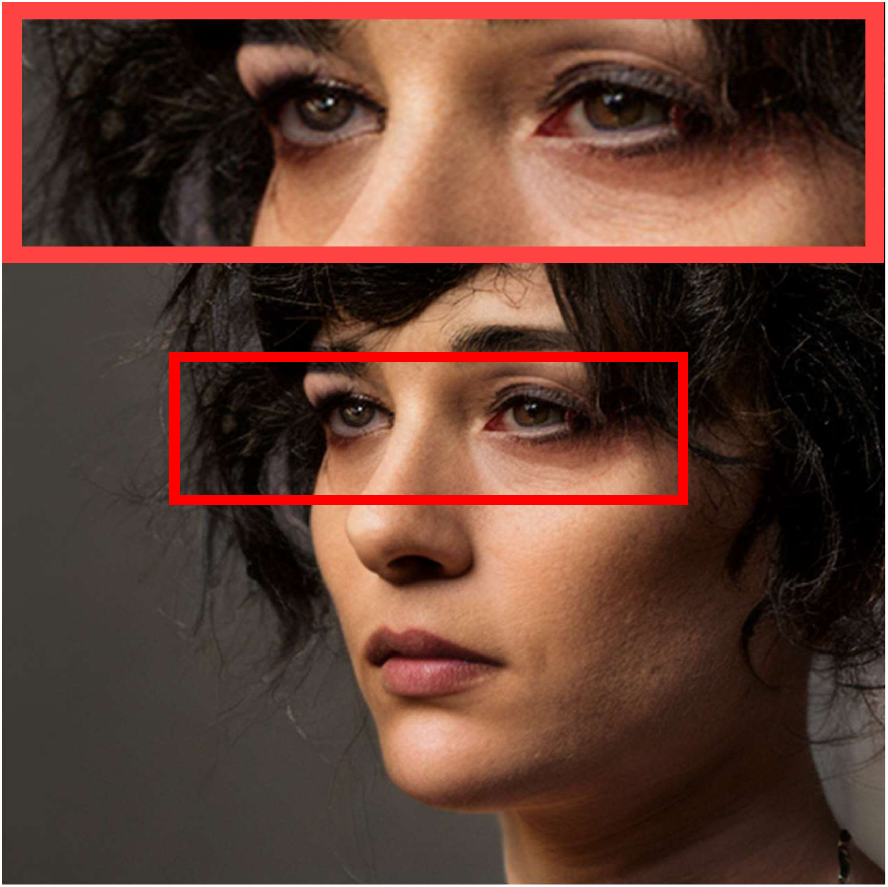}  
\hspace{-4.6mm} &
\includegraphics[width=0.1092\linewidth]{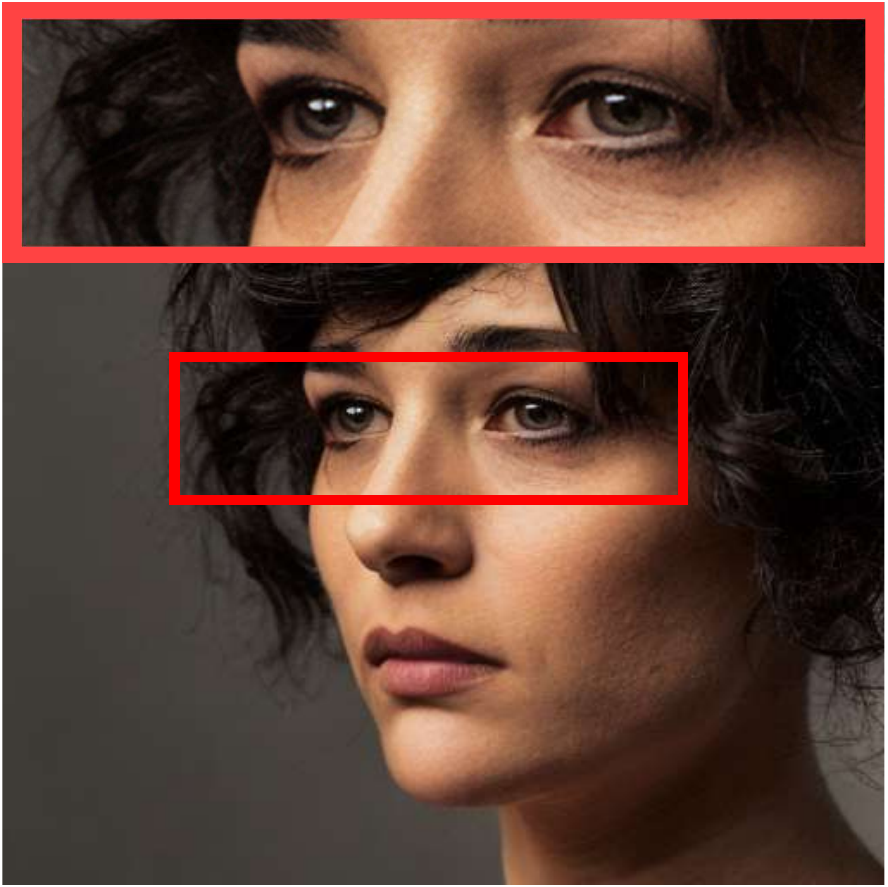} 
\\
Input ($\times$ 4) \hspace{-4.6mm} &
No Prompt \hspace{-4.6mm} &
\makecell{\emph{`no glasses'}} \hspace{-4.6mm} &
Input ($\times$ 8) \hspace{-4.6mm} &
No Prompt \hspace{-4.6mm} &
\makecell{\emph{`no glasses'}} \hspace{-4.6mm} &
Input \hspace{-4.6mm} &
No Prompt \hspace{-4.6mm} &
\makecell{\emph{`blue eyes'}} 
\\ 
\end{tabular}
\end{adjustbox}
\end{tabular}
\vspace{-2.mm}
\caption{We investigate the following options for attribute prompt control. First of all, the model becomes increasingly dependent on attribute prompt as input deterioration increases (case 1 \& case 2). Second, the input attribute tag does not have a control role if it is not present in \cref{tab-a} (case 3).}
\label{fig:D_non}
\vspace{-2.mm}
\end{figure*}

Moreover, as depicted in \cref{fig:D_non} case 1 and case 2, with increasing levels of degradation, the model's reliance on attribute prompts for control becomes more apparent, leading to greater flexibility. 
This observation, a logical experimental outcome, confirms the model's fidelity to LQ inputs during recovery.
Specifically, attribute prompts that starkly contradict the LQ input do not influence the effect, which aligns with our expectations. 
The primary function of attribute labels, we contend, is to facilitate more efficient and effective image restoration, rather than to focus on image editing and control. 
This is intrinsic to the core objective of real-world face restoration. In our method, all attribute labels listed in \cref{tab-a}, including 'black hair', 'brown hair', and others, do not possess the ability to control recovery but rather aid the model in interpreting the LQ input. 
These insights robustly underscore the effectiveness of our approach.

\section{User Study}
Currently, the relevance and efficacy of metrics such as PSNR, SSIM, and LPIPS require evaluation. In this study, a User study was conducted as an alternative metric for assessing image restoration quality. The study concentrated on two primary questions: (1) How does our model without reference images perform in terms of restoring image quality versus reducing facial illusions compared to previous methods? (2) Does the addition of reference image and identity information in guiding restoration result in images that are closer to the Ground Truth compared to the model without reference images? Two sets of questionnaires were prepared, and the study was conducted with 50 participants. Participants were presented with random, anonymous options for their selection. For question (1), our model was compared with DiffBIR \cite{diffbir}, VQFR \cite{gu2022vqfr}, and CodeFormer \cite{coderformer}, focusing on selecting images with better quality and fewer hallucinations, without providing Ground Truth images. This comparison involved 50 sets of images. For question (2), a self-comparison approach was adopted. Specifically, ground truth images were provided, and participants were asked to choose between restoration results with and without reference images, assessing them based on their proximity and realism to the ground truth. In this experiment, 50 pairs of synthetically degraded images were compared.

\begin{figure}[h]
  \captionsetup{font={small}, skip=6pt}
  \includegraphics[width=1\textwidth]{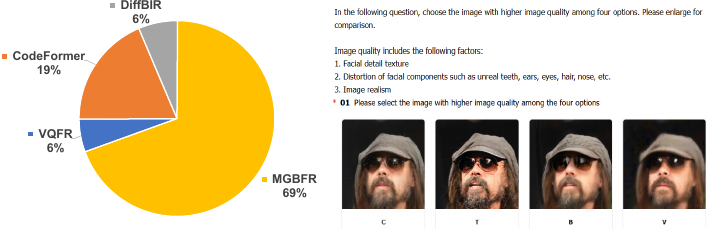}
  \caption{Results and question details of user study.}
  \label{fig:E1}
\end{figure}

\begin{figure}[h]
  \captionsetup{font={small}, skip=6pt}
  \includegraphics[width=1\textwidth]{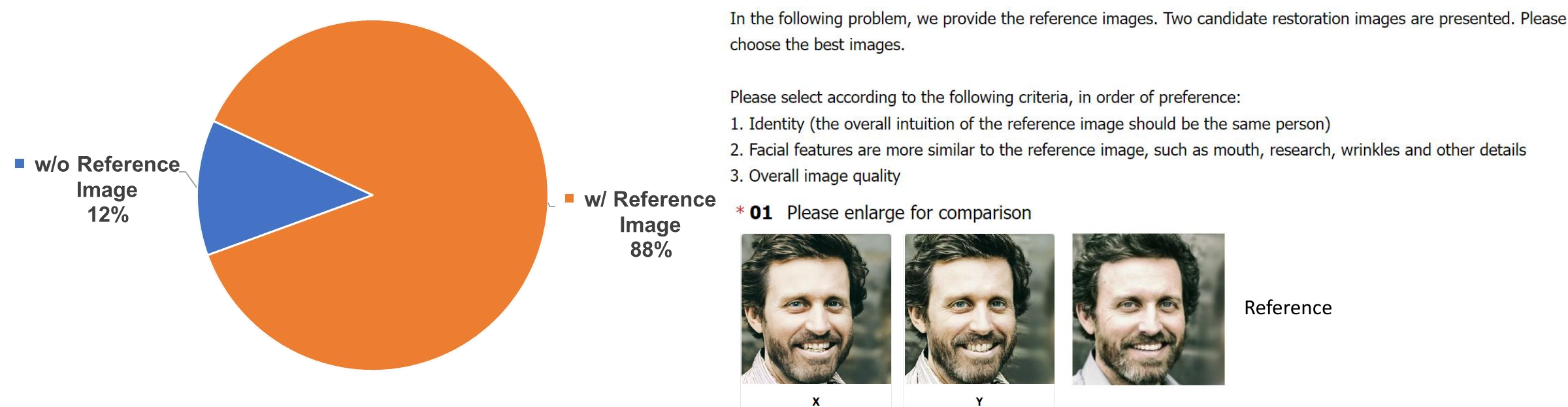}
  \caption{Results and question details of user study.}
  \label{fig:E2}
\end{figure}

Subsequently, the first part of the user study, focusing on the improvement of our model in terms of image quality and the reduction of facial illusion, is discussed. The results and detailed information of this study segment are presented in \cref{fig:E1} and \cref{fig:E2}. It was observed that the majority of the 50 participants favored our model for its superior image quality and minimal facial illusions. Reflecting on the recovery results of the advanced method CodeFormer, illustrated in \cref{sec:c_base}, it is noted that while CodeFormer achieves relatively good quality in restored images, considerable facial illusions persist, particularly around the mouth and eyes. In contrast, our method consistently produces high-quality, realistic facial images with minimal facial illusion. These findings underscore the our model's capability to reduce illusion and enhance image quality through negative prompts. Specifically, supported by the diffusion model and LR control adapter, our model is adept at generating realistic high-quality restorations influenced by negative prompts, and it effectively minimizes facial illusions by utilizing an optimal amount of attribute prompts. The synergy of these elements paves the way for further exploration in MGFR.

It is noteworthy that our two-part User study also corresponds to the two-stage development process of the MGFR model. For the second part, the first User study has demonstrated the superior performance of our model, as shown in the figure. Participants generally agreed that adding a guide to the reference image would further achieve superior visual effects.

\section{Ablation Study for Negative Prompt}
\label{sec:Appen_prompt}
\begin{figure}[t]
  \captionsetup{font={small}, skip=6pt}
  \includegraphics[width=1\textwidth]{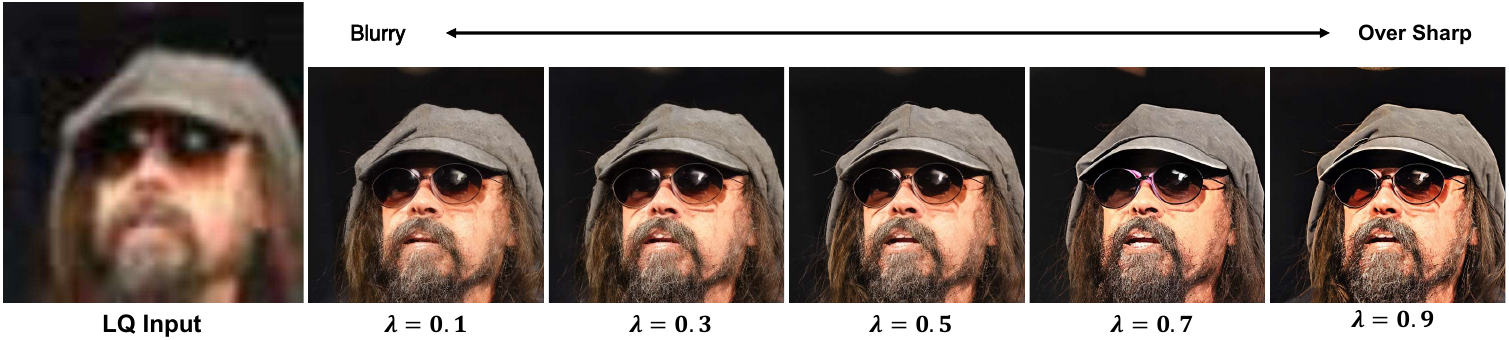}
  \caption{Influence of hyperparameters on recovery effect in CFG. The smaller $\lambda$ does not get a clear recovery result and the huge $\lambda$ causes the recovered image to be over sharp.}
  \label{fig:ANP}
  
\end{figure}

For negative prompts, we introduce two hyperparameters, $\lambda_{na}$ and $\lambda_{nq}$. However, we find that the changes of the two values tend to have the same effect on the restored images. Thus, we keep $\lambda_{na} = 0.5 $ and $\lambda_{nq} = 0.5 $ during reasoning. Here we will represent the values of $\lambda_{na}$ and $\lambda_{nq}$ with $\lambda$ to show the qualitative comparison results under different hyperparameters in \cref{fig:ANP}.

\section{Impact Statements}
\label{sec:impact}
Controlled generation technology, as a pivotal innovation in the field of diffusion models, exerts a significant impact across multiple sectors of society. In the creative industries, it enables artists and designers to realize complex visions with unprecedented precision and flexibility, fostering innovation in digital art, design, and multimedia content creation. In commercial applications, controlled generation technology enhances marketing strategies by offering more targeted and dynamic advertising visuals, effectively engaging consumers. Additionally, its influence extends to education and training, where it can revolutionize teaching methods and materials, especially in visually-dependent disciplines, by generating customized educational content and simulations.

The work presented in this paper aims to advance machine learning and computer vision. This method can provide the public with better face processing effects and has greater social value. However, the technique is designed to process facial information, inevitably involving facial attributes such as race and privacy risks. We are aware of these risks. Our research uses publicly available data and images accompanied by captions. We are also wary of potentially discriminatory attribute descriptions in our research. Our method also provides control over face restoration, which reduces the possibility of our method outputting harmful information.

\section{More Qualitative Comparisons For MGFR Model}
\begin{figure}[h]
  \captionsetup{font={small}, skip=4pt}
  \centering
  \includegraphics[width=1\textwidth]{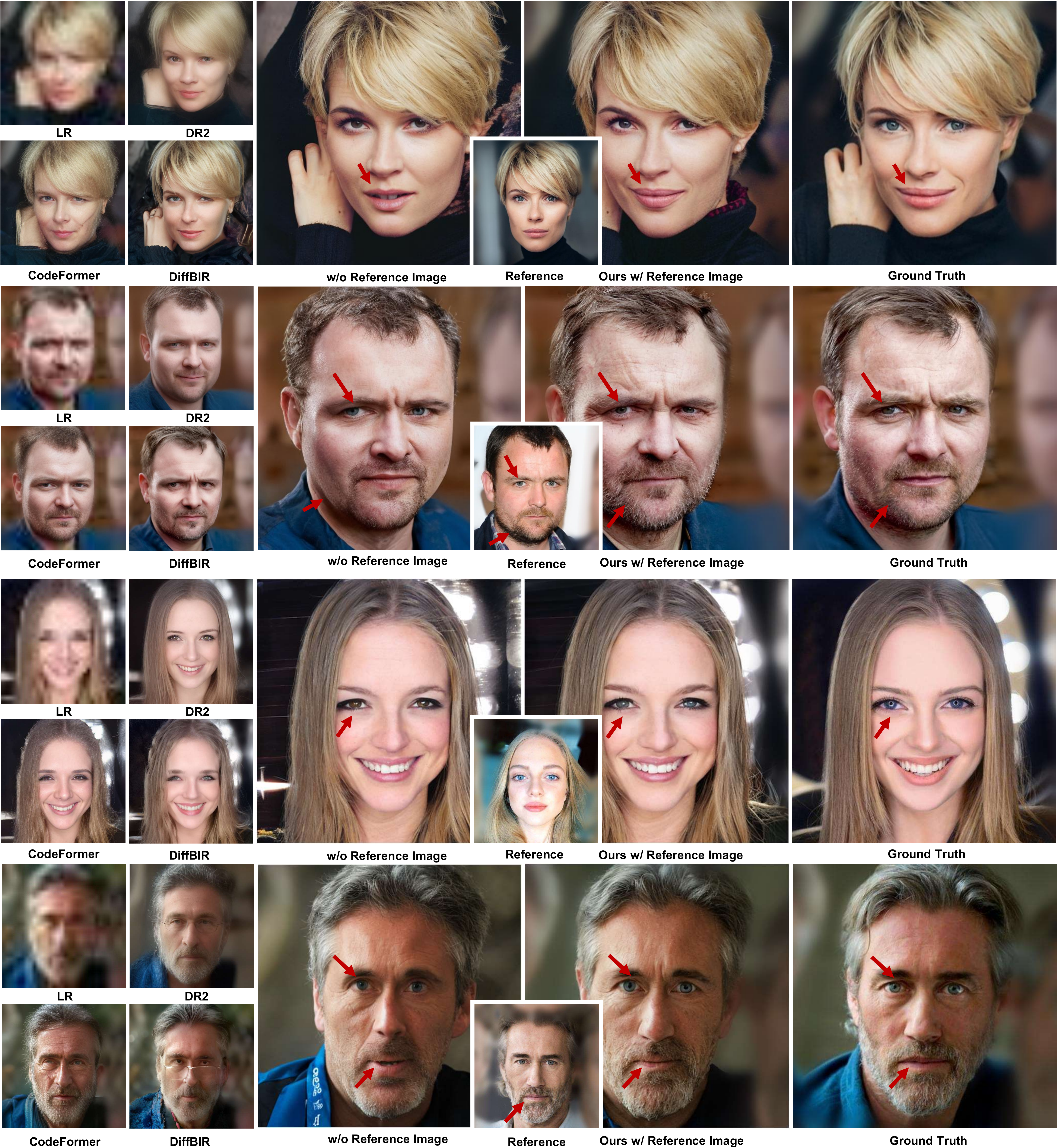}
  \caption{More qualitative comparisons for MGFR with reference image and ID guidance on synthetic dataset in Reface-Test dataset. Zoom in for best view.}
  \label{fig26}
  
\vspace{-5mm}
\end{figure}

\cref{fig26} displays the qualitative comparison results between the proposed MGFR model and other advanced methods. The ``w/o Reference Image'' represents the restoration results of our model after initial training. The use of the negative intuition strategy and attribute prompts significantly reduces the false illusions in face images and substantially enhances overall quality. Subsequently, the inclusion of additional multi-modal information, such as reference images and identity information, can achieve superior visual effects.

\add{\section{Scalability of MGFR for real-world video face restoration}
The proposed MGFR framework shows significant potential for real-world video-based face recovery tasks. Unlike single-image restoration, video restoration poses the unique challenge of ensuring temporal consistency. To address this, our method leverages the recovered output of the previous frame as a reference for the current frame. This approach aligns seamlessly with our model architecture, which integrates high-quality continuous frame references into guided restoration. Additionally, as our model does not require strict alignment between the reference and low-quality inputs, it effectively handles natural variations in pose and expression commonly found in consecutive video frames, surpassing previous reference-based face restoration models. By leveraging temporal dependencies between frames, the proposed method ensures identity consistency and high-quality recovery in video sequences. Future work could enhance this approach by integrating explicit temporal models or constraints, such as optical flow guidance, to better handle motion artifacts and dynamic variations in video data. Unlike single-image restoration based on reference images, video data offers more diverse and abundant training samples, which we believe will further unlock the potential of our proposed model. This will be a key focus of our future work.}

\vspace{-2mm}
\add{\section{Model stability}}
 \begin{figure*}[h]
\captionsetup{font={small}, skip=12pt}
\scriptsize
\begin{tabular}{ccc}
\hspace{-0.55cm}
\begin{adjustbox}{valign=t}
\begin{tabular}{c}
\end{tabular}
\end{adjustbox}
\begin{adjustbox}{valign=t}
\begin{tabular}{ccccccccc}
\includegraphics[width=0.12\linewidth]{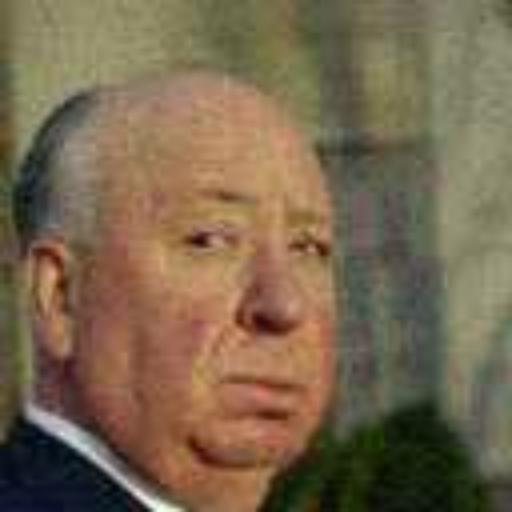} \hspace{-4mm} &
\includegraphics[width=0.12\linewidth]{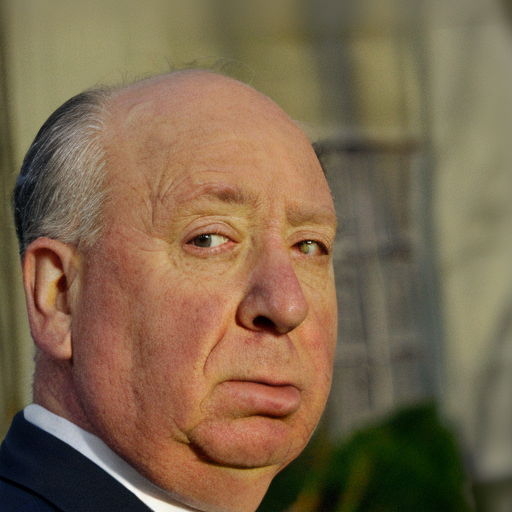} \hspace{-4mm} &
\includegraphics[width=0.12\linewidth]{Add_more/STB/HR-1.png} \hspace{-4mm} &
\includegraphics[width=0.12\linewidth]{Add_more/STB/HR-1.png} \hspace{-4mm} &
\includegraphics[width=0.12\linewidth]{Add_more/STB/HR-1.png} \hspace{-4mm} &
\includegraphics[width=0.12\linewidth]{Add_more/STB/HR-1.png} \hspace{-4mm} &
\includegraphics[width=0.12\linewidth]{Add_more/STB/HR-1.png} \hspace{-4mm} &
\includegraphics[width=0.12\linewidth]{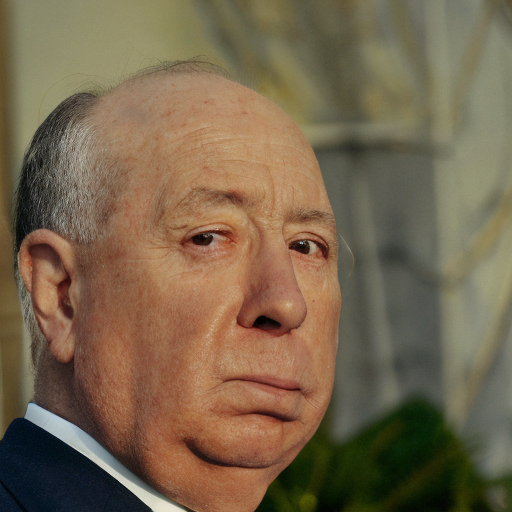}
\\
LR \hspace{-4.6mm} &
Seed 2982 \hspace{-4.6mm} &
Seed 78 \hspace{-4.6mm} &
seed 11281 \hspace{-4.6mm} &
seed 15 \hspace{-4.6mm} &
seed 424 \hspace{-4.6mm} &
seed 9168 \hspace{-4.6mm} &
GT
\\
\end{tabular}
\end{adjustbox}
\end{tabular}
\vspace{-2.mm}
\caption{\textbf{Model Stability Analysis}. The recovery results of MGFR remain consistent across different random seeds, eliminating the need for selection among multiple input outcomes.}
\label{fig:a-s}
\vspace{-5mm}
\end{figure*}

\add{\section{Brief Overview of Evaluation Metrics}
For quantitative comparison, the selected image quality evaluation metrics include full-reference metrics PSNR, SSIM, and LPIPS \cite{zhang2018unreasonable}. \cite{yu2024scaling,jinjin2020pipal} experiment initially confirmed that as image restoration quality improves, the reference utility of metrics such as PSNR, SSIM, and LPIPS needs to be re-evaluated, necessitating the selection of more effective evaluation indicators. Therefore, we introduce three non-reference metrics—ManIQA \cite{yang2022maniqa}, ClipIQA \cite{wang2023exploring}, and MUSIQ \cite{ke2021musiq}—in this work. }

\add{
A summary of each evaluation metric is provided below.
\begin{itemize}
    \item \textbf{SSIM} is a key metric for assessing image restoration quality, measuring the similarity between the restored and original images based on brightness, contrast, and structural information. It has been widely used in previous face image restoration tasks \cite{diffbir,dr2,gfp,gpen,coderformer,gu2022vqfr,yu2024scaling,chan2021glean,chen2021progressive,9921338,Li_2020_CVPR,li2020blind,wang2021towards,li2020enhanced,teng2022blind}.
    \item \textbf{PSNR} is a metric derived from the mean square error (MSE), calculated as the logarithmic ratio of the maximum possible pixel value to the error. The results are expressed in decibels (dB), where higher values signify better image quality. It has been widely used in previous face image restoration tasks \cite{diffbir,dr2,gfp,gpen,coderformer,gu2022vqfr,Dogan_2019_CVPR_Workshops,yu2024scaling,chan2021glean,chen2021progressive,9921338,Li_2020_CVPR,li2020blind,wang2021towards,li2020enhanced,teng2022blind}.
    \item \textbf{LPIPS} quantifies image differences by extracting features from deep neural networks and measuring the distances between these features. This metric better captures perceptual changes in image details and textures. Previous studies have emphasized image similarity metrics aligned with human visual perception \cite{diffbir,dr2,gfp,gpen,coderformer,gu2022vqfr,yu2024scaling,chan2021glean,chen2021progressive,9921338,Li_2020_CVPR,li2020blind,wang2021towards,li2020enhanced,teng2022blind}.
    \item \textbf{ManIQA} maps images into a low-dimensional manifold space and analyzes their feature distribution and location to assess image quality. This approach demonstrates a high correlation with perceived quality, and its effectiveness has been validated in \cite{yu2024scaling}.
    \item \textbf{MUSIQ} implements a multi-scale feature extraction mechanism designed to capture the quality characteristics of images across varying resolutions and perceptual scales for effective image quality evaluation, and its effectiveness has been validated in \cite{yu2024scaling}.
    \item \textbf{ClipIQA} leverages the robust vision-language priors embedded within the CLIP model. The focus is on enhancing the capability to evaluate both quality perception (seeing) and abstract perception (feeling) of visual content. This approach's effectiveness has been demonstrated in \cite{yu2024scaling}.
\end{itemize}
}

\add{\section{Qualitative Comparison with SUPIR}}
\label{sec:supir}
 \begin{figure*}[h]
\captionsetup{font={small}, skip=12pt}
\scriptsize
\begin{tabular}{ccc}
\hspace{-0.85cm}
\begin{adjustbox}{valign=t}
\begin{tabular}{c}
\end{tabular}
\end{adjustbox}
\begin{adjustbox}{valign=t}
\begin{tabular}{cccccccc}
\includegraphics[width=0.244\linewidth]{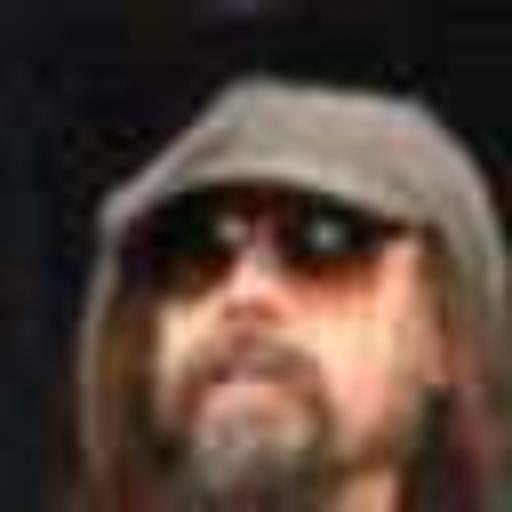} \hspace{-4mm} &
\includegraphics[width=0.244\linewidth]{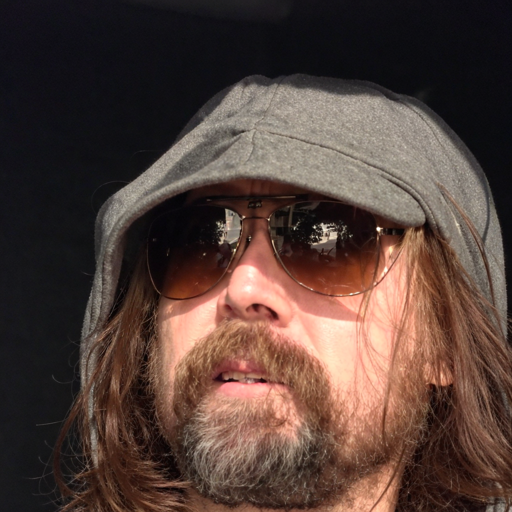}   \hspace{-4mm} &
\includegraphics[width=0.244\linewidth]{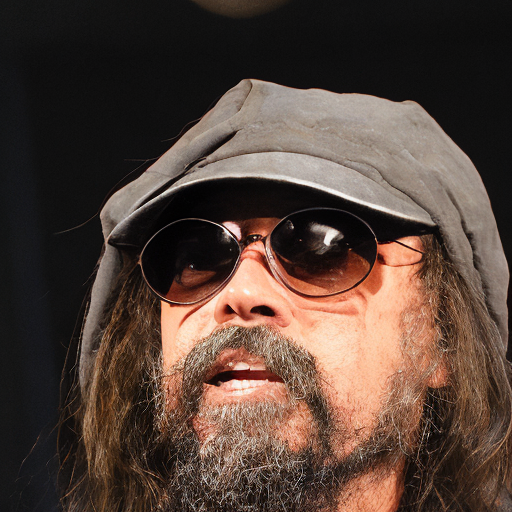}  \hspace{-4mm} &
\includegraphics[width=0.244\linewidth]{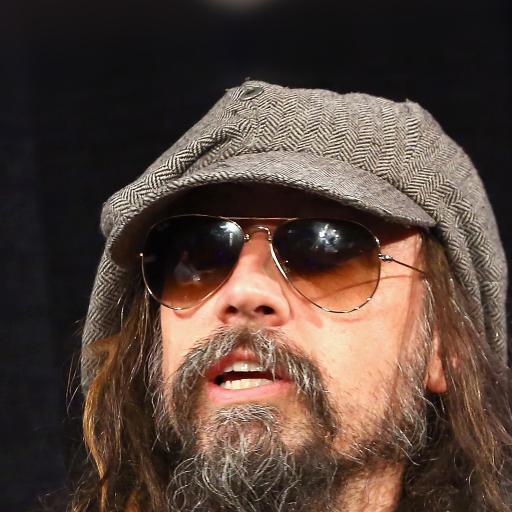} 
\end{tabular}
\end{adjustbox}
\vspace{0.1mm}
\\
\hspace{-0.55cm}
\begin{adjustbox}{valign=t}
\begin{tabular}{c}
\end{tabular}
\end{adjustbox}
\begin{adjustbox}{valign=t}
\begin{tabular}{cccccccc}
\includegraphics[width=0.244\linewidth]{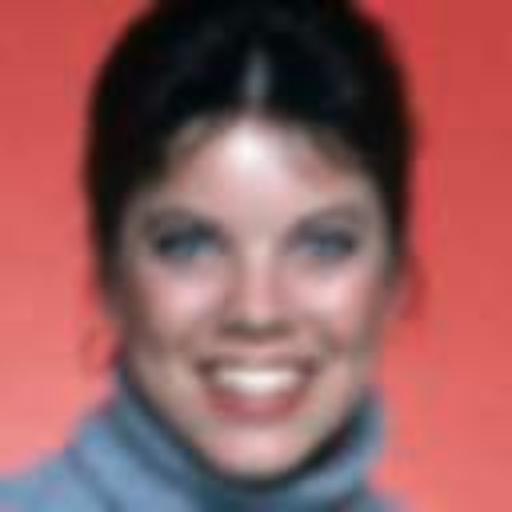} \hspace{-4mm} &
\includegraphics[width=0.244\linewidth]{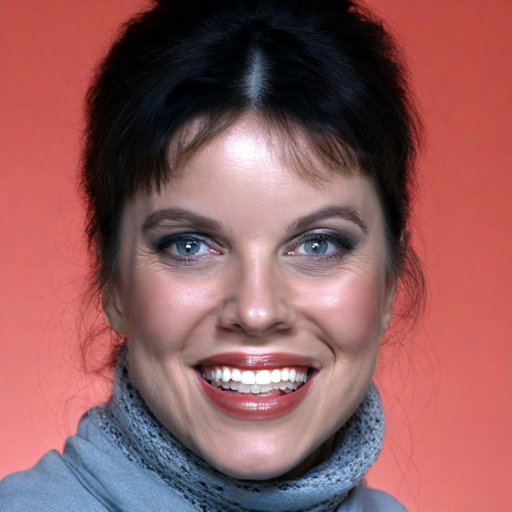}   \hspace{-4mm} &
\includegraphics[width=0.244\linewidth]{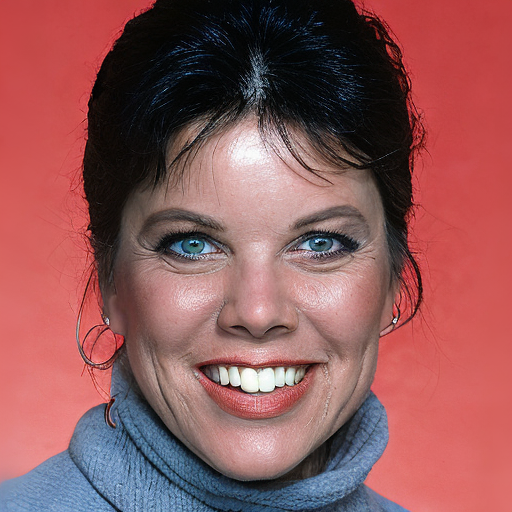}  \hspace{-4mm} &
\includegraphics[width=0.244\linewidth]{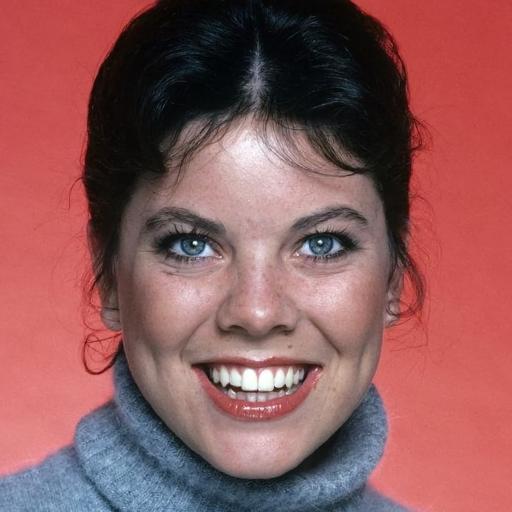} &  
\\
\includegraphics[width=0.244\linewidth]{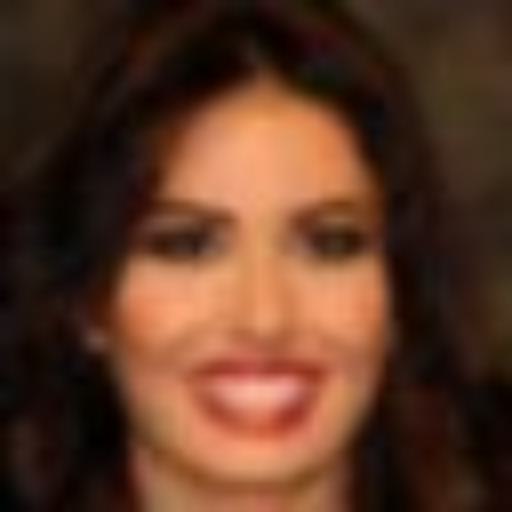} \hspace{-4mm} &
\includegraphics[width=0.244\linewidth]{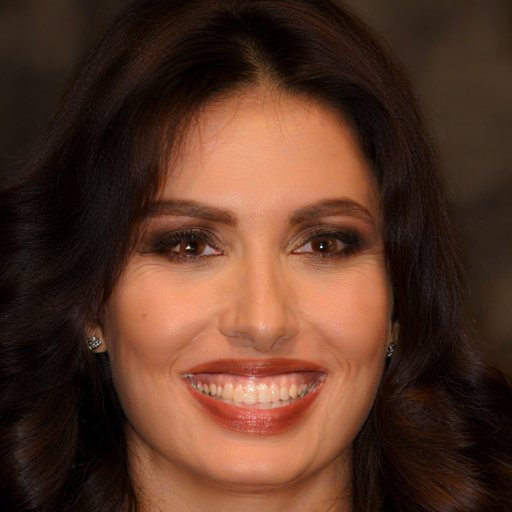}   \hspace{-4mm} &
\includegraphics[width=0.244\linewidth]{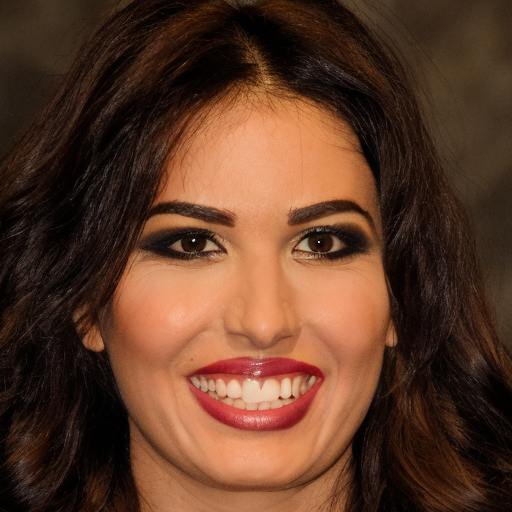}  \hspace{-4mm} &
\includegraphics[width=0.244\linewidth]{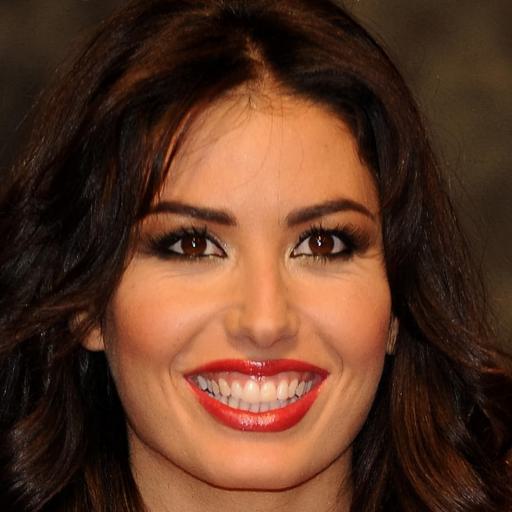} &  
\\
LR \hspace{-4mm} &
SUPIR \hspace{-4mm} &
Ours w/o Ref. \hspace{-4mm} &
GT \hspace{-4mm} &
\\
\end{tabular}
\end{adjustbox}
\end{tabular}
\vspace{-5.mm}
\caption{Qualitative comparisons with SUPIR \cite{yu2024scaling} for our text-guided baseline model on synthetic dataset under moderate degradation in CelebA-Test dataset. Zoom in for the best view.}
\label{fig:supit-1}
\vspace{-2.mm}
\end{figure*}

 \begin{figure*}[h]
\captionsetup{font={small}, skip=14pt}
\scriptsize
\begin{tabular}{ccc}
\hspace{-0.55cm}
\begin{adjustbox}{valign=t}
\begin{tabular}{c}
\end{tabular}
\end{adjustbox}
\begin{adjustbox}{valign=t}
\begin{tabular}{cccccccc}
\includegraphics[width=0.1933\linewidth]{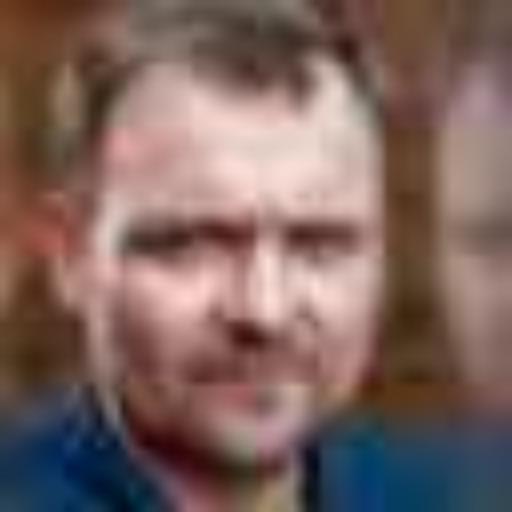} \hspace{-4mm} &
\includegraphics[width=0.1933\linewidth]{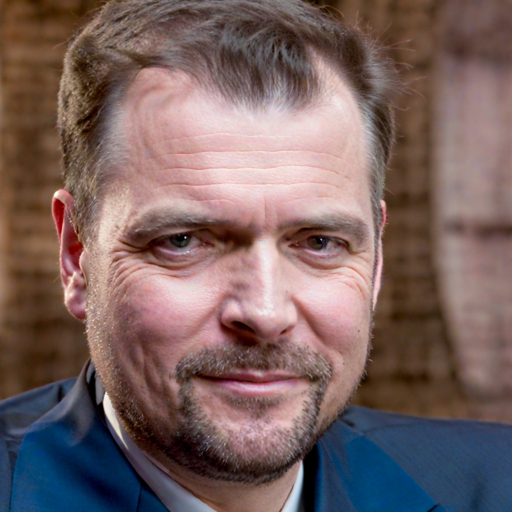}    \hspace{-4mm} &
\includegraphics[width=0.1933\linewidth]{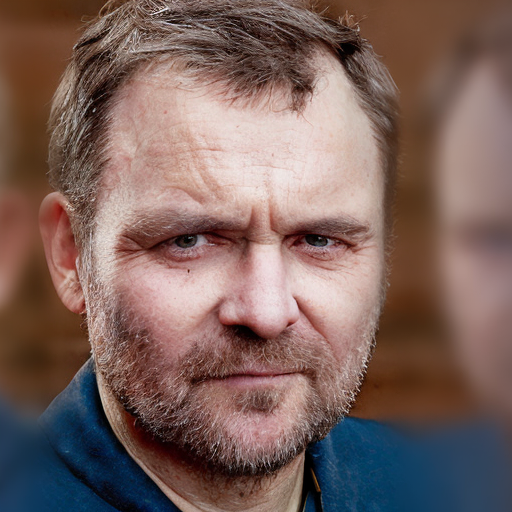}  \hspace{-4mm} &
\includegraphics[width=0.1933\linewidth]{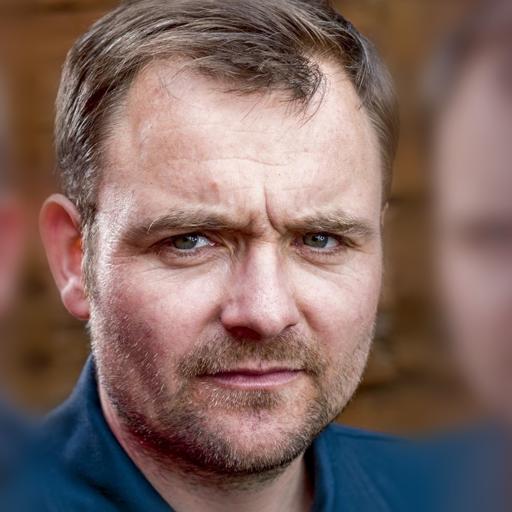}  \hspace{-4mm} &
\includegraphics[width=0.1933\linewidth]{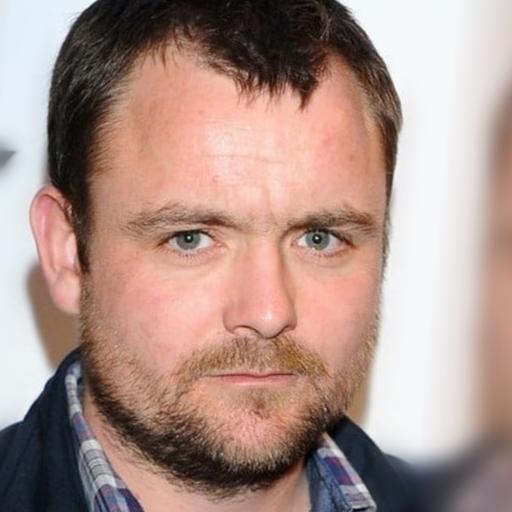} 
\end{tabular}
\end{adjustbox}
\vspace{0.1mm}
\\
\hspace{-0.55cm}
\begin{adjustbox}{valign=t}
\begin{tabular}{c}
\end{tabular}
\end{adjustbox}
\begin{adjustbox}{valign=t}
\begin{tabular}{cccccccc}
\includegraphics[width=0.1933\linewidth]{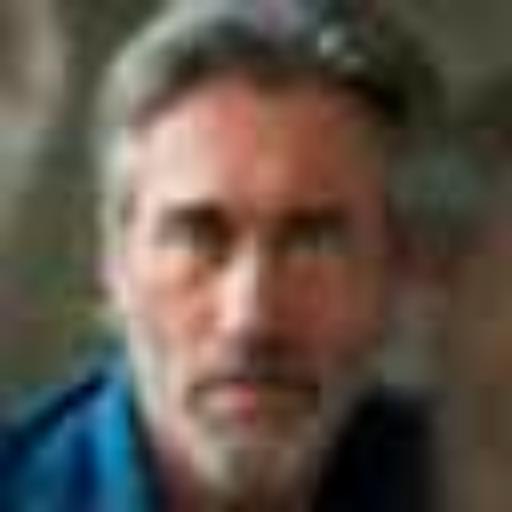} \hspace{-4mm} &
\includegraphics[width=0.1933\linewidth]{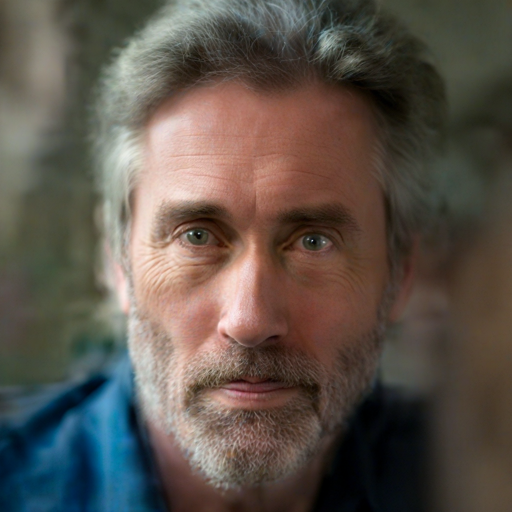}    \hspace{-4mm} &
\includegraphics[width=0.1933\linewidth]{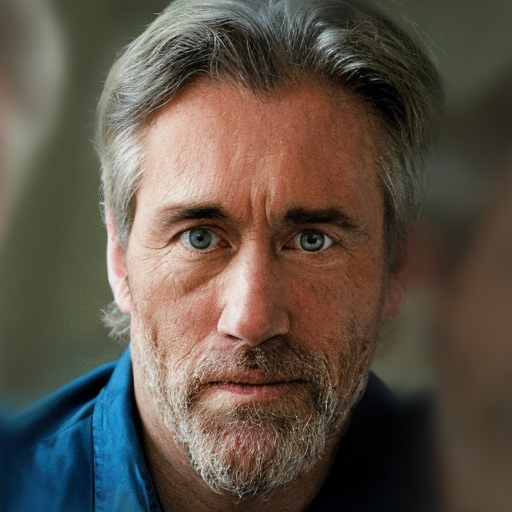}  \hspace{-4mm} &
\includegraphics[width=0.1933\linewidth]{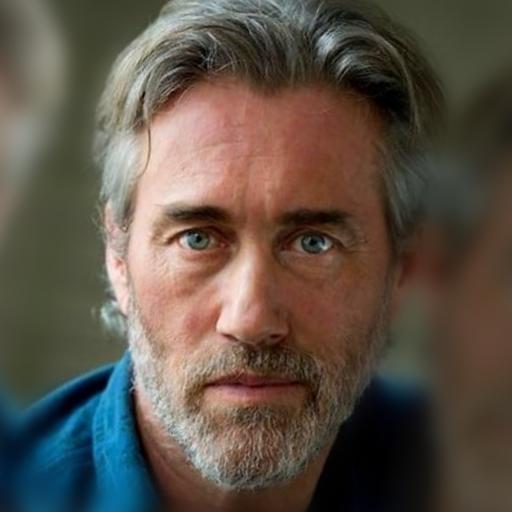}  \hspace{-4mm} &
\includegraphics[width=0.1933\linewidth]{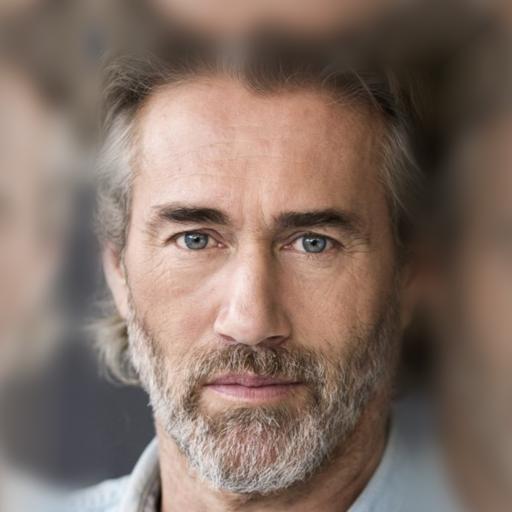} 
\\
LR \hspace{-4mm} &
SUPIR \hspace{-4mm} &
Ours w/ Reference \hspace{-4mm} &
GT \hspace{-4mm} &
Reference

\\
\end{tabular}
\end{adjustbox}
\end{tabular}
\vspace{-5.mm}
\caption{Qualitative comparisons with SUPIR \cite{yu2024scaling} for MGFR on synthetic dataset under moderate degradation in Reface-Test dataset. Zoom in for best view.}
\label{fig:supir-1}
\vspace{-2.mm}
\end{figure*}

 \begin{figure*}[h]
\captionsetup{font={small}, skip=14pt}
\scriptsize
\begin{tabular}{ccc}
\hspace{-0.55cm}
\begin{adjustbox}{valign=t}
\begin{tabular}{c}
\end{tabular}
\end{adjustbox}
\begin{adjustbox}{valign=t}
\begin{tabular}{cccccccc}
\includegraphics[width=0.1933\linewidth]{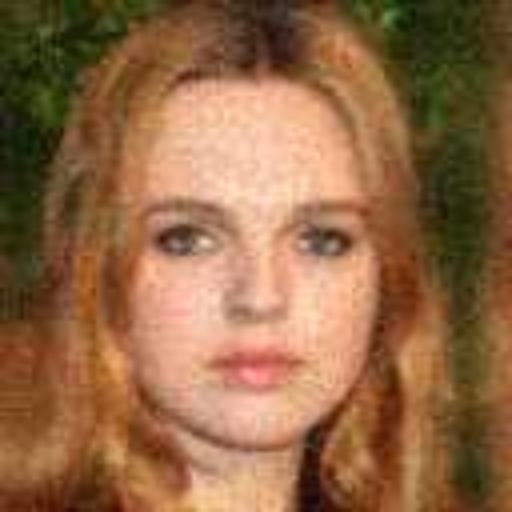} \hspace{-4mm} &
\includegraphics[width=0.1933\linewidth]{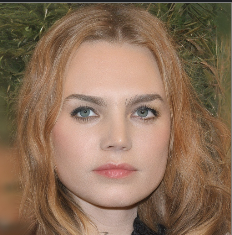}    \hspace{-4mm} &
\includegraphics[width=0.1933\linewidth]{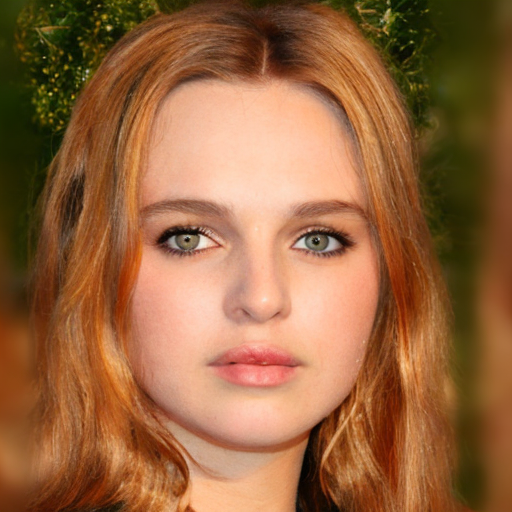}  \hspace{-4mm} &
\includegraphics[width=0.1933\linewidth]{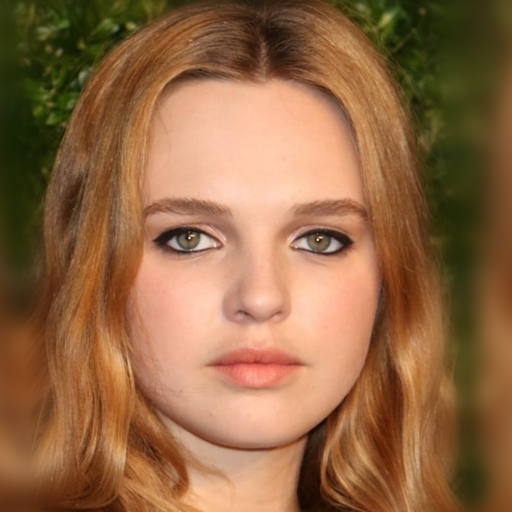}  \hspace{-4mm} &
\includegraphics[width=0.1933\linewidth]{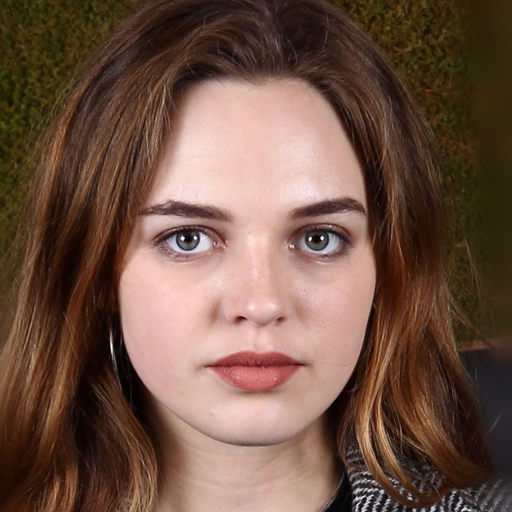} 
\\
LR \hspace{-4mm} &
Ours w/ SCA \hspace{-4mm} &
Ours w/ DCA \hspace{-4mm} &
GT \hspace{-4mm} &
Reference

\\
\end{tabular}
\end{adjustbox}
\end{tabular}
\vspace{-5.mm}
\caption{Ablation experiments comparing the reception of multi-modal information using a single control adapter (SCA) versus a dual control adapter (DCA) revealed that SCA led to reduced recovery performance and increased chromatic aberration.}
\label{fig:one-1}
\vspace{-2.mm}
\end{figure*}

\end{document}